\documentclass[10pt,twocolumn,letterpaper]{article}

\usepackage[pagenumbers]{cvpr} 

\usepackage[dvipsnames]{xcolor}

\definecolor{light-grey}{RGB}{120, 120, 120}
\definecolor{BrickRed}{RGB}{140, 0, 0}
\definecolor{LighterBlue}{RGB}{47, 140, 255}
\definecolor{lightblue}{RGB}{200, 200, 255}

\definecolor{green}{rgb}{0.21,0.74,0.49}
\newcommand{\best}[1]{{\textbf{#1}}}

\newcommand{\tablefont}[0]{\fontsize{8.5pt}{8.5pt}\selectfont}

\newcommand{\vertex}{\ensuremath{\text{v}}}
\newcommand{\imgspace}{\ensuremath{\mathcal{I}}}
\newcommand{\uvspace}{\ensuremath{\mathcal{U}}}
\newcommand{\uvcoord}{\ensuremath{\nu}}
\newcommand{\flowmap}{\ensuremath{\mathbf{F}}}
\newcommand{\uvtoimgmap}{\ensuremath{\mathbf{F}}}

\newcommand{\alignment}{\ensuremath{A}}

\definecolor{cvprblue}{rgb}{0.21,0.49,0.74}
\usepackage[pagebackref,breaklinks,colorlinks,citecolor=cvprblue]{hyperref}

\usepackage{multirow}
\usepackage{algorithm}
\usepackage[noend]{algpseudocode}
\usepackage{gensymb}
\usepackage{comment}
\usepackage{svg}
\usepackage{amsmath}
\usepackage{siunitx}
\usepackage[normalem]{ulem} 
\usepackage{makecell}
\usepackage{bm}

\makeatletter
\renewcommand{\paragraph}{%
  \@startsection{paragraph}{4}%
  {\z@}{1.5ex \@plus 1ex \@minus .2ex}{-1em}
  {\normalfont\normalsize\bfseries}%
}
\makeatother

\makeatletter
\def\BState{\State\hskip-\ALG@thistlm}
\makeatother

\def\paperID{4162} 
\def\confName{CVPR}
\def\confYear{2024}

\title{3D Face Tracking from 2D Video through Iterative Dense UV to Image Flow}

\author{
{Felix Taubner \hspace{0.5cm}
Prashant Raina \hspace{0.5cm}
Mathieu Tuli \hspace{0.5cm}
Eu Wern Teh \hspace{0.5cm}
Chul Lee \hspace{0.5cm}
Jinmiao Huang}\\
LG Electronics \\ 
{\tt\small \{prashant.raina, mathieu.tuli, euwern.teh, clee.lee\}@lge.com}
}

\begin{document}
\maketitle

    

\begin{abstract}
When working with 3D facial data, improving fidelity and avoiding the uncanny valley effect is critically dependent on accurate 3D facial performance capture. Because such methods are expensive and due to the widespread availability of 2D videos, recent methods have focused on how to perform monocular 3D face tracking. However, these methods often fall short in capturing precise facial movements due to limitations in their network architecture, training, and evaluation processes. Addressing these challenges, we propose a novel face tracker, \textbf{\textit{FlowFace}}, that introduces an innovative 2D alignment network for dense per-vertex alignment. Unlike prior work, FlowFace is trained on high-quality 3D scan annotations rather than weak supervision or synthetic data. Our 3D model fitting module jointly fits a 3D face model from one or many observations, integrating existing neutral shape priors for enhanced identity and expression disentanglement and per-vertex deformations for detailed facial feature reconstruction. Additionally, we propose a novel metric and benchmark for assessing tracking accuracy. Our method exhibits superior performance on both custom and publicly available benchmarks.
We further validate the effectiveness of our tracker by generating high-quality 3D data from 2D videos, which leads to performance gains on downstream tasks.

\vspace{-0.5cm}
\end{abstract}    
\section{Introduction}
\label{sec:intro}

Access to 3D face tracking data lays the foundation for many computer graphics tasks such as 3D facial animation, 3D human avatar reconstruction, and expression transfer. Obtaining high visual fidelity, portraying subtle emotional cues, and preventing the uncanny valley effect in these downstream tasks is reliant on high motion capture accuracy. As a result, a common approach to generating 3D face tracking data is to use 3D scans and visual markers however, this process is cost-intensive. To alleviate this burden, building computational models to obtain 3D faces from monocular 2D videos and images has cemented its importance in recent years and seen great progress \cite{3ddfa_v2, deca, emoca, hrn, dense_landmarks_microsoft, mica, face2face}. 
Nevertheless, three issues persist:
First, current methods rely heavily on sparse landmarks and photometric similarity, which is computationally expensive and ineffective in ensuring accurate face motion. 
Second, the monocular face tracking problem is both ill-posed and contains a large solution space dependent on camera intrinsics, pose, head shape, and expression \cite{monocular_survey}. 
Third, current benchmarks for this task neglect the temporal aspect of face tracking and do not adequately evaluate facial motion capture accuracy.

To address the aforementioned issues, 
we introduce a novel 3D face tracking model called \textbf{FlowFace}, consisting of a versatile two-stage pipeline: 
A 2D alignment network that predicts the screen-space positions of each vertex of a 3D morphable model~\cite{3dmm} (3DMM) and an optimization module that jointly fits this model across multiple views by minimizing an alignment energy function.
Unlike traditional methods that rely on sparse landmarks and photometric consistency, FlowFace uses only 2D alignment as input signal, similar to recent work \cite{dense_landmarks_microsoft}. This alleviates the computational burden of inverse rendering and allows joint reconstruction using a very large number of observations.
We enhance previous work in four ways: (1) The 2D alignment network features a novel architecture with a vision-transformer backbone and an iterative, recurrent refinement block. 
(2) In contrast to previous methods that use weak supervision or synthetic data, the alignment network is trained using high-quality annotations from 3D scans. 
(3) The alignment network predicts dense, per-vertex alignment instead of key-points, which enables the reconstruction of finer details. 
(4) We integrate an off-the-shelf neutral shape prediction model to improve identity and expression disentanglement. 

In addition, we present the screen-space motion error (SSME) as a novel face tracking metric. Based on optical flow, SSME computes and contrasts screen-space motion, aiming to resolve the limitation observed in existing evaluation methods. These often rely on sparse key points, synthetic annotations, or RGB/3D reconstruction errors, and lack a thorough and comprehensive measurement of temporal consistency. Using the Multiface \cite{multiface} dataset, we develop a 3D face tracking benchmark
around this metric.

Finally, through extensive experiments on available benchmarks, we show that our method significantly outperforms the state-of-the-art on various tasks.
To round off our work, we demonstrate how our face tracker can positively affect the performance of downstream tasks, including speech-driven 3D facial animation and 3D head avatar synthesis. 
Specifically, we demonstrate how our method can be used to generate high-quality data --- comparable to studio-captured data --- for both these tasks by using it to augment existing models to achieve state-of-the-art results.

\section{Related Work}
\label{sec:related_work}

\paragraph{Uncalibrated 3D Face Reconstruction.}

Previous work reconstructing 3D face shapes from uncalibrated 2D images or video fall into two broad categories:

\textbf{Optimization-based methods} recover face shape and motion by jointly optimizing 3D model parameters to fit the 2D observations.
They traditionally treat this optimization as an inverse rendering problem \cite{old_photometric_0, old_photometric_2, old_photometric_1, face2face, mica, show_body_tracker, imavatar},
using sparse key-points as guidance. Typically, they employ geometric priors such as 3DMMs~\cite{3dmm, FLAME, baselface, facescape, realy_benchmark}, texture models, simplified illumination models, and temporal priors. Some methods use additional constraints such as depth \cite{face2face} or optical flow \cite{face_tracking_with_optical_flow}. \cite{monocular_survey} and \cite{reconstruction_survey} present detailed surveys of such methods.
Most methods use 3DMMs to disentangle shape and expression components. MPT \cite{mica} is the first method to integrate metrical head shape priors predicted by a deep neural network (DNN). 
However, photometric and sparse landmark supervision is not sufficient to obtain consistent and accurate face alignment, especially in areas not covered by landmarks and or of low visual saliency.
More recently, \cite{dense_landmarks_microsoft} proposes to use only 2D face alignment (dense landmarks) as supervision, avoiding the computationally expensive inverse rendering process. Our method extends this idea with an improved 2D alignment module, better shape priors, and per-vertex deformation. 

\textbf{Regression-based methods} train DNNs to directly predict face reconstructions from single images \cite{now_benchmark, emoca, deep3d, mgcnet, sadrnet, hrn, 3ddfa_v2, hiface_microsoft, albedogan}. This reconstruction includes information such as pose, 3DMM components, and sometimes texture. Typically, convolutional networks like image classification networks~\cite{resnet, mobilenetv2} or encoder-decoder networks \cite{prnet} are used. Due to the lack of large-scale 2D to 3D annotations, these methods typically rely on photometric supervision for their training. Some methods propose complex multi-step network architectures \cite{hrn, sadrnet} to improve reconstruction. \cite{hrn} use additional handcrafted losses to improve alignment, whereas \cite{hiface_microsoft} use synthetic data and numerous of landmarks. More recently, \cite{tokenface} proposes to use vision-transformers to improve face reconstruction.

\paragraph{2D Face Alignment.}

Traditional 2D face alignment methods predict a sparse set of manually defined landmarks. These methods typically involve convolutional DNNs to predict heat maps for each landmark \cite{star_keypoints, face_alignment_network, spiga}. Sparse key-points are not sufficient to describe full face motion, and heat maps make it computationally infeasible to predict a larger number of key-points. \cite{dense_landmarks_microsoft} and \cite{mediapipe} achieve pseudo-dense alignment by using classifier networks to directly predict a very large number of landmarks. \cite{densereg} predict the UV coordinates in image space and then map the vertices onto the image. Just like \cite{prnet} and \cite{sadrnet}, our method predicts a per-pixel dense mapping between the UV space of a face model and the image space. However, we set our method apart by using better network architectures with vision-transformers and real instead of synthetic data. 

\paragraph{Evaluation of Face Trackers.}

Prior work evaluates face tracking and reconstruction using key-point accuracy \cite{dense_landmarks_microsoft, sadrnet, prnet, 3ddfa_v2, aflw_dataset}, depth \cite{mica, face2face}, photometric \cite{mica, face2face} or 3D reconstruction \cite{facescape, realy_benchmark, face_tracking_with_optical_flow} errors. Sparse key-points are usually manually-annotated, difficult to define without ambiguities \cite{star_keypoints}, and insufficient to describe the full motion of the face. Dense key-points \cite{aflw_dataset} are difficult to compare between models using different mesh topologies.
Photometric errors \cite{mica, face2face, tokenface} are unsuitable since a perfect solution already exists within the input data, and areas with low visual saliency are neglected. A fair comparison of depth errors \cite{mica, face2face} is only possible for methods using a pre-calibrated, perspective camera model. Methods that evaluate 3D reconstruction errors have to rigidly align the target and predicted mesh to fairly evaluate results \cite{facescape, realy_benchmark, now_benchmark}, which causes valuable tracking information such as pose and intrinsics to be lost. Most importantly, depth and 3D reconstruction metrics neglect motion tangential to the surface normal. 
In contrast, our proposed metric measures the dense face motion in screen space, which is topology-independent and eliminates the need for rigid alignment.
\section{Method}
\label{sec:method}

\begin{figure*}[ht]
\centering
\includegraphics[width=0.95\textwidth]{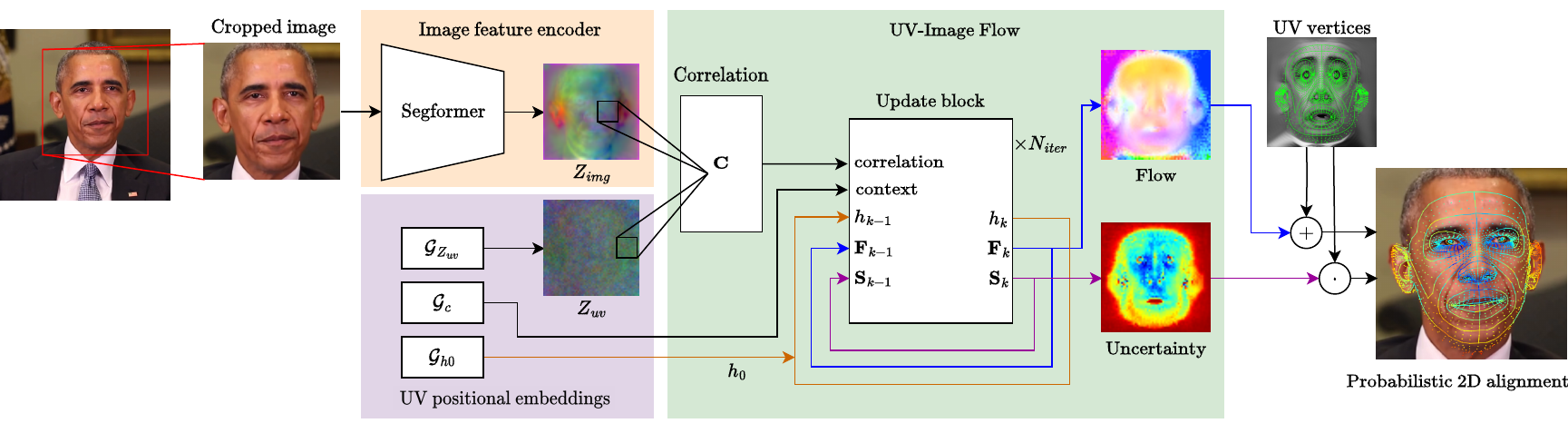}
\caption{An overview of the proposed 2D alignment network architecture. A feature encoder transforms the image into a latent feature map that is then iteratively aligned with a learned UV positional embedding map by the recurrent update block.}
\label{fig:2d_alignment}
\vspace{-0.4cm}
\end{figure*}

Our 3D face tracking pipeline consists of two stages: The first stage is predicting a dense 2D alignment of the face model, and the second stage is fitting a parametric 3D model to this alignment.

\subsection{Dense 2D Face Alignment Network}
\label{sec:2dalign}

\subsubsection{Network Architecture}

The 2D alignment module is responsible for predicting the probabilistic location --- in image space ---  of each vertex of our face model.
As in \cite{dense_landmarks_microsoft}, the 2D alignment of each vertex is represented as a random variable $\alignment_i= \{ \mu_i,\sigma_i \}$.
$\mu_i=[x_i,y_i] \in \imgspace$ is the expected vertex position in image space $\imgspace \in [0, D_\textit{img}]^2$, and $\sigma_i \in \mathbb{R}_{>0}$ is its uncertainty, modeled as the standard deviation of a circular 2D Gaussian density function. 
As an intermediate step, for each iteration $k$, the alignment network predicts a dense UV to image correspondence map $\flowmap_k:\uvspace \rightarrow \imgspace$ and uncertainty map $\mathbf{S}_k$.
$\flowmap_k$ maps any point in UV space $\uvspace \in [0, D_\textit{uv}]^2$ to a position in image space through a pixel-wise offset, which we call \textit{UV-image flow}. This network consists of three parts (\cref{fig:2d_alignment}):
\begin{enumerate}
    \item An image feature encoder producing a latent feature map of the target image.
    \item A positional encoding module that produces learned positional embeddings in UV space.
    \item An iterative, recurrent optical flow module that predicts the probabilistic UV-image flow.
\end{enumerate}
The image space position and uncertainty of each vertex is then bi-linearly sampled from the intermediate correspondence and uncertainty map for each iteration: 
\begin{equation}
    \mu_{i,k} = \uvcoord_i + \flowmap_k ( \uvcoord_i ) \quad\mathrm{and}\quad \sigma_{i,k} = \mathbf{S}_k ( \uvcoord_i )
\end{equation}
where $\uvcoord_i \in \uvspace$ denotes the pre-defined UV coordinate of each vertex. These are manually defined by a 3D artist.

\paragraph{Image feature encoder.}
To obtain the input to the image encoder $\mathcal{F}$, we use SFD \cite{sfd_face} to detect a square face bounding box from the target image and enlarge it by 20\%. We then crop the image to the bounding box and resize it to $D_\textit{img}$.
We use Segformer \cite{segformer} as the backbone, and replace the final classification layer with a linear layer to produce a 128-dimensional feature encoding. We further down-sample it to attain a final image feature map $Z_\textit{img} \in \mathbb{R}^{D_\textit{uv} \times D_\textit{uv} \times 128}$ through average pooling. With image $\mathbf{I}$ and network parameters $\theta_\mathcal{F}$, this is defined as:
\begin{equation}
    Z_\textit{img} = \mathcal{F} (\mathbf{I}, \theta_\mathcal{F})
\end{equation}

\paragraph{UV positional encoding module.}
We use a set of modules $\mathcal{G}$ with identical architecture to generate learned positional embeddings in UV-space. 
Each module is comprised of a multi-scale texture pyramid and a pixel-wise linear layer. This pyramid consists of four trainable textures with 32 channels and squared resolutions of $D_\textit{uv}$, $\frac{D_\textit{uv}}{2}$, $\frac{D_\textit{uv}}{4}$, and $\frac{D_\textit{uv}}{8}$ respectively. Each texture is upsampled to $D_\textit{uv}$ through bi-linear interpolation before concatenating them along the channel dimension. The concatenated textures are then passed through a pixel-wise linear layer to produce the UV positional embeddings. 
The multi-scale setup ensures structural consistency in UV space (closer pixels in UV should have similar features). 
We use 3 of these modules: $ \mathcal{G}_{Z_\textit{uv}} $ to generate a UV feature map $Z_\textit{uv}$, $ \mathcal{G}_{c} $ to generator a context map $c$, and $ \mathcal{G}_{h_0}$ to generate an initial hidden state $h_0$. With corresponding network parameters $\theta_{\mathcal{G}_{Z_\textit{uv}}}$, $\theta_{\mathcal{G}_{c}}$ and $\theta_{\mathcal{G}_{h_0}}$, this is described as:
\begin{equation}
    Z_\textit{uv} = \mathcal{G} (\theta_{\mathcal{G}_{Z_\textit{uv}}}); \hspace{0.5em} c = \mathcal{G} (\theta_{\mathcal{G}_{c}})  ; \hspace{0.5em} h_0 = \mathcal{G} (\theta_{\mathcal{G}_{h_0}})
\end{equation}

\paragraph{UV-image flow.}
The RAFT~\cite{raft_optical_flow} network is designed to predict the optical flow between two images. It consists of a correlation block that maps the latent features encoded from each image into a 4D correlation volume. A context encoder initializes the hidden state of a recurrent update block and provides it with additional context information. The update block then iteratively refines a flow estimate while sampling the correlation volume. 

We adapt this network to predict the UV-image flow $\flowmap \in \mathbb{R}^{D_\textit{uv} \times D_\textit{uv} \times 2}$. We directly pass $Z_\textit{uv}$ and $Z_\textit{img}$ to the correlation block $\mathbf{C}$. 
We use the context map $c$ and initial hidden state $h_0$ from the positional encoding modules for the update module $\mathbf{U}$. We modify the update module to also predict a per-iteration uncertainty in addition to the flow estimate,
by duplicating the flow prediction head to predict a 1-channel uncertainty map $\mathbf{S} \in \mathbb{R}_{>0}^{D_\textit{uv} \times D_\textit{uv}}$. An exponential operation is applied to ensure positive values. The motion encoder head is adjusted to accept the uncertainty as an input. 
The modified RAFT network then works as follows: For each iteration $k$, the recurrent update module performs a look-up in the correlation volume, context map $c$, previous hidden state $h_{k-1}$, previous flow $\flowmap_{k-1}$ and previous uncertainty $\mathbf{S}_{k-1}$. It outputs the refined flow estimate $\flowmap_{k}$ and uncertainty $\mathbf{S}_{k}$ and the subsequent hidden state $h_{k}$. Formally,
\begin{equation}
    \flowmap_k, \mathbf{S}_{k}, h_{k} = \mathbf{U} (\mathbf{C} (Z_\textit{uv}, Z_\textit{img}), c, \flowmap_{k-1}, \mathbf{S}_{k-1}, h_{k-1}, \theta_{\mathbf{U}})
\end{equation}%
with update module weights $\theta_{\mathbf{U}}$.
For a detailed explanation of our modified RAFT, we defer to \cite{raft_optical_flow} and \cref{supp:2d_alignment}.

\subsubsection{Loss Functions}

We supervise our network with Gaussian negative log-likelihood (GNLL) both on the probabilistic per-vertex positions and the dense UV-image flow. For each iteration $k$ of the update module, we apply the per-vertex loss function:
\vspace{-0.2cm}
\begin{equation}
    \textit{L}_{k}^\textit{vertex} =  \sum_{i=1}^{N_\textit{v}}{\lambda_i ( \log ( \sigma_{i,k}^2) + \frac{\parallel \mu_{i,k} - \mu_{i}' \parallel^2}{2 \sigma_{i,k}^2})}
\vspace{-0.1cm}
\end{equation}
where $\lambda_\text{i}$ is a pre-defined vertex weight and $\mu_{i}'$ is the ground truth vertex position.
We encourage our network to predict coherent flow and uncertainty maps in areas with no vertices by applying the GNLL loss for each pixel $p$ in UV space:
\vspace{-0.2cm}
\begin{equation}
    \textit{L}_{k}^\textit{dense} = \sum_{p \in |\uvspace|} \lambda_p ( \log ( \mathbf{S}_{k,p} ^2 ) + \frac{ \parallel \flowmap_{k,p} - \flowmap'_p \parallel^2}{ 2 \mathbf{S}_{k,p}^2}) 
\vspace{-0.2cm}
\end{equation}
where $\lambda_p$ is a pre-defined per-pixel weight and $\flowmap'$ is the ground truth UV-image flow. 
The final loss is a weighted sum of these losses, with a decay factor for each iteration of $\alpha=0.8$ and a dense weight of $\lambda_\textit{dense}=0.01$:
\vspace{-0.2cm}
\begin{equation}
    \text{Loss} = \sum_{k=1}^{N_\textit{iter}} \alpha ^ {N_\textit{iter}-k} (\textit{L}_{k}^\textit{vertex}+ \lambda_\textit{dense} \textit{L}_{k}^\textit{dense})
\end{equation}

\subsection{3D Model Fitting}
\label{sec:correspondence}

\begin{figure}[h]
\centering
\includegraphics[trim={0 0 1cm 0},clip,width=0.5\textwidth]{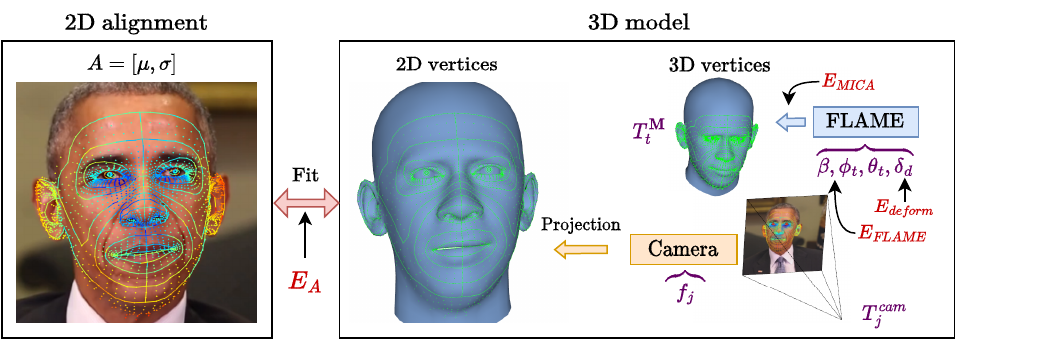}
\caption{An illustration of the 3D model fitting process.}
\label{fig:2d_4d_illustration}
\vspace{-0.5cm}
\end{figure}

As in \cite{dense_landmarks_microsoft}, the 3D reconstruction is obtained by jointly fitting a 3D head model and camera parameters to the predicted 2D alignment observations for the entire sequence. This is done by optimizing the energy function $\textit{E}(\Phi;\alignment)$ w.r.t to the model parameters $\Phi$ and alignment $\alignment$ (see \cref{fig:2d_4d_illustration}).
These parameters and the energy terms are defined below.

\subsubsection{Tracking Model and Parameters}
The tracking model consists of a 3D head model and a camera model.
A tracking sequence contains $\textit{C}$ cameras, $\textit{F}$ frames with a total of $\textit{C} \times \textit{F}$ images.

\paragraph{3D head model.}
We use FLAME \cite{FLAME} as our 3D head model $\mathbf{M}$. This model consists of $N_\textit{v}=5023$ vertices, which are controlled by identity shape parameters $\boldsymbol{\beta} \in \mathbb{R}^{300}$, expression shape parameters $\boldsymbol{\phi} \in \mathbb{R}^{100}$ and $K=5$ skeletal joint poses $\boldsymbol{\theta} \in \mathbb{R}^{3K + 3} $ (including the root translation) through linear blend skinning \cite{linearblendskinning}. We ignore root, neck and jaw pose and use the FLAME2023 model, which includes deformations due to jaw rotation within the expression blend-shapes. We also introduce additional static per-vertex deformations $\delta_d \in \mathbb{R}^{N_\textit{v} \times 3}$ to enhance identity shape detail. 
The local head model vertices can be expressed using its parameters as follows:
\begin{equation}
    \mathbf{M}(\boldsymbol{\beta}, \boldsymbol{\delta}_d, \boldsymbol{\phi}, \boldsymbol{\theta}) = \textit{FLAME}( \boldsymbol{\beta}, \boldsymbol{\phi}, \boldsymbol{\theta} ) + \boldsymbol{\delta}_\text{d}
\end{equation}
The rigid transform $\mathbf{T}^\mathbf{M} \in \mathbb{R}^{3 \times 4}$ represents the head pose, which transforms head model vertices $i$ into world space for each frame $t$:
\begin{equation}
    \mathbf{x}^\text{3D}_{i,t} = \mathbf{T}^\mathbf{M}_{t} \mathbf{M}_{i}
\end{equation}

\paragraph{Camera model.}
The cameras are described by the world-to-camera rigid transform $\mathbf{T}_\textit{cam} \in \mathbb{R}^{3 \times 4}$ and the pinhole camera projection matrix $\mathbf{K} \in \mathbb{R}^{3 \times 3}$ defined by a single focal length $\textit{\textbf{f}} \in \mathbb{R}$ parameter. The camera model defines the image-space projection of the 3D vertices in camera $j$: 
\begin{equation}
\label{eq:camera_projection}
    \mathbf{x}^\text{2D}_{i,j,t}= \mathbf{K}_j \mathbf{T}_j^\textit{cam} \mathbf{x}^\text{3D}_{i,t}
\end{equation}

\paragraph{Parameters.}
The parameters $\Psi$ consist of the head model and camera parameters, which are optimized to minimize $\textit{E}(\Phi;\alignment)$. The camera parameters can be fixed to known values, if the calibration is available. Expression and poses vary for each frame $t$, whereas camera, identity shape, and deformation parameters are shared over the sequence.
\begin{equation}
    \boldsymbol{\Psi} = \{ \boldsymbol{\beta}, \Phi_{\textit{F} \times | \boldsymbol{\phi} |}, \boldsymbol{\Theta}_{\textit{F} \times | \boldsymbol{\theta} |}, \boldsymbol{\delta}_{\text{d}}; \mathbf{T}_{\textit{F} \times 3 \times 4}^\mathbf{M}; \mathbf{T}_{\textit{C} \times 3 \times 4}^\textit{cam}, \textit{\textbf{f}}_\textit{C} \}
\end{equation}

\subsubsection{Energy Terms}

The energy function is defined as:
\begin{equation}
\textit{E}(\Phi;\alignment) = \textit{E}_\alignment + \textit{E}_\textit{FLAME} + \textit{E}_\textit{temp} + \textit{E}_\textit{MICA} + \textit{E}_\text{deform}
\end{equation}

\textbf{$\textit{E}_{\alignment}$} encourages 2D alignment: 
\vspace{-0.2cm}
\begin{equation}
    \textit{E}_\alignment = \sum_{i, j, t}^{N_\textit{v}, \textit{C}, \textit{F}} \lambda_i {\frac{\parallel\mathbf{x}^\text{2D}_{i,j,t} - \mathbf{\mu}_{i,j,t}\parallel^2}{2\sigma_{i,j,t}^2}}
\vspace{-0.1cm}
\end{equation}
where for vertex $i$ seen by camera $j$ in frame $t$. $\mu_{i,j,t}$ and $\sigma_{i,j,t}$ is the 2D location and uncertainty predicted by the final iteration of our 2D alignment network, and $\mathbf{x}^\text{2D}_{i,j,t}$ (\cref{eq:camera_projection}) is the 2D camera projection of that vertex. 

\textbf{$\textit{E}_\textit{FLAME}$} $ = \lambda_\textit{FLAME} (\parallel \beta \parallel^2 + \parallel \Phi \parallel^2) $ encourages the optimizer to explain the data with smaller identity and expression parameters. This leads to face shapes that are statistically more likely \cite{FLAME, deca, emoca, mica} and a more accurate 3D reconstruction. We do not penalize joint rotation, face translation or rotation. 

\textbf{$\textit{E}_\textit{temp}$} applies a loss on the acceleration of the 3D position $\mathbf{x}^\text{3D}_{i,t}$ of every vertex of the 3D model to prevent jitter and encourage a smoother, more natural face motion:
\begin{equation}
    \textit{E}_\textit{temp} = \lambda_\textit{temp} \sum_{i, j, t=2}^{N_\textit{v}, \textit{C}, \textit{F}-1} {\parallel \mathbf{x}^\text{3D}_{j,t-1} - 2 \mathbf{x}^\text{3D}_{j,t} + \mathbf{x}^\text{3D}_{j,t+1} \parallel^2}
\end{equation}

\textbf{$\textit{E}_\textit{MICA}$} $=\lambda_\textit{MICA} \parallel \mathbf{M}_{\boldsymbol{\Phi}=0,\boldsymbol{\theta}=0} - \mathbf{M}_\textit{MICA} \parallel^2$ provides a 3D neutral geometry prior for the optimizer to enable a better disentanglement between identity and expression components. It consists of the L2 distance of the neutral head model vertices to the MICA \cite{mica} template $\mathbf{M}_\textit{MICA}$.
This template is computed by predicting the average neutral head vertices using the MICA model \cite{mica} for all frames of the sequence. The term also enables a more accurate 3D reconstruction since the model can rely on MICA predictions where the alignment is uncertain, such as in the depth direction or for occluded vertices. In areas of confident alignment, the MICA prediction can be refined.

\textbf{$\textit{E}_\textit{deform}$} $=\lambda_\textit{deform} \parallel \boldsymbol{\delta}_{\text{d}} \parallel^2$ encourages per-vertex deformations to be small w.r.t. the FLAME model.

\subsection{Multiface Face Tracking Benchmark}

Our monocular 3D face tracking benchmark focuses on 3D reconstruction and motion capture accuracy.
To evaluate these, we use our proposed screen space motion error (SSME) and the scan-to-mesh chamfer distance (CD). 

\paragraph{Screen Space Motion Error.}

\newcommand{\ssvertex}{\textbf{v}_\text{img}}

\begin{figure}[h]
\centering
\vspace{-0.4cm}
\includegraphics[width=0.5\textwidth]{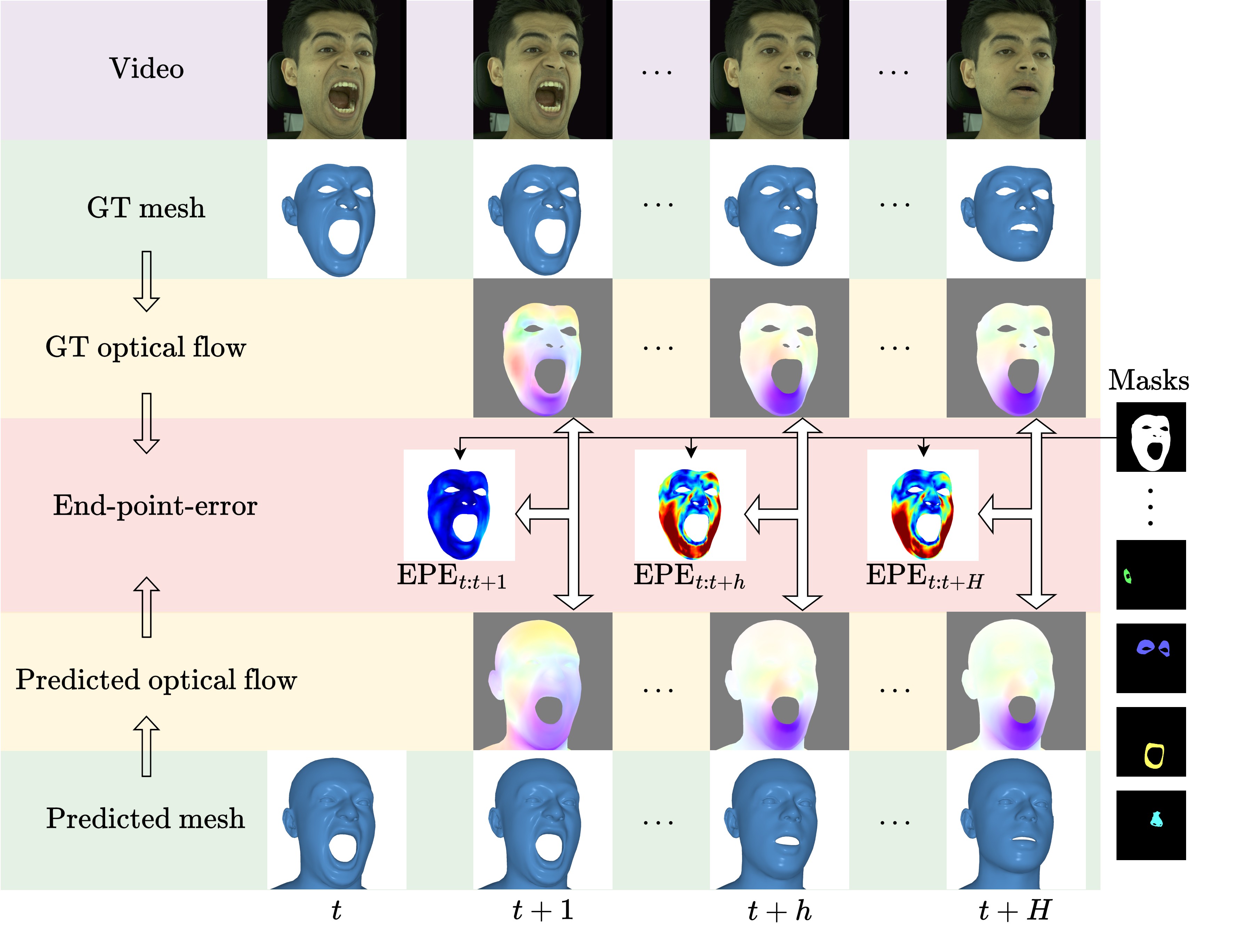}
\caption{An illustration of the EPE computation for each frame. }
\label{fig:ssme_illustration}
\end{figure}

To define the \textbf{S}creen \textbf{S}pace \textbf{M}otion \textbf{E}rror (SSME),
we reformulate face tracking as an optical flow prediction problem over a set of time windows. First, we project the ground truth mesh and predicted mesh into screen space using the respective camera model. 
Then, we use the screen space coordinates to compute the ground truth optical flow $\textbf{f}'_{t :  t+h}$ and predicted optical flow $\textbf{f}_{t : t+h}$ from frame $t$ to frame $t+h$ for each frame $t \in [1, \hdots, F]$ and a sequence of frame windows $h=[1,...,\textit{N}_\textit{H}]$. For each frame and frame window, the average end-point-error $\textit{EPE}_{t : t+h}$ is computed by averaging the L2-distance between ground truth and predicted optical flow for each pixel (see \cref{fig:ssme_illustration}). 
\begin{equation}
    \textit{EPE}_{t: t + h} = \parallel V \odot (\textbf{f}_{t : t + h} - \textbf{f}'_{t : t + h}) \parallel ^2
\end{equation}
where $V$ is a mask to separate different face regions and $\odot$ is the Hadamard product.
See \cref{fig:ssme_illustration} for a visual reference.

The screen space motion error $\textit{SSME}_h$ for frame window $h$ is then defined as the mean of all EPEs over all frames $t$ where frame $t+h$ exists: 
\vspace{-0.2cm}
\begin{equation}
    \textit{SSME}_h = \frac{1}{\textit{F}-h} \sum_{t=1}^{t + h\leq \textit{F}}  \textit{EPE}_{t : t+h}
\vspace{-0.1cm}
\end{equation}
Finally, to summarize tracking performance in one value, we compute the average screen space motion error $\overline{\textit{SSME}}$ over all frame windows as
\vspace{-0.2cm}
\begin{equation}
    \overline{\textit{SSME}} = \sum_{h=1}^{\textit{N}_\textit{H}} \textit{SSME}_h
\vspace{-0.1cm}
\end{equation} 

In other words, $\overline{\textit{SSME}}$ measures the average trajectory accuracy of each pixel over a time horizon of $N_H$ frames. We choose a maximum frame window of $N_\textit{H}=30$ (1 second) since most human expressions are performed within this time frame. 
Because the screen space motion is directly affected by most face-tracking parameters such as intrinsics, pose, and face shape, it also measures their precision in a holistic manner. 
In contrast to prior works and benchmarks that use sparse key-points, SSME covers the motion of all visible face regions and is invariant to mesh topology. As it operates in screen space, it does not require additional alignment and works with all camera models, unlike 3D reconstruction or depth errors.
In our benchmark, we evaluate SSME over a set of masks for semantically meaningful face regions (face, eyes, nose, mouth, and ears) (\cref{fig:ssme_illustration}), permitting a more nuanced analysis of the tracking performance. 

\paragraph{3D Reconstruction.}
To complete our benchmark, we additionally measure the chamfer distance (CD) to account for the depth dimension. Similar to \cite{now_benchmark}, the tracked mesh is rigidly aligned to the ground truth mesh using 7 key-points and ICP. Then, the distance of each ground truth vertex with respect to the predicted mesh is computed and averaged. For a detailed explanation, we defer to the NoW benchmark \cite{now_benchmark}. Just like the SSME, we evaluate the CD for the same set of face regions to provide a more detailed analysis of reconstruction accuracy, similar to \cite{realy_benchmark}. 

\paragraph{Multiface Dataset.}
We build our benchmark around the Multiface dataset \cite{multiface}. Multiface consists of multi-view videos with high quality topologically consistent 3D registrations. High-resolution videos are captured at 30 FPS from a large variety of calibrated views. We limit the evaluation data to a manageable size by carefully selecting a subset of 86 sequences with a diverse set of view directions and facial performances (see \cref{supp:multiface}). 

\section{Experiments}

\paragraph{Training data.}
To train the 2D alignment network, we use a combined dataset made up of FaceScape \cite{facescape}, Stirling \cite{stirling}, and FaMoS \cite{tempeh}. 
Where a FLAME~\cite{FLAME} registration is not available, we fit the FLAME template mesh to the 3D scan through semi-automatic key-point annotation and commercial topology fitting software. For an accurate capture of face motion, we auto-annotate expression scans with additional key-points propagated with optical flow (more information in \cref{supp:training}). The ground truth image space vertex positions $\mu'$ are obtained by projecting the vertices of the fitted FLAME mesh into screen space using the available camera calibrations.

\paragraph{Training strategy for 2D alignment network.}
We use Segformer-b5 (pre-trained on ImageNet~\cite{imagenet}) as our backbone, with $D_\textit{img}=512$, $D_\textit{uv}=64$ and $N_\textit{iter}=3$. We use the RAFT-L configuration for the update module and keep its hyperparameters when possible \cite{raft_optical_flow}. 
We optimize the model for 6 epochs using the AdamW optimizer \cite{adamw}, an initial learning rate of $\num{1e-4}$ and a decay of $0.1$ every 2 epochs. We use image augmentation such as random scaling, rotation, and color corruption~\cite{dense_landmarks_microsoft}, synthetic occlusions \cite{occlusions} and synthetic backgrounds (see \cref{supp:training}).

\paragraph{3D model fitting.}
To minimize the energy function and obtain tracking parameters, we use
the AdamW optimizer with an initial learning rate of $\num{1e-2}$ and a automatic learning rate scheduler with a decay factor of $0.5$ and patience of 30 steps, until convergence. We enable $\boldsymbol{\delta}_d$ only for multi-view reconstruction, and only for the nose region. 

\paragraph{Baselines.} We implement and test against the most recent publicly available methods for single image regression-based approaches 3DDFAv2 \cite{3ddfa_v2}, SADRNet \cite{sadrnet}, PRNet \cite{prnet}, DECA (coarse) \cite{deca}, EMOCA (coarse) \cite{emoca}, and HRN \cite{hrn}.
We extend the ability of these methods to use temporal priors by applying a simple temporal Gaussian filter to the screen-space vertices. We also include the popular photometric optimization-based approach MPT \cite{mica}. Lastly, we compare against the key-point-only optimization-based method \textit{Dense} proposed by \cite{dense_landmarks_microsoft} on public benchmarks.

\subsection{Multiface Benchmark}

\begin{figure} 
\centering
\begin{subfigure}{.25\textwidth}
  \centering
  \includegraphics[width=1\linewidth]{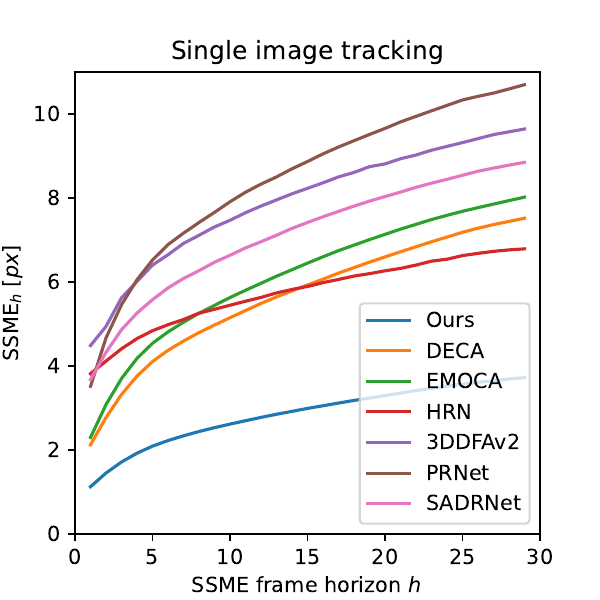}
  \label{fig:multiface_plot_no_temp}
\end{subfigure}%
\begin{subfigure}{.25\textwidth}
  \centering
  \includegraphics[width=1\linewidth]{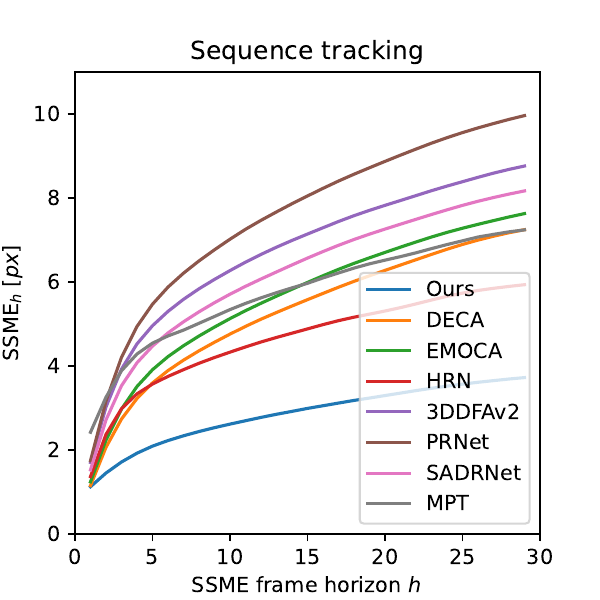}
  \label{fig:multiface_plot_with_temp}
\end{subfigure}
\caption{$\textit{SSME}_h$ plotted over all frame horizons for each evaluated tracker for single-image and full sequence tracking (right). Lower $\textit{SSME}_h$ in smaller frame horizons $h$ (left in the graph) means short-term temporal stability while lower $\textit{SSME}_h$ in larger frame horizons (right in the graph) means better long-term tracking consistency. Our tracker performs significantly better over every time horizon. }
\label{fig:multiface_plot}
\vspace{-0.3cm}
\end{figure}

We divide our Multiface benchmark into two categories: Without temporal information sharing, where each method is restricted to operate on single images, and with (both forward and backward) temporal information sharing, where each method is allowed to use the entire sequence as observations.
Our method significantly outperforms the best publicly available method by 54\% w.r.t. face-region $\overline{\textit{SSME}}$ on both on single-image and by 46\% on sequence prediction. This confirms the superior 2D alignment accuracy of our method. Despite using only 2D alignment as supervision, our method performs 8\% better in terms of 3D reconstruction (CD) than the photometric optimization approach MPT \cite{mica} (see \cref{tab:multiface_quantitative}. To our surprise, MPT performs inferior w.r.t. motion error than some regression-based models --- this is likely due to uniform lighting and texture in the Multiface dataset. Qualitative results \cref{tab:multiface_qualitative} confirm that methods using photometric errors (DECA, HRN, MPT) perform inferior w.r.t. screen space motion in areas without key-point supervision such as cheeks and forehead.  
Plotting the $\textit{SSME}_h$ over different time windows $h$ (see \cref{fig:multiface_plot}) gives a previously unseen overview of temporal stability. Regression-based methods suffer from high short-term error ($\textit{SSME}_1$) which is due to temporal instability and jitter. As expected, introducing temporal smoothing improves this issue and the overall $\overline{\textit{SSME}}$ for these methods. Our method achieves very low short-term SSME even with single image prediction, which indicates the high robustness and accuracy of the alignment network. As expected, introducing temporal priors reduces $\overline{\textit{SSME}}$. 

\subsection{FaceScape Benchmark}

\begin{table}[h]
\vspace{-0.2cm}
\begin{center}
\tablefont
\setlength{\tabcolsep}{2pt}
\begin{tabular}{l|c|c}
\toprule
Method  & CD $\downarrow$  (mm) & NME $\downarrow$ (rad)      \\ 
\midrule
MGCNet~\cite{mgcnet}   & 4.00                        & 0.093                         \\ 
PRNet~\cite{prnet}    & 3.56                         & 0.126                         \\ 
SADRNet~\cite{sadrnet}  & 6.75                         & 0.133                         \\ 
DECA~\cite{deca}     & 4.69                         & 0.108                         \\ 
3DDFAv2~\cite{3ddfa_v2} & 3.60                         & 0.096                         \\ 
HRN~\cite{hrn}      & 3.67                         & 0.087                         \\ 
\midrule
Ours     & \best{2.21}                & \best{0.083}                \\ 
\bottomrule
\end{tabular}
\end{center}
\caption{Results on the FaceScape benchmark \cite{facescape}. }
\label{tab:fscape_benchmark}
\vspace{-0.2cm}
\end{table}

\newcommand{\tabimg}[1]{\includegraphics[width=1cm]{figures/mq/#1}}

\begin{figure*}[th]
\tablefont
\setlength{\tabcolsep}{1pt}
\renewcommand{\arraystretch}{0}
\centering
\begin{tabular}{c c | c c c | c c c |c c c | c c c}
Image & 3D & 3D & CD & SSME & 3D & CD & SSME & 3D & CD & SSME & 3D & CD & SSME \\[0.1cm]

\tabimg{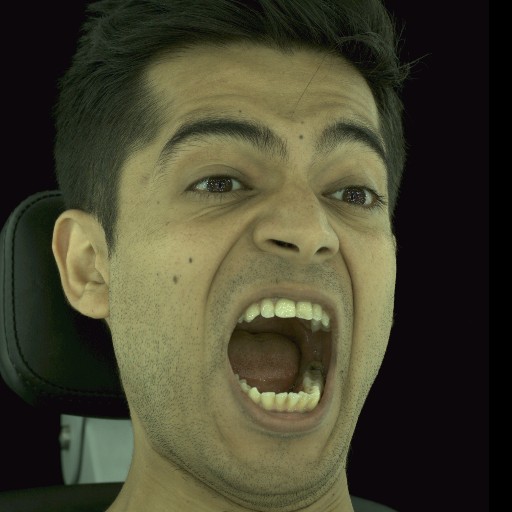} & \tabimg{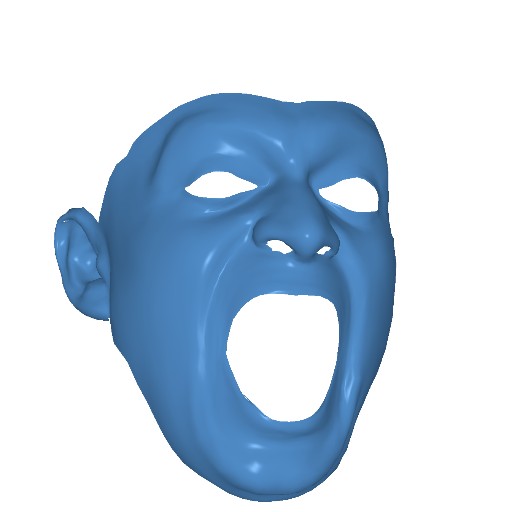} & \tabimg{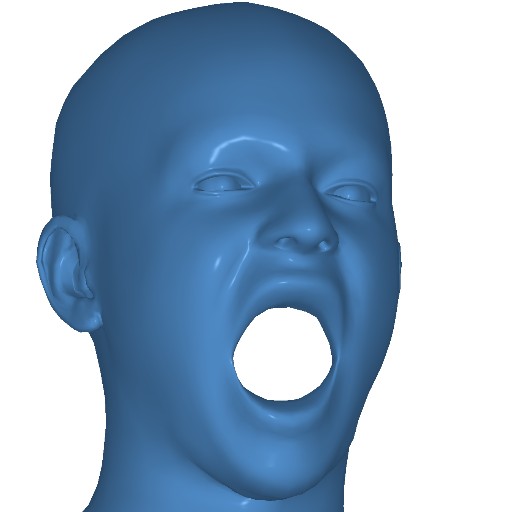} & \tabimg{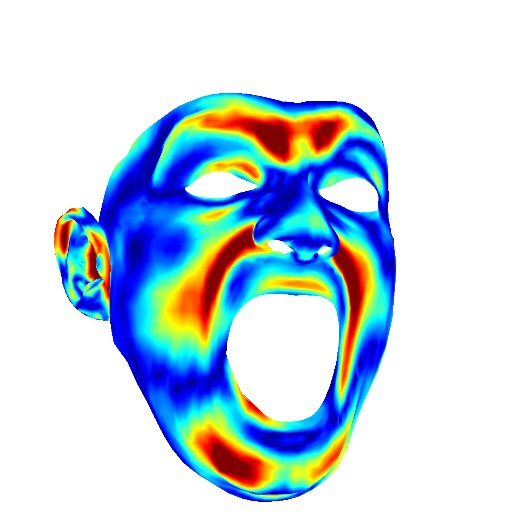} & \tabimg{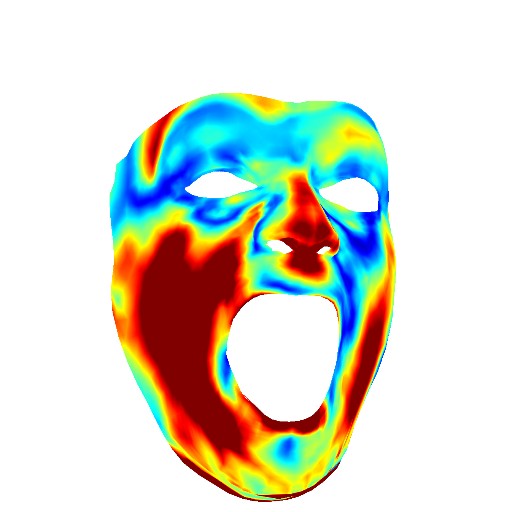} & \tabimg{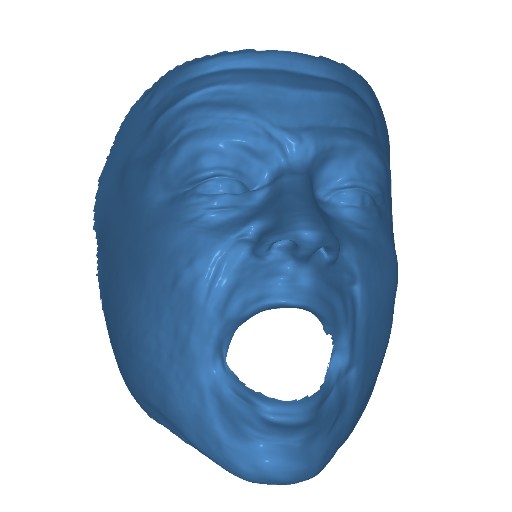} & \tabimg{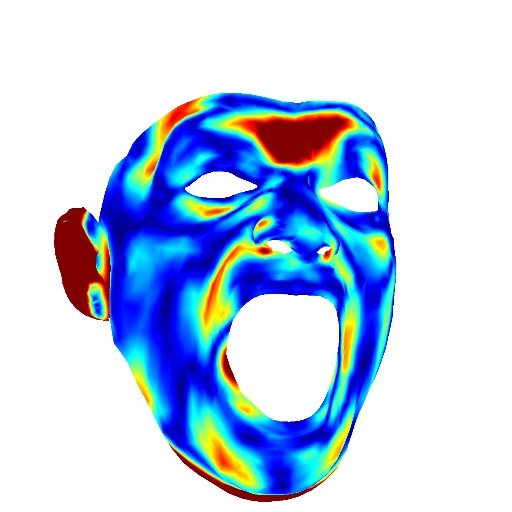} & \tabimg{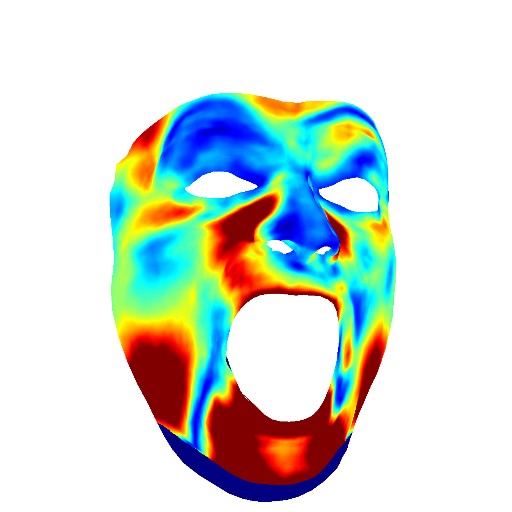} & \tabimg{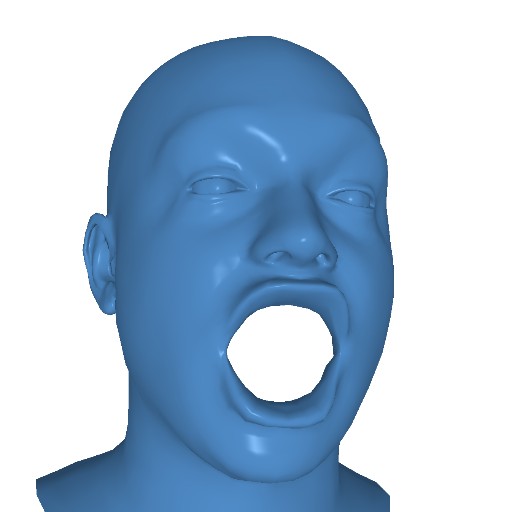} & \tabimg{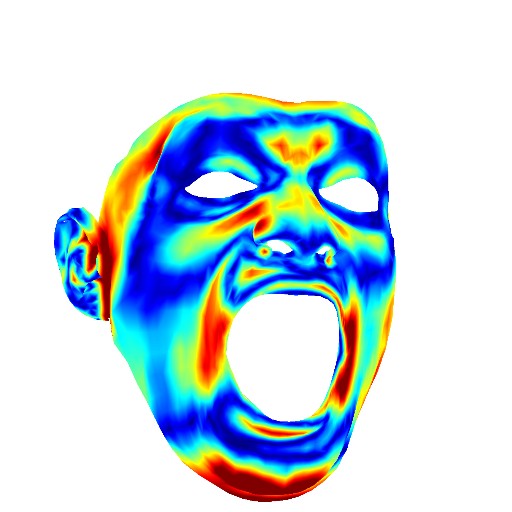} & \tabimg{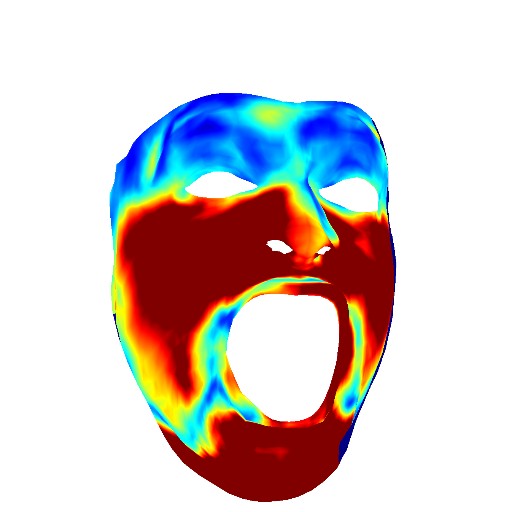}  & \tabimg{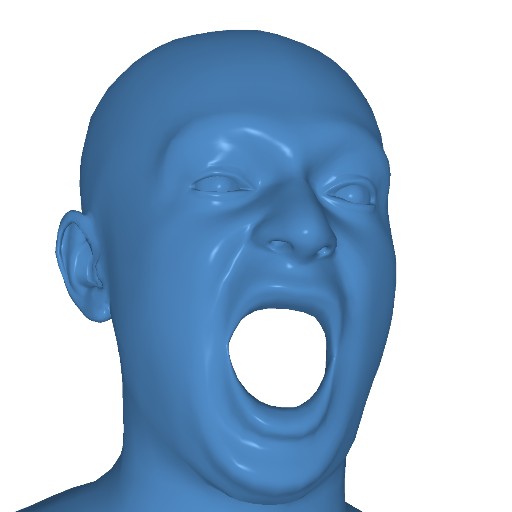} & \tabimg{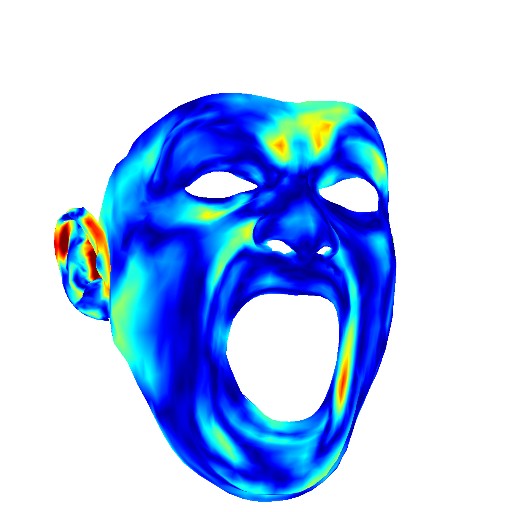} & \tabimg{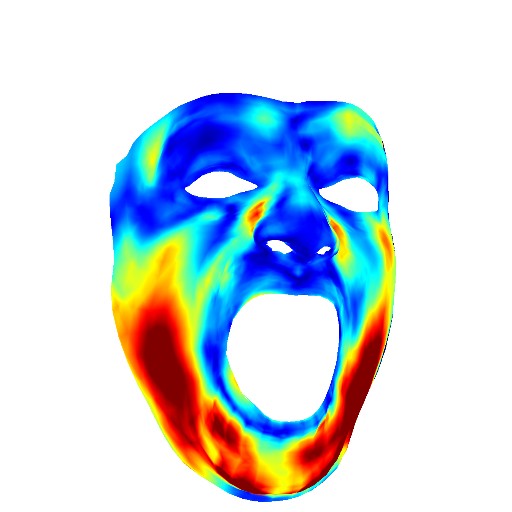} \\[0cm]
\tabimg{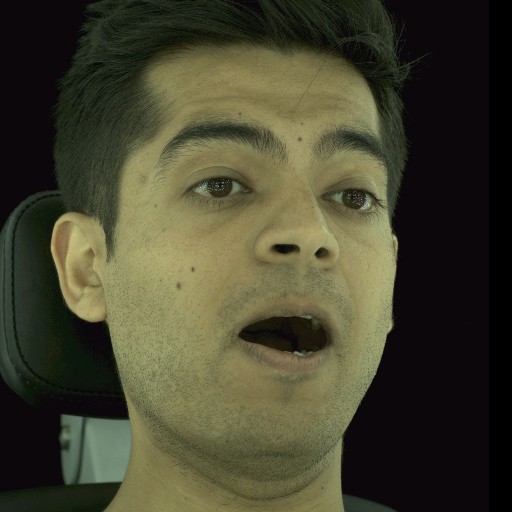} & \tabimg{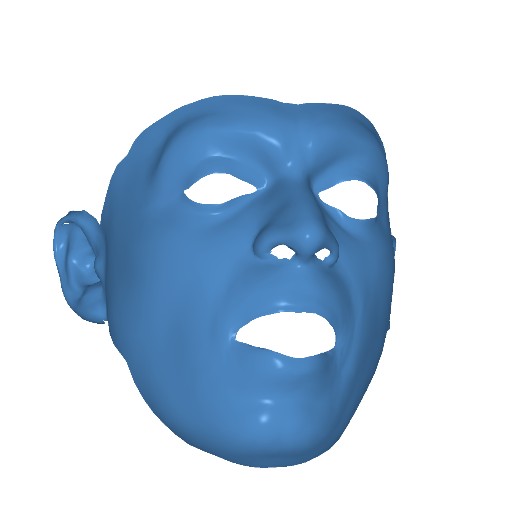} & \tabimg{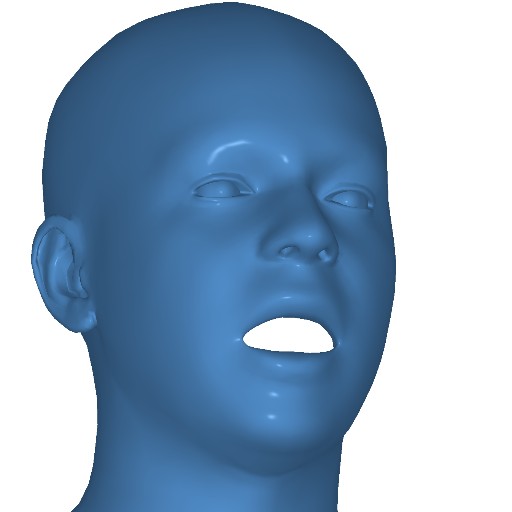} & \tabimg{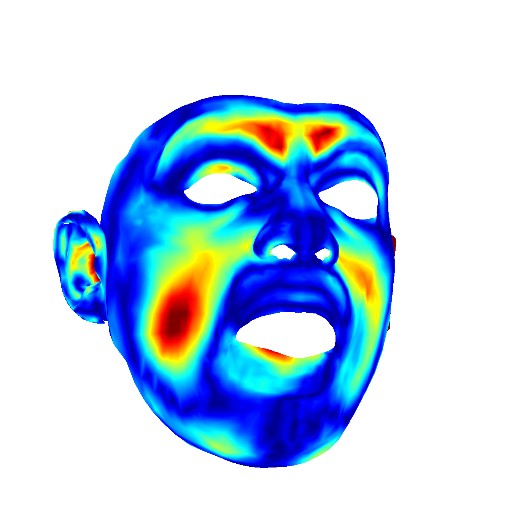} & \tabimg{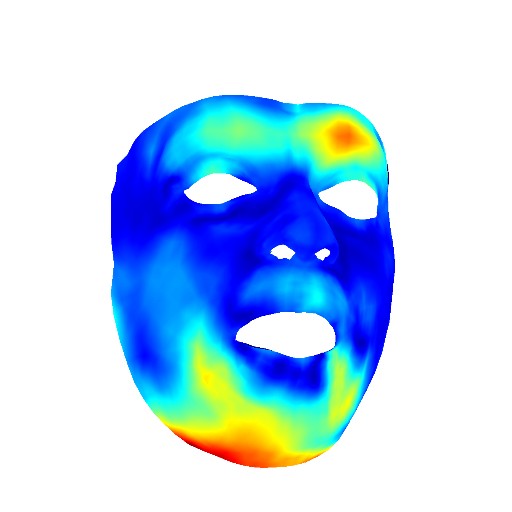} & \tabimg{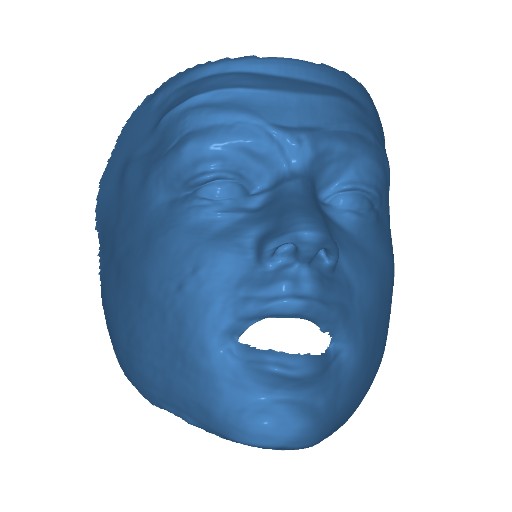} & \tabimg{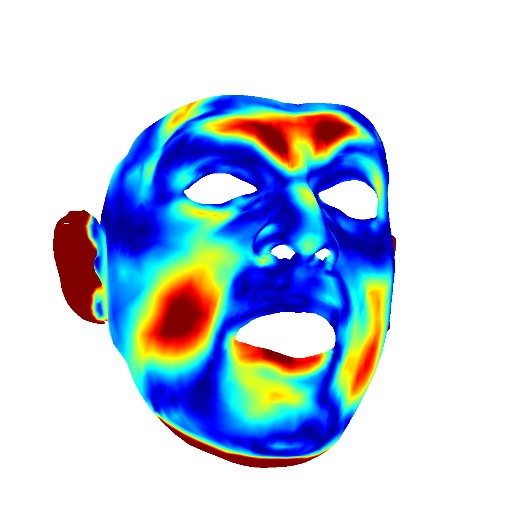} & \tabimg{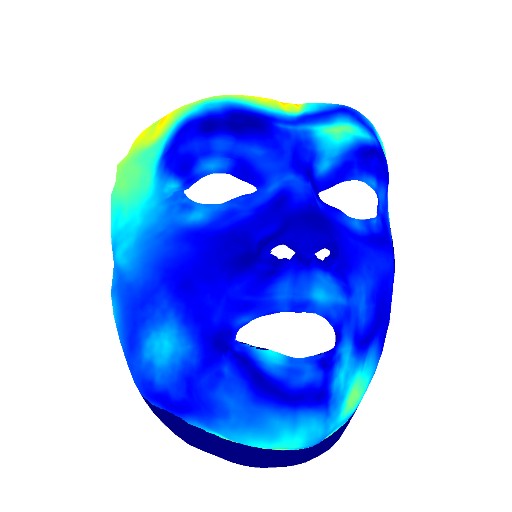} & \tabimg{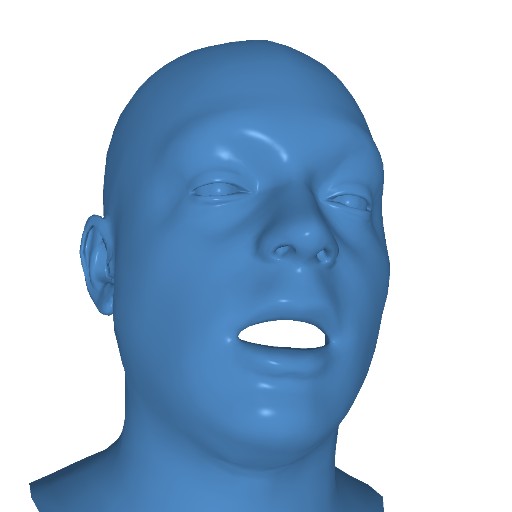} & \tabimg{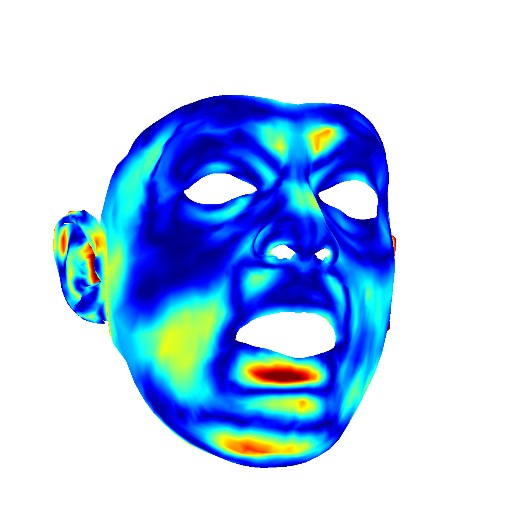} & \tabimg{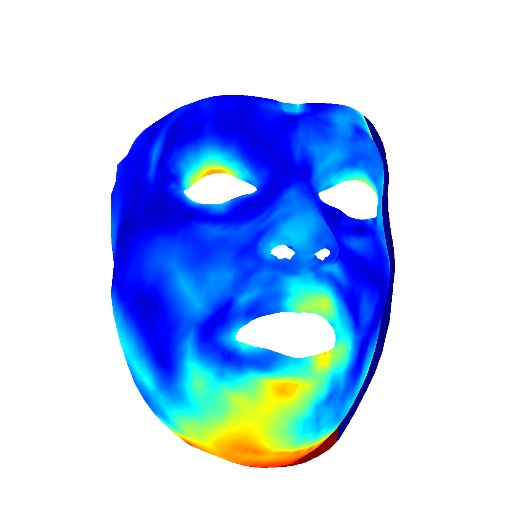}  & \tabimg{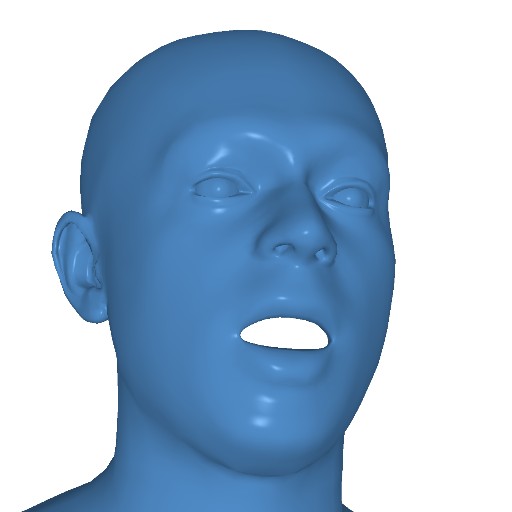} & \tabimg{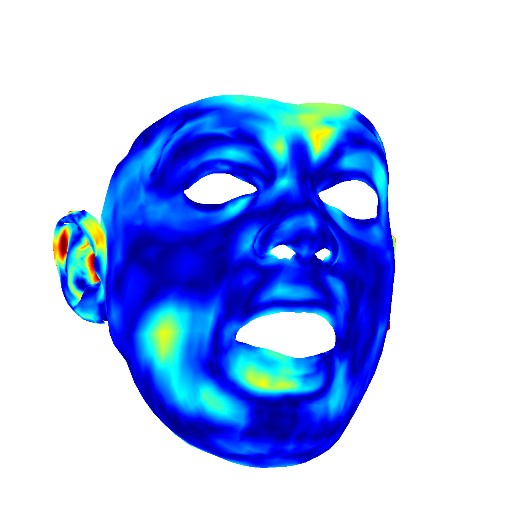} & \tabimg{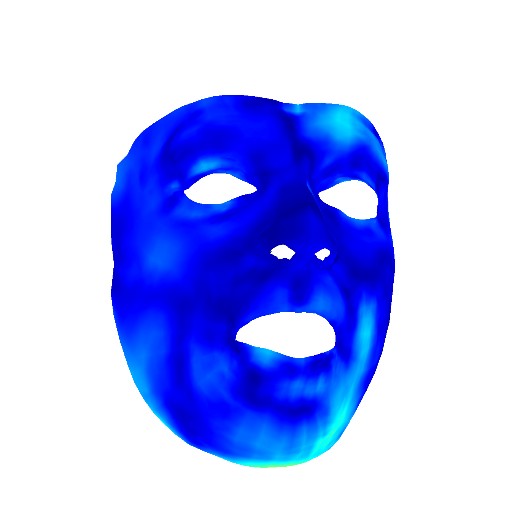} \\[0cm]
\tabimg{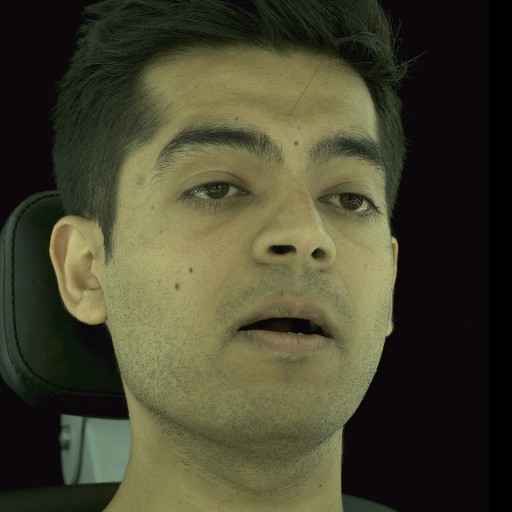} & \tabimg{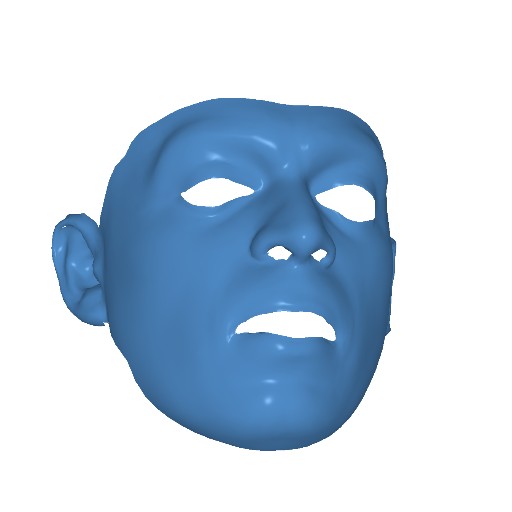} & \tabimg{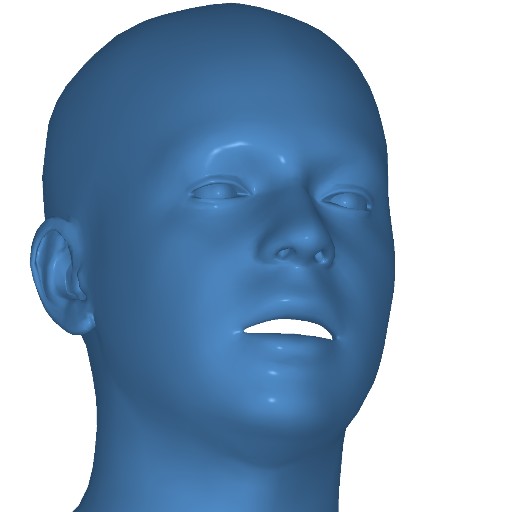} & \tabimg{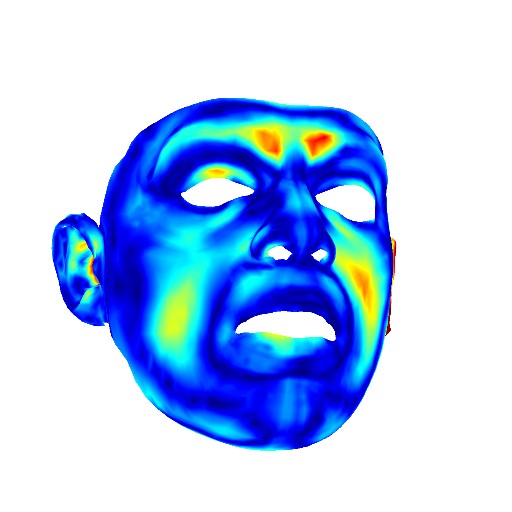} & \tabimg{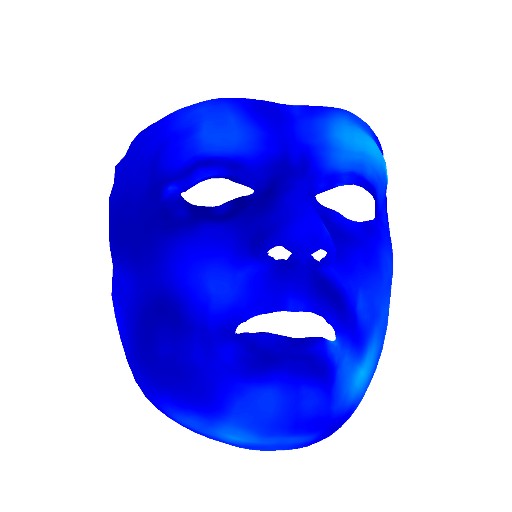} & \tabimg{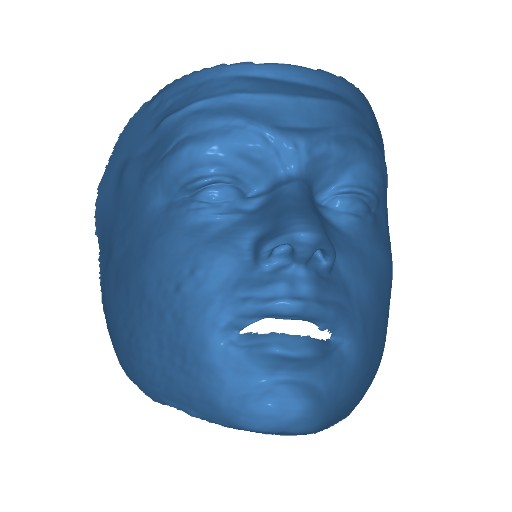} & \tabimg{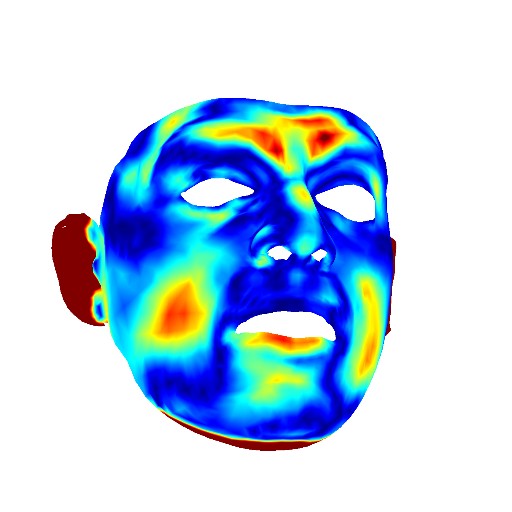} & \tabimg{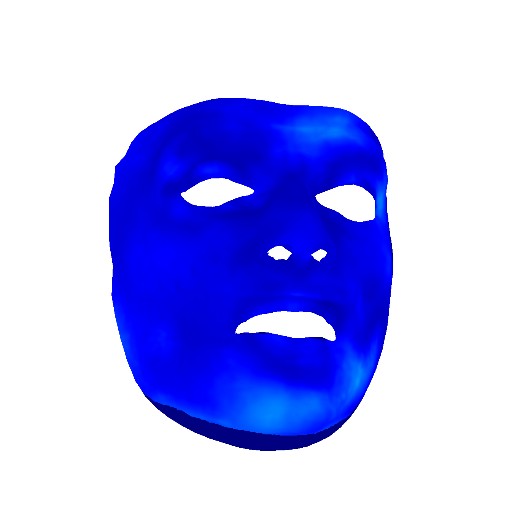} & \tabimg{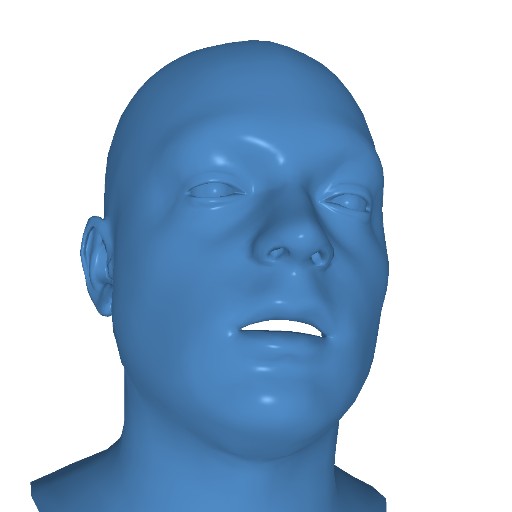} & \tabimg{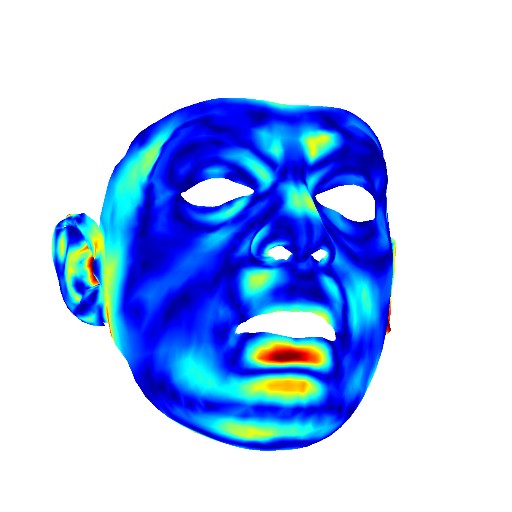} & \tabimg{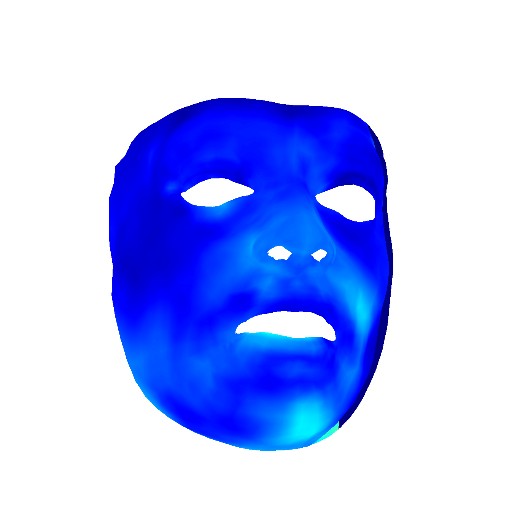}  & \tabimg{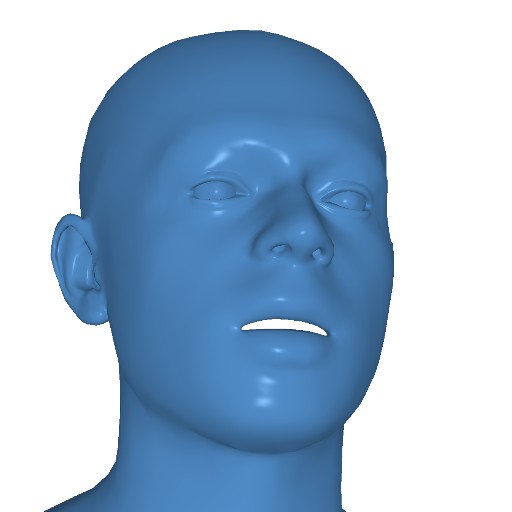} & \tabimg{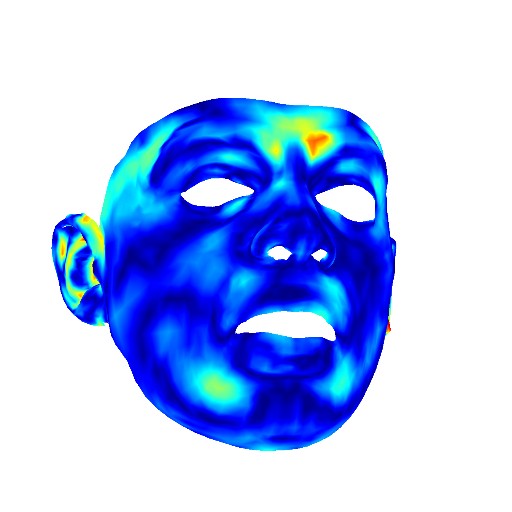} & \tabimg{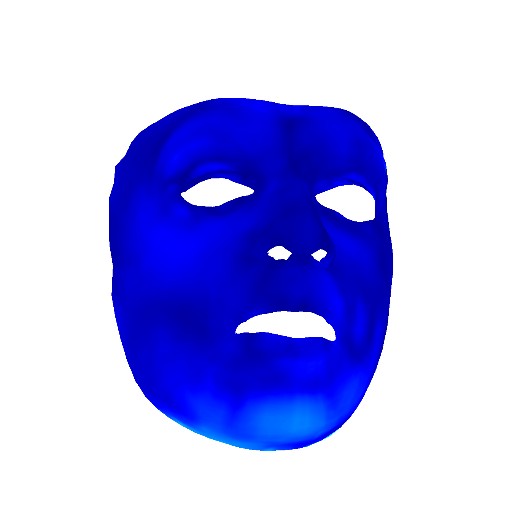} \\[0.1cm]

\tabimg{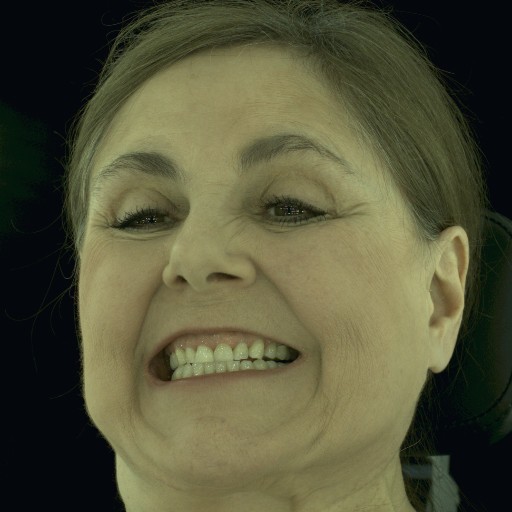} & \tabimg{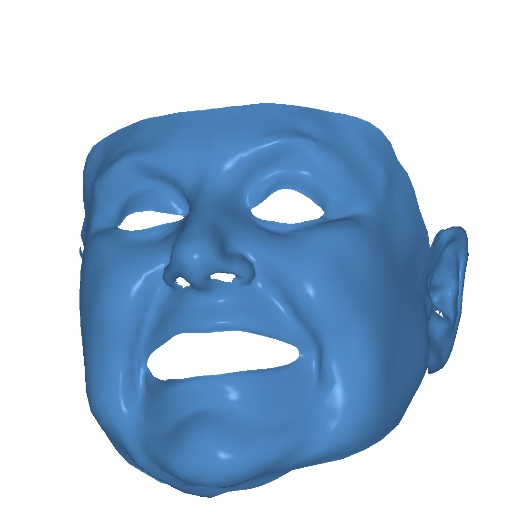} & \tabimg{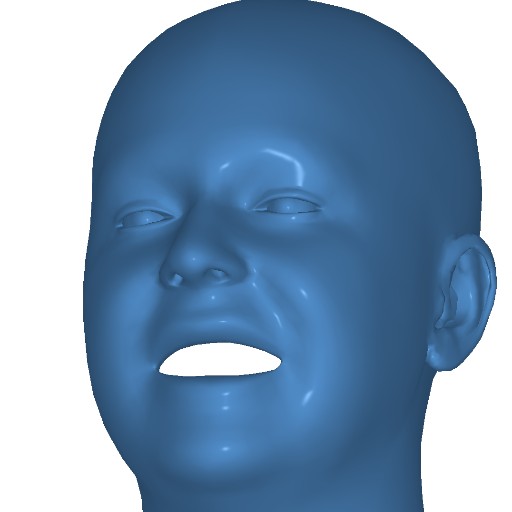} & \tabimg{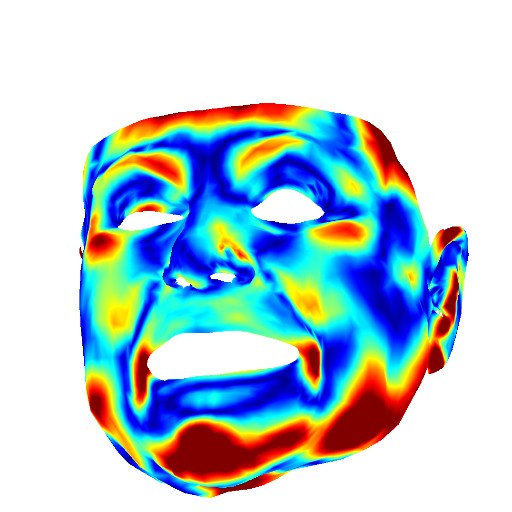} & \tabimg{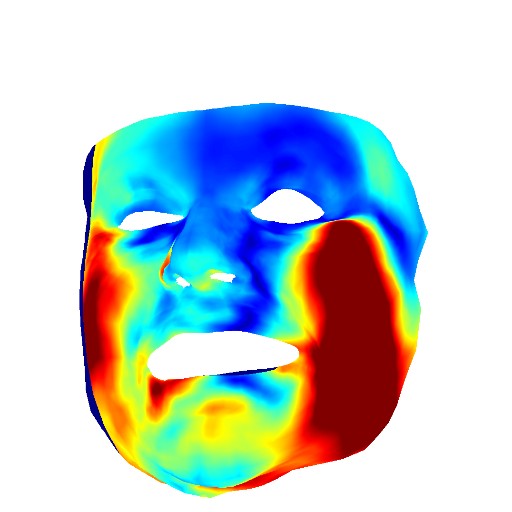} & \tabimg{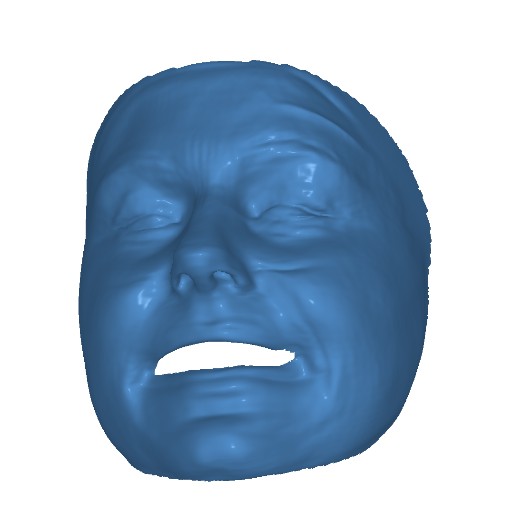} & \tabimg{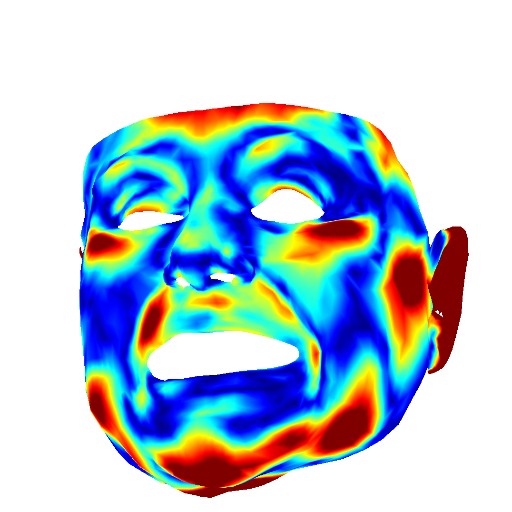} & \tabimg{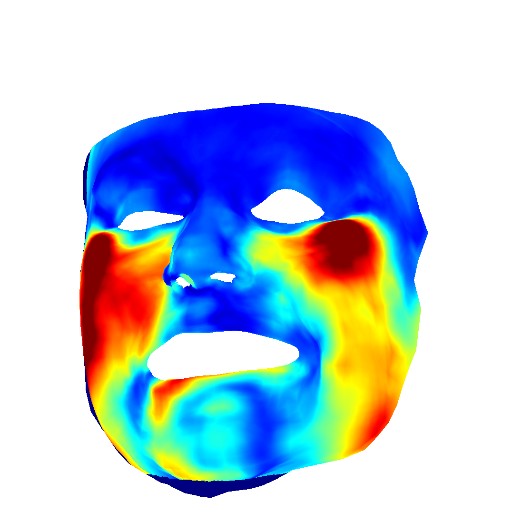} & \tabimg{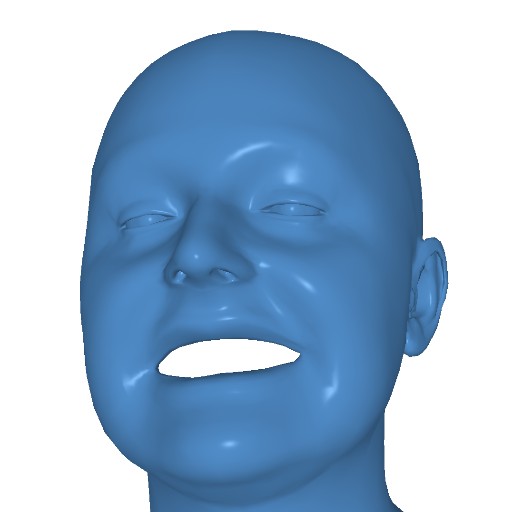} & \tabimg{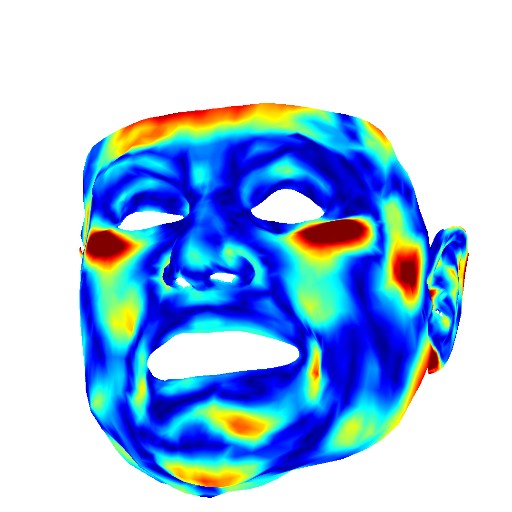} & \tabimg{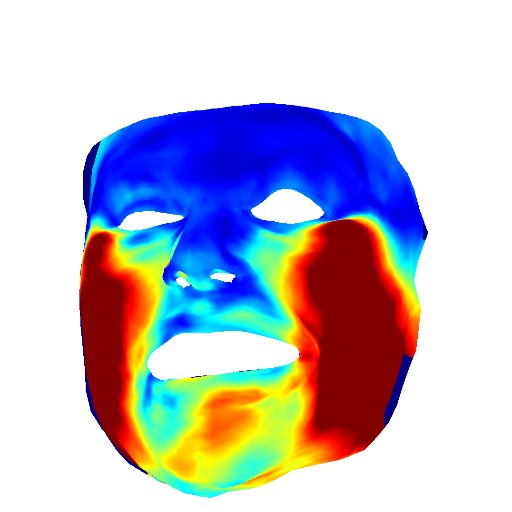}  & \tabimg{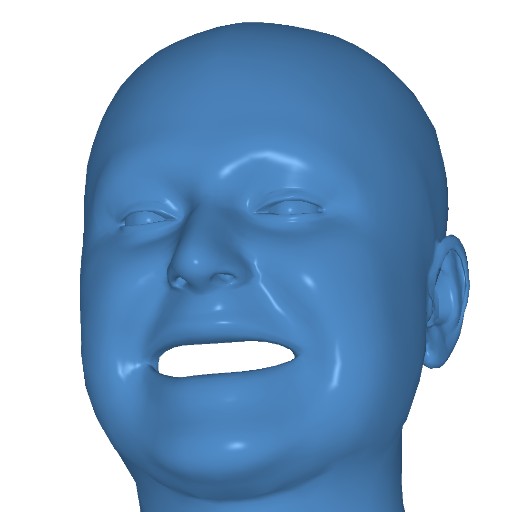} & \tabimg{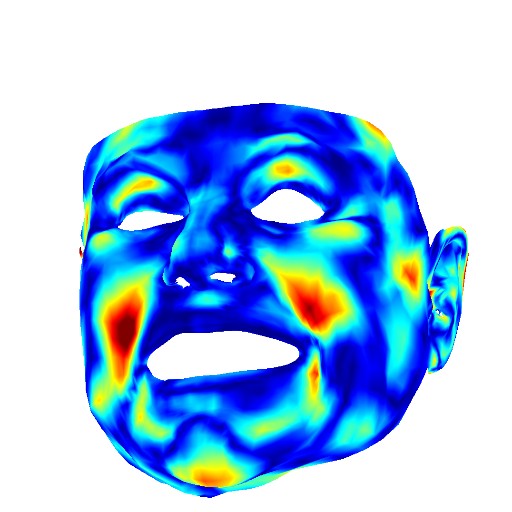} & \tabimg{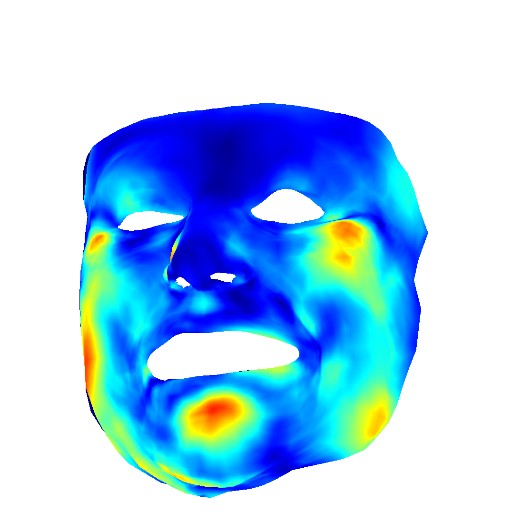} \\
\tabimg{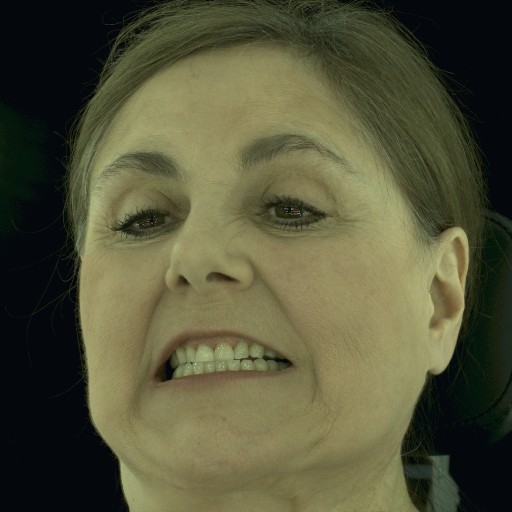} & \tabimg{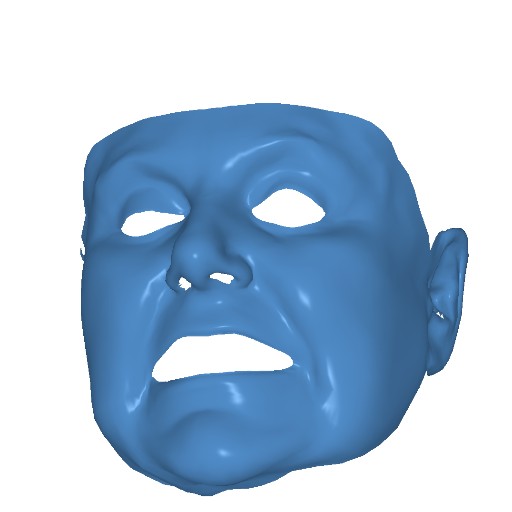} & \tabimg{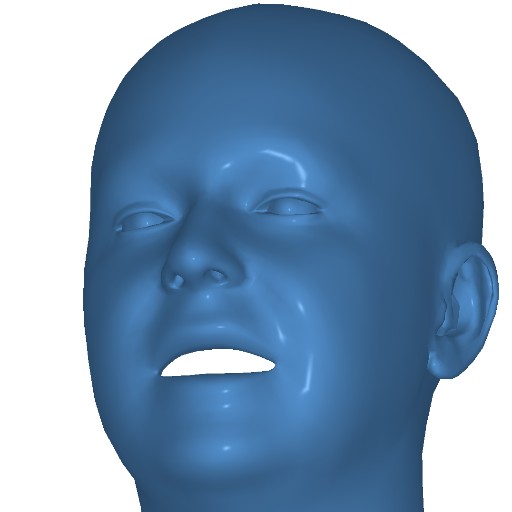} & \tabimg{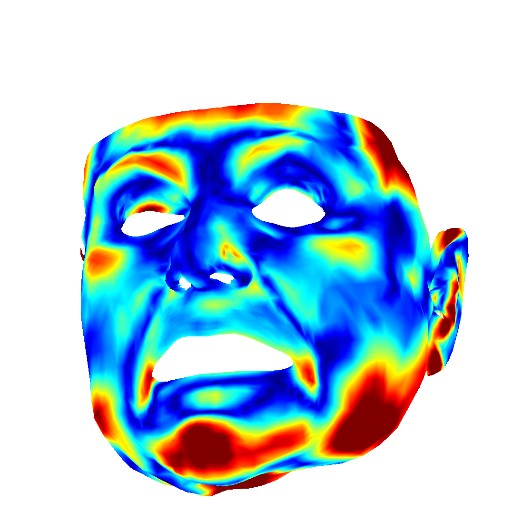} & \tabimg{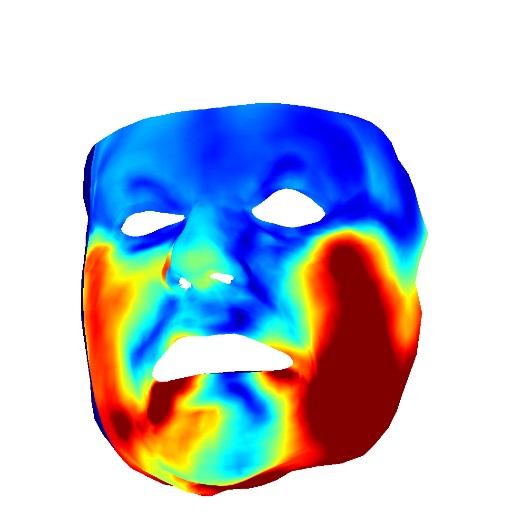} & \tabimg{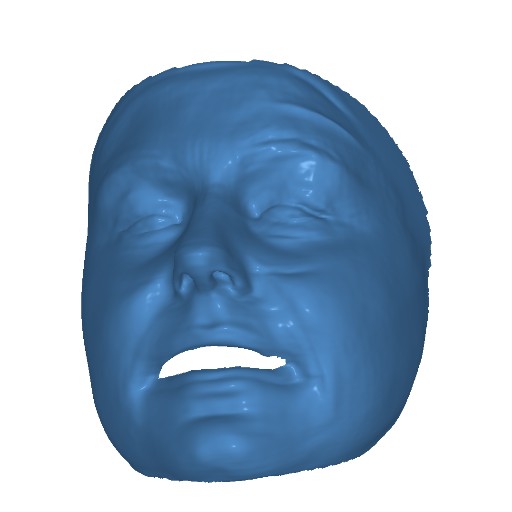} & \tabimg{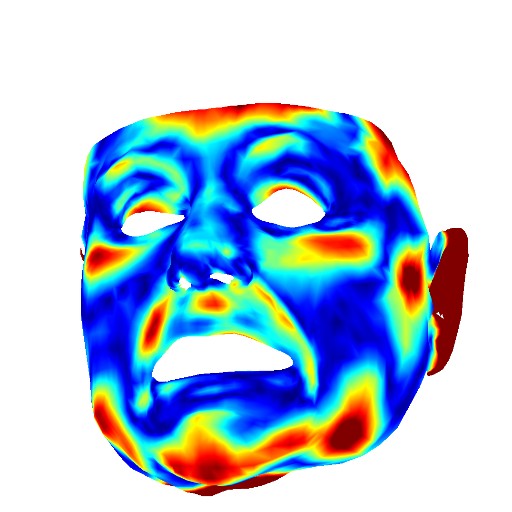} & \tabimg{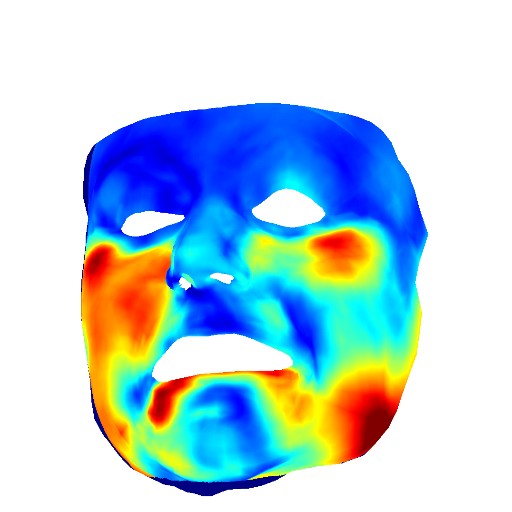} & \tabimg{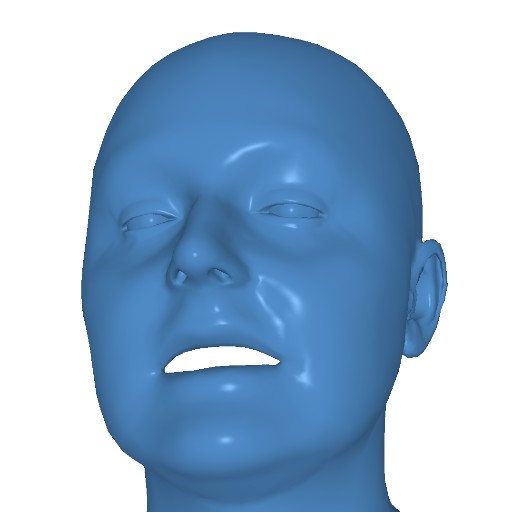} & \tabimg{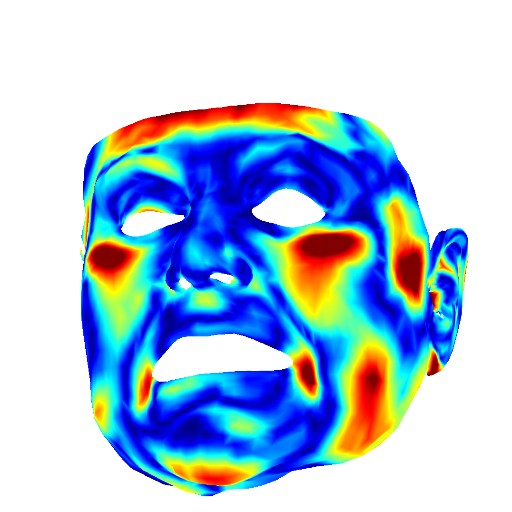} & \tabimg{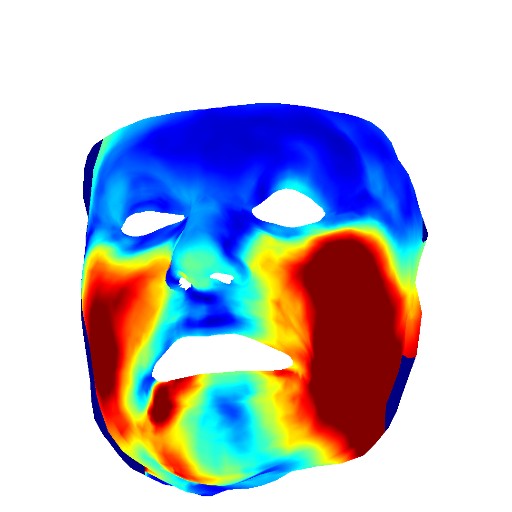}  & \tabimg{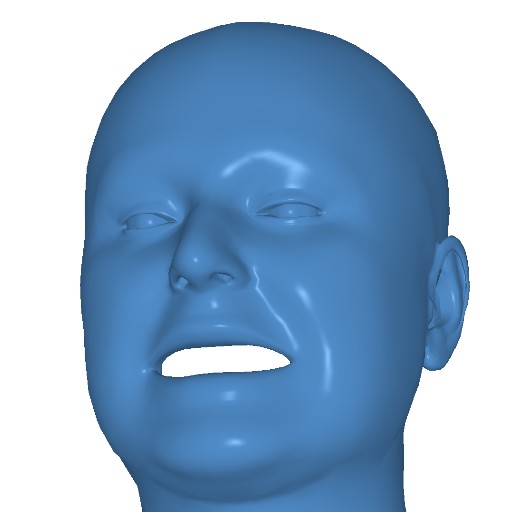} & \tabimg{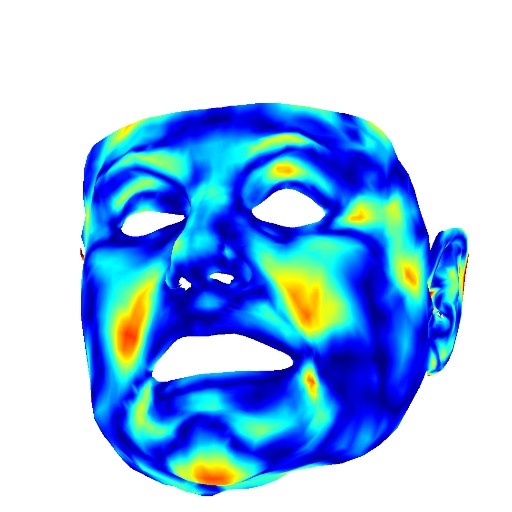} & \tabimg{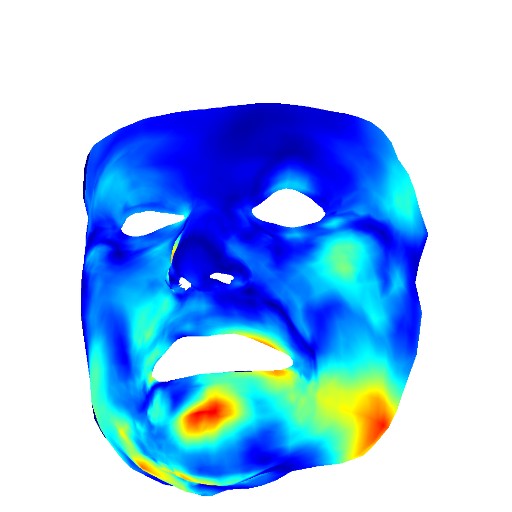} \\
\tabimg{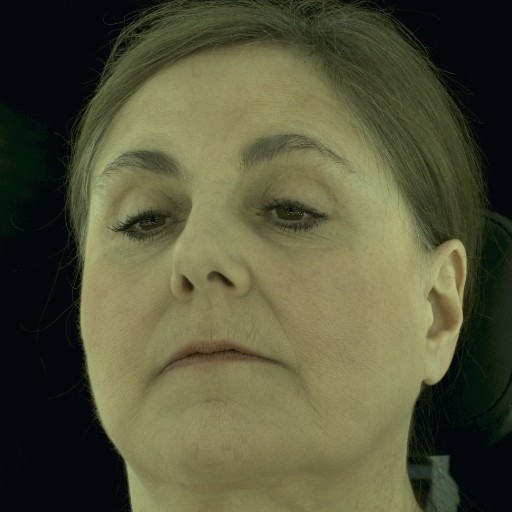} & \tabimg{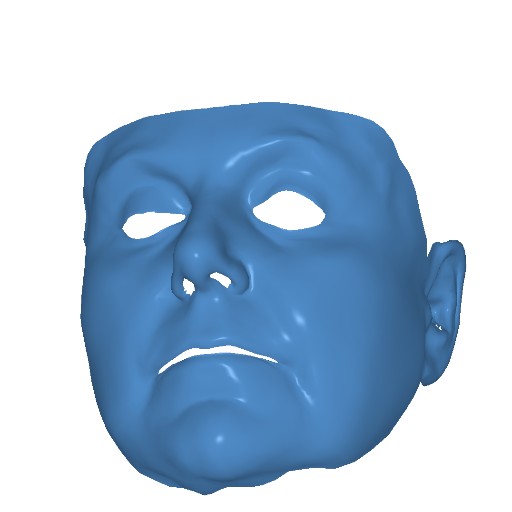} & \tabimg{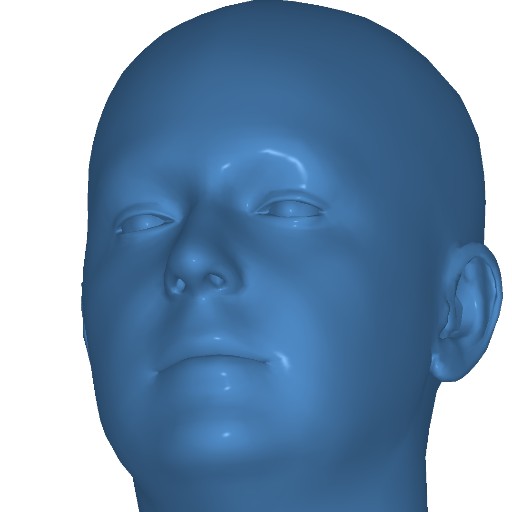} & \tabimg{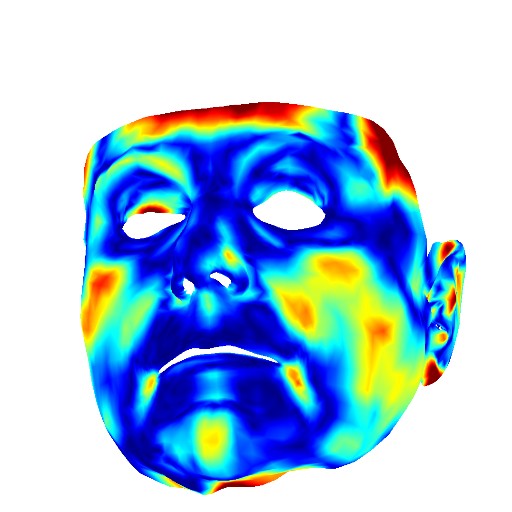} & \tabimg{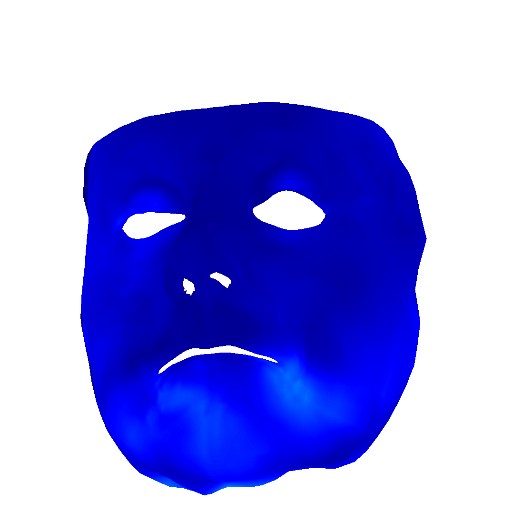} & \tabimg{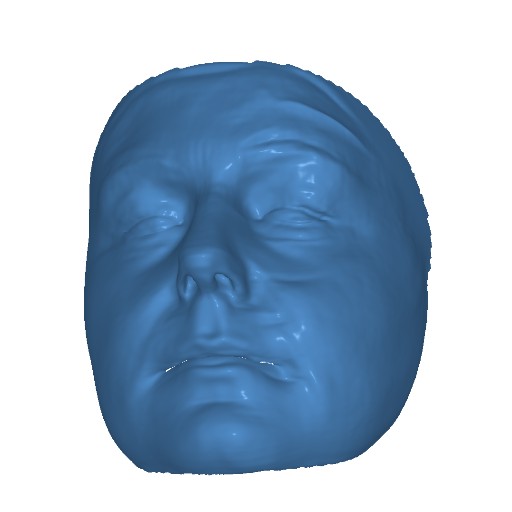} & \tabimg{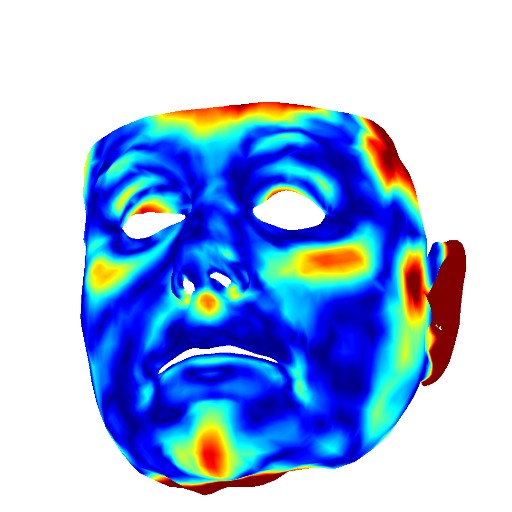} & \tabimg{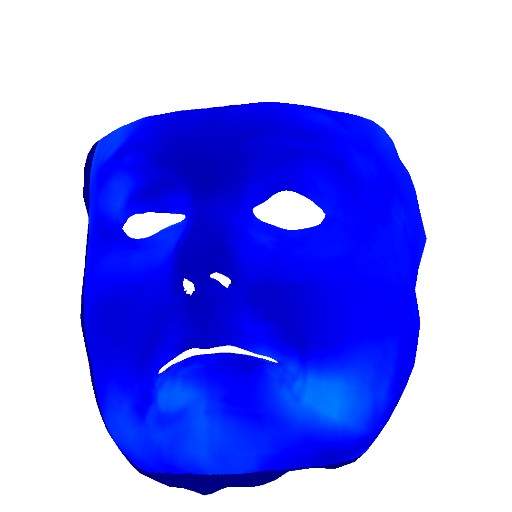} & \tabimg{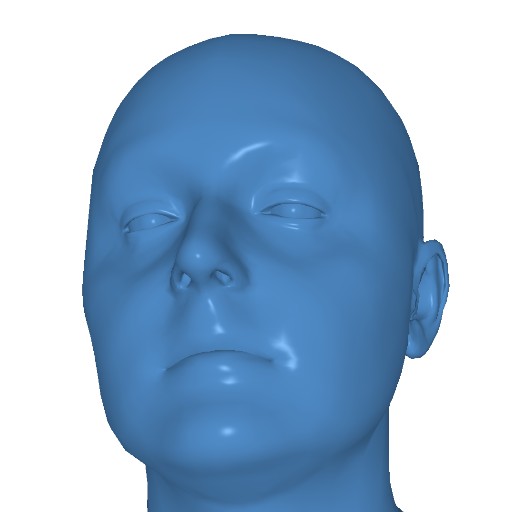} & \tabimg{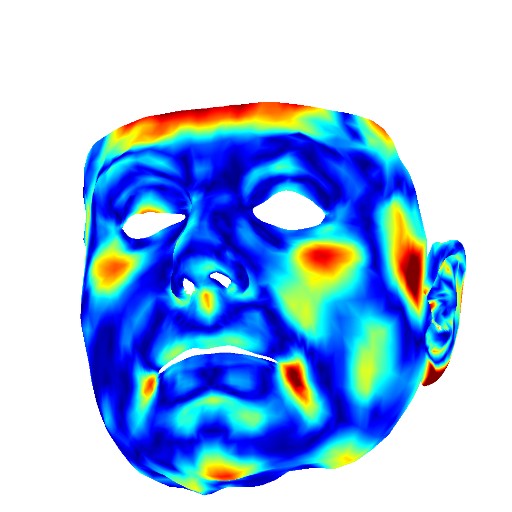} & \tabimg{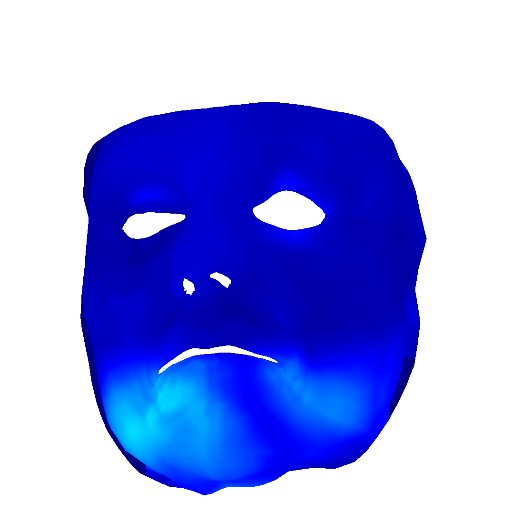}  & \tabimg{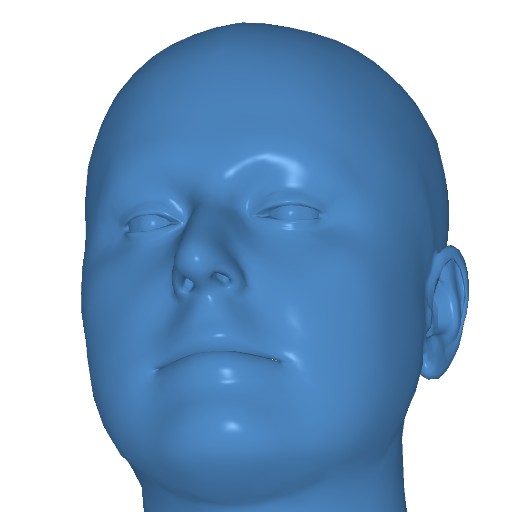} & \tabimg{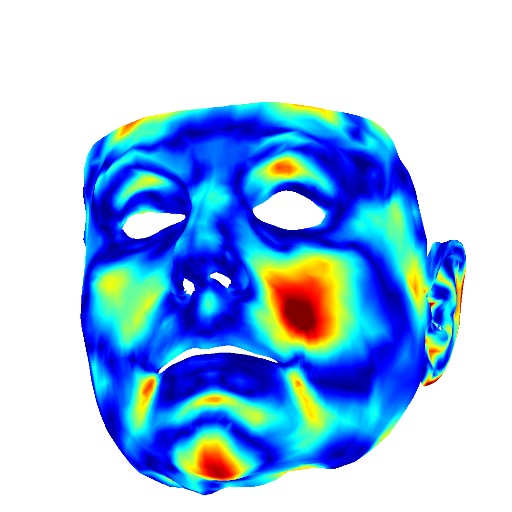} & \tabimg{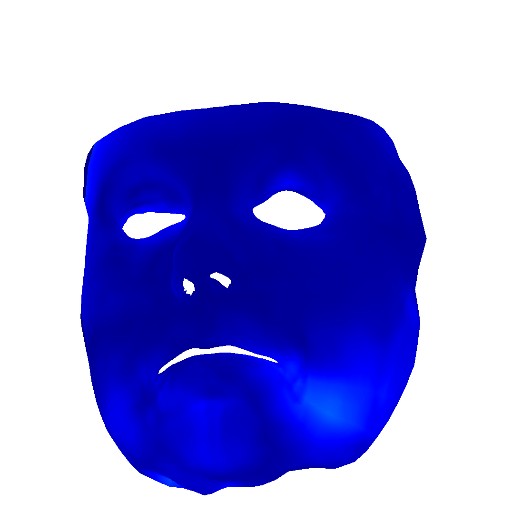} \\[0.1cm]
\multicolumn{2}{c|}{GT} & \multicolumn{3}{c|}{DECA} & \multicolumn{3}{c|}{HRN} & \multicolumn{3}{c|}{MPT} & \multicolumn{3}{c}{Ours} \\
\end{tabular}
\vspace{0.3cm}
\caption{Qualitative results on two sequences (top and bottom 3 rows) of our Multiface benchmark. Warmer colors represent high error, while colder colors represent low error. DECA \cite{deca}, HRN \cite{hrn}, and MPT \cite{mica} struggle with motion in the cheek and forehead region, which is visible in the SSME error plot (right columns). Despite using only 2D alignment as supervision, our method achieves a better 3D reconstruction (CD) (center columns).}
\label{tab:multiface_qualitative}
\vspace{-0.3cm}
\end{figure*}

\begin{table*}[th]
\tablefont
\setlength{\tabcolsep}{3pt}
\begin{center}
\begin{tabular}{ l|c c c c c|c c c c c | c c c c c | c c c c c  } 
& \multicolumn{10}{c|}{\textbf{No temporal information sharing (single image)}} & \multicolumn{10}{c}{\textbf{With temporal information sharing (sequence)}} \\
\toprule
\multirow{2}*{Method} & \multicolumn{5}{c|}{CD (mm) $\downarrow$} & \multicolumn{5}{c|}{$\overline{\text{SSME}}$ (px) $\downarrow$} & \multicolumn{5}{c|}{CD (mm) $\downarrow$} & \multicolumn{5}{c}{$\overline{\text{SSME}}$ (px) $\downarrow$}  \\
                & face   & mouth   & nose  & eyes   & ears   & face   & mouth   & nose  & eyes   & ears    & face   & mouth   & nose  & eyes   & ears   & face   & mouth   & nose  & eyes   & ears \\
\midrule
 DECA~\cite{deca}   & 1.37 &    1.29 &    1.32 &    1.08 &    2.68 &    5.66 &    6.16 &    3.60 &    4.25 &    8.34  &    1.37 &    1.29 &    1.32 &    1.08 &    2.68 &    5.26 &    6.12 &    3.22 &    3.87 &    7.10 \\
 EMOCA~\cite{emoca}  & 1.47 &    1.46 &    1.49 &    1.10 &    2.71 &    6.14 &    7.32 &    3.99 &    4.26 &    8.55  &    1.47 &    1.46 &    1.49 &    1.10 &    2.71 &    5.63 &    6.95 &    3.56 &    3.87 &    7.28 \\
  HRN~\cite{hrn}   & 1.49 &    1.39 &    1.24 &    1.09 &       - &    5.75 &    6.04 &    4.20 &    4.84 &       - &    1.49 &    1.39 &    1.24 &    1.09 &       - &    4.63 &    5.39 &    3.02 &    3.68 &       - \\
3DDFAv2~\cite{3ddfa_v2}   & 1.53 &    1.52 &    1.59 &    1.24 &       - &    7.91 &    9.47 &    6.65 &    6.55 &       - &    1.53 &    1.52 &    1.59 &    1.24 &       - &    6.71 &    8.43 &    5.43 &    5.44 &       - \\
PRNet~\cite{prnet}   & 1.55 &    1.59 &    1.50 &    1.28 &       - &    8.45 &   10.66 &    5.98 &    6.03 &       - &    1.55 &    1.59 &    1.50 &    1.28 &       - &    7.54 &    9.80 &    5.25 &    5.35 &       -\\
SADRNet~\cite{sadrnet}   & 1.49 &    1.52 &    1.49 &    1.22 &       - &    7.11 &    8.21 &    5.15 &    5.53 &       - &    1.49 &    1.52 &    1.49 &    1.22 &       - &    6.18 &    7.46 &    4.31 &    4.72 &       - \\
  MPT~\cite{mica} & - &    - &    - &    - &       - &    - &    - &    - &    - &       -  &    1.30 &    1.47 &    1.11 &    0.96 &       - &    5.74 &    7.34 &    4.64 &    4.01 &       -  \\
  \midrule
 Ours   & \best{1.20} &    \best{1.3} &    \best{1.05} &    \best{0.97} &    \best{2.34} &    
 \best{2.58} &    \best{3.14} &    \best{1.33} &    \best{2.07} &    \best{1.72}  & 
 \best{1.19} &    \best{1.31} &    \best{1.04} &    \best{0.96} &    \best{2.34} &    
 \best{2.50} &    \best{3.16} &    \best{1.27} &    \best{2.03} &    \best{1.68} \\   
\bottomrule
\end{tabular}
\end{center}
\caption{Results on our Multiface tracking benchmark with and without temporal information sharing. Our method consistently outperforms previous methods on every single category, metric and face region. }
\label{tab:multiface_quantitative}
\vspace{-0.5cm}
\end{table*}

We also compare our method on the FaceScape benchmark~\cite{facescape}, which measures 3D reconstruction accuracy from 2D images under large view (up to 90\degree) and expression variations. 
On this benchmark, we outperform the best previous regression-based methods by 38\% in terms of CD and 4.6\% in terms of mean normal error (NME) \cref{tab:fscape_benchmark}. This shows that our method can accurately reconstruct faces even under large view deviations.

\subsection{Now Challenge}

\begin{table} 
\begin{center}
\tablefont
\setlength{\tabcolsep}{2pt}
\begin{tabular}{l|ccc|ccc}
&\multicolumn{3}{c}{\textbf{Single-view}} &\multicolumn{3}{c}{\textbf{Multi-view}}\\
\toprule
\multirow{2}*{Method} & \multicolumn{3}{c|}{Error (mm) $\downarrow$}&\multicolumn{3}{c}{Error (mm) $\downarrow$}  \\
 & Median                  & Mean                    & Std & Median                  & Mean                    & Std     \\ 
\midrule
MGCNet~\cite{mgcnet}   & 1.31                         & 1.87                          & 2.63 & - & - & -      \\ 
PRNet~\cite{prnet}    & 1.50                         & 1.98                          & 1.88 & - & - & -       \\ 
DECA~\cite{deca}     & 1.09                         & 1.38                          & 1.18 & - & - & -      \\ 
Deep3D~\cite{deep3d}   & 1.11                         & 1.41                          & 1.21 & 1.08                         & 1.35                          & 1.15        \\ 
Dense~\cite{dense_landmarks_microsoft}    & 1.02                         & 1.28                          & 1.08  
& 0.81  & 1.01   & 0.84    \\ 
MICA~\cite{mica}    & 0.90                         & 1.11                          & 0.92       & - & - & -\\ 
TokenFace~\cite{tokenface} & \best{0.76}                        & \best{0.95}                          & \best{0.82} & - & - & - \\
\midrule
Ours     & 0.87                         & 1.07                         & 0.88 & \best{0.71}                & \best{0.88}                 & \best{0.73} \\ 
\bottomrule
\end{tabular}
\end{center}
\caption{Results on the NoW Challenge \cite{now_benchmark}. Multi-view evaluation is done as in \cite{dense_landmarks_microsoft}. Multi-view results for \cite{deep3d} and \cite{dense_landmarks_microsoft} are reported by \cite{dense_landmarks_microsoft}. }
\label{tab:now_benchmark}
\vspace{-0.5cm}
\end{table}

The NoW benchmark is a public benchmark for evaluating neutral head reconstruction from 2D images captured indoors and outdoors, with different expressions, and under variations in lighting conditions and occlusions. 
We evaluate our method on the non-metrical challenge (\cref{tab:now_benchmark}). For single-view reconstruction, our model outperforms our neutral shape predictor MICA \cite{mica} by 4\% on mean scan-to-mesh distance. 
For the multi-view case, we outperform the baseline \textit{Dense}~\cite{dense_landmarks_microsoft} by 13\%, likely due to our method's high 2D alignment accuracy, better neutral shape priors, and per-vertex deformations. TokenFace~\cite{tokenface} performs better for the single-view case, however, their predictions could be integrated into our pipeline since they use the FLAME topology. 
Importantly, our network is able to generalize to these in-the-wild images despite being trained only on in-the-lab data captured under controlled lighting conditions. 
An important sub-task for 3D face trackers is to disentangle the identity and expression components of the face shape. The outstanding results on the NoW benchmark indicate the ability of our tracker to accomplish this.

\subsection{Downstream Tasks} 

In the following, we show how we enhance downstream models using our face tracker.

\paragraph{3D Head Avatar Synthesis.}

Recent head avatar synthesis methods heavily rely on photometric head trackers to generate face alignment priors \cite{insta, pointavatar, nha}. 
INSTA~\cite{insta}, a top-performing model, uses MPT~\cite{mica}.
We modify INSTA by replacing their tracker with ours. We compare our enhanced FlowFace-INSTA to the baseline MPT-INSTA. On their publicly available dataset, we outperform MPT-INSTA by 10.5\% on perceptual visual fidelity (LPIPS). On our Multiface benchmark videos, we outperform MPT-INSTA by 20.3\% on LPIPS. Detailed results can be viewed in \cref{supp:head_avatar}. These results demonstrate how better face trackers can directly improve performance on down-stream tasks which highlights the importance of our research. 

\paragraph{Speech-driven 3D facial animation.}
The field of speech-driven facial animation often suffers from data sparsity \cite{vocaset, codetalker, faceformer}. To alleviate this issue, we generate 3D face meshes using the multi-view video dataset MEAD~\cite{mead}. In using this generated dataset to augment the training of the  state-of-the-art model CodeTalker~\cite{codetalker} (see \cref{supp:audio2face}), we are able to improve from a lip vertex error of $\num{3.13e-5}$ to $\num{2.85e-5}$ on the VOCASET benchmark \cite{vocaset}, an $8.8\%$ improvement. This underlines the benefit of high-accuracy video face trackers for large-scale data generation.

\subsection{2D Alignment}
\label{sec:2d_alignment_comparison}

To show the benefit of our 2D alignment model architecture, we conduct an evaluation on our validation set, which consists of 84 subjects of our dataset. We implement the dense landmark model of \cite{dense_landmarks_microsoft} (ResNet-101 backbone) and adapt it to output FLAME vertex alignment and uncertainty. We also implement PRNet~\cite{prnet} and modify it in the same way. We retrain each method on our training set. In evaluate the 2D alignment accuracy with respect to normalized mean error (NME) of every vertex in the face area (\cref{fig:flame_vertices}, green vertices). With an NME of $1.30$, our method performs signficantly better than the ResNet architecture of \textit{Dense}~\cite{dense_landmarks_microsoft} ($\text{NME}=1.63$), and PRNet ($\text{NME}=2.52$). We note that the accuracy of uncertainty cannot be evaluated with NME. A qualitative comparison can be viewed in \cref{fig:align_qualitative}.

\subsection{Ablation Studies}

\paragraph{2D alignment network.} To analyze the effect of different feature encoder backbones, we replace our backbone with different variations of the Segformer model and also test the CNN-based backbone BiSeNet-v2~\cite{bisenet_v2} (see \cref{tab:multiface_ablation}). As expected, vision-transformer-based networks show better performance. 
Experimenting with the number of iterations $N_\textit{iter}$ for the update module, we find that multiple iterations instead of one improves the performance. 
Finally, we confirm the superior performance of our 2D alignment network compared to the ResNet-101-based network of \cite{dense_landmarks_microsoft} mentioned in \cref{sec:2d_alignment_comparison}.

\begin{table}[t]
\tablefont
\begin{center}
\begin{tabular}{l|c|c|c|c|c}
\toprule
Backbone     & $N_\textit{iter}$  & \#Param  & latency (ms) & CD$\downarrow$ & $\overline{\textit{SSME}}$$\downarrow$ \\ 
\midrule     
ResNet-101   & ---              & 73.4M    &   9    & 1.54           & 3.90                 \\ 
BiSeNet-v2   & 3                & 17.6M    &  23    & 1.21           & 3.52                 \\ 
MiT-b1       & 3                & 17.3M    &  29    & 1.22           & 3.21                           \\
MiT-b2       & 3                & 31.0M    &  46     & 1.20           & 2.78                           \\ 
MiT-b5       & 1                & 88.2M    &  66     &  1.25           & 2.70                 \\ 
MiT-b5       & 2                & 88.2M    &  71     &  1.21           & 2.61                 \\ 
MiT-b5       & 3                & 88.2M    &  75     &  1.18           & 2.58                 \\ 
MiT-b5       & 4                & 88.2M    &  80     &  1.23           & 2.62             \\ 
\bottomrule
\end{tabular}
\end{center}
\caption{ Ablations for backbone architectures and hyper-parameters of the 2D alignment network on our Multiface benchmark. Latency is evaluated on a \textit{Quadro RTX 5000} GPU.}
\label{tab:multiface_ablation}
\vspace{-0.3cm}
\end{table}

\paragraph{3D model fitting.} We show in \cref{tab:now_ablation} the benefit of integrating the MICA neutral shape prediction on the NoW Challenge validation set. The significant performance gain on single-image predictions shows that our 3D tracking pipeline can integrate MICA predictions very well, even improving them. We also show the benefit of predicting a dense face alignment in conjunction with per-vertex deformations in multi-view settings. This shows that our 2D alignment is precise enough to predict face shapes that lie outside of the FLAME blend-shape space, which previous optimization-based methods \cite{mica,dense_landmarks_microsoft} cannot achieve. For a qualitative analysis, see \cref{supp:results}.

\begin{table} [ht]
\tablefont
\setlength{\tabcolsep}{3pt}
\begin{center}
\begin{tabular}{l|ccc|ccc}
&\multicolumn{3}{c}{\textbf{Single-view}}&\multicolumn{3}{c}{\textbf{Multi-view}} \\
\toprule
\multirow{2}*{Method} & \multicolumn{3}{c|}{Error (mm) $\downarrow$} & \multicolumn{3}{c}{Error (mm) $\downarrow$}  \\
 & Median                  & Mean                    & Std   & Median                  & Mean                    & Std   \\ 
\midrule
Ours w/o MICA           & 0.99  & 1.23 & 1.03 & 0.71 & 0.88 & 0.76       \\ 
MICA only               & 0.91  & 1.13 & 0.94 & - & - & -        \\
Ours w/o $\delta_d$   & - & - & -             & 0.68 & 0.84 & 0.72          \\
Ours                    & 0.82  & 1.02 & 0.85 & 0.67 & 0.83 & 0.71 \\ 
\bottomrule
\end{tabular}
\end{center}
\caption{Ablations for the 3D model fitting module on single and multi-view reconstruction on the NoW validation set.}
\label{tab:now_ablation}
\vspace{-0.3cm}
\end{table}
\section{Conclusion and Future Work}

This paper presents a state-of-the-art face tracking pipeline with a highly robust and accurate 2D alignment module. Its performance is thoroughly validated on a variety of benchmarks and downstream tasks. 
However, the proposed two-stage pipeline is not fully differentiable, which prevents end-to-end learning. Furthermore, our training data is limited to data captured in-the-lab.
In future work, we intend to extend the alignment network to directly predict depth as well, obviating the need for the 3D model fitting step. Synthetic datasets~\cite{dense_landmarks_microsoft} could alleviate the data issue.

We're confident that our tracker will accelerate research in downstream tasks by generating large-scale face capture data using readily available video datasets~\cite{voxceleb1,voxceleb2,celebvtext}. We also believe that our novel motion capture evaluation benchmark will focus and align future research efforts to create even more accurate methods. 




{
    \small
    \bibliographystyle{ieeenat_fullname}
    \bibliography{main}

\begin{thebibliography}{58}
\providecommand{\natexlab}[1]{#1}
\providecommand{\url}[1]{\texttt{#1}}
\expandafter\ifx\csname urlstyle\endcsname\relax
  \providecommand{\doi}[1]{doi: #1}\else
  \providecommand{\doi}{doi: \begingroup \urlstyle{rm}\Url}\fi

\bibitem[sti()]{stirling}
Stirling/esrc 3d face database.
\newblock \url{https://pics.stir.ac.uk/ESRC/}.
\newblock Accessed: 2023-10-25.

\bibitem[Blanz and Vetter(1999)]{3dmm}
Volker Blanz and Thomas Vetter.
\newblock A morphable model for the synthesis of 3d faces.
\newblock In \emph{Proceedings of the 26th Annual Conference on Computer Graphics and Interactive Techniques}, page 187–194, USA, 1999. ACM Press/Addison-Wesley Publishing Co.

\bibitem[Bolkart et~al.(2023)Bolkart, Li, and Black]{tempeh}
Timo Bolkart, Tianye Li, and Michael~J. Black.
\newblock Instant multi-view head capture through learnable registration.
\newblock In \emph{Conference on Computer Vision and Pattern Recognition (CVPR)}, pages 768--779, 2023.

\bibitem[Bulat and Tzimiropoulos(2017)]{face_alignment_network}
Adrian Bulat and Georgios Tzimiropoulos.
\newblock How far are we from solving the 2d \& 3d face alignment problem? (and a dataset of 230,000 3d facial landmarks).
\newblock In \emph{International Conference on Computer Vision}, 2017.

\bibitem[Cao et~al.(2018)Cao, Chai, Woodford, and Luo]{face_tracking_with_optical_flow}
Chen Cao, Menglei Chai, Oliver Woodford, and Linjie Luo.
\newblock Stabilized real-time face tracking via a learned dynamic rigidity prior.
\newblock \emph{ACM Trans. Graph.}, 37\penalty0 (6), 2018.

\bibitem[Chai et~al.(2022)Chai, Zhang, Ren, Kang, Xu, Zhe, Yuan, and Bao]{realy_benchmark}
Zenghao Chai, Haoxian Zhang, Jing Ren, Di Kang, Zhengzhuo Xu, Xuefei Zhe, Chun Yuan, and Linchao Bao.
\newblock Realy: Rethinking the evaluation of 3d face reconstruction, 2022.

\bibitem[Chai et~al.(2023)Chai, Zhang, He, Tan, Baltrušaitis, Wu, Li, Zhao, Yuan, and Bian]{hiface_microsoft}
Zenghao Chai, Tianke Zhang, Tianyu He, Xu Tan, Tadas Baltrušaitis, HsiangTao Wu, Runnan Li, Sheng Zhao, Chun Yuan, and Jiang Bian.
\newblock Hiface: High-fidelity 3d face reconstruction by learning static and dynamic details, 2023.

\bibitem[Chung et~al.(2018)Chung, Nagrani, and Zisserman]{voxceleb2}
J.~S. Chung, A. Nagrani, and A. Zisserman.
\newblock Voxceleb2: Deep speaker recognition.
\newblock In \emph{INTERSPEECH}, 2018.

\bibitem[Cudeiro et~al.(2019)Cudeiro, Bolkart, Laidlaw, Ranjan, and Black]{vocaset}
Daniel Cudeiro, Timo Bolkart, Cassidy Laidlaw, Anurag Ranjan, and Michael Black.
\newblock Capture, learning, and synthesis of {3D} speaking styles.
\newblock In \emph{Proceedings IEEE Conf. on Computer Vision and Pattern Recognition (CVPR)}, pages 10101--10111, 2019.

\bibitem[Danecek et~al.(2022)Danecek, Black, and Bolkart]{emoca}
Radek Danecek, Michael~J. Black, and Timo Bolkart.
\newblock Emoca: Emotion driven monocular face capture and animation, 2022.

\bibitem[Deng et~al.(2009)Deng, Dong, Socher, Li, Li, and Fei-Fei]{imagenet}
Jia Deng, Wei Dong, Richard Socher, Li-Jia Li, Kai Li, and Li Fei-Fei.
\newblock Imagenet: A large-scale hierarchical image database.
\newblock In \emph{2009 IEEE conference on computer vision and pattern recognition}, pages 248--255. Ieee, 2009.

\bibitem[Deng et~al.(2019)Deng, Yang, Xu, Chen, Jia, and Tong]{deep3d}
Yu Deng, Jiaolong Yang, Sicheng Xu, Dong Chen, Yunde Jia, and Xin Tong.
\newblock Accurate 3d face reconstruction with weakly-supervised learning: From single image to image set.
\newblock In \emph{IEEE Computer Vision and Pattern Recognition Workshops}, 2019.

\bibitem[Fan et~al.(2021)Fan, Lin, Saito, Wang, and Komura]{faceformer}
Yingruo Fan, Zhaojiang Lin, Jun Saito, Wenping Wang, and Taku Komura.
\newblock Faceformer: Speech-driven 3d facial animation with transformers.
\newblock \emph{arXiv preprint arXiv:2112.05329}, 2021.

\bibitem[Feng et~al.(2020)Feng, Feng, Black, and Bolkart]{deca}
Yao Feng, Haiwen Feng, Michael~J. Black, and Timo Bolkart.
\newblock Learning an animatable detailed 3d face model from in-the-wild images.
\newblock \emph{CoRR}, abs/2012.04012, 2020.

\bibitem[Garrido et~al.(2016{\natexlab{a}})Garrido, Zollh\"{o}fer, Casas, Valgaerts, Varanasi, P\'{e}rez, and Theobalt]{old_photometric_2}
Pablo Garrido, Michael Zollh\"{o}fer, Dan Casas, Levi Valgaerts, Kiran Varanasi, Patrick P\'{e}rez, and Christian Theobalt.
\newblock Reconstruction of personalized 3d face rigs from monocular video.
\newblock \emph{ACM Trans. Graph.}, 35\penalty0 (3), 2016{\natexlab{a}}.

\bibitem[Garrido et~al.(2016{\natexlab{b}})Garrido, Zollh\"{o}fer, Wu, Bradley, P\'{e}rez, Beeler, and Theobalt]{old_photometric_0}
Pablo Garrido, Michael Zollh\"{o}fer, Chenglei Wu, Derek Bradley, Patrick P\'{e}rez, Thabo Beeler, and Christian Theobalt.
\newblock Corrective 3d reconstruction of lips from monocular video.
\newblock \emph{ACM Trans. Graph.}, 35\penalty0 (6), 2016{\natexlab{b}}.

\bibitem[Grassal et~al.(2021)Grassal, Prinzler, Leistner, Rother, Nie{\ss}ner, and Thies]{nha}
Philip-William Grassal, Malte Prinzler, Titus Leistner, Carsten Rother, Matthias Nie{\ss}ner, and Justus Thies.
\newblock Neural head avatars from monocular rgb videos.
\newblock \emph{arXiv preprint arXiv:2112.01554}, 2021.

\bibitem[Grishchenko et~al.(2020)Grishchenko, Ablavatski, Kartynnik, Raveendran, and Grundmann]{mediapipe}
Ivan Grishchenko, Artsiom Ablavatski, Yury Kartynnik, Karthik Raveendran, and Matthias Grundmann.
\newblock Attention mesh: High-fidelity face mesh prediction in real-time.
\newblock \emph{CoRR}, abs/2006.10962, 2020.

\bibitem[Guo et~al.(2020)Guo, Zhu, Yang, Fan, Lei, and Li]{3ddfa_v2}
Jianzhu Guo, Xiangyu Zhu, Yang Yang, Yang Fan, Zhen Lei, and Stan Li.
\newblock \emph{Towards Fast, Accurate and Stable 3D Dense Face Alignment}, pages 152--168.
\newblock 2020.

\bibitem[Güler et~al.(2017)Güler, Trigeorgis, Antonakos, Snape, Zafeiriou, and Kokkinos]{densereg}
Rıza~Alp Güler, George Trigeorgis, Epameinondas Antonakos, Patrick Snape, Stefanos Zafeiriou, and Iasonas Kokkinos.
\newblock Densereg: Fully convolutional dense shape regression in-the-wild, 2017.

\bibitem[He et~al.(2015)He, Zhang, Ren, and Sun]{resnet}
Kaiming He, Xiangyu Zhang, Shaoqing Ren, and Jian Sun.
\newblock Deep residual learning for image recognition, 2015.

\bibitem[IEE(2009)]{baselface}
\emph{A 3D Face Model for Pose and Illumination Invariant Face Recognition}, Genova, Italy, 2009. IEEE.

\bibitem[Kingma and Ba(2015)]{ba2015adam}
Diederik~P. Kingma and Jimmy Ba.
\newblock Adam: {A} method for stochastic optimization.
\newblock In \emph{3rd International Conference on Learning Representations, {ICLR} 2015, San Diego, CA, USA, May 7-9, 2015, Conference Track Proceedings}, 2015.

\bibitem[Lei et~al.(2023)Lei, Ren, Feng, Cui, and Xie]{hrn}
Biwen Lei, Jianqiang Ren, Mengyang Feng, Miaomiao Cui, and Xuansong Xie.
\newblock A hierarchical representation network for accurate and detailed face reconstruction from in-the-wild images, 2023.

\bibitem[Lewis et~al.(2000)Lewis, Cordner, and Fong]{linearblendskinning}
J.~P. Lewis, Matt Cordner, and Nickson Fong.
\newblock Pose space deformation: A unified approach to shape interpolation and skeleton-driven deformation.
\newblock In \emph{Proceedings of the 27th Annual Conference on Computer Graphics and Interactive Techniques}, page 165–172, USA, 2000. ACM Press/Addison-Wesley Publishing Co.

\bibitem[Li et~al.(2017)Li, Bolkart, Black, Li, and Romero]{FLAME}
Tianye Li, Timo Bolkart, Michael.~J. Black, Hao Li, and Javier Romero.
\newblock Learning a model of facial shape and expression from {4D} scans.
\newblock \emph{ACM Transactions on Graphics, (Proc. SIGGRAPH Asia)}, 36\penalty0 (6):\penalty0 194:1--194:17, 2017.

\bibitem[Loshchilov and Hutter(2017)]{adamw}
Ilya Loshchilov and Frank Hutter.
\newblock Fixing weight decay regularization in adam.
\newblock \emph{CoRR}, abs/1711.05101, 2017.

\bibitem[Morales et~al.(2020)Morales, Piella, and Sukno]{reconstruction_survey}
Araceli Morales, Gemma Piella, and Federico~M. Sukno.
\newblock Survey on 3d face reconstruction from uncalibrated images.
\newblock \emph{CoRR}, abs/2011.05740, 2020.

\bibitem[Nagrani et~al.(2019)Nagrani, Chung, Xie, and Zisserman]{voxceleb1}
Arsha Nagrani, Joon~Son Chung, Weidi Xie, and Andrew Zisserman.
\newblock Voxceleb: Large-scale speaker verification in the wild.
\newblock \emph{Computer Science and Language}, 2019.

\bibitem[Prados-Torreblanca et~al.(2022)Prados-Torreblanca, Buenaposada, and Baumela]{spiga}
Andrés Prados-Torreblanca, José~M Buenaposada, and Luis Baumela.
\newblock Shape preserving facial landmarks with graph attention networks.
\newblock In \emph{33rd British Machine Vision Conference 2022, {BMVC} 2022, London, UK, November 21-24, 2022}. {BMVA} Press, 2022.

\bibitem[Rai et~al.(2023)Rai, Gupta, Pandey, Carrasco, Takagi, Aubel, Kim, Prakash, and de~la Torre]{albedogan}
Aashish Rai, Hiresh Gupta, Ayush Pandey, Francisco~Vicente Carrasco, Shingo~Jason Takagi, Amaury Aubel, Daeil Kim, Aayush Prakash, and Fernando de~la Torre.
\newblock Towards realistic generative 3d face models, 2023.

\bibitem[Ruan et~al.(2021)Ruan, Zou, Wu, Wu, and Wang]{sadrnet}
Zeyu Ruan, Changqing Zou, Longhai Wu, Gangshan Wu, and Limin Wang.
\newblock {SADRNet}: Self-aligned dual face regression networks for robust 3d dense face alignment and reconstruction.
\newblock \emph{{IEEE} Transactions on Image Processing}, 30:\penalty0 5793--5806, 2021.

\bibitem[Sandler et~al.(2019)Sandler, Howard, Zhu, Zhmoginov, and Chen]{mobilenetv2}
Mark Sandler, Andrew Howard, Menglong Zhu, Andrey Zhmoginov, and Liang-Chieh Chen.
\newblock Mobilenetv2: Inverted residuals and linear bottlenecks, 2019.

\bibitem[Sanyal et~al.(2019)Sanyal, Bolkart, Feng, and Black]{now_benchmark}
Soubhik Sanyal, Timo Bolkart, Haiwen Feng, and Michael Black.
\newblock Learning to regress 3d face shape and expression from an image without 3d supervision.
\newblock In \emph{Proceedings IEEE Conf. on Computer Vision and Pattern Recognition (CVPR)}, 2019.

\bibitem[Shang et~al.(2020)Shang, Shen, Li, Zhou, Zhen, Fang, and Quan]{mgcnet}
Jiaxiang Shang, Tianwei Shen, Shiwei Li, Lei Zhou, Mingmin Zhen, Tian Fang, and Long Quan.
\newblock Self-supervised monocular 3d face reconstruction by occlusion-aware multi-view geometry consistency.
\newblock \emph{arXiv preprint arXiv:2007.12494}, 2020.

\bibitem[Teed and Deng(2020)]{raft_optical_flow}
Zachary Teed and Jia Deng.
\newblock {RAFT:} recurrent all-pairs field transforms for optical flow.
\newblock \emph{CoRR}, abs/2003.12039, 2020.

\bibitem[Thies et~al.(2020)Thies, Zollhöfer, Stamminger, Theobalt, and Nießner]{face2face}
Justus Thies, Michael Zollhöfer, Marc Stamminger, Christian Theobalt, and Matthias Nießner.
\newblock Face2face: Real-time face capture and reenactment of rgb videos, 2020.

\bibitem[Tianke et~al.(2023)Tianke, Xuangeng, Yunfei, Lijian, Zhendong, Zhengzhuo, Chengkun, Fei, Changyin, Chun, and Li]{tokenface}
Zhang Tianke, Chu Xuangeng, Liu Yunfei, Lin Lijian, Yang Zhendong, Xu Zhengzhuo, Cao Chengkun, Yu Fei, Zhou Changyin, Yuan Chun, and Yu Li.
\newblock Accurate 3d face reconstruction with facial component tokens.
\newblock In \emph{Proceedings of the IEEE/CVF International Conference on Computer Vision (ICCV)}, 2023.

\bibitem[Voo et~al.(2022)Voo, Jiang, and Loy]{occlusions}
Kenny T.~R. Voo, Liming Jiang, and Chen~Change Loy.
\newblock Delving into high-quality synthetic face occlusion segmentation datasets.
\newblock In \emph{Proceedings of the IEEE/CVF Conference on Computer Vision and Pattern Recognition (CVPR) Workshops}, 2022.

\bibitem[Wang et~al.(2020)Wang, Wu, Song, Yang, Wu, Qian, He, Qiao, and Loy]{mead}
Kaisiyuan Wang, Qianyi Wu, Linsen Song, Zhuoqian Yang, Wayne Wu, Chen Qian, Ran He, Yu Qiao, and Chen~Change Loy.
\newblock Mead: A large-scale audio-visual dataset for emotional talking-face generation.
\newblock In \emph{ECCV}, 2020.

\bibitem[Wang and Solomon(2019)]{prnet}
Yue Wang and Justin~M. Solomon.
\newblock Prnet: Self-supervised learning for partial-to-partial registration, 2019.

\bibitem[Wood et~al.(2022)Wood, Baltrusaitis, Hewitt, Johnson, Shen, Milosavljevic, Wilde, Garbin, Raman, Shotton, Sharp, Stojiljkovic, Cashman, and Valentin]{dense_landmarks_microsoft}
Erroll Wood, Tadas Baltrusaitis, Charlie Hewitt, Matthew Johnson, Jingjing Shen, Nikola Milosavljevic, Daniel Wilde, Stephan Garbin, Chirag Raman, Jamie Shotton, Toby Sharp, Ivan Stojiljkovic, Tom Cashman, and Julien Valentin.
\newblock 3d face reconstruction with dense landmarks, 2022.

\bibitem[Wu et~al.(2016)Wu, Bradley, Gross, and Beeler]{old_photometric_1}
Chenglei Wu, Derek Bradley, Markus Gross, and Thabo Beeler.
\newblock An anatomically-constrained local deformation model for monocular face capture.
\newblock \emph{ACM Trans. Graph.}, 35\penalty0 (4), 2016.

\bibitem[Wuu et~al.(2022)Wuu, Zheng, Ardisson, Bali, Belko, Brockmeyer, Evans, Godisart, Ha, Huang, Hypes, Koska, Krenn, Lombardi, Luo, McPhail, Millerschoen, Perdoch, Pitts, Richard, Saragih, Saragih, Shiratori, Simon, Stewart, Trimble, Weng, Whitewolf, Wu, Yu, and Sheikh]{multiface}
Cheng-hsin Wuu, Ningyuan Zheng, Scott Ardisson, Rohan Bali, Danielle Belko, Eric Brockmeyer, Lucas Evans, Timothy Godisart, Hyowon Ha, Xuhua Huang, Alexander Hypes, Taylor Koska, Steven Krenn, Stephen Lombardi, Xiaomin Luo, Kevyn McPhail, Laura Millerschoen, Michal Perdoch, Mark Pitts, Alexander Richard, Jason Saragih, Junko Saragih, Takaaki Shiratori, Tomas Simon, Matt Stewart, Autumn Trimble, Xinshuo Weng, David Whitewolf, Chenglei Wu, Shoou-I Yu, and Yaser Sheikh.
\newblock Multiface: A dataset for neural face rendering.
\newblock In \emph{arXiv}, 2022.

\bibitem[Xie et~al.(2021)Xie, Wang, Yu, Anandkumar, Alvarez, and Luo]{segformer}
Enze Xie, Wenhai Wang, Zhiding Yu, Anima Anandkumar, Jose~M Alvarez, and Ping Luo.
\newblock Segformer: Simple and efficient design for semantic segmentation with transformers.
\newblock In \emph{Neural Information Processing Systems (NeurIPS)}, 2021.

\bibitem[Xing et~al.(2023)Xing, Xia, Zhang, Cun, Wang, and Wong]{codetalker}
Jinbo Xing, Menghan Xia, Yuechen Zhang, Xiaodong Cun, Jue Wang, and Tien-Tsin Wong.
\newblock Codetalker: Speech-driven 3d facial animation with discrete motion prior, 2023.

\bibitem[Yang et~al.(2020)Yang, Zhu, Wang, Huang, Shen, Yang, and Cao]{facescape}
Haotian Yang, Hao Zhu, Yanru Wang, Mingkai Huang, Qiu Shen, Ruigang Yang, and Xun Cao.
\newblock Facescape: a large-scale high quality 3d face dataset and detailed riggable 3d face prediction, 2020.

\bibitem[Yi et~al.(2023)Yi, Liang, Liu, Cao, Wen, Bolkart, Tao, and Black]{show_body_tracker}
Hongwei Yi, Hualin Liang, Yifei Liu, Qiong Cao, Yandong Wen, Timo Bolkart, Dacheng Tao, and Michael~J. Black.
\newblock Generating holistic 3d human motion from speech, 2023.

\bibitem[Yu et~al.(2020)Yu, Gao, to~the~baseline MPT-INSTA Jingbo~Wang, Yu, Shen, and Sang]{bisenet_v2}
Changqian Yu, Changxin Gao, FlowFace-INSTA to~the~baseline MPT-INSTA Jingbo~Wang, Gang Yu, Chunhua Shen, and Nong Sang.
\newblock Bisenet {V2:} bilateral network with guided aggregation for real-time semantic segmentation.
\newblock \emph{CoRR}, abs/2004.02147, 2020.

\bibitem[Yu et~al.(2023)Yu, Zhu, Jiang, Loy, Cai, and Wu]{celebvtext}
Jianhui Yu, Hao Zhu, Liming Jiang, Chen~Change Loy, Weidong Cai, and Wayne Wu.
\newblock {CelebV-Text}: A large-scale facial text-video dataset.
\newblock In \emph{CVPR}, 2023.

\bibitem[Zhang et~al.(2017)Zhang, Zhu, Lei, Shi, Wang, and Li]{sfd_face}
Shifeng Zhang, Xiangyu Zhu, Zhen Lei, Hailin Shi, Xiaobo Wang, and Stan~Z. Li.
\newblock S$^3$fd: Single shot scale-invariant face detector, 2017.

\bibitem[Zheng et~al.(2021)Zheng, Abrevaya, Chen, B{\"{u}}hler, Black, and Hilliges]{imavatar}
Yufeng Zheng, Victoria~Fern{\'{a}}ndez Abrevaya, Xu Chen, Marcel~C. B{\"{u}}hler, Michael~J. Black, and Otmar Hilliges.
\newblock I {M} avatar: Implicit morphable head avatars from videos.
\newblock \emph{CoRR}, abs/2112.07471, 2021.

\bibitem[Zheng et~al.(2023)Zheng, Yifan, Wetzstein, Black, and Hilliges]{pointavatar}
Yufeng Zheng, Wang Yifan, Gordon Wetzstein, Michael~J. Black, and Otmar Hilliges.
\newblock Pointavatar: Deformable point-based head avatars from videos.
\newblock In \emph{Proceedings of the IEEE/CVF Conference on Computer Vision and Pattern Recognition (CVPR)}, 2023.

\bibitem[Zhou et~al.(2023)Zhou, Li, Liu, Wang, Yu, and Ji]{star_keypoints}
Zhenglin Zhou, Huaxia Li, Hong Liu, Nanyang Wang, Gang Yu, and Rongrong Ji.
\newblock Star loss: Reducing semantic ambiguity in facial landmark detection, 2023.

\bibitem[Zhu et~al.(2015)Zhu, Lei, Liu, Shi, and Li]{aflw_dataset}
Xiangyu Zhu, Zhen Lei, Xiaoming Liu, Hailin Shi, and Stan~Z. Li.
\newblock Face alignment across large poses: {A} 3d solution.
\newblock \emph{CoRR}, abs/1511.07212, 2015.

\bibitem[Zielonka et~al.(2022{\natexlab{a}})Zielonka, Bolkart, and Thies]{insta}
Wojciech Zielonka, Timo Bolkart, and Justus Thies.
\newblock Instant volumetric head avatars.
\newblock \emph{2023 IEEE/CVF Conference on Computer Vision and Pattern Recognition (CVPR)}, pages 4574--4584, 2022{\natexlab{a}}.

\bibitem[Zielonka et~al.(2022{\natexlab{b}})Zielonka, Bolkart, and Thies]{mica}
Wojciech Zielonka, Timo Bolkart, and Justus Thies.
\newblock Towards metrical reconstruction of human faces, 2022{\natexlab{b}}.

\bibitem[Zollh{\"o}fer et~al.(2018)Zollh{\"o}fer, Thies, Bradley, Garrido, Beeler, P{\'e}erez, Stamminger, Nie{\ss}ner, and Theobalt]{monocular_survey}
Michael Zollh{\"o}fer, Justus Thies, Darek Bradley, Pablo Garrido, Thabo Beeler, Patrick P{\'e}erez, Marc Stamminger, Matthias Nie{\ss}ner, and Christian Theobalt.
\newblock State of the art on monocular 3d face reconstruction, tracking, and applications.
\newblock 2018.

\end{thebibliography}
}

\appendix
\clearpage
\setcounter{page}{1}
\maketitlesupplementary

\section{Overview}
In the following, we describe in detail the architecture of our 2D alignment network. We also show the datasets used to train the 2D alignment network, how they are annotated and how we augment the data. Furthermore, we provide details of our Multiface benchmark dataset. Through various visualizations of additional results, we show and compare the accuracy of our model. Lastly, we explain in detail our experiments on the downstream tasks head avatar synthesis and speech-driven 3D face animation. 

\section{2D Alignment Network Architecture Details}
\label{supp:2d_alignment}

As mentioned in the paper, our 2D alignment network consists of three parts: an image feature encoder, UV feature generators and a UV-image flow prediction module. This setup allows us to build on extensive research the fields of image feature encoding and optical flow prediction. 

\subsection{Image feature encoder}

\begin{figure}[h]
\centering
\begin{subfigure}[t]{0.25\textwidth}
  \begin{center}
  \includegraphics[width=0.75\linewidth]{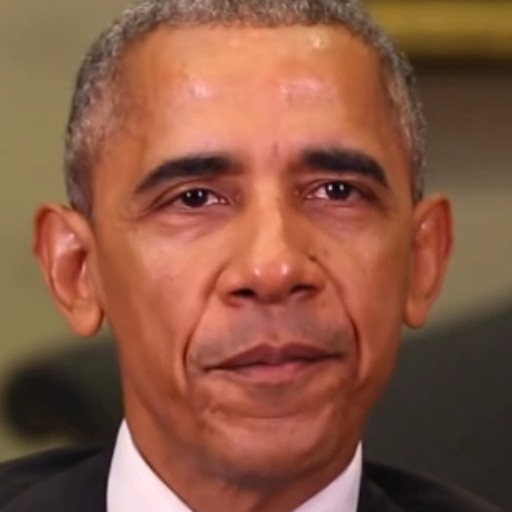}
  \caption{Input image}
  \label{fig:latent_orig_img}
  \end{center}
\end{subfigure}%
\begin{subfigure}[t]{0.25\textwidth}
  \begin{center}
  \includegraphics[width=0.75\linewidth]{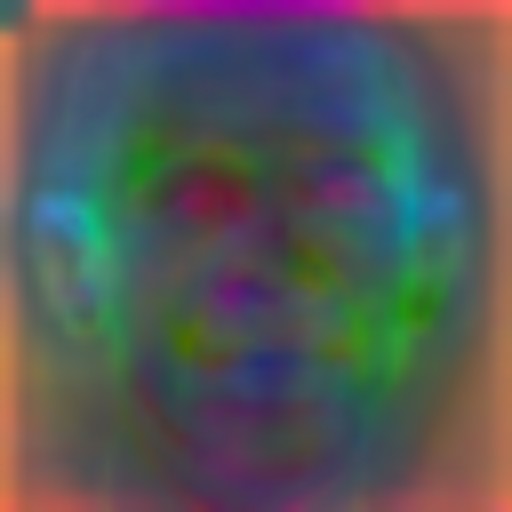}
  \caption{Latent feature map }
  \label{fig:latent_latent_img}
  \end{center}
\end{subfigure}
\vspace{0.2cm}
\caption{Visualization of the latent feature encoding $Z_\textit{img}$ (b) of the corresponding input image (a) using PCA. The first three principal components are colored in red, green and blue respectively. This visualization shows that our image feature encoder learns to produce some sort of semantic information. It also suggests that the network attends to visually salient areas such as tip of the ear (light blue), eyebrows (green), or silhouette (green and purple). }
\label{fig:latent_vis}
\end{figure}

To produce accurate and semantically meaningful features, we use a state-of-the-art semantic segmentation model as our feature encoder. As mentioned in the paper, we select the vision-transformer-based Segformer~\cite{segformer}, which has demonstrated top results in semantic segmentation benchmarks. It is pre-trained on ImageNet~\cite{imagenet}, which enables us to transfer large-scale image knowledge for enhanced feature generation. We show that this network can predict meaningful information by visualizing the generated latent feature map in \cref{fig:latent_vis}. 

\subsection{UV-image flow prediction}

\begin{figure}
\centering
\includegraphics[width=0.95\linewidth]{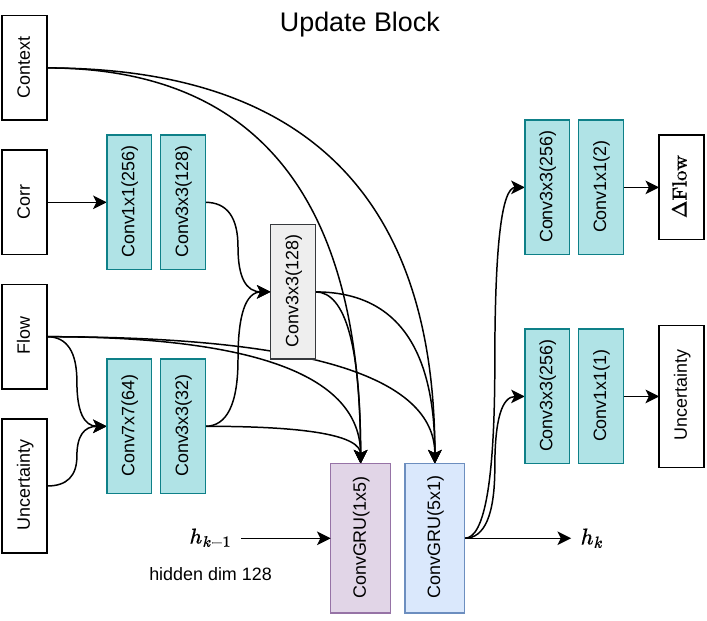}
\vspace{0.2cm}
\caption{An overview of our modified RAFT update module. We include the previous uncertainty prediction in the motion encoder (on the left) and output the updated  uncertainty using an additional output block (on the right). Context and initial hidden code are generated by our UV feature generators. }
\label{fig:flowface_raft}
\end{figure}

For our UV-image flow prediction module, we adapt RAFT~\cite{raft_optical_flow}. This model has shown excellent results on optical flow prediction, and demonstrated great capability for generalization due to its clever network design. The multi-scale 4D correlation volume allows the network to \textit{correlate} and associate features across large pixel offsets. The recurrent update block mimics an iterative optimization process, where a flow estimate is refined with each iteration. In our 2D alignment network, RAFT is modified to not predict the optical flow between two images, but the per-pixel offset between the UV space and image space. As mentioned in the paper, we add the capability to predict the UV-image flow uncertainty. In \cref{fig:flowface_raft}, we show the specific modifications we made to the RAFT module to also output uncertainty. 

Offloading the alignment task to this UV-flow prediction network allows the image feature encoder to focus on both high and low-level features (see \cref{fig:latent_vis}). The flow prediction module can then use these features to align the UV space with pixel-level accuracy.

\subsection{UV positional encoding module}

\begin{figure}
\centering
\includegraphics[width=0.95\linewidth]{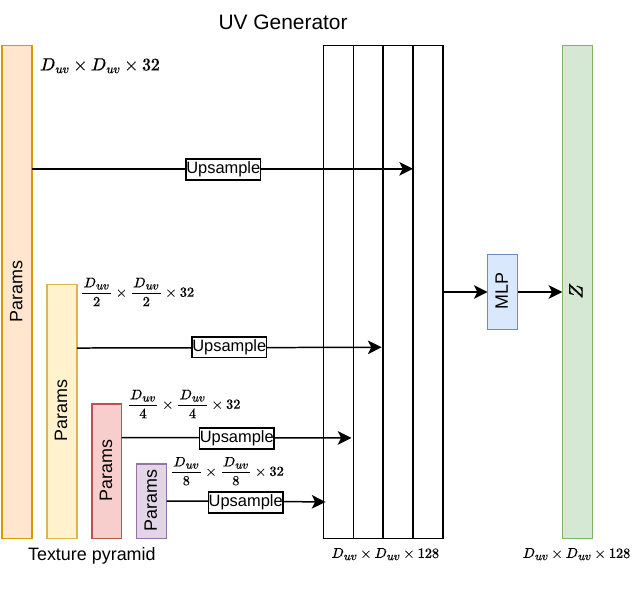}
\vspace{0.2cm}
\caption{The architecture our UV positional encoding modules. A parameter texture pyramid (left) is upsampled to UV dimensions, concatenated (center) and then processed by a linear layer (right). We deploy three of these generators to generate positional embeddings that are used as UV features for the RAFT correlation block, and context and hidden code for the RAFT update block. }
\label{fig:uv_generator}
\end{figure}

To generate UV space features, initial hidden code and a context map for the update module, we use three identical multi-scale positional encoding modules. The architecture of these modules is shown in \cref{fig:uv_generator}.

\section{Multiface Benchmark Dataset}
\label{supp:multiface}

\begin{figure}
\centering
\begin{subfigure}[t]{0.32\linewidth}
  \begin{center}
  \includegraphics[width=0.95\linewidth]{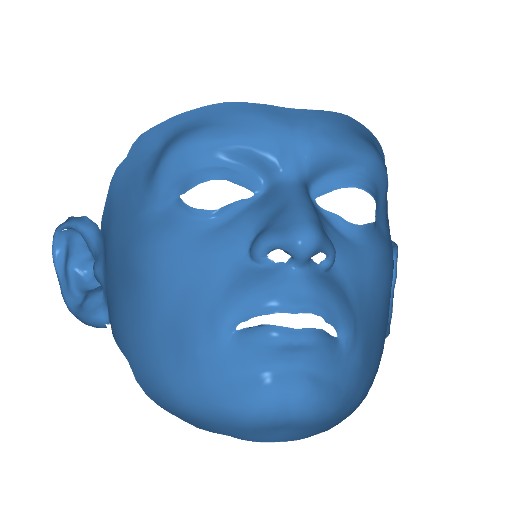}
  \caption{GT Mesh}
  \end{center}
\end{subfigure}%
\begin{subfigure}[t]{0.32\linewidth}
  \begin{center}
  \includegraphics[width=0.95\linewidth]{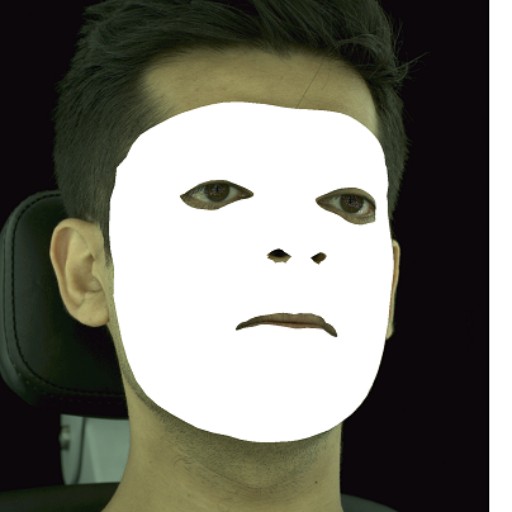}
  \caption{face}
  \end{center}
\end{subfigure}
\begin{subfigure}[t]{0.32\linewidth}
  \begin{center}
  \includegraphics[width=0.95\linewidth]{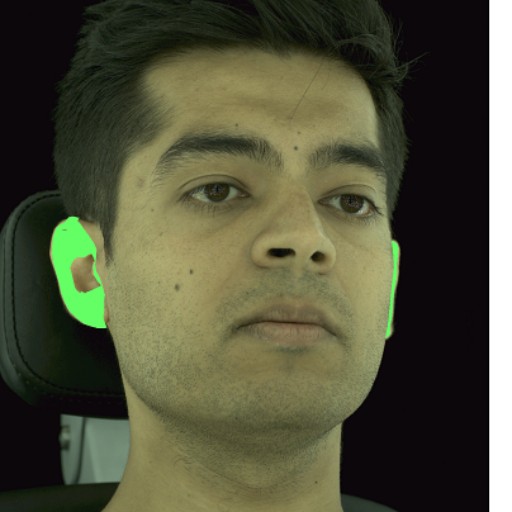}
  \caption{ears}
  \end{center}
\end{subfigure}
\begin{subfigure}[t]{0.32\linewidth}
  \begin{center}
  \includegraphics[width=0.95\linewidth]{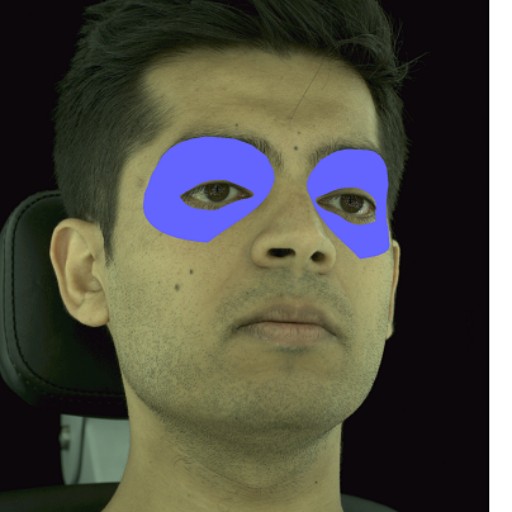}
  \caption{eyes}
  \end{center}
\end{subfigure}
\begin{subfigure}[t]{0.32\linewidth}
  \begin{center}
  \includegraphics[width=0.95\linewidth]{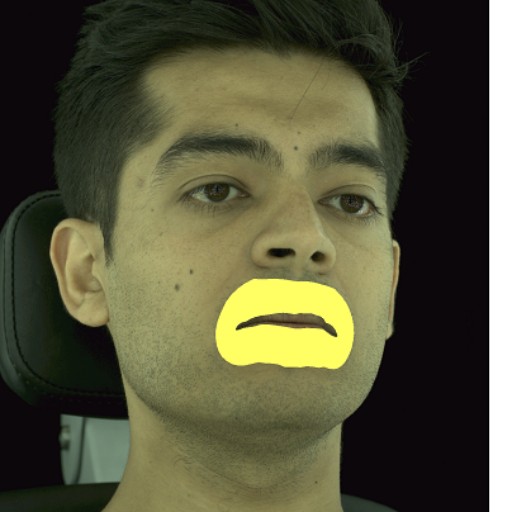}
  \caption{mouth}
  \end{center}
\end{subfigure}
\begin{subfigure}[t]{0.32\linewidth}
  \begin{center}
  \includegraphics[width=0.95\linewidth]{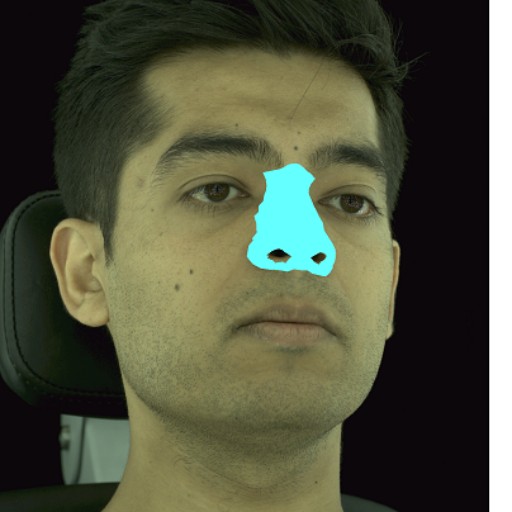}
  \caption{nose}
  \end{center}
\end{subfigure}
\vspace{0.2cm}
\caption{Visualization of the masks used to compute our metrics for the Multiface benchmark. Masks are generated by selecting vertices from the topologically uniform ground truth mesh (a). We select masks for the face (b), ear (c), eye (d), mouth (e) and nose (f) region. }
\label{fig:multiface_masks}
\end{figure}

As mentioned in the paper, we select a subset of 86 sequences of the Mulitface~\cite{multiface} dataset. This subset consists of 10 subjects with 8 or 9 sequences each and a randomly selected camera view. Each sequence consists of one facial performance that is approximately 2 to 4 seconds in length. We select a diverse set of facial performances, including extreme ones (scream, cheeks blowing) and more common ones (speaking, blinking). The camera view is constrained to face the subject with a maximum horizontal viewing angle of 60\degree and a maximum vertical viewing angle of 35\degree. Example sequences for each subject are shown in \cref{fig:multiface_sequences}. In the Multiface dataset, each frame of every sequence is annotated with a topologically uniform ground truth mesh. We use this mesh to compute the ground truth optical flow for the screen space motion error, and the chamfer distance. We also generate the semantic masks using this ground truth mesh by selecting corresponding vertices as shown in \cref{fig:multiface_masks}. 

\newcommand{\multiimg}[1]{\includegraphics[width=1.2cm]{figures/multiface_supp/sequences/#1}}

\begin{figure}[th]
\tablefont
\setlength{\tabcolsep}{0pt}
\renewcommand{\arraystretch}{0.5}
\centering
\begin{tabular}{c c c c c c c}
\multiimg{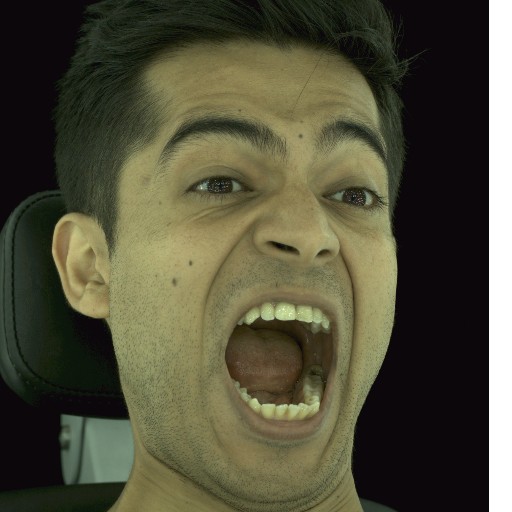} & \multiimg{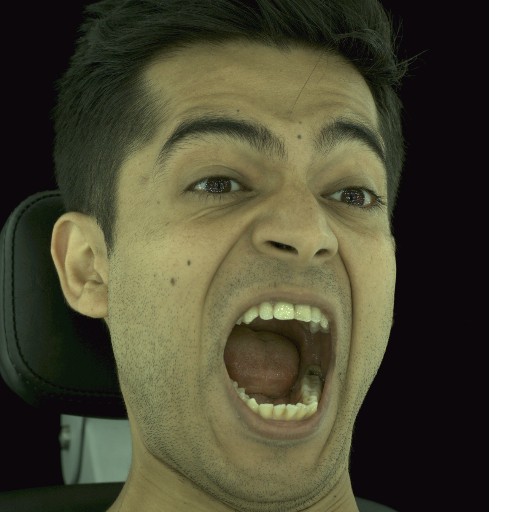} & \multiimg{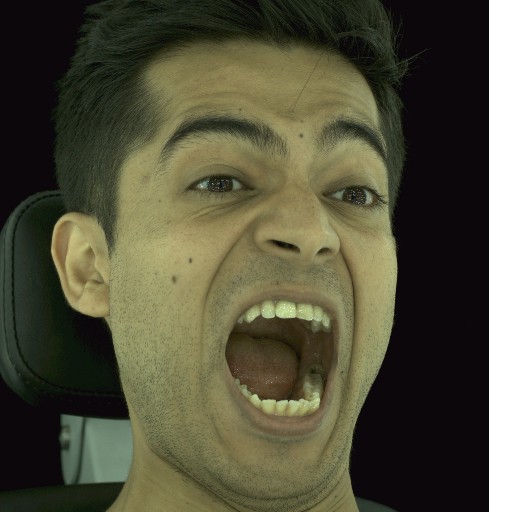} & \multiimg{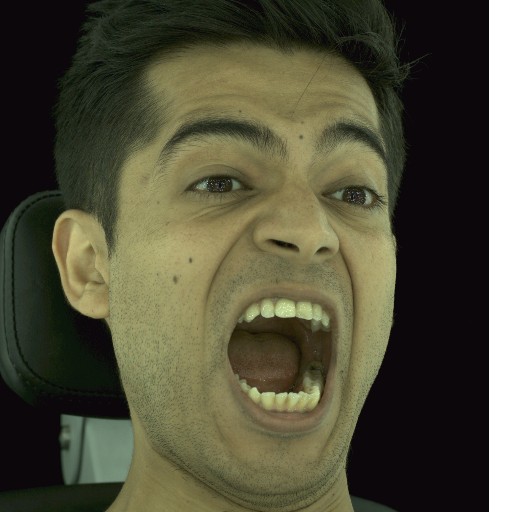} & \multiimg{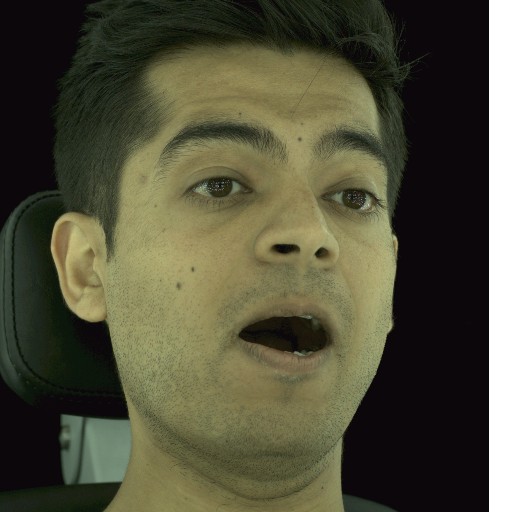} & \multiimg{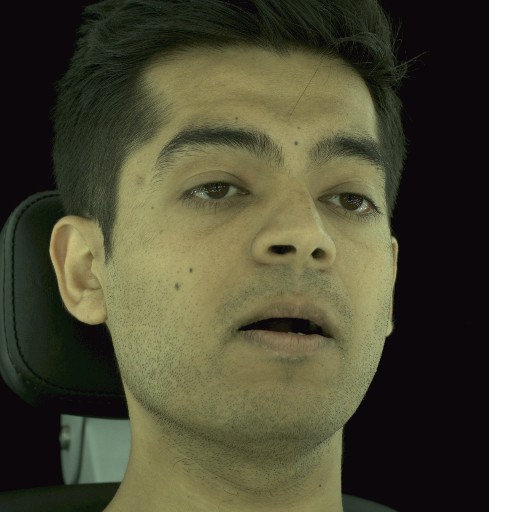} & \multiimg{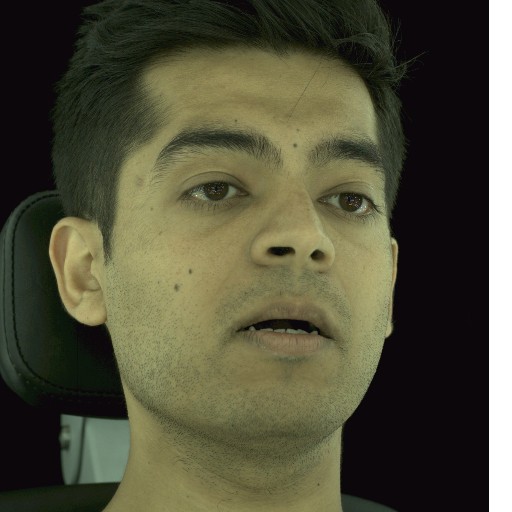} \\
\multiimg{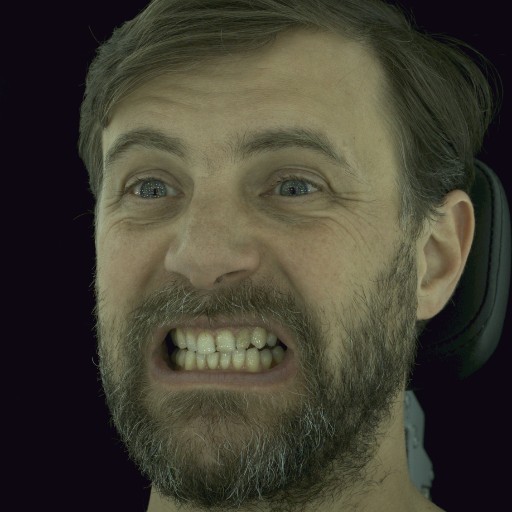} & \multiimg{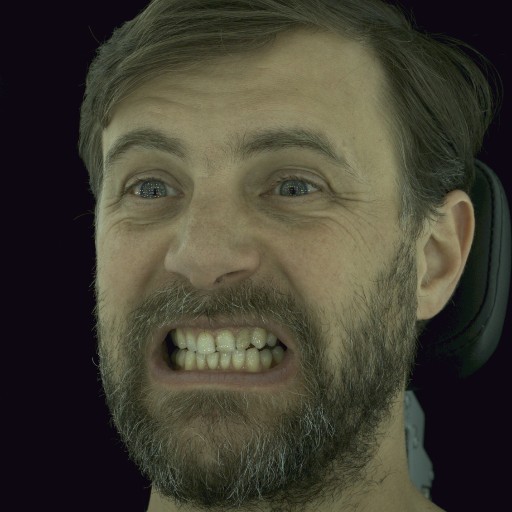} & \multiimg{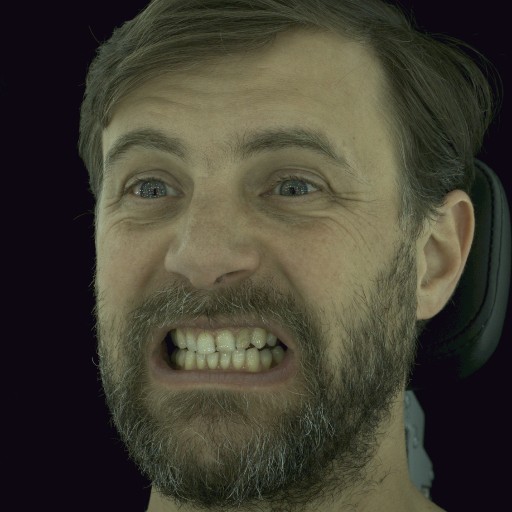} & \multiimg{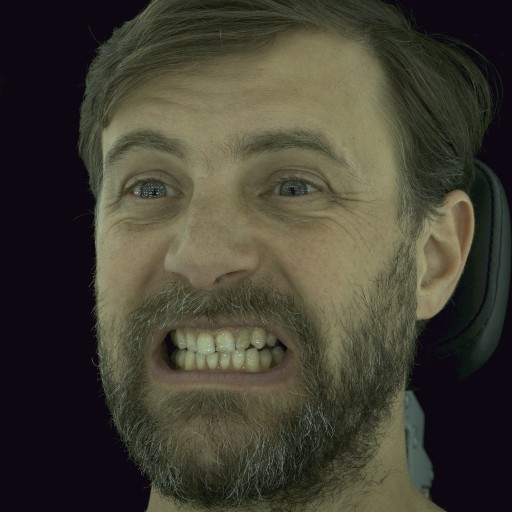} & \multiimg{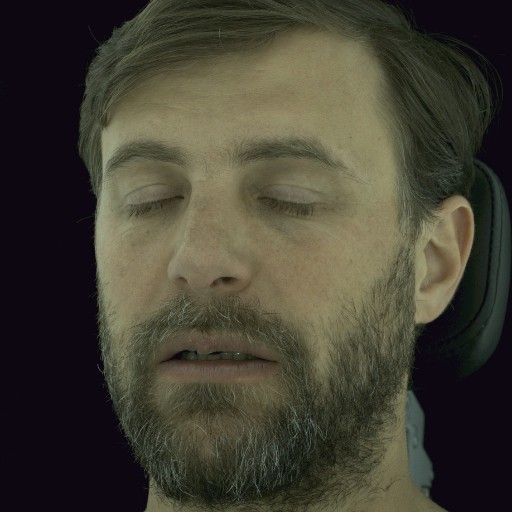} & \multiimg{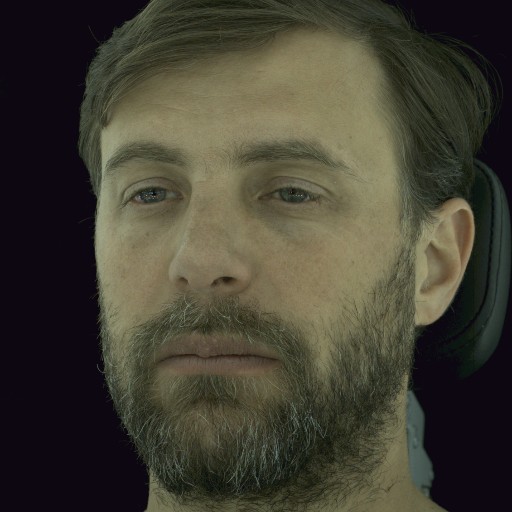} & \multiimg{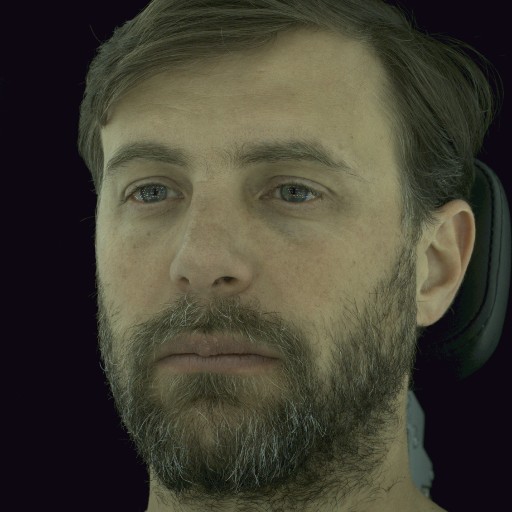} \\
\multiimg{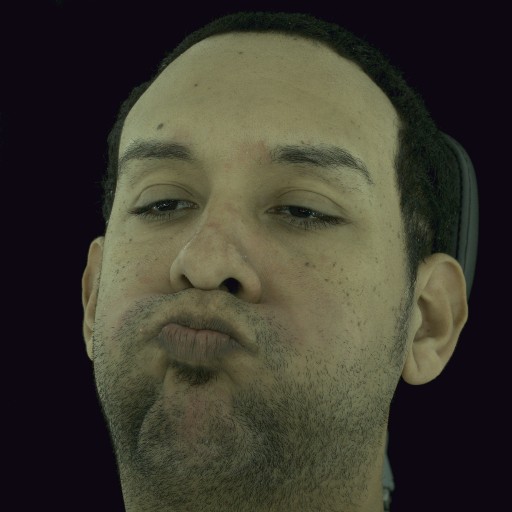} & \multiimg{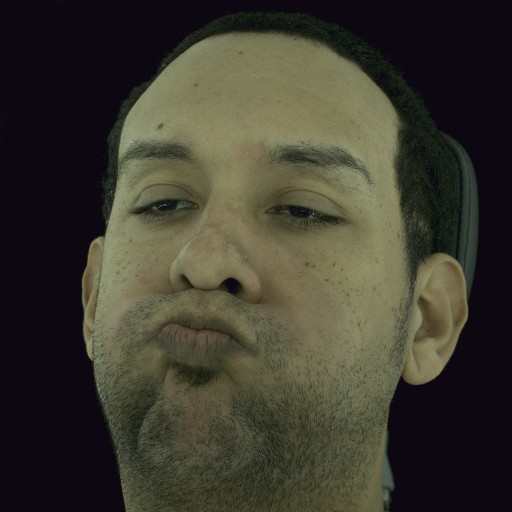} & \multiimg{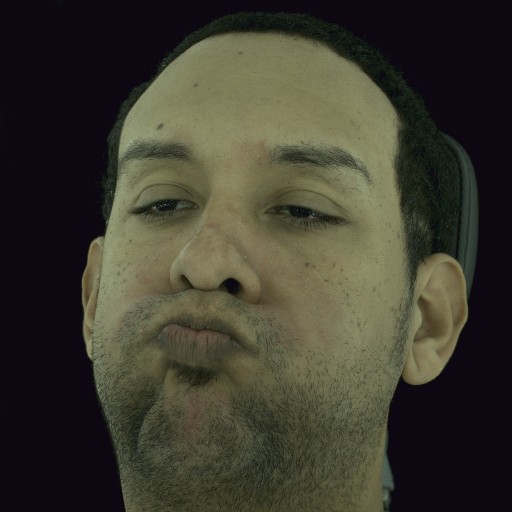} & \multiimg{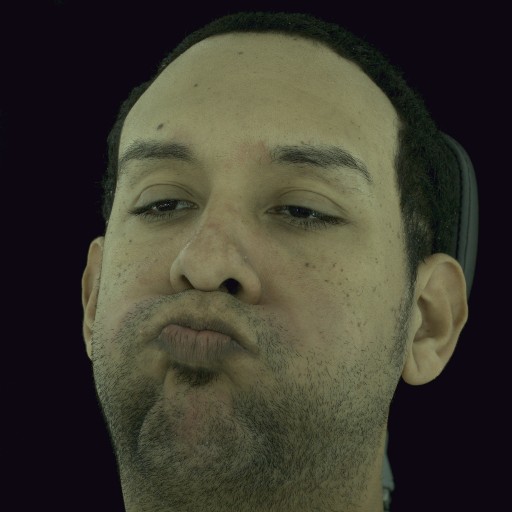} & \multiimg{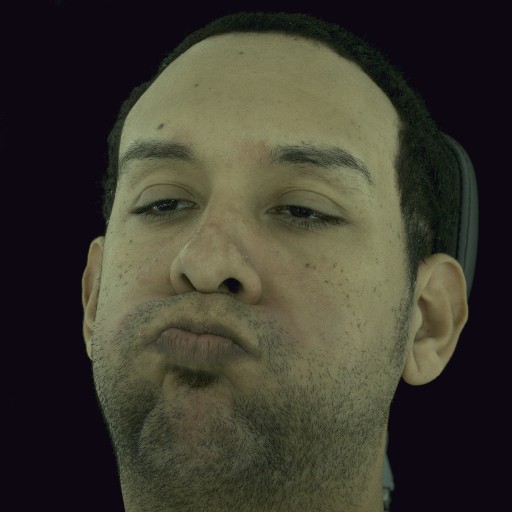} & \multiimg{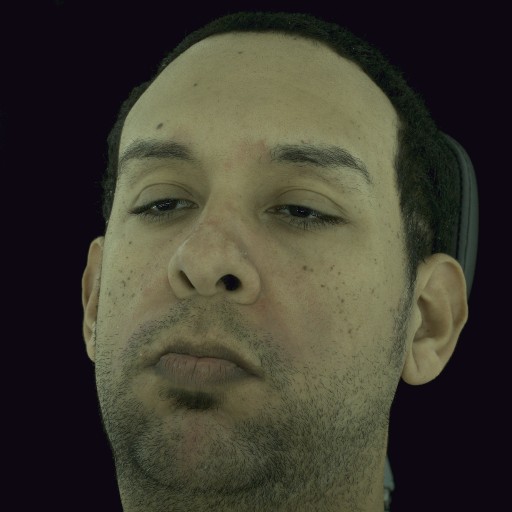} & \multiimg{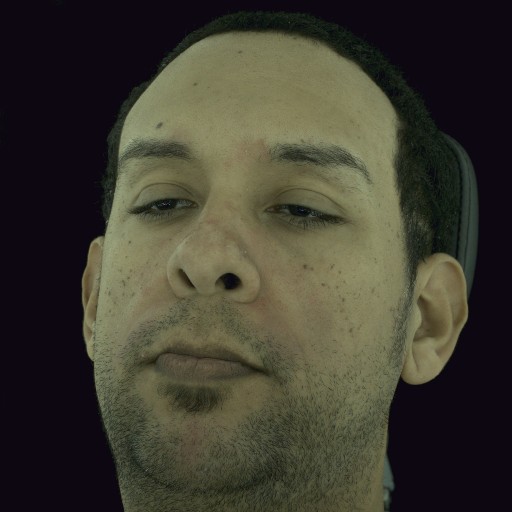} \\
\multiimg{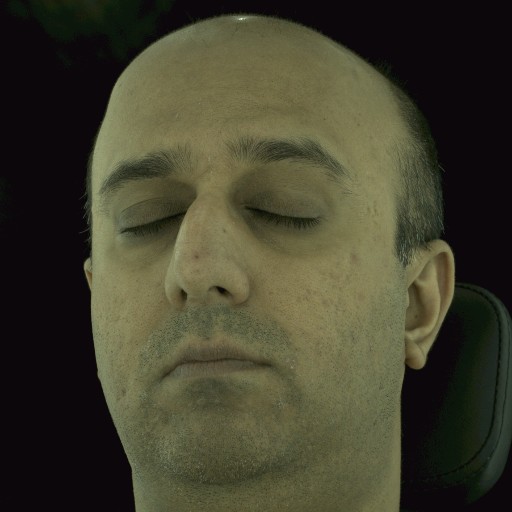} & \multiimg{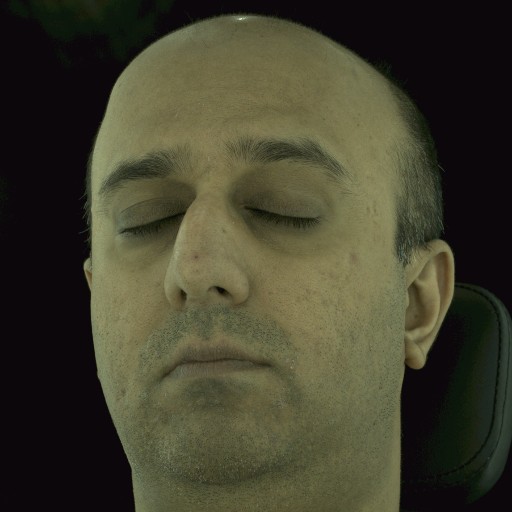} & \multiimg{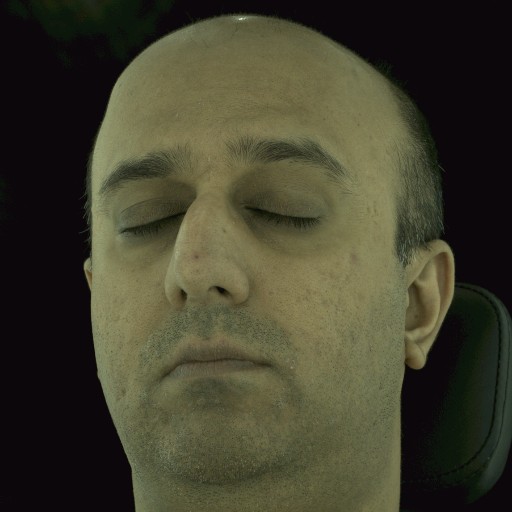} & \multiimg{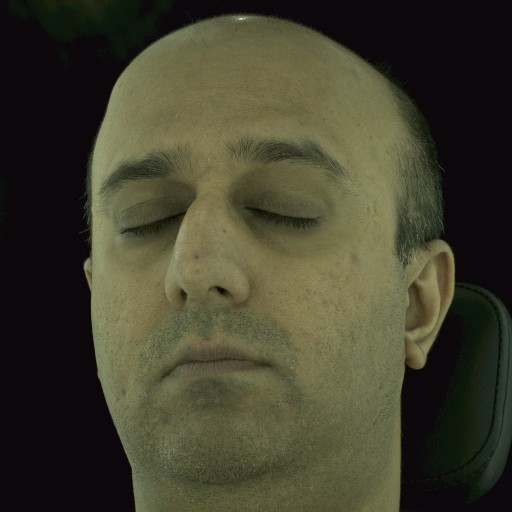} & \multiimg{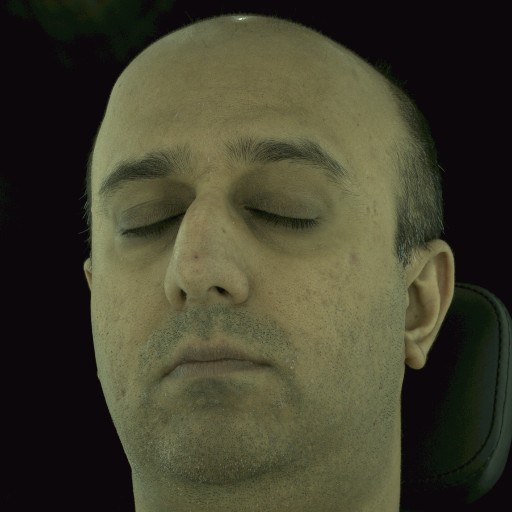} & \multiimg{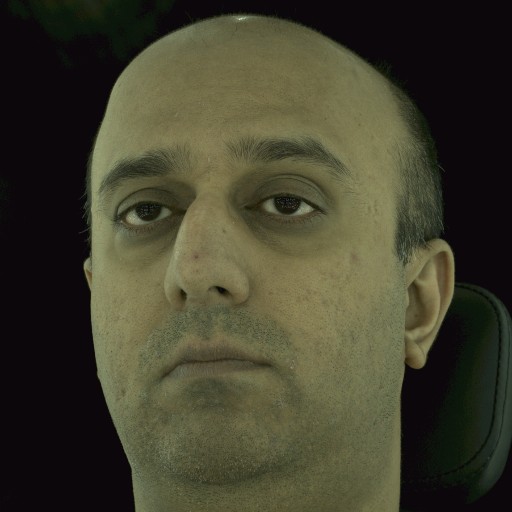} & \multiimg{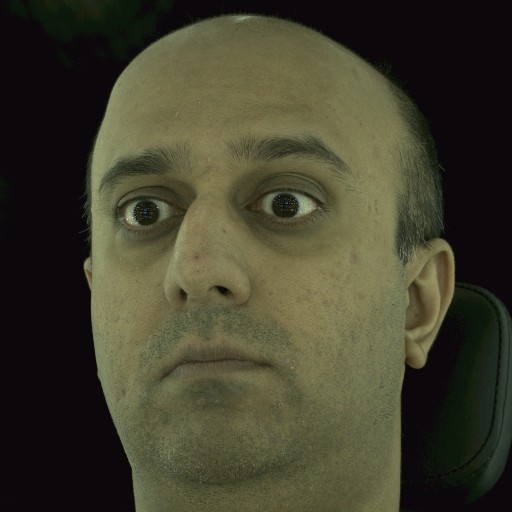}  \\
\multiimg{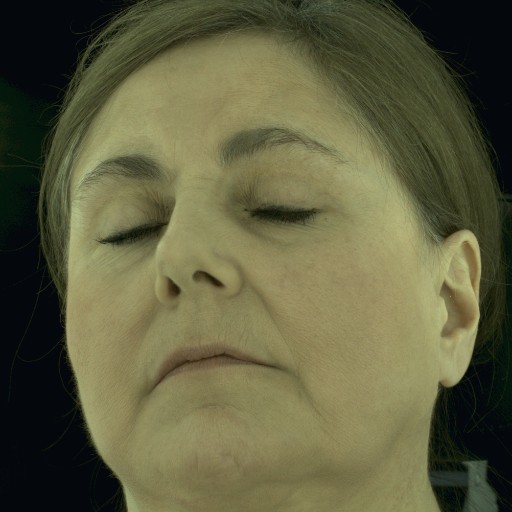} & \multiimg{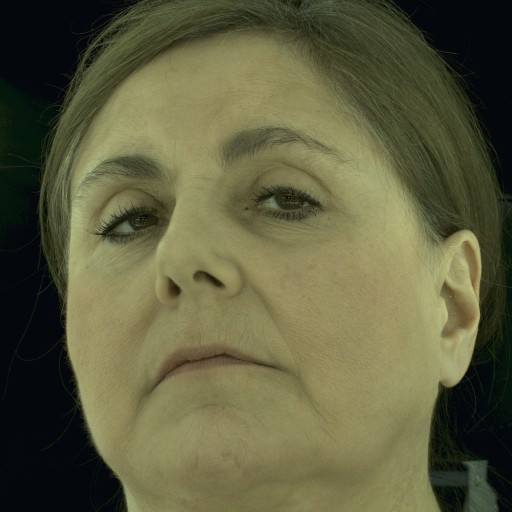} & \multiimg{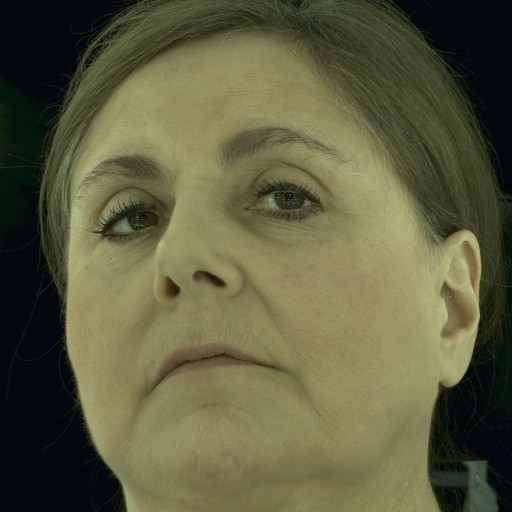} & \multiimg{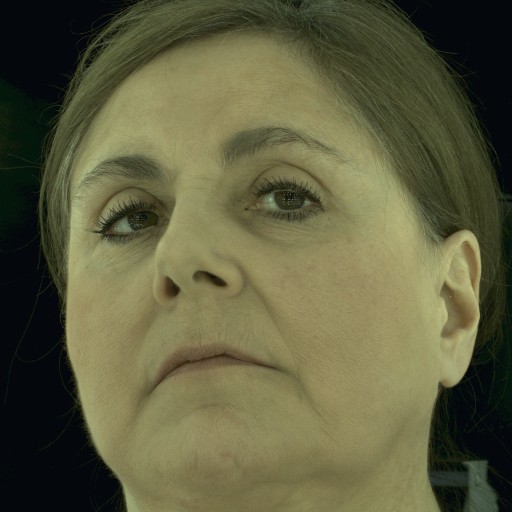} & \multiimg{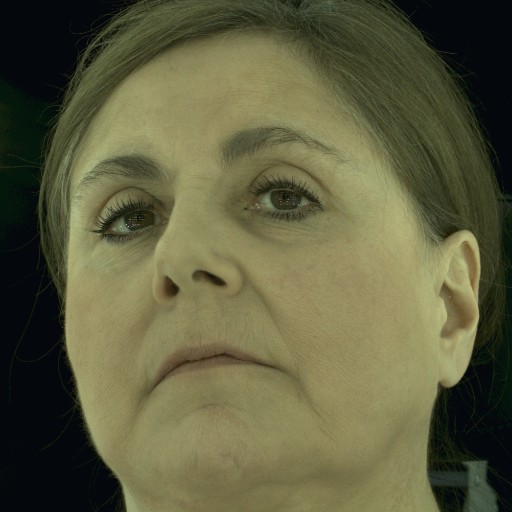} & \multiimg{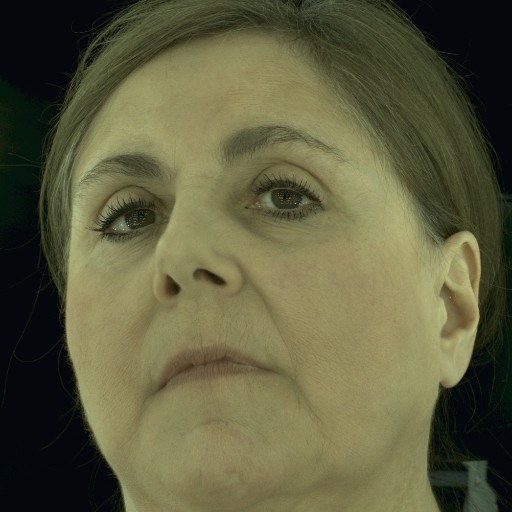} & \multiimg{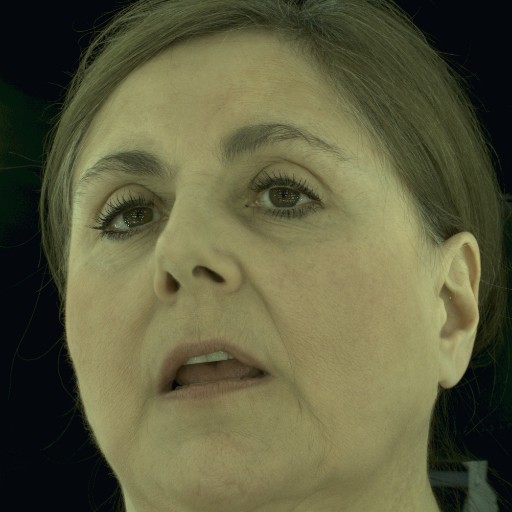} \\
\multiimg{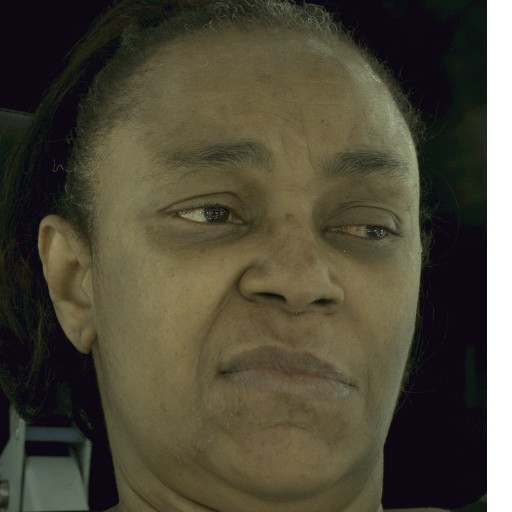} & \multiimg{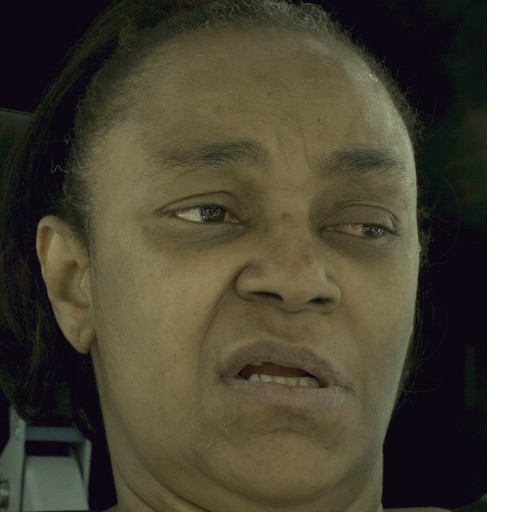} & \multiimg{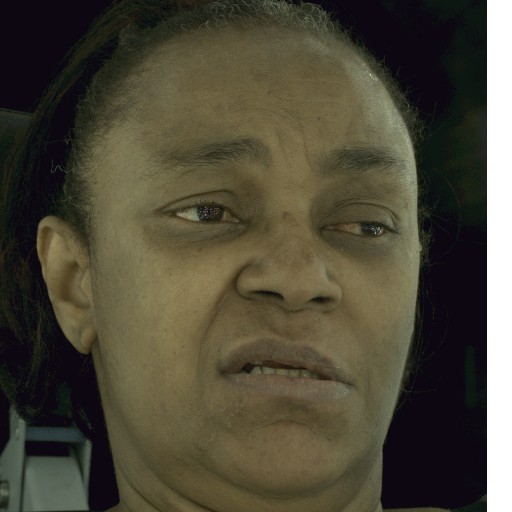} & \multiimg{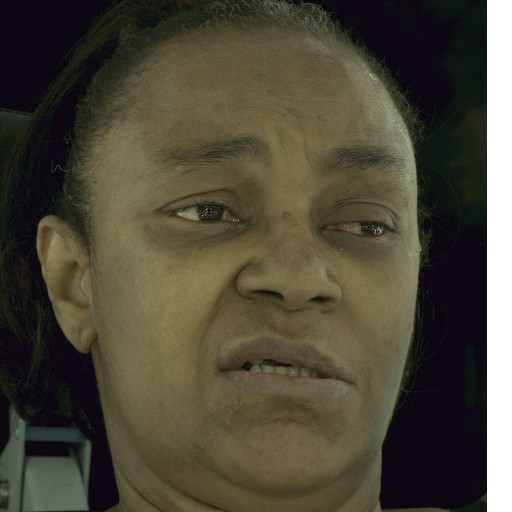} & \multiimg{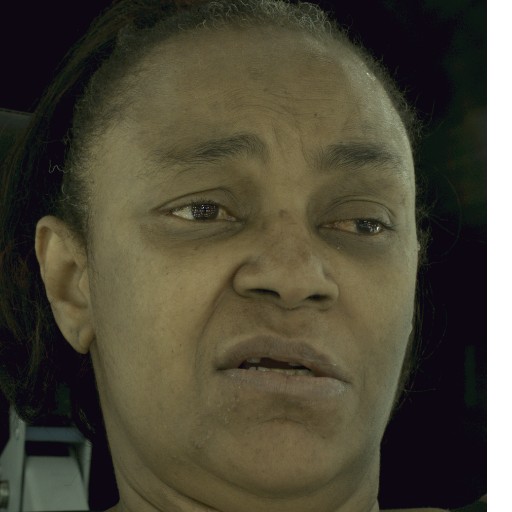} & \multiimg{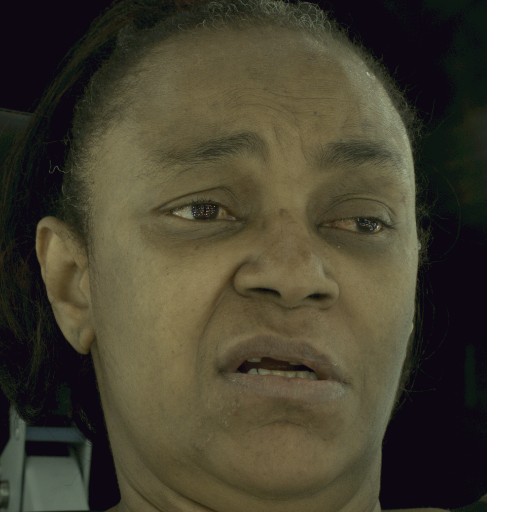} & \multiimg{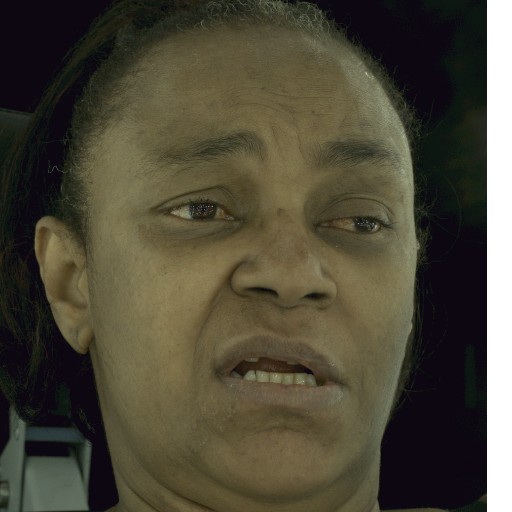} \\
\multiimg{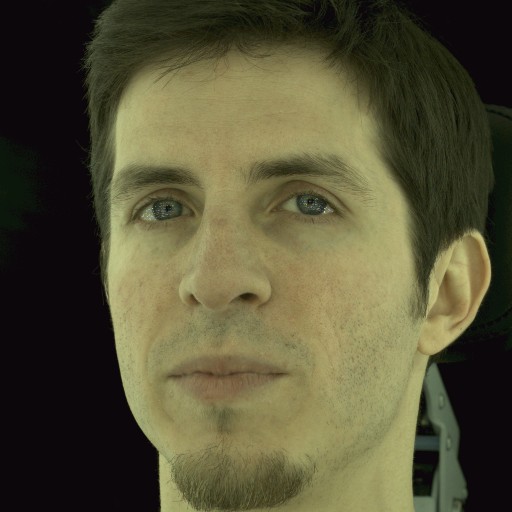}  & \multiimg{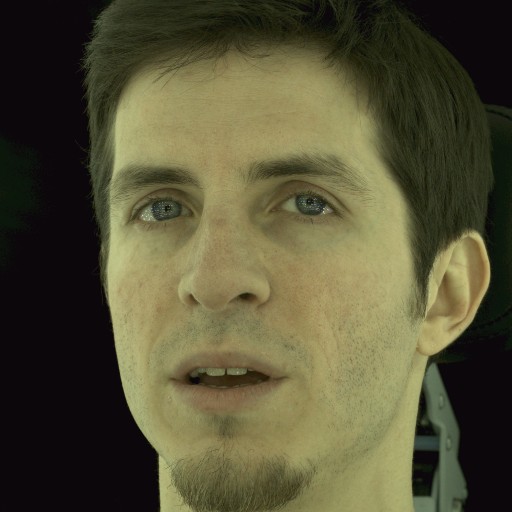} & \multiimg{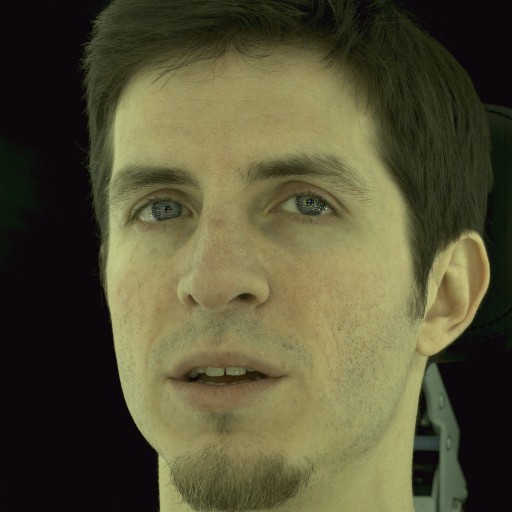} & \multiimg{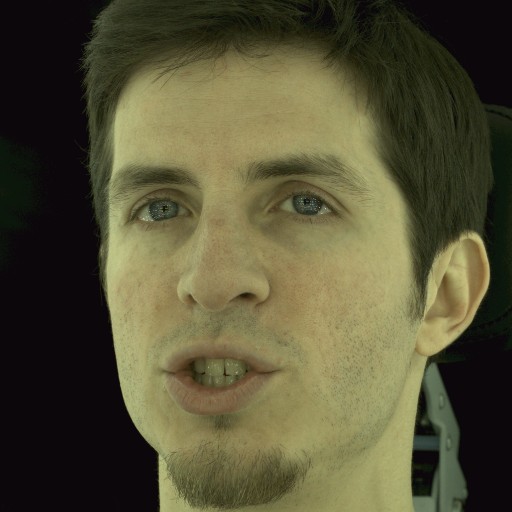} & \multiimg{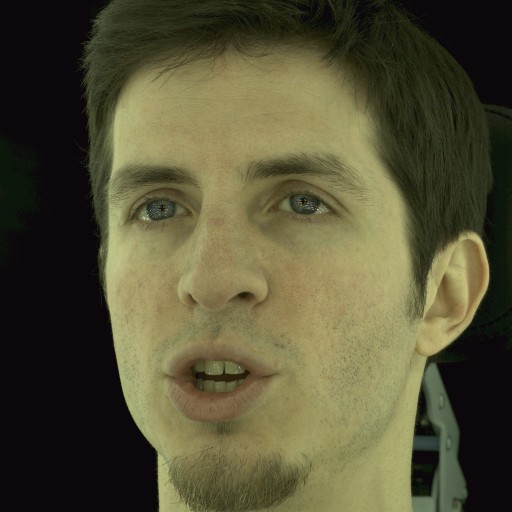} & \multiimg{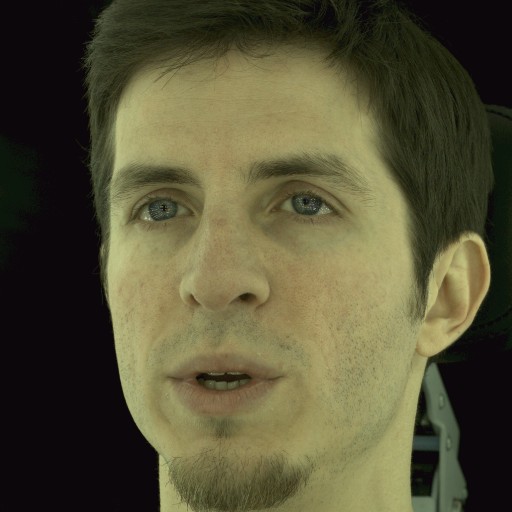} & \multiimg{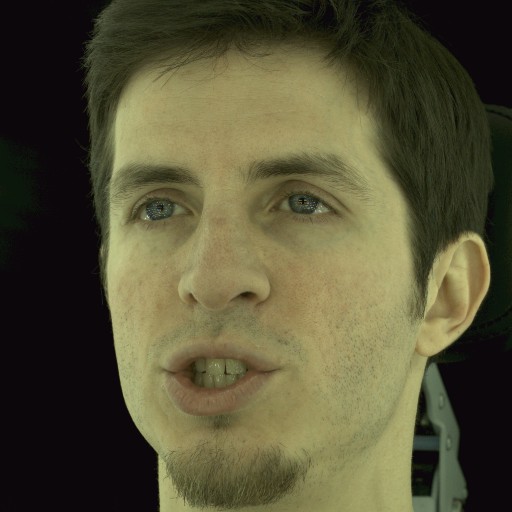} \\
\multiimg{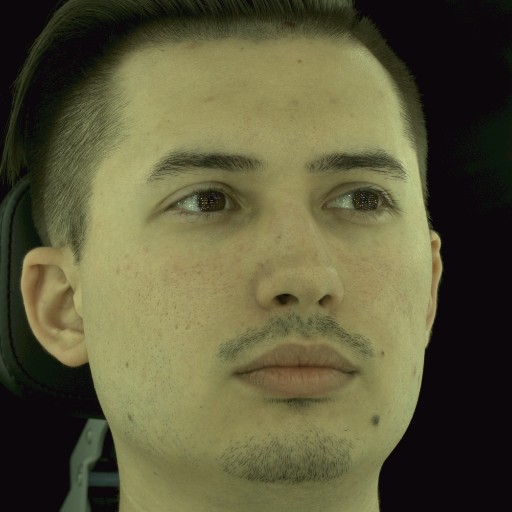} & \multiimg{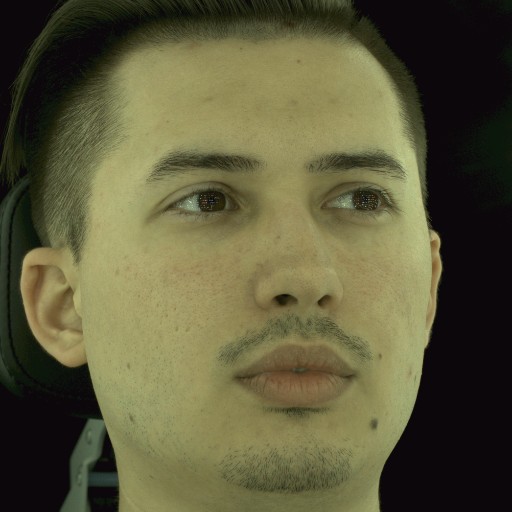} & \multiimg{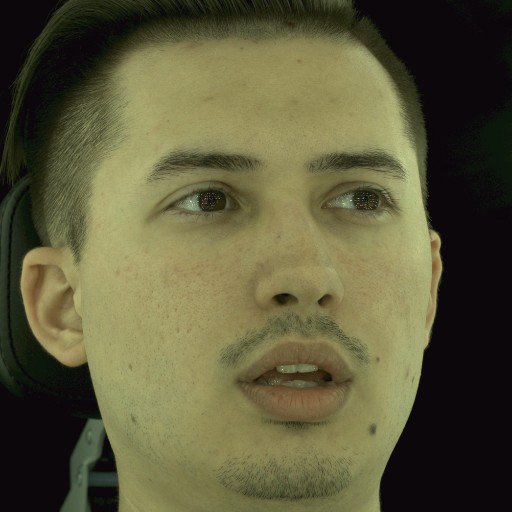} & \multiimg{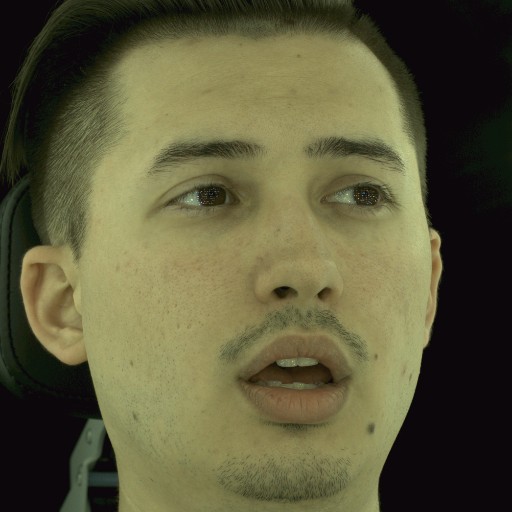} & \multiimg{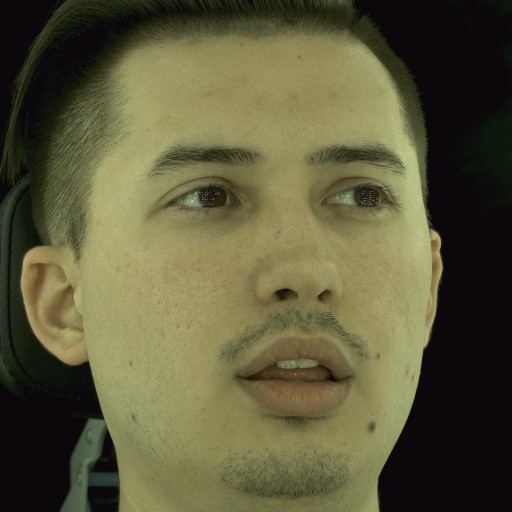} & \multiimg{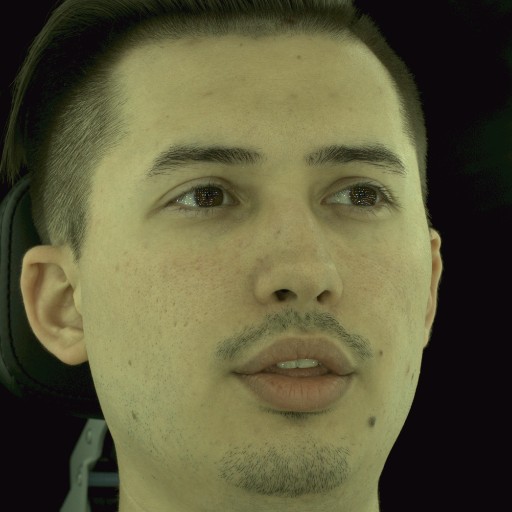} & \multiimg{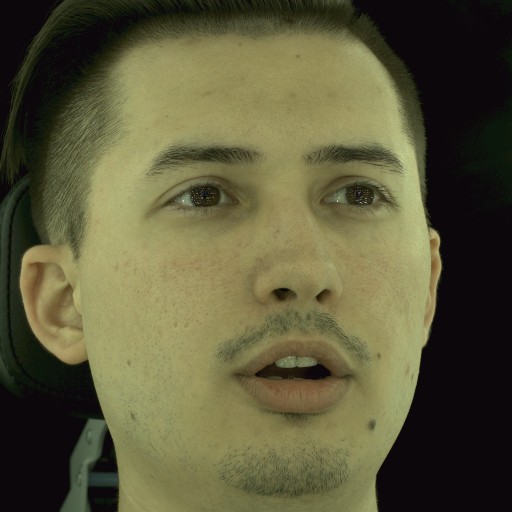} \\
\multiimg{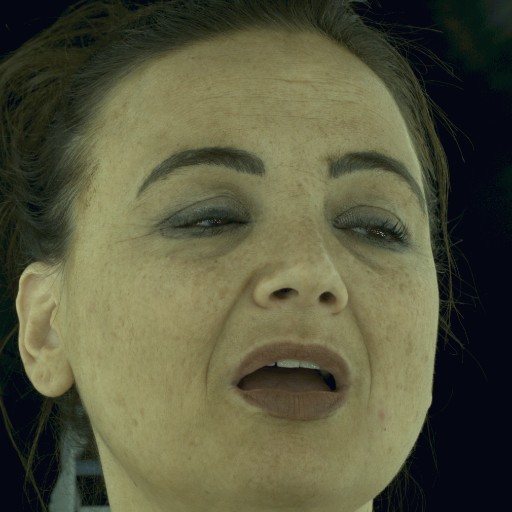} & \multiimg{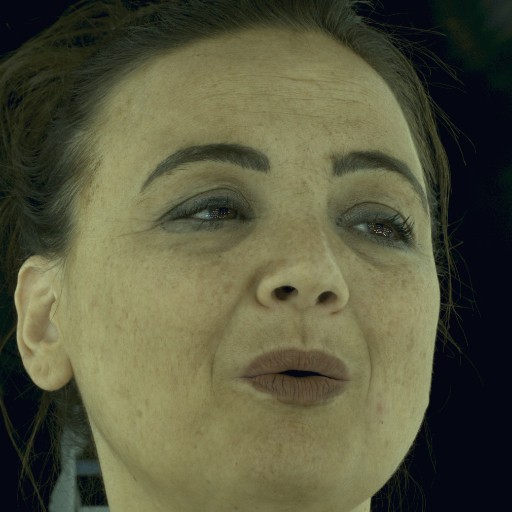} & \multiimg{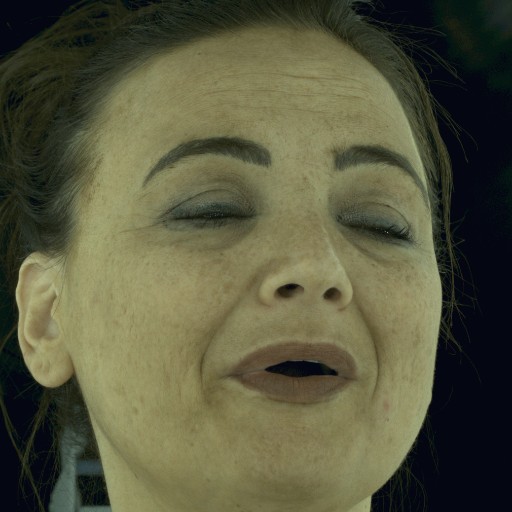} & \multiimg{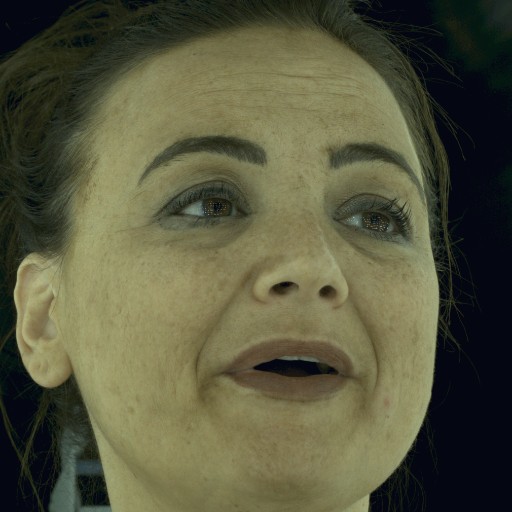} & \multiimg{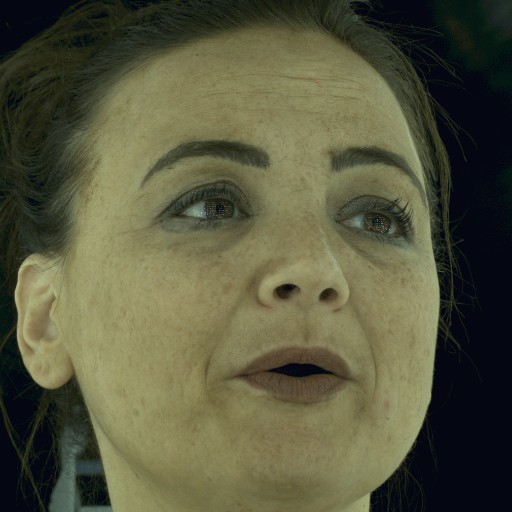} & \multiimg{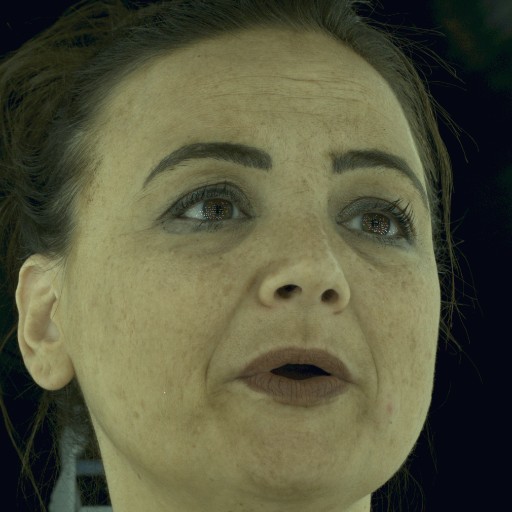} & \multiimg{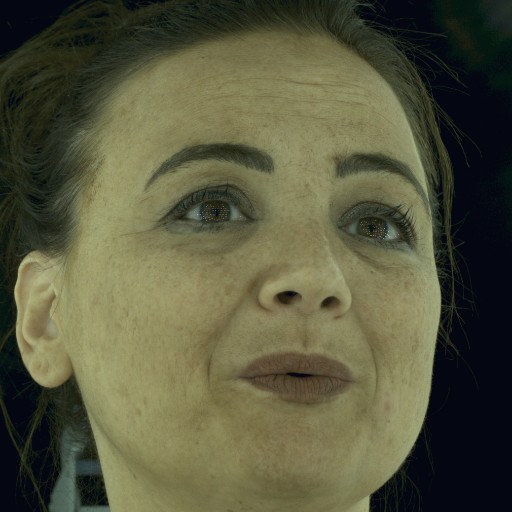} \\
\multiimg{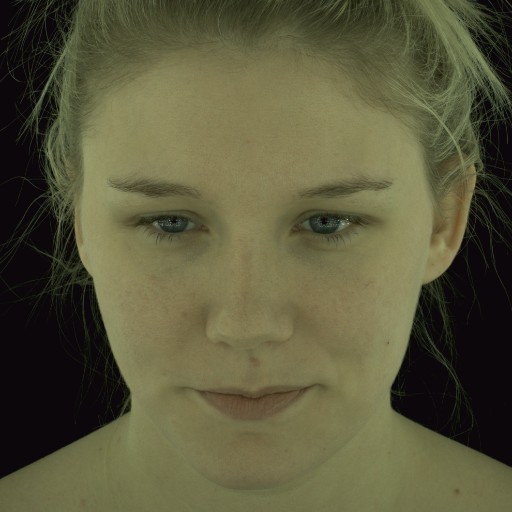}  & \multiimg{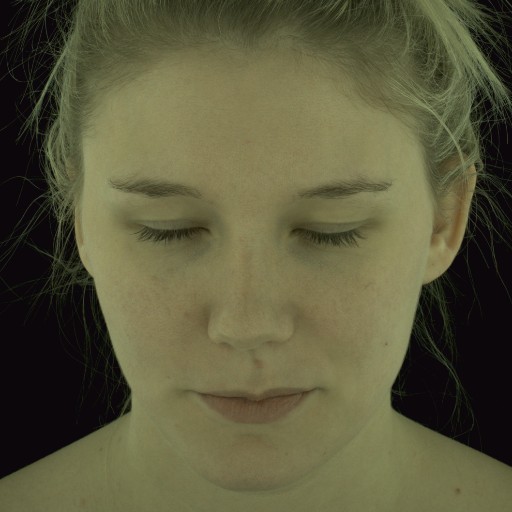} & \multiimg{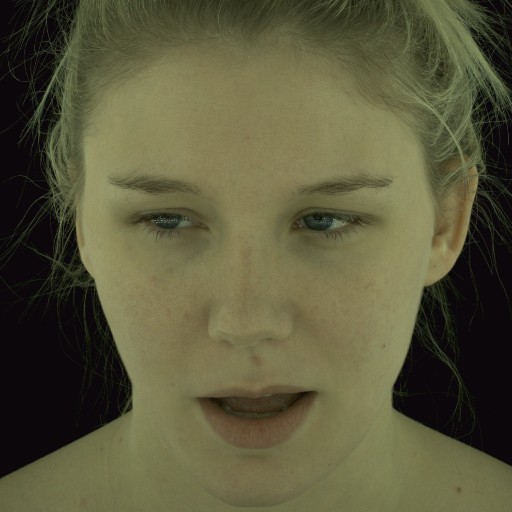} & \multiimg{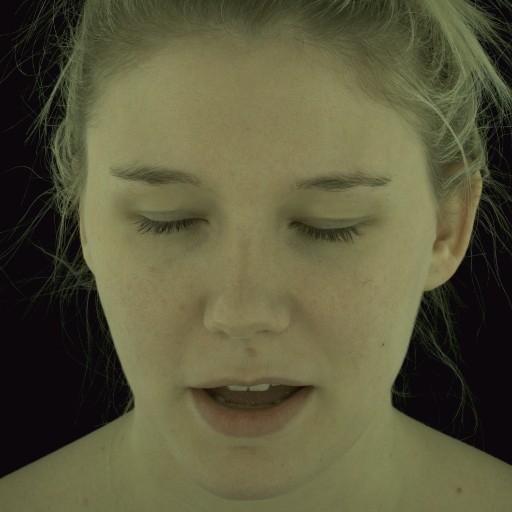} & \multiimg{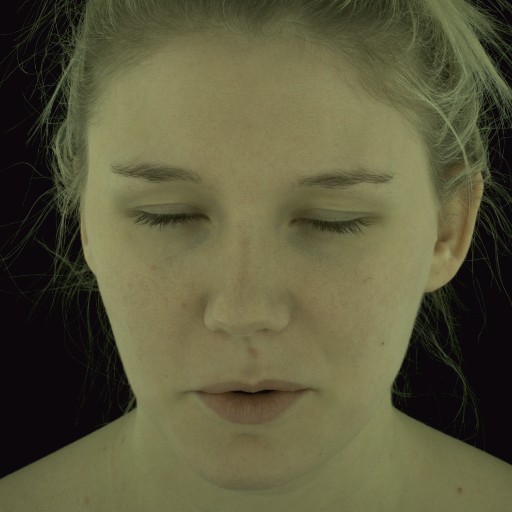} & \multiimg{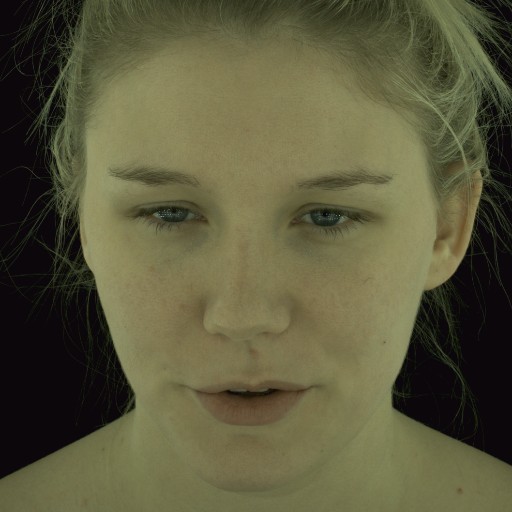} & \multiimg{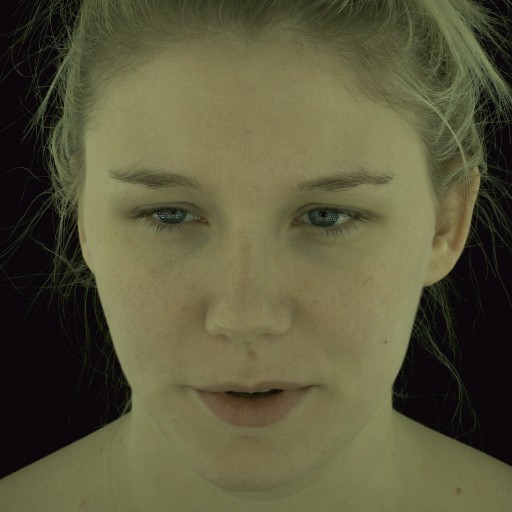} \\
\end{tabular}
\vspace{0.2cm}
\caption{Extracts from one sequence for each subject of our Multiface~\cite{multiface} subset. Our benchmark contains a variety of expressions from diverse subjects and view directions. }
\label{fig:multiface_sequences}
\end{figure}

\section{Datasets and Training}
\label{supp:training}

As previously mentioned, we use the FaceScape~\cite{facescape}, Stirling~\cite{stirling} and FaMoS~\cite{tempeh} dataset to train our 2D alignment module. 

The FaceScape dataset contains 20 expressions performed by 360 subjects with a very large number of calibrated camera views (more than 40) and 3D scans obtained using photogrammetry. To train the network to be robust to large view-deviations, we select views with up to a 90\degree  horizontal and 45\degree  vertical deviation from frontal view of the face. 

The Stirling dataset contains textured 3D scans of 8 expressions performed by 140 subjects. These scans are generated by a calibrated stereo camera setup. We use the two views from the stereo camera, and generate 30 additional synthetic views. These views are generated with random focal lengths and random view directions. As in the FaceScape dataset, these view deviations are as high as 90\degree horizontally and 45\degree vertically.

The FaMoS dataset contains 95 subjects with 28 motion sequences each. It comes with high-quality FLAME registrations generated with the help of facial markers. It contains 6 RGB camera views, of which we use the forward facing ones. 
To balance this dataset, we keep only every $10^\text{th}$ frame. 

\subsection{Scan registration}

\begin{figure*}[ht]
\centering
\includegraphics[width=0.95\textwidth]{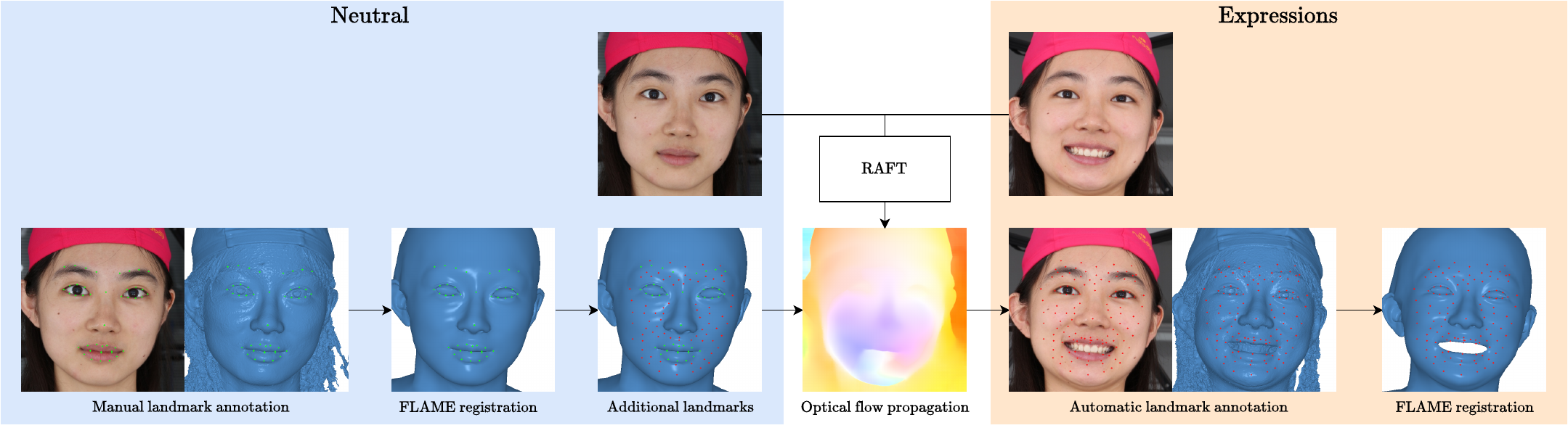}
\vspace{0.2cm}
\caption{An overview of our scan annotation process. First, 44 landmarks (marked in green) are manually annotated for the neutral scans of each subject. The FLAME topology mesh is then fitted onto this scan. For each expression, landmarks pre-selected on the topology mesh (marked in red) are projected into screen space and propagated using optical flow. With these propagated landmarks, the topology mesh is fitted onto the expression scans. This optical flow assisted registration pipeline ensures accurate skin deformations tangential to the scan surface. }
\label{fig:fscape_annotation}
\end{figure*}

Since FLAME~\cite{FLAME} mesh registrations are not available for the FaceScape and Stirling datasets, we generate them using a semi-automatic annotation process to ensure high accuracy and consistency. For each subject in the datasets, we do the following: First, we manually annotate 44 landmarks (eyebrows, eyes, nose and lips) of the neutral scans of each subject. We then use commercial software to fit the FLAME topology mesh onto this scan with these landmarks as guidance. After the registration of the neutral mesh, we append landmarks pre-selected on the topology mesh to the manually annotated landmarks. We also compute the optical flow between the frontal view of the neutral face and each expression using the original RAFT~\cite{raft_optical_flow} model. The manually selected and automatically added landmarks are then propagated to the expression images using this optical flow. After manual correction on propagation failures, these landmarks are used to fit the topology mesh onto the expression scans. Using optical flow to propagate the landmarks ensures that the skin deformation along the surface tangent is precisely tracked across the scans. This in turn enables our network to accurately predict skin deformations. See \cref{fig:fscape_annotation} for an overview of this annotation process, and \cref{fig:registration_vis} for example registration results.

\newcommand{\dataimg}[1]{\includegraphics[width=2.3cm]{figures/dataset/#1}}

\begin{figure*}
\tablefont
\setlength{\tabcolsep}{1pt}
\renewcommand{\arraystretch}{0}
\centering
\begin{tabular}{c c c c c c }
\dataimg{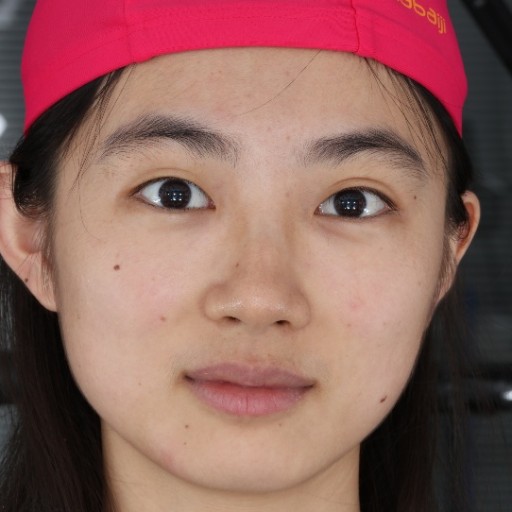} & \dataimg{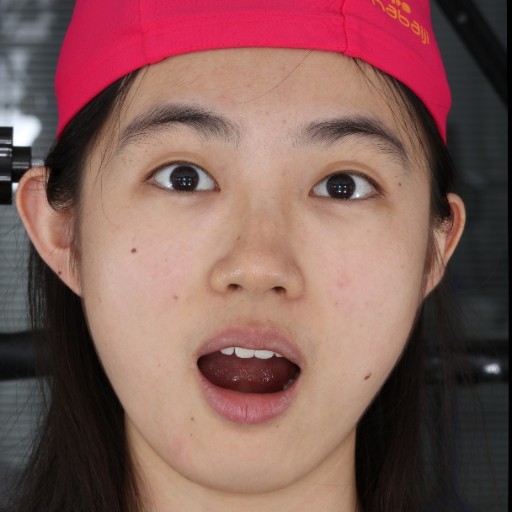} & \dataimg{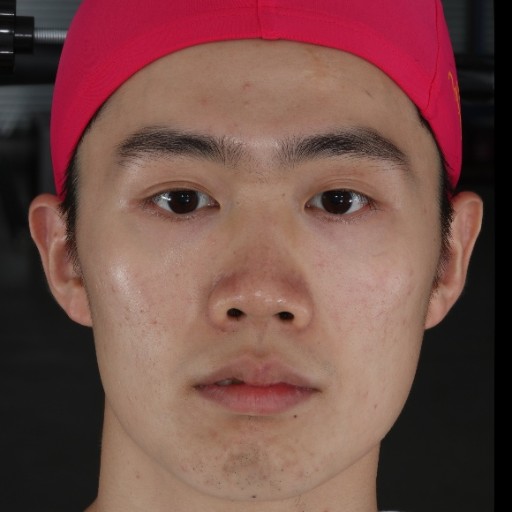} & \dataimg{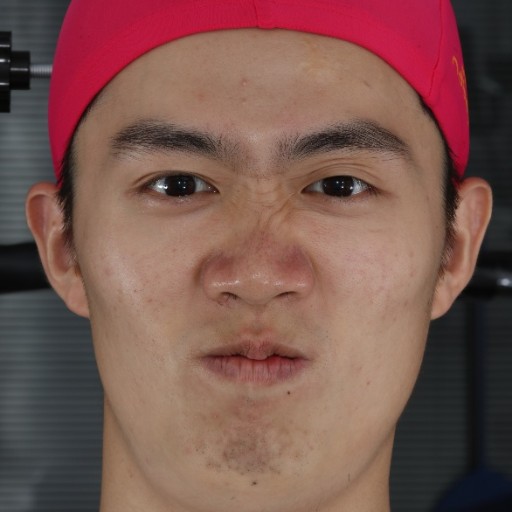} & \dataimg{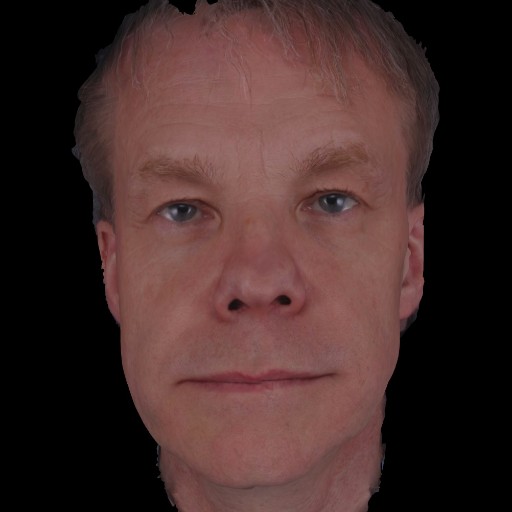} & \dataimg{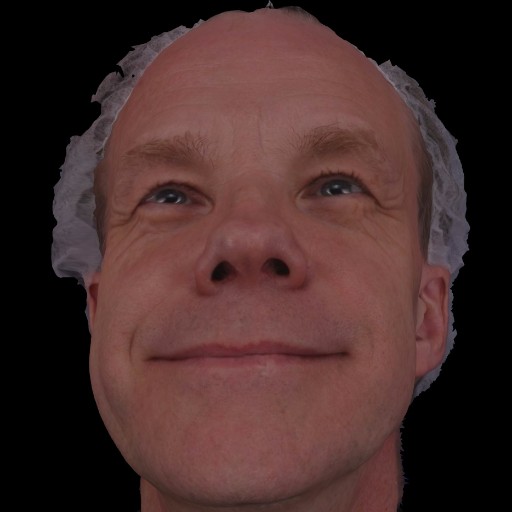} \\

\dataimg{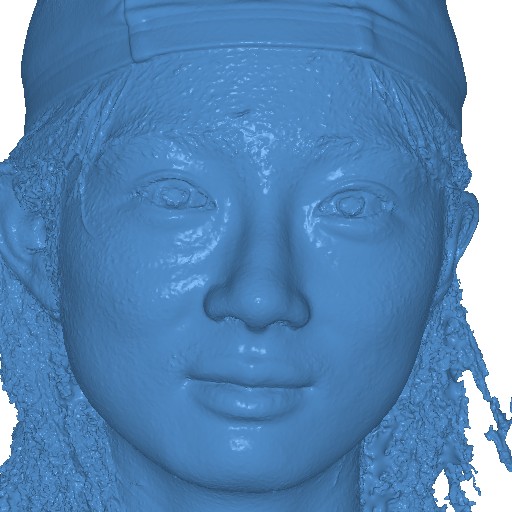} & \dataimg{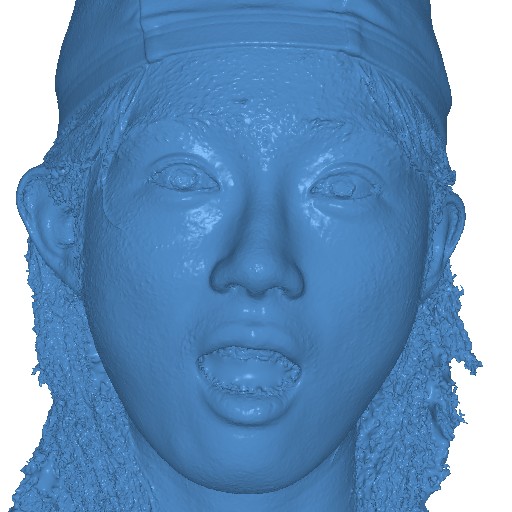} & \dataimg{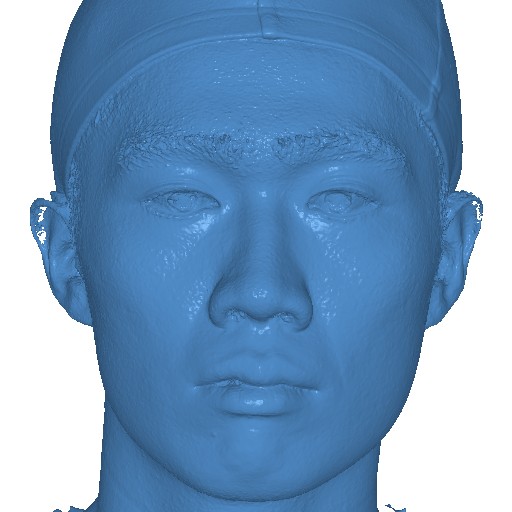} & \dataimg{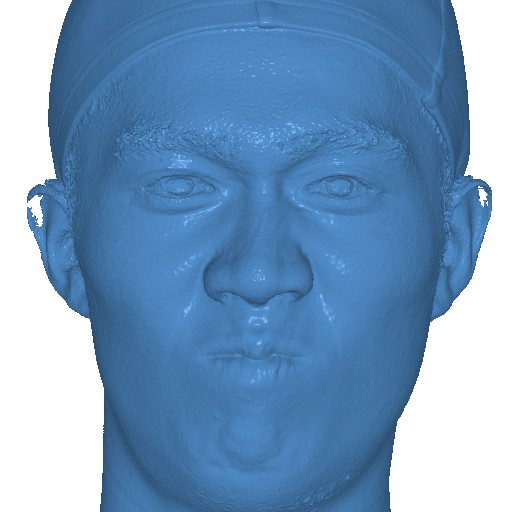} & \dataimg{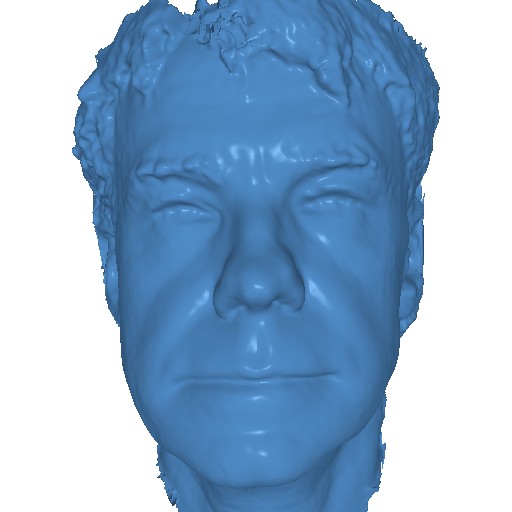} & \dataimg{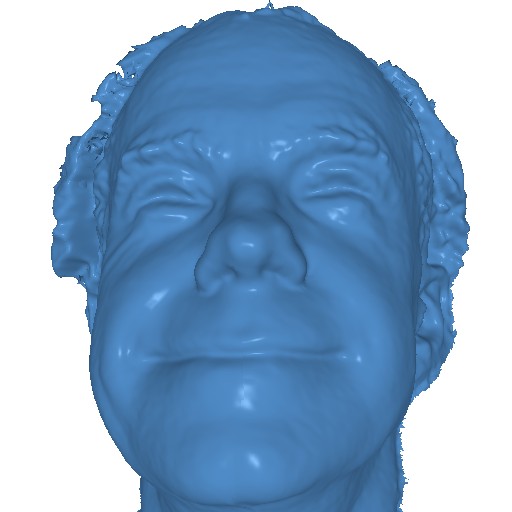} \\

\dataimg{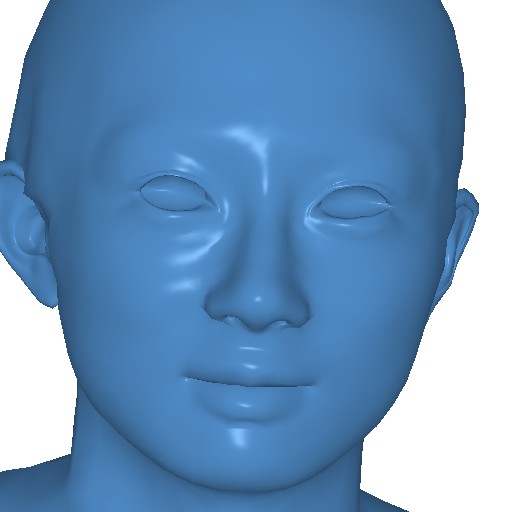} & \dataimg{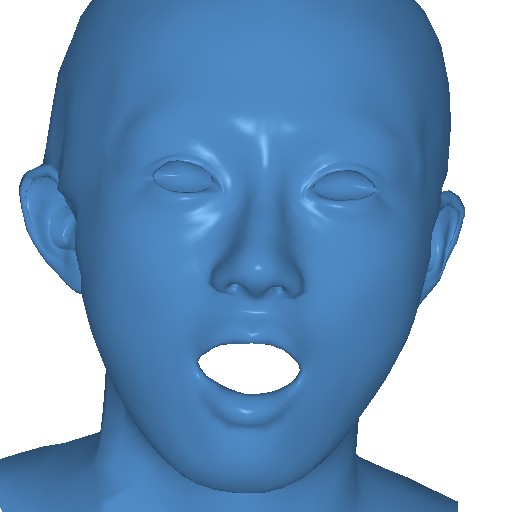} & \dataimg{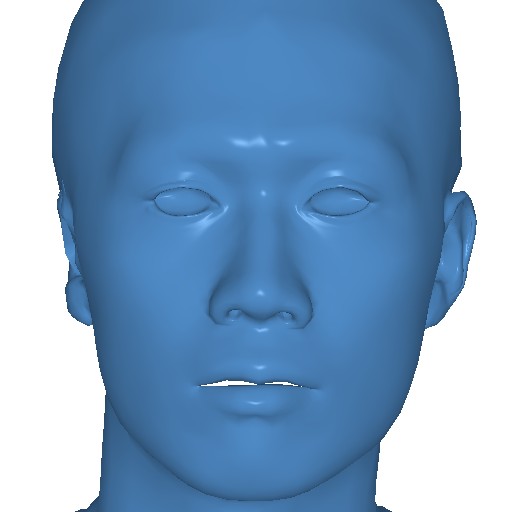} & 
\dataimg{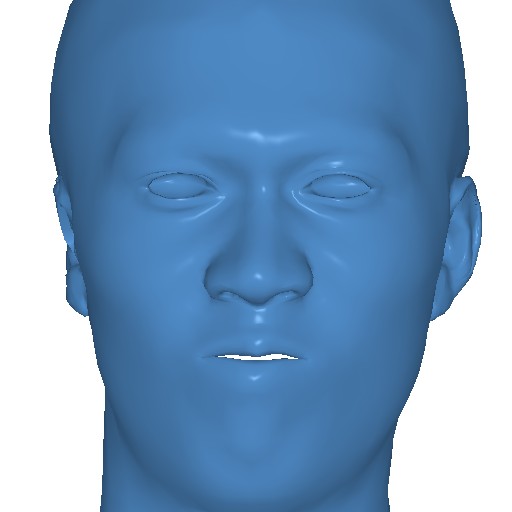} & \dataimg{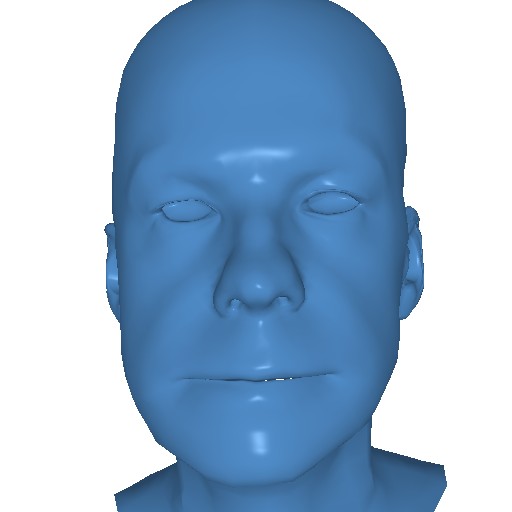} & \dataimg{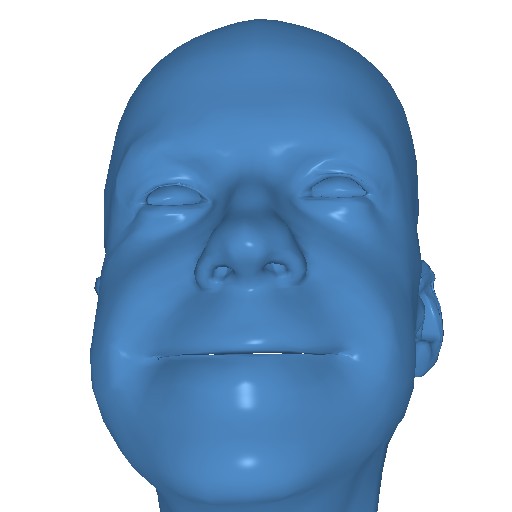} \\
\end{tabular}
\vspace{0.3cm}
\caption{Example FLAME~\cite{FLAME} registrations from the FaceScape~\cite{facescape} (four columns on the left) and Stirling~\cite{stirling} (two columns on the right) dataset. Top row contains the ground truth images, middle row contains ground truth scans and bottom row contains the fitted FLAME meshes.  For the Stirling dataset, we generate synthetic views using the available colored 3D scans. }
\label{fig:registration_vis}
\end{figure*}

\subsection{Data augmentation}

\begin{figure}
\centering
\begin{subfigure}[t]{0.24\linewidth}
  \begin{center}
  \includegraphics[width=0.95\linewidth]{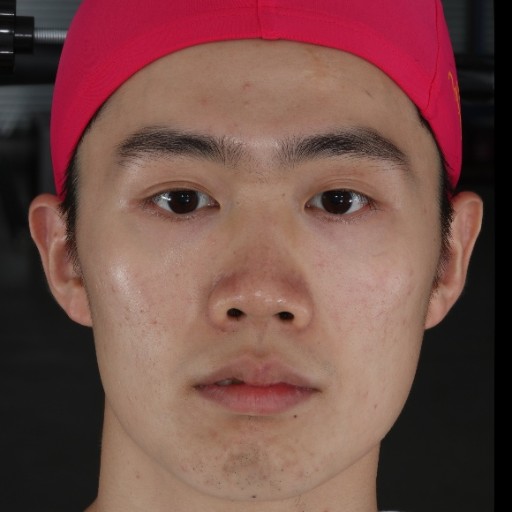}
  \caption{}
  \label{fig:aug_orig}
  \end{center}
\end{subfigure}%
\begin{subfigure}[t]{0.24\linewidth}
  \begin{center}
  \includegraphics[width=0.95\linewidth]{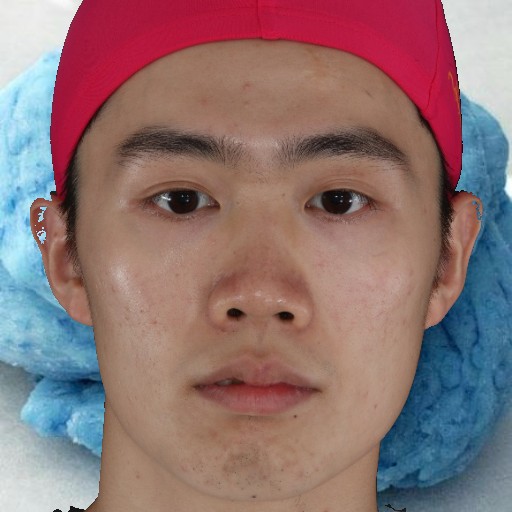}
  \caption{}
  \label{fig:aug_bg}
  \end{center}
\end{subfigure}
\begin{subfigure}[t]{0.24\linewidth}
  \begin{center}
  \includegraphics[width=0.95\linewidth]{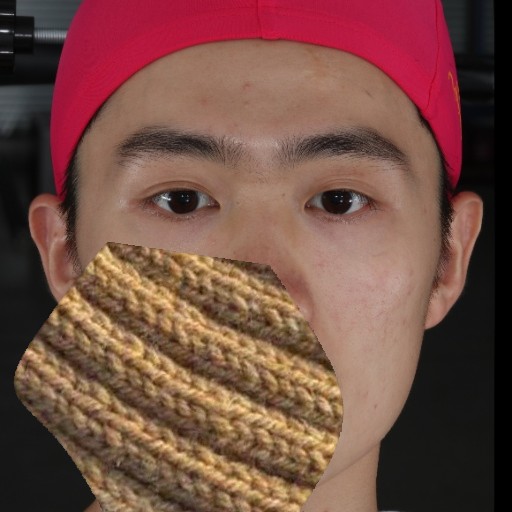}
  \caption{}
  \label{fig:aug_occ}
  \end{center}
\end{subfigure}
\begin{subfigure}[t]{0.24\linewidth}
  \begin{center}
  \includegraphics[width=0.95\linewidth]{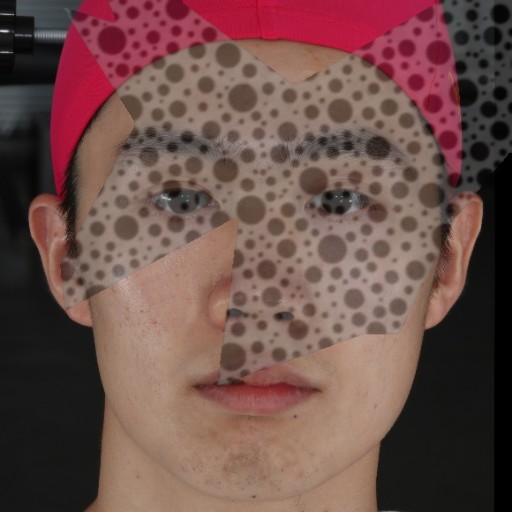}
  \caption{ }
  \label{fig:aug_trans_occ}
  \end{center}
\end{subfigure}
\vspace{0.3cm}
\caption{Examples of our data augmentation: random background (b), random occlusions (c) and random semi-transparent occlusions (d). The original image is shown in (a).}
\label{fig:augmentation}
\end{figure}

All of the above mentioned datasets contain only images captured in controlled, occlusion-free environments. Subjects are wearing hair caps, special lighting ensure uniform illumination and the background is dark and clutter-free. To make our model more robust to outdoor environments and occlusions due to hair, glasses, etc., we deploy three types of data-augmentation (see \cref{fig:augmentation}). First, we use common image-based augmentation techniques such as Gaussian noise, color shift, gray-scale, random rotations, translations and scale. Second, we deploy background augmentation. This is done by replacing the background of the ground truth image (computed using the ground truth scan mesh) with randomly selected images from the Describable Texture Dataset (DTD). Lastly, we include occlusion augmentation using the technique described by \cite{occlusions}. Random masks are generated to partially occlude the face. We extend this technique to also generate semi-transparent occlusions to simulate lighting effects and transparent objects. 

\subsection{Vertex weights}

\begin{figure}[h]
\centering
\begin{subfigure}{0.32\linewidth}
  \begin{center}
  \includegraphics[width=0.95\linewidth]{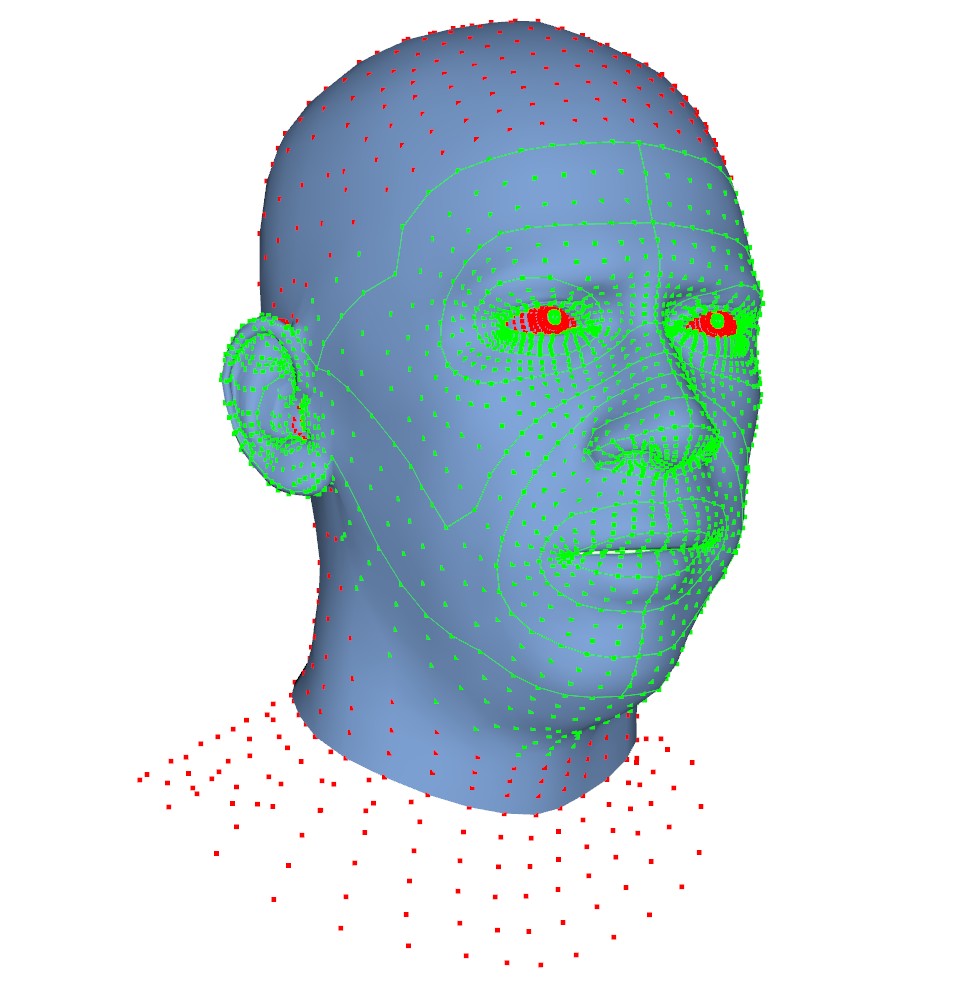}
  \caption{}
  \label{fig:flame_vertex_mask}
  \end{center}
\end{subfigure}%
\begin{subfigure}{0.32\linewidth}
  \begin{center}
  \includegraphics[width=0.95\linewidth]{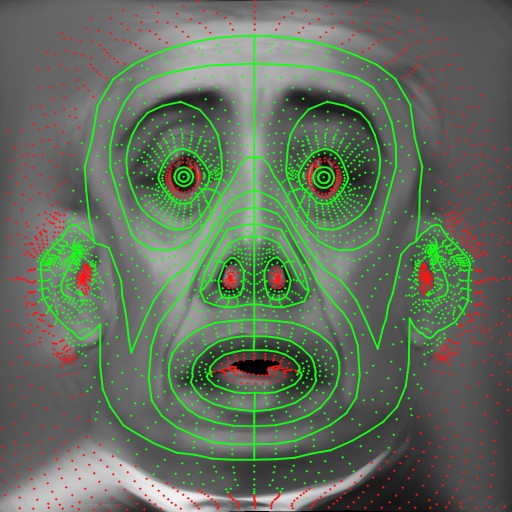}
  \caption{}
  \label{fig:flame_uv_verts}
  \end{center}
\end{subfigure}
\begin{subfigure}{0.32\linewidth}
  \begin{center}
  \includegraphics[width=0.95\linewidth]{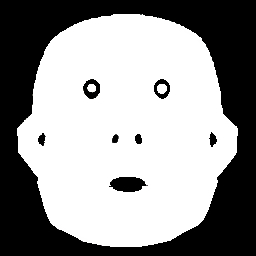}
  \caption{}
  \label{fig:flame_uv_mask}
  \end{center}
\end{subfigure}
\vspace{0.2cm}
\caption{Visualization of our FLAME~\cite{FLAME} head model vertices and vertex weights. The FLAME model contains 5023 3D vertices (a) and their corresponding coordinates in UV space (b). We set $\lambda_i=1$ for the vertices shown in green and $\lambda_i=0.005$ for the vertices shown in red. (c) shows the UV weight map used for the dense loss. We set $\lambda_p=1$ for the areas shown in white and $\lambda_p=0.005$ for the areas shown in black. }
\label{fig:flame_vertices}
\end{figure}

For the training of our 2D alignment model and model fitting, we focus on the vertices of the face and ear region. To this end, we introduced the per-vertex weights $\lambda_i$ and dense per-pixel UV weight mask $\lambda_p$ in \cref{sec:2dalign}. These weights are visualized in \cref{fig:flame_vertices}. For vertices and pixels in the face and ear area, we set a weight of $\lambda_i=1$ and $\lambda_p=1$, and for all other vertices and pixels we set $\lambda_i=0.005$ and $\lambda_p=0.005$. 

\section{Additional Results}
\label{supp:results}

In this section, we show additional results to demonstrate the performance of our method. 

\begin{figure}[h]
\centering
\begin{tabular}{c c }
\setlength{\tabcolsep}{0pt}

\includegraphics[width=4cm]{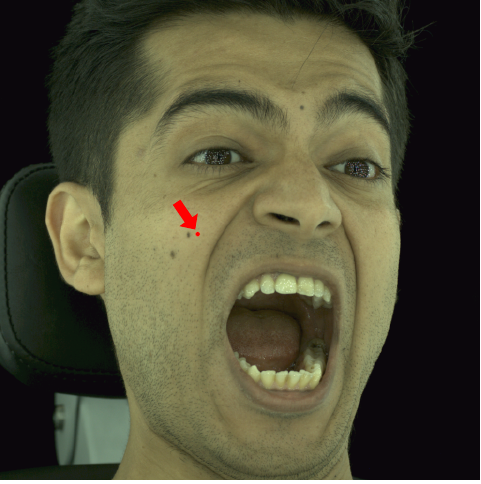} &
  \includegraphics[width=4cm]{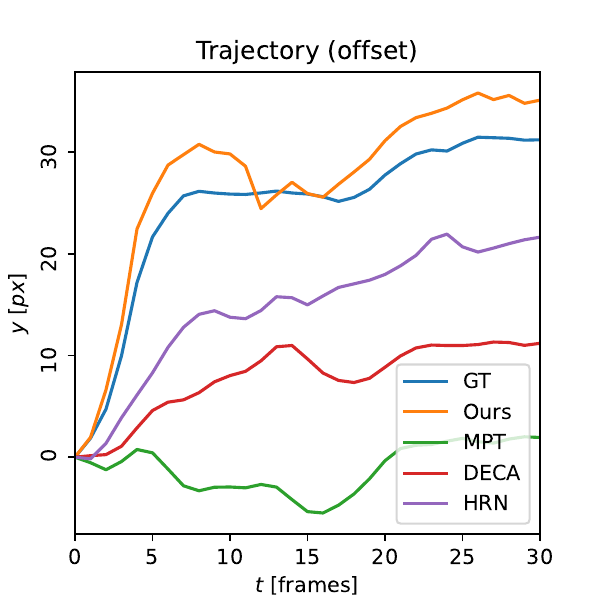} \\
  
\includegraphics[width=4cm]{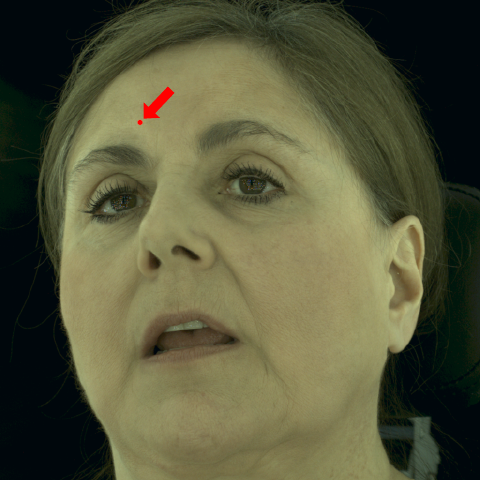} &
  \includegraphics[width=4cm]{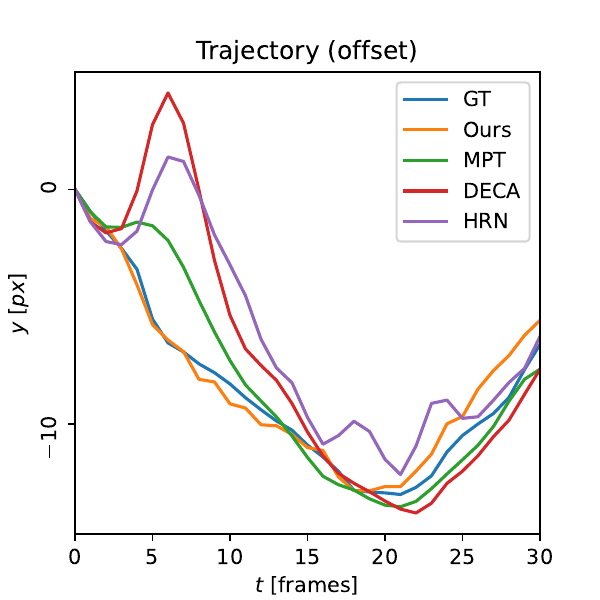} \\

Input image & Predicted trajectory 

\end{tabular}
\vspace{0.2cm}
\caption{ A visualization of the pixel-wise motion trajectory error for some methods. The ground truth and the predicted trajectory for a pixel (denoted in the images on the left side with a red dot and arrow) is plotted over the next 30 frames (right side). It is apparent that our model can track face motion more accurately, even in areas that are not visually salient such as the forehead (top row) or the cheek (bottom row). The fact that this motion error is not measured by previous metrics prompts the need for our screen space motion error (SSME). }
\label{fig:trajectory_vis}
\end{figure}

In \cref{fig:trajectory_vis}, we show how our tracker more accurately predicts the per-pixel trajectory than previous methods. This temporal accuracy is not measured by previous methods, which underlines the importance of our new SSME metric. 

\begin{figure}[h]
\centering
\includegraphics[width=0.99\linewidth]{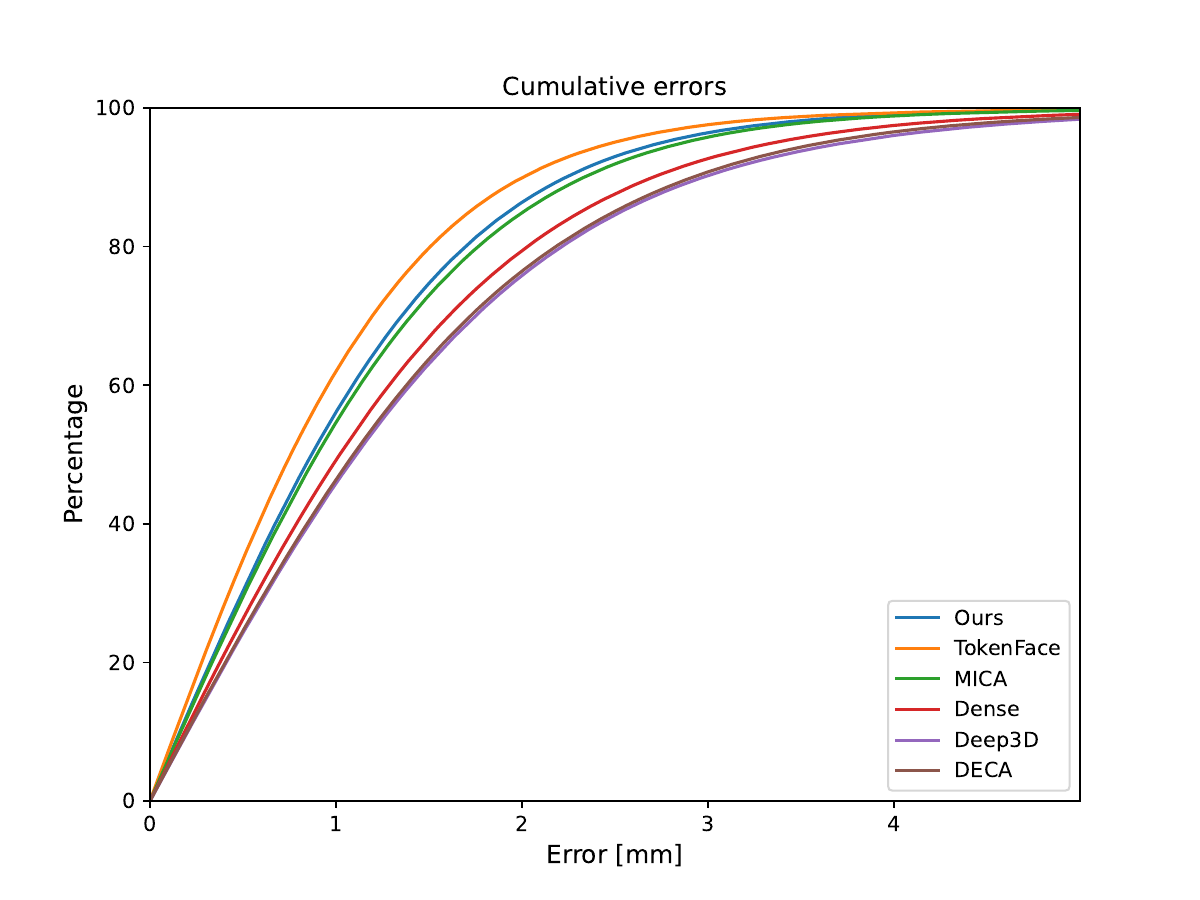}
\vspace{0.2cm}
\caption{The cumulative error plot on the NoW Challenge~\cite{now_benchmark} (single-view) of our method and recent methods. Competitive results show that our face tracker can disentangle expression and neutral shape and accurately reconstruct faces even with in-the-wild images.}
\label{fig:now_cumulative}
\end{figure}

The cumulative error of our method on the NoW Challenge~\cite{now_benchmark} are plotted and compared in \cref{fig:now_cumulative}.
In \cref{fig:now_qualitative_single} we qualitatively show the effects of ablations to our 3D model fitting method on the NoW single-view benchmark. In \cref{fig:now_qualitative_multi}, we show the importance of per-vertex deformations on the NoW multi-view benchmark.

In \cref{fig:align_qualitative}, we show a qualitative comparison between our dense 2D alignment network architecture and the ResNet-101 architecture of \cite{dense_landmarks_microsoft}. 

\newcommand{\alignimgone}[1]{\includegraphics[trim={4cm 4cm 4cm 4cm},clip,width=4cm]{figures/2d_align/#1}}
\newcommand{\alignimgtwo}[1]{\includegraphics[trim={4cm 4cm 4cm 4cm},clip,width=4cm]{figures/2d_align/#1}}

\begin{figure}[h]
\tablefont
\setlength{\tabcolsep}{1pt}
\renewcommand{\arraystretch}{0}
\centering
\begin{tabular}{c c}
\alignimgone{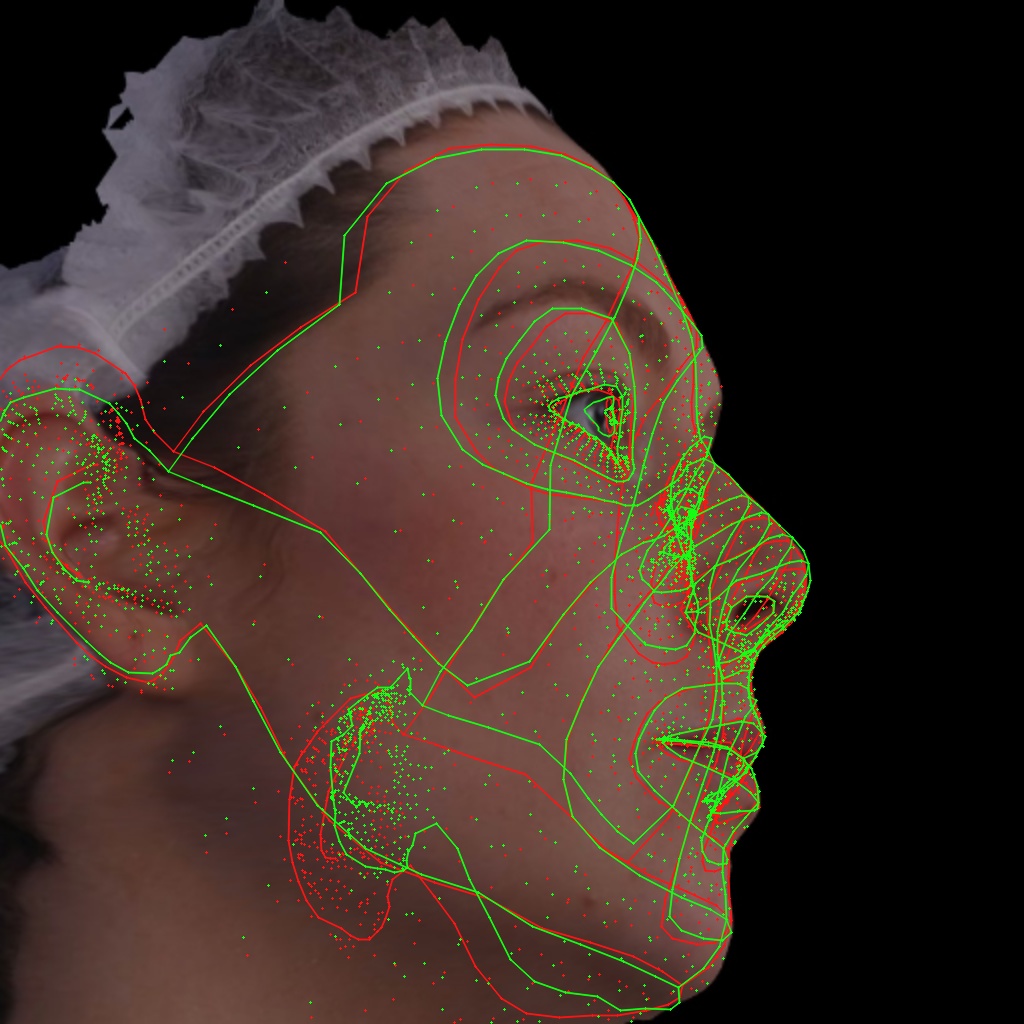} & \alignimgone{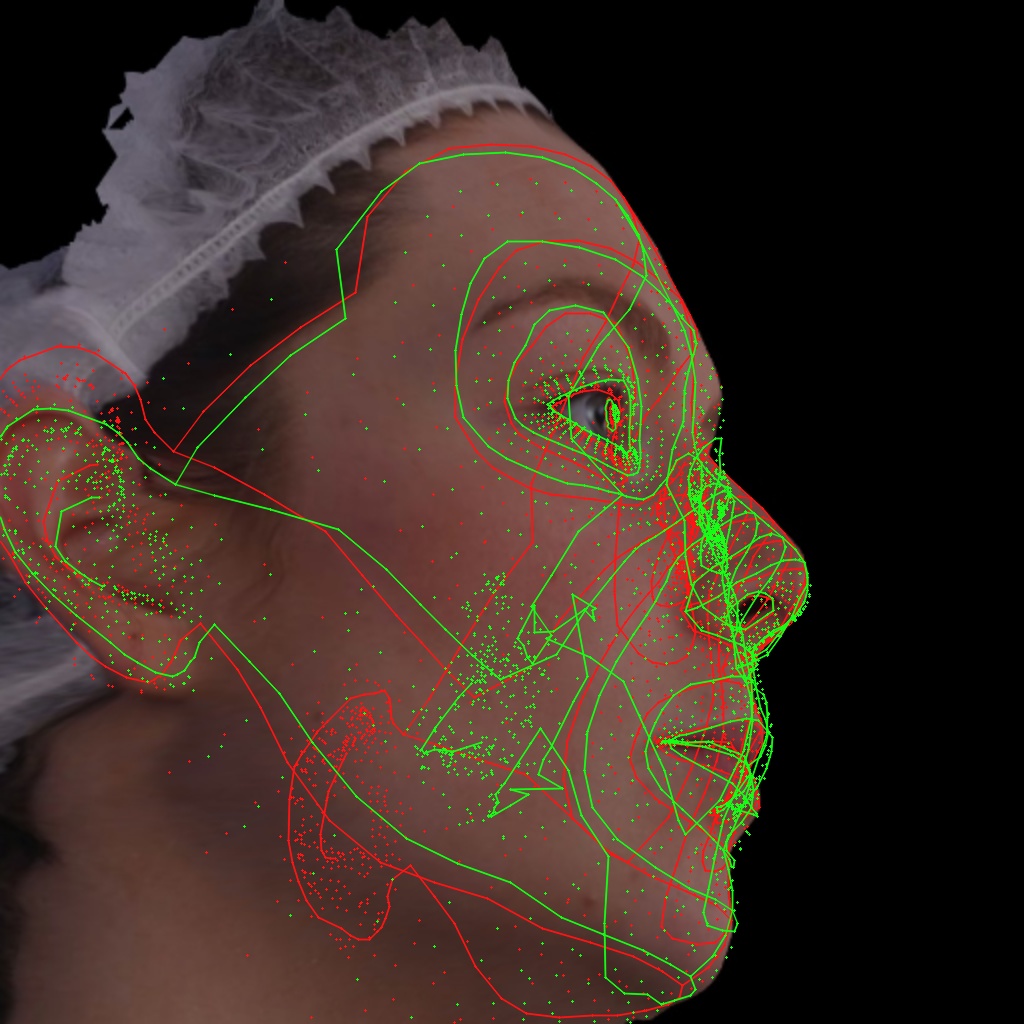} \\
\alignimgtwo{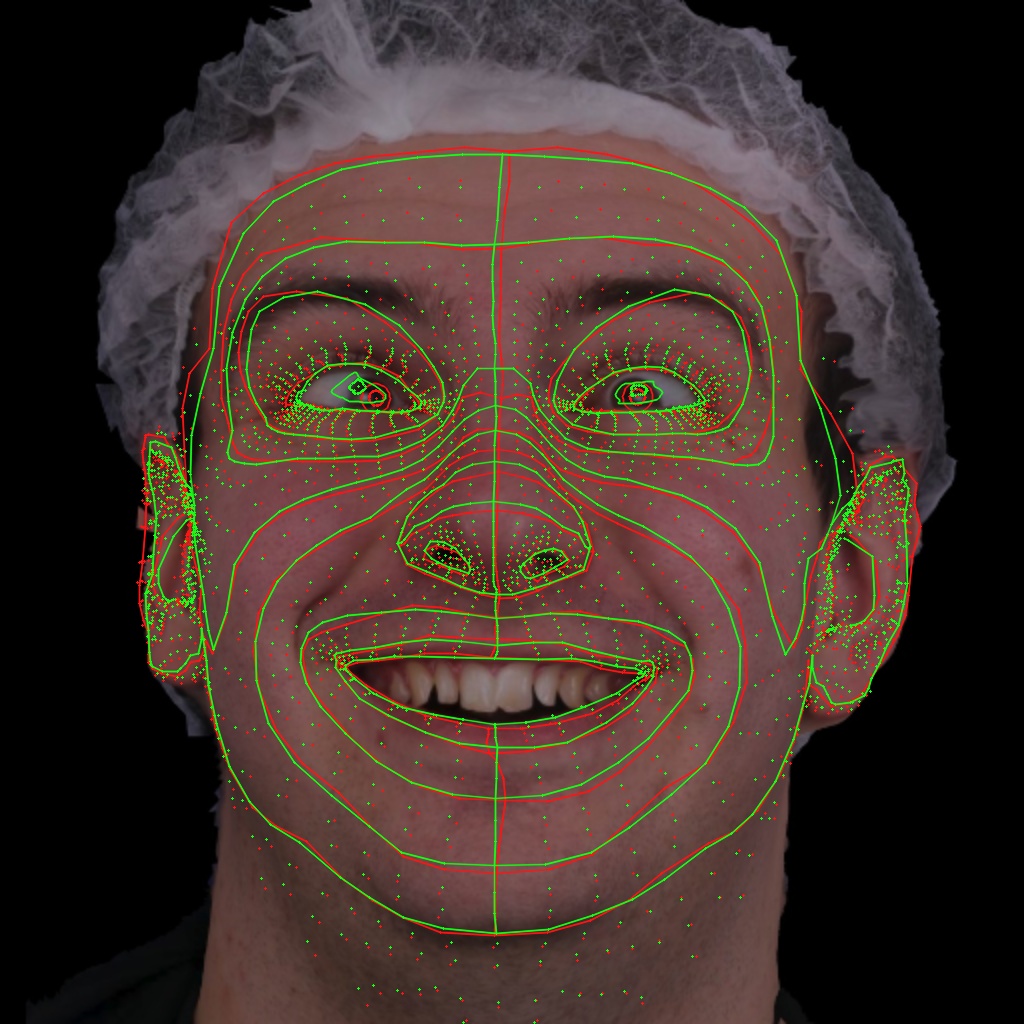} & \alignimgtwo{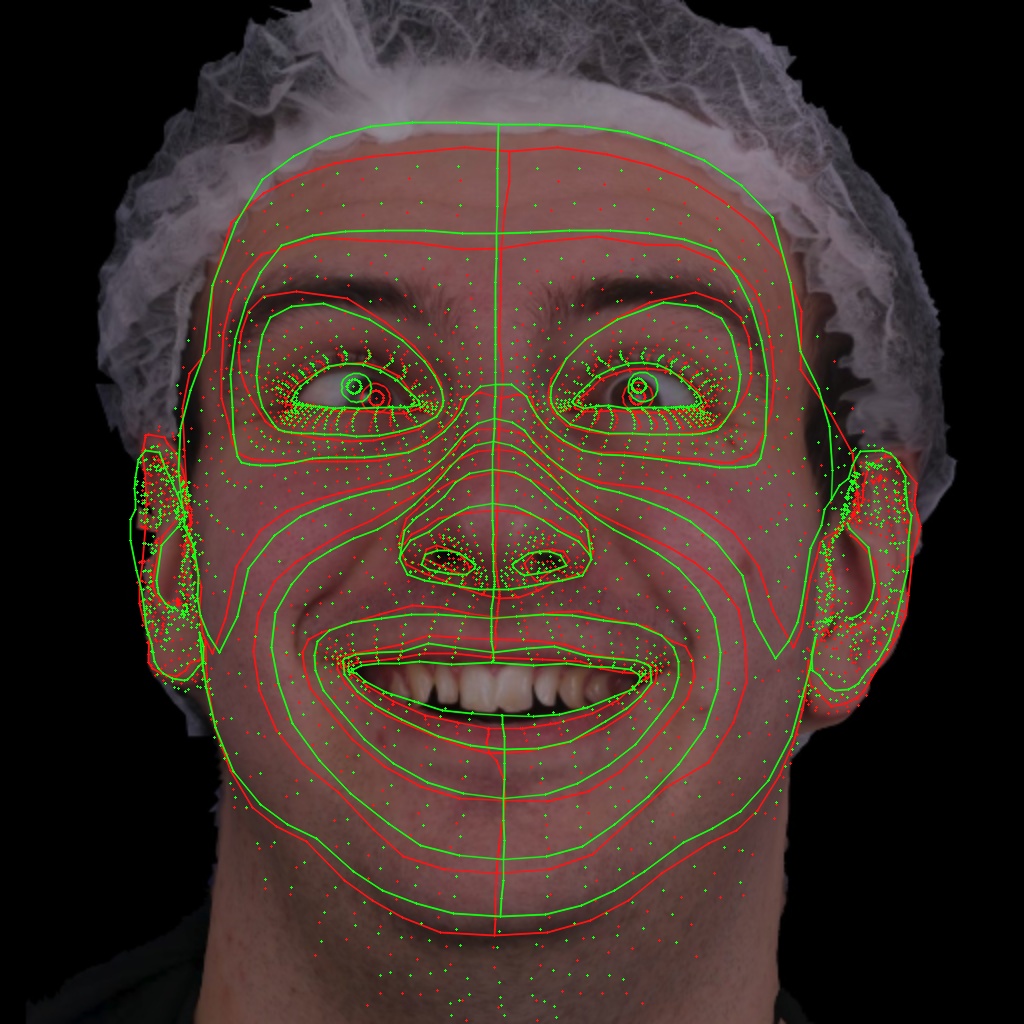}  \\ [0.1cm]
Ours & Dense [41] \\[0.1cm]
\end{tabular}
\caption{Qualitative comparison between our dense 2D alignment network architecture and the ResNet architecture of \cite{dense_landmarks_microsoft}. Red denotes ground truth alignment, green denotes predicted alignment. Our alignment network (left column) shows a significantly better alignment than \cite{dense_landmarks_microsoft} (right column) in areas such as the nose and lip contour (top row) and mouth and cheek region (bottom row). }
\label{fig:align_qualitative}
\vspace{-0.6cm}
\end{figure}

In the video \texttt{extreme\_expressions.mp4} (included in the supplementary material), we show how our tracker can handle extreme view deviations and expressions. Note the accuracy of our predicted 2D alignment and 3D model despite challenging facial motions. 
Finally, we show the qualitative performance of our tracker compared to other methods on in-the-wild images in \cref{fig:ffhq_qualitative_1} and \cref{fig:ffhq_qualitative_2}.

\section{Computational Complexity}
The tracking of 520 frames with 17 cameras takes 36 minutes on a \textit{Quadro RTX 5000} GPU, where MICA, face detection and 2D alignment take 15 minutes, and 3D model fitting takes 21 minutes. For this sequence, the GPU memory requirement is 4.5~GB. We note that our focus is not speed, but accuracy for offline 3D data generation.

\section{3D Head Avatar Synthesis}
\label{supp:head_avatar}

To evaluate the downstream performance of FlowFace on 3D head avatar synthesis, we choose the recent state-of-the-art method INSTA~\cite{insta}. INSTA learns a high-quality deformable NeRF from a tracked video of a moving head, which can be animated in real time using a proxy FLAME morphable head model. The original implementation of INSTA uses head tracking data provided by MPT~\cite{mica}. We therefore refer to the baseline implementation as MPT-INSTA and our combination of FlowFace output with INSTA as FlowFace-INSTA.

We minimally modify INSTA by replacing their tracker with ours. As recommended by the authors of INSTA, the C++ version of the public implementation of INSTA is used for all experiments. For each frame of the dataset, the INSTA implementation expects to be provided with camera intrinsics and pose, a 3D head mesh, FLAME expression blendshape coefficients, a depth map covering the face, and a semantic segmentation map. As described in \cref{sec:correspondence}, our method provides almost all the information we need to generate the necessary frame data. The only data not generated from our tracker's output is the semantic segmentation maps, for which we followed the INSTA implementation and generated them using BiSeNet~\cite{bisenet_v2}.

\begin{figure}[h]
\tablefont
\setlength{\tabcolsep}{1pt}
\renewcommand{\arraystretch}{1}
\centering
\begin{tabular}{c c c c }

{\centering \rotatebox{90}{Reconstruction}} & \multicolumn{3}{c}{ \includegraphics[trim={0 3cm 0 1cm},clip,width=\linewidth]{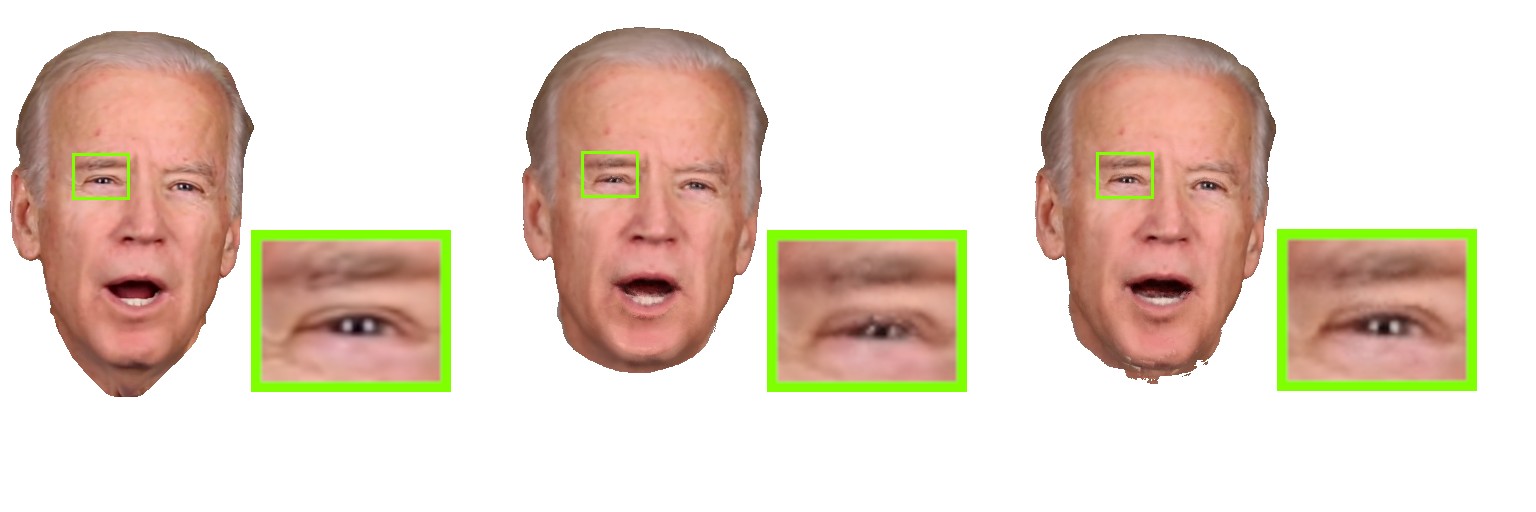}} \\
\rotatebox{90}{ \quad \quad Tracking} & 
\includegraphics[trim={1cm 2cm 1cm 1cm},clip,width=0.33\linewidth,]{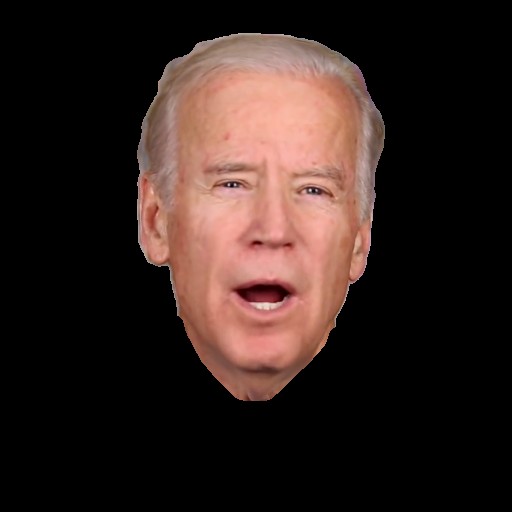} & \includegraphics[trim={1cm 2cm 1cm 1cm},clip,width=0.33\linewidth]{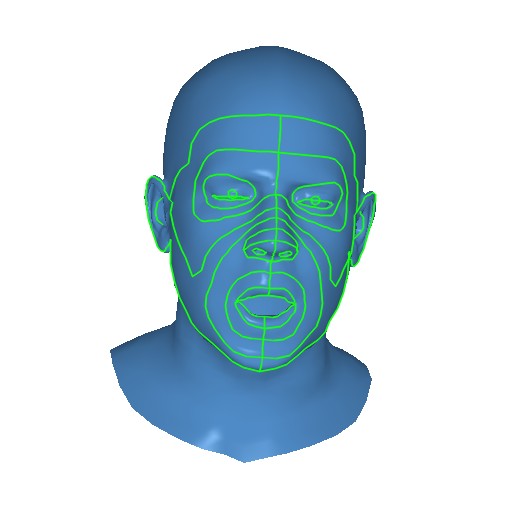}  & \includegraphics[trim={1cm 2cm 1cm 1cm},clip,width=0.33\linewidth]{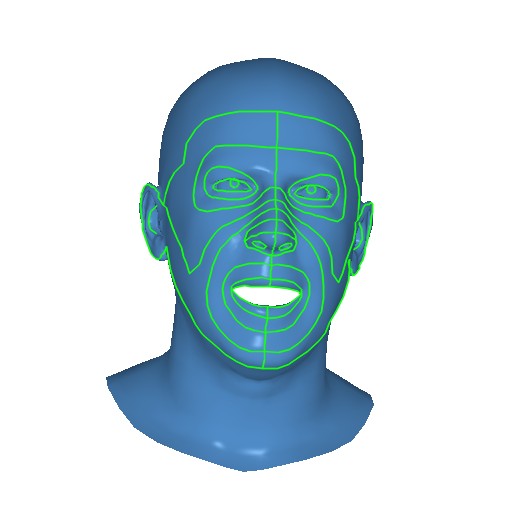} \\
& GT & MPT-INSTA & FlowFace-INSTA \\
\hline

\rotatebox{90}{\quad \quad Reconstruction} & \multicolumn{3}{c}{ \includegraphics[trim={0 0 0 -1cm},clip,width=\linewidth]{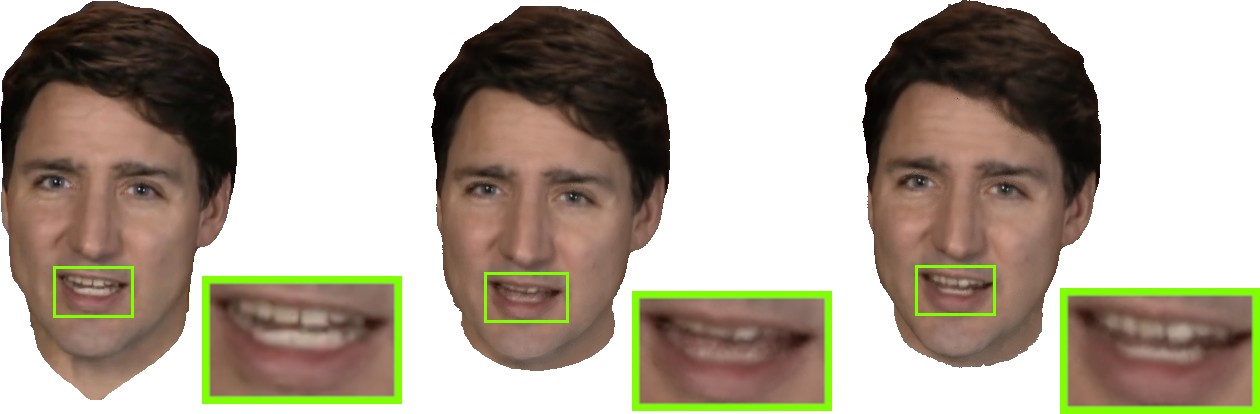}} \\
\rotatebox{90}{\quad \quad  \quad  Tracking} & 
\includegraphics[trim={1cm 0 1cm 1cm},clip,width=0.33\linewidth]{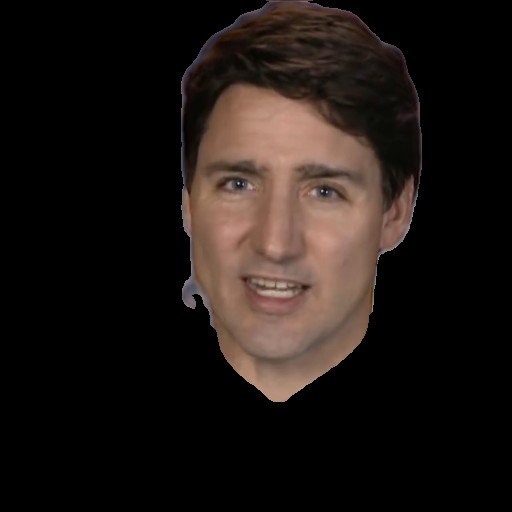} & 
\includegraphics[trim={1cm 0 1cm 1cm},clip,width=0.33\linewidth]{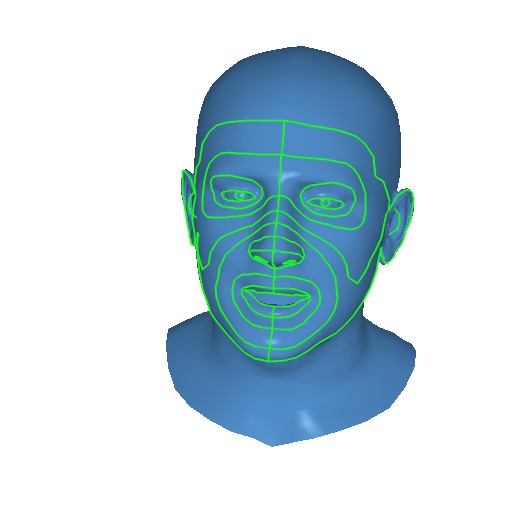}  & 
\includegraphics[trim={1cm 0 1cm 1cm},clip,width=0.33\linewidth]{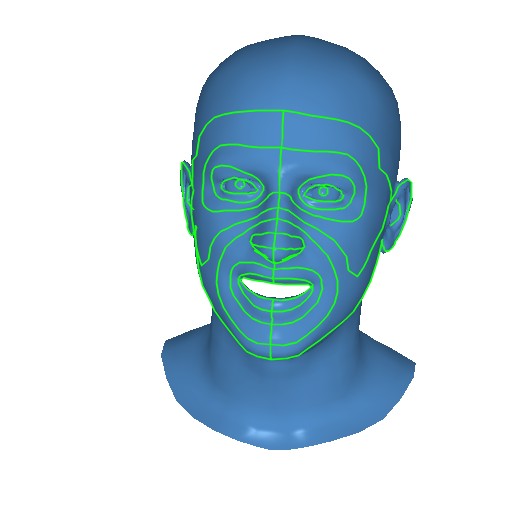} \\

& GT & MPT-INSTA & FlowFace-INSTA \\

\end{tabular}
\vspace{0.3cm}
\caption{ Qualitative comparison of INSTA results using MPT~\cite{mica} (center column) and FlowFace (right column) as face tracker. More accurate and more consistent tracking throughout the train and test images by our tracker leads to a more accurate and detailed reconstruction. }
\label{fig:insta_qualitative_1}
\end{figure}

\begin{figure}[h]
\tablefont
\setlength{\tabcolsep}{1pt}
\renewcommand{\arraystretch}{1}
\centering
\begin{tabular}{c c c c }

\rotatebox{90}{\quad Reconstruction} & \multicolumn{3}{c}{ \includegraphics[trim={0 3cm 0 1cm},clip,width=\linewidth]{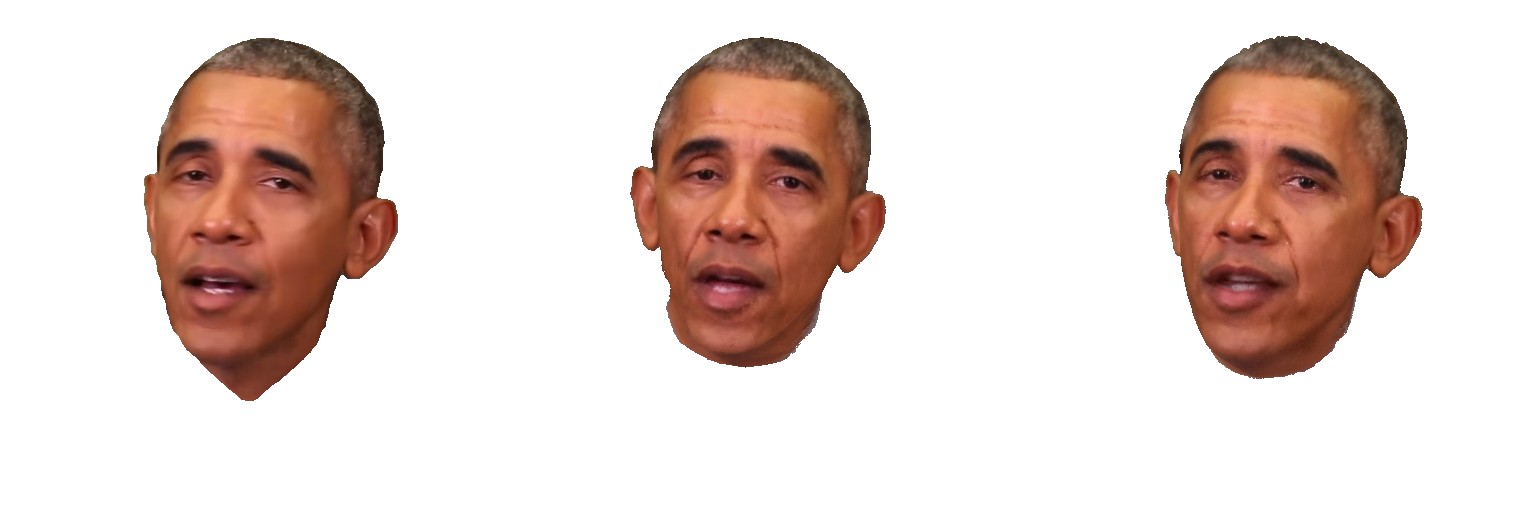}} \\
\rotatebox{90}{\quad \quad \quad Tracking} & 
\includegraphics[trim={1cm 2cm 1cm 1cm},clip,width=0.33\linewidth,]{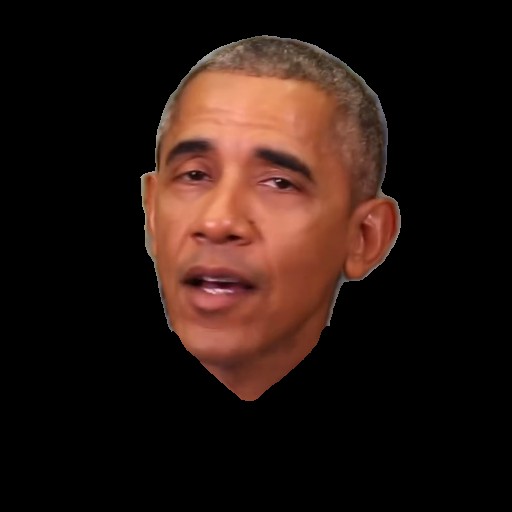} & \includegraphics[trim={1cm 2cm 1cm 1cm},clip,width=0.33\linewidth]{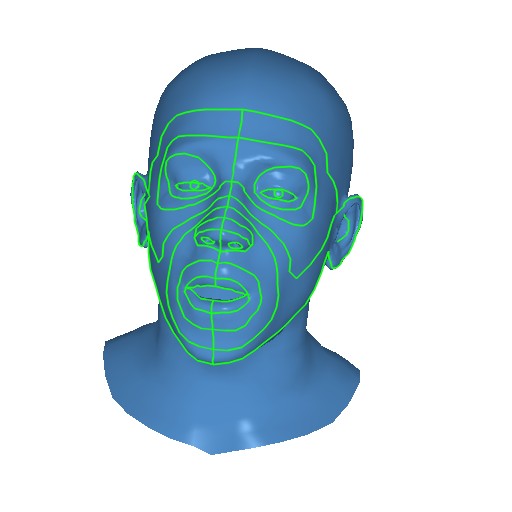}  & \includegraphics[trim={1cm 2cm 1cm 1cm},clip,width=0.33\linewidth]{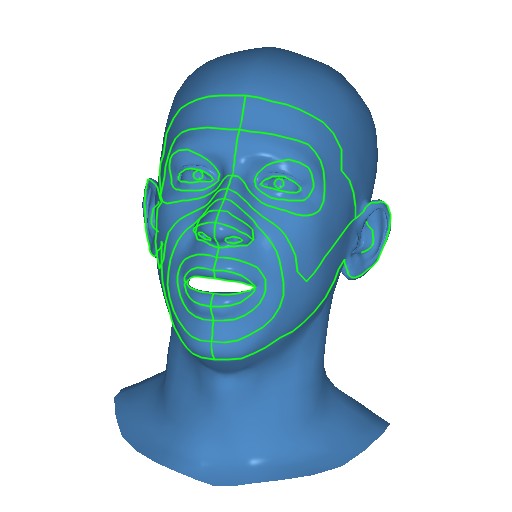} \\

& GT & MPT-INSTA & FlowFace-INSTA \\

\end{tabular}
\vspace{0.3cm}
\caption{ Examples of large photometric errors due to failure of the MPT~\cite{mica} tracker. The tracked pose of the head (center column, bottom) by MPT is inaccurate, which leads to a misalignment of the reconstructed image (center column, top) and the ground truth (left column). This is likely due to the motion blur present in the ground truth image. Our tracker (left column) can still accurately predict the head pose, resulting in a better reconstruction. }
\label{fig:insta_qualitative_2}
\end{figure}

We use two sets of data to compare our enhanced FlowFace-INSTA to the baseline MPT-INSTA. One dataset is the full set of 10 videos released with INSTA, where we adopt the same splits for training and testing frames. The training and testing splits cover two distinct intervals of each video with no overlap. We use the pre-trained INSTA models provided by the authors to predict images for the testing frames which represent the output of MPT-INSTA. As seen from the image quality metrics in \cref{table-insta-insta_data}, FlowFace-INSTA improves LPIPS by 10.5\%, with slight improvements in other metrics (PSNR, SSIM and MS-SSIM) as well. Qualitatively, we observe that the improved tracking accuracy of FlowFace result in  higher-quality reconstruction of the eyes and mouth as well as slightly sharper overall reconstruction, visible in facial skin and stubble (see \cref{fig:insta_qualitative_1}). These relatively subtle improvements could account for the superior perceptual quality indicated by LPIPS. We also notice that FlowFace robustly tracks the head in portions of the video where MPT fails, as shown in \cref{fig:insta_qualitative_2}. A video comparing MPT-INSTA and FLowFace-INSTA is included with the supplementary material (\texttt{avatar\_synthesis\_compare.mp4}).

The second evaluation dataset is the aforementioned subset of 86 videos from the MultiFace dataset~\cite{multiface}. MultiFace does not provide its own training and testing splits for the videos in the dataset. We observe that in many video sequences, the subject would perform certain expressions and then transition to a neutral pose towards the end of the sequence. The videos are also very short, being only a few seconds long. This means that unlike in the first dataset, the latter portion of each video is biased toward neutral expressions and would not provide an adequate test set. Therefore, we take the middle 20\% of each video as the test set for that sequence, and use the remainder for training INSTA. 
For both MPT and FlowFace, we perform head tracking on the video and train 86 INSTA models separately on each sequence, without mixing frames of the same subject from different cameras or sequences. The computed image quality metrics are given in \cref{table-insta-multiface}. FlowFace-INSTA shows a significant improvement of 20.3\% for LPIPS over MPT-INSTA. Other common image quality metrics are either slightly better or comparable.

Aside from the photometric reconstruction quality, we also show in \cref{fig:expression_transfer} that our tracker can be used to transfer motion and expressions between a driver video of a person and an INSTA model trained on FlowFace tracking data. A video with an example of expression transfer is included with the supplementary material (\texttt{expression\_transfer.mp4}).

\newcommand{\expimgdriver}[1]{\includegraphics[trim={0 0 18.2cm 0},clip,width=2.4cm]{figures/insta/exp_transfer/#1.jpg}}
\newcommand{\expimgresult}[1]{\includegraphics[trim={18.2cm 0 0 0},clip,width=2.4cm]{figures/insta/exp_transfer/#1.jpg}}

\begin{figure*}[h]
\tablefont
\setlength{\tabcolsep}{1pt}
\renewcommand{\arraystretch}{1}
\centering
\begin{tabular}{c | c c c c c c c}

Target &

\rotatebox{90}{\quad \quad Driver} & 
\expimgdriver{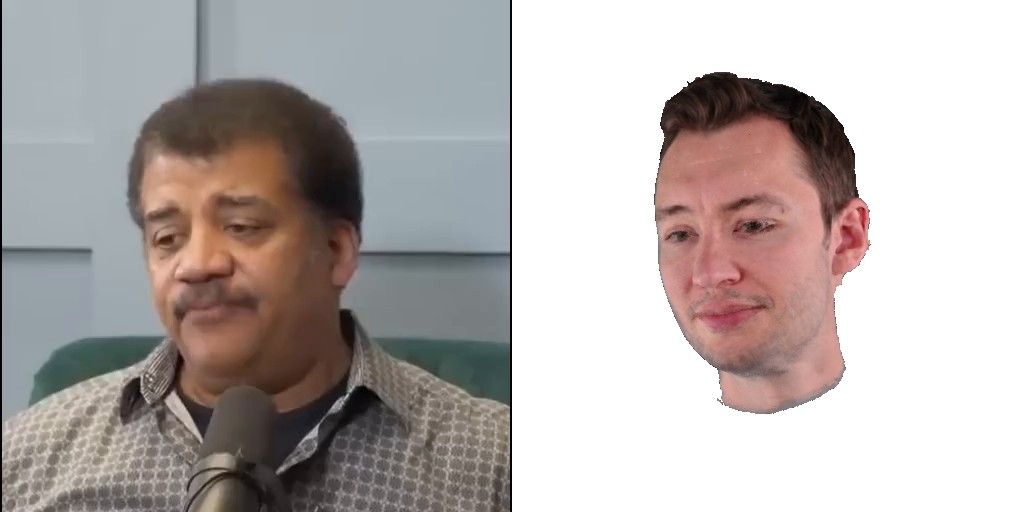} & \expimgdriver{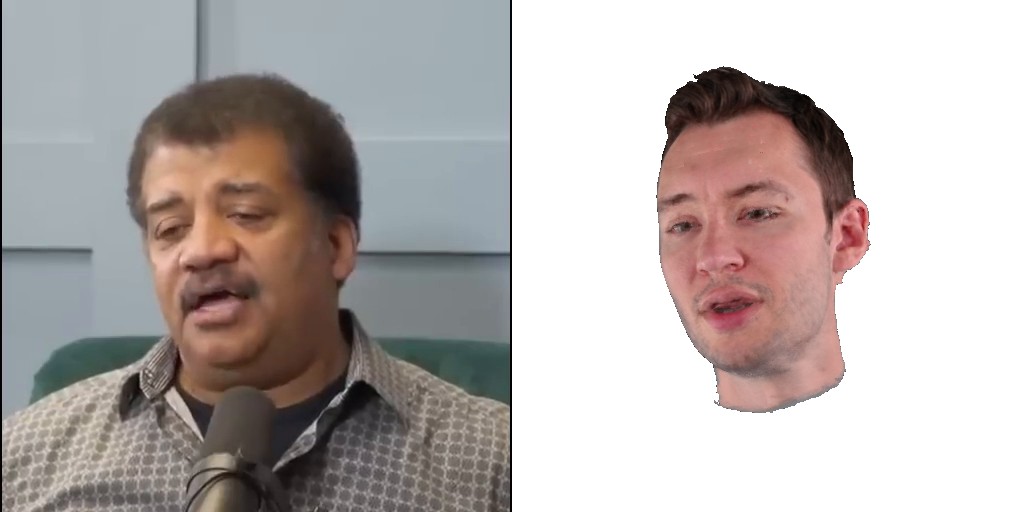} & \expimgdriver{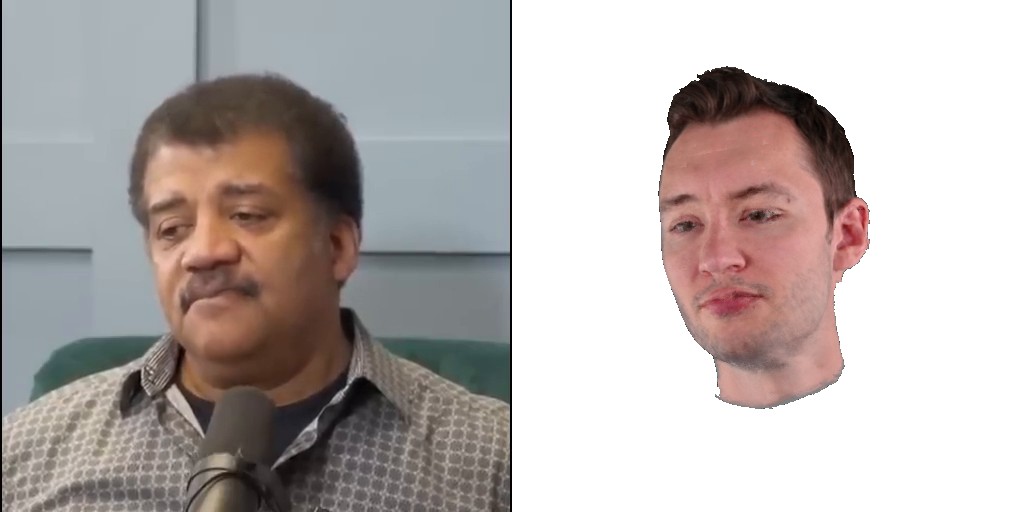} & \expimgdriver{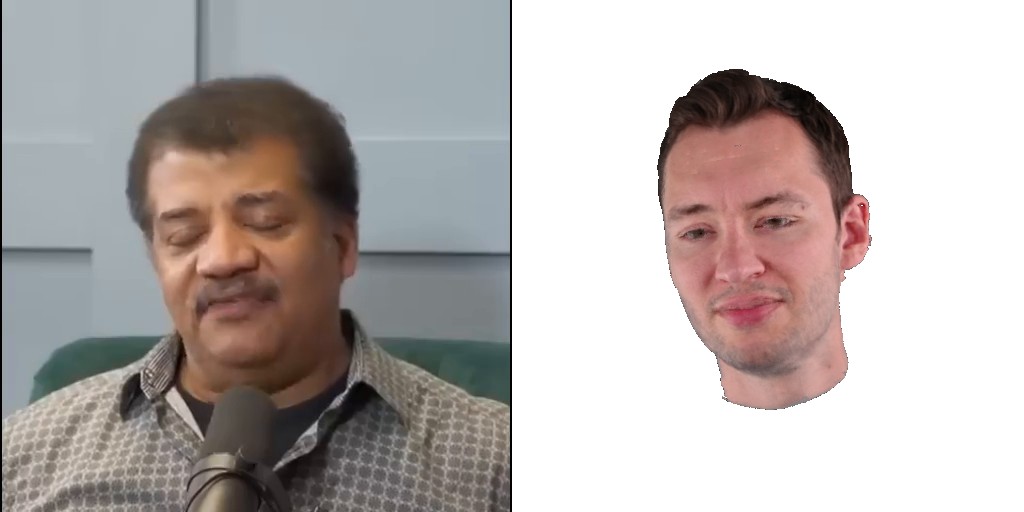} & \expimgdriver{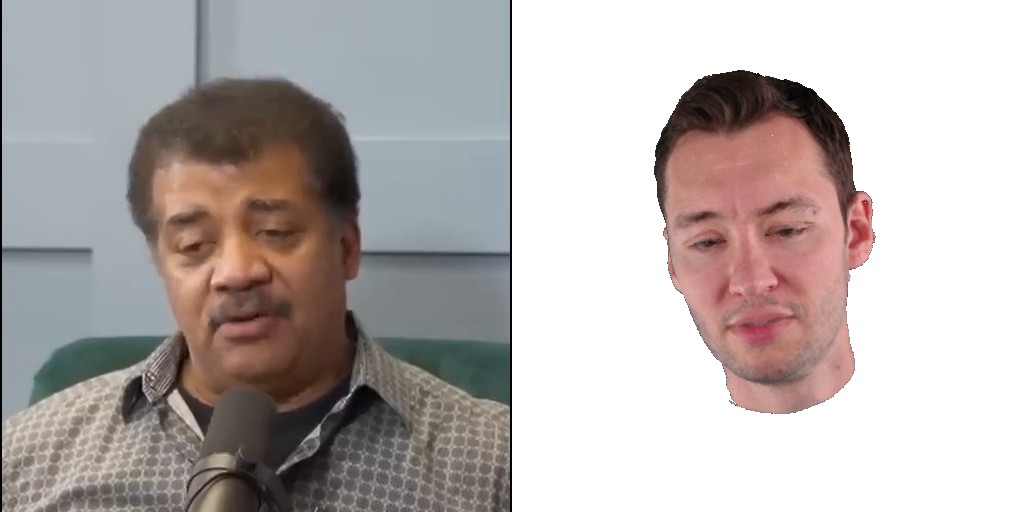} & \expimgdriver{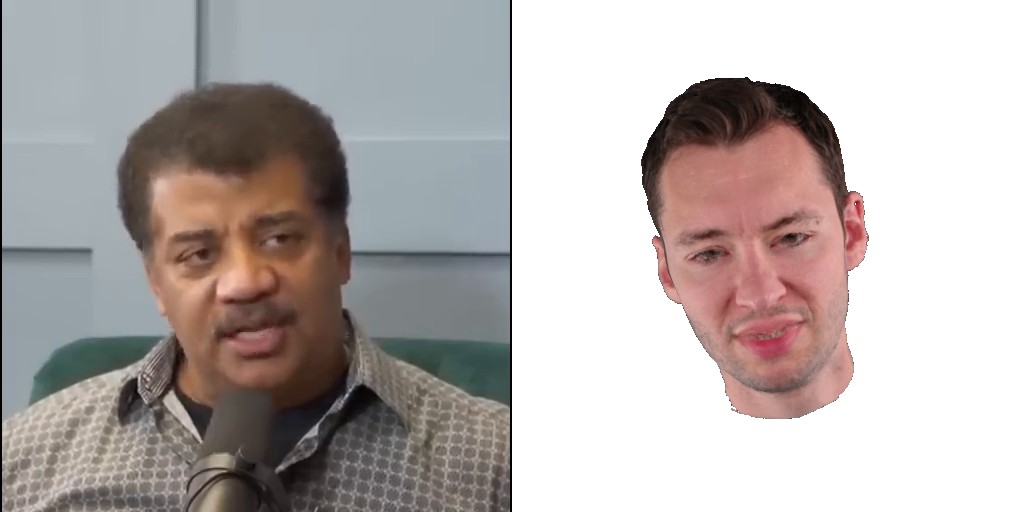}
\\

\includegraphics[trim={0 0 0 0},clip,width=2cm]{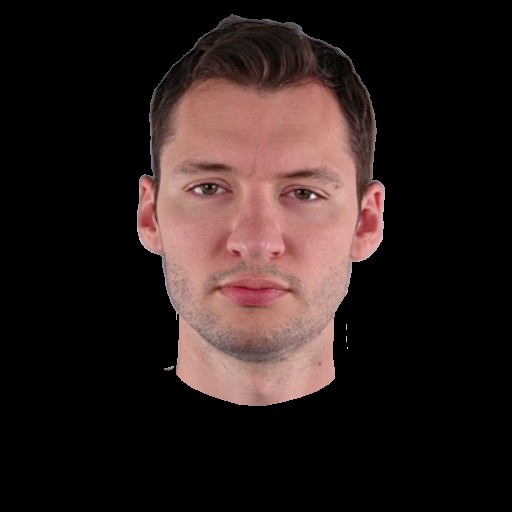}

& \rotatebox{90}{\quad \quad \quad Result} & 
\expimgresult{00000} & \expimgresult{00050} & \expimgresult{00100} & \expimgresult{00155} & \expimgresult{00200} & \expimgresult{00250}
\\


\end{tabular}
\vspace{0.3cm}
\caption{ Expression transfer using our tracker and FlowFace-INSTA. First, an INSTA~\cite{insta} avatar reconstruction is generated using a video of the target subject. Then, the driving face is reconstructed from a video using our face tracker. The expression and pose are extracted from the driving sequence and inserted into the target avatar and novel views are synthesized. }
\label{fig:expression_transfer}
\end{figure*}

\begin{table}[]
\tablefont
\renewcommand{\arraystretch}{1.2}
\centering
\label{tab:insta_public}
\begin{tabular}{l|c|c|c|c}
\hline
Tracker & PSNR$\uparrow$          & SSIM$\uparrow$           & MS-SSIM$\uparrow$        & LPIPS$\downarrow$           \\ \hline
MPT~\cite{mica}    & 31.5          & 0.949          & 0.973          & 0.0410          \\ 
Ours   & \textbf{31.9} & \textbf{0.953} & \textbf{0.977} & \textbf{0.0367} \\ \hline
\end{tabular}
\caption{\label{table-insta-insta_data} Downstream avatar synthesis results on videos released with INSTA. By replacing the tracker used in INSTA~\cite{insta}, we achieve significantly better perceptual similarity (LPIPS). }
\end{table}

\begin{table}[]
\tablefont
\renewcommand{\arraystretch}{1.2}
\centering
\label{tab:insta_multiface}
\begin{tabular}{l|c|c|c|c}
\hline
Tracker & PSNR$\uparrow$          & SSIM$\uparrow$           & MS-SSIM$\uparrow$        & LPIPS$\downarrow$         \\ \hline
MPT~\cite{mica}    & \textbf{20.2}          & \textbf{0.885}          & 0.939          & 0.1821          \\ 
Ours   & 20.1 & 0.884 & \textbf{0.945} & \textbf{0.1452} \\ \hline
\end{tabular}
\caption{\label{table-insta-multiface} Downstream avatar synthesis results on MultiFace dataset videos.}
\end{table}

\section{Speech-Driven 3D Facial Animation}
\label{supp:audio2face}



\subsection{Generating Data}

We apply our facial reconstruction method on the popular MEAD \cite{mead} dataset to generate 3D-MEAD, a speech to 3D facial animation dataset. MEAD is a multi-view talking-face video corpus with $43$ English speakers, speaking $40$ unique sequences with $8$ different emotions. For the purposes of this work, we focus only on the \textit{neutral} emotion. We split training, validation, and testing sets into $27$, $8$, and $8$ speakers, yielding $1080$, $320$, and $320$ animation sequences, respectively. We also generate a training subset of only $8$ speakers from the same set of $27$ speakers for certain studies. In all subsets, there is an equal (when possible) split of female and male speakers.
The dataset contains 7 uncalibrated multi-view videos for each sequence, and we use 4 of these to track the face. An example of our multi-view tracking on the MEAD dataset can be viewed in \cref{fig:mead_example} and in the supplementary videos (\texttt{mead\_tracking.mp4}).
In the MEAD dataset, images of the subjects with neutral expressions are not available. However, typical face animation models such as CodeTalker~\cite{codetalker} require the neutral reconstruction. We can generate this reconstruction with the accurate neutral shape and expression disentanglement of our tracker. 

\begin{figure*}[h]
\tablefont
\setlength{\tabcolsep}{1pt}
\renewcommand{\arraystretch}{0}
\centering
\begin{tabular}{c }
\includegraphics[width=\linewidth]{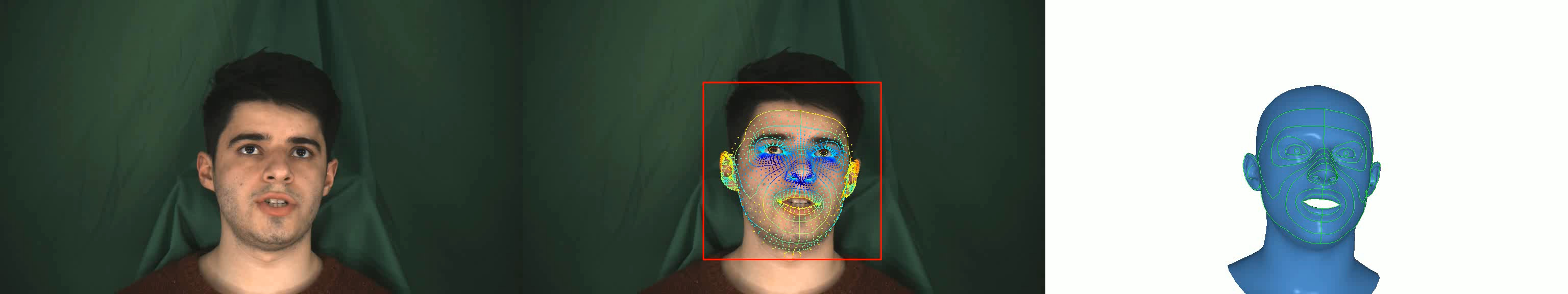} \\
\includegraphics[width=\linewidth]{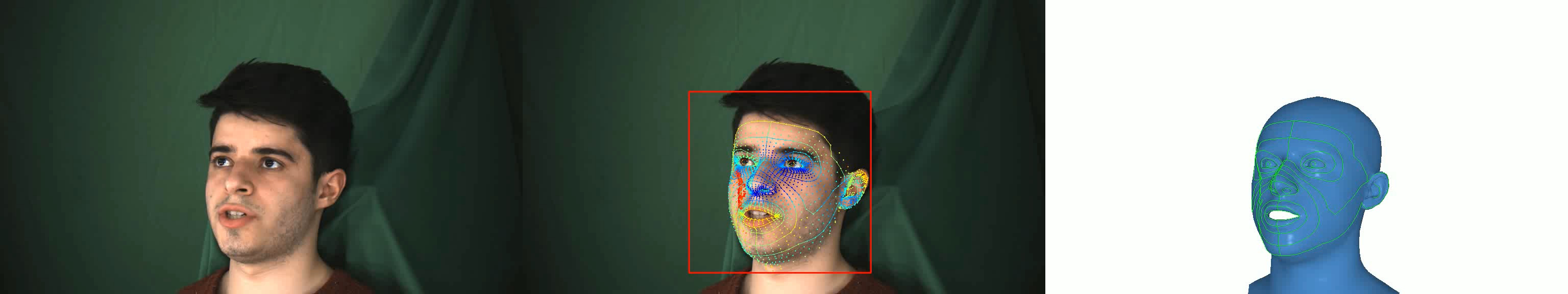} \\ 
\includegraphics[width=\linewidth]{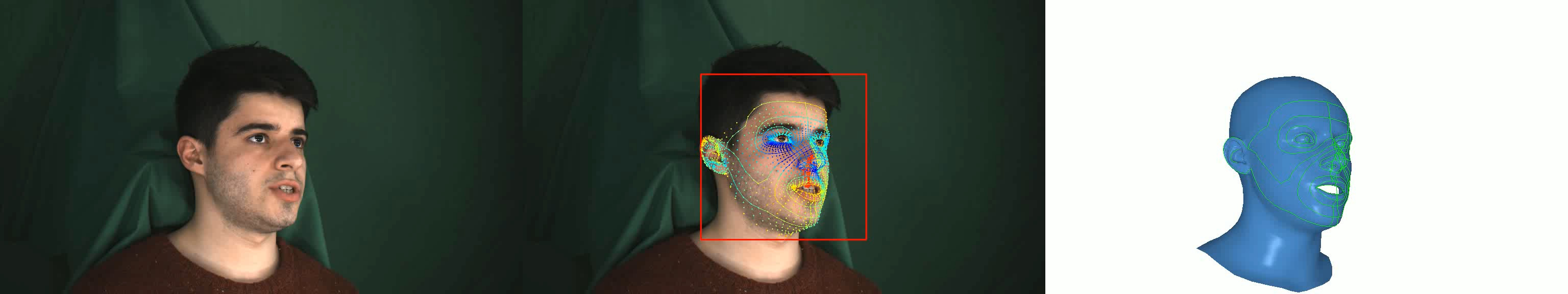} \\
\includegraphics[width=\linewidth]{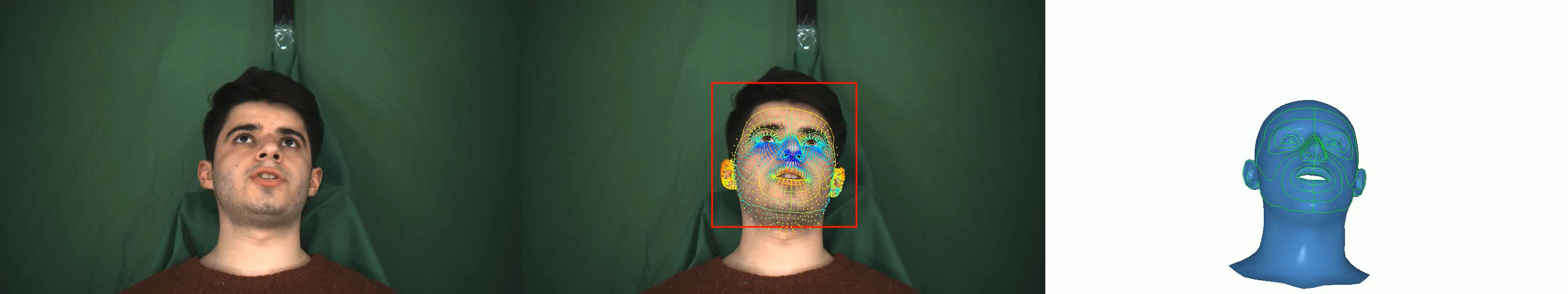} 
\end{tabular}
\vspace{0.3cm}
\caption{3D data generation using the MEAD~\cite{mead} dataset. Our face tracker can seamlessly integrate multiple-view video to improve 3D face tracking. Extrinsics, intrinsics and the 3D face model (right column) are simultaneously optimized to fit the predicted 2D alignment (center column) in our 3D model fitting module. We utilize 4 cameras for each sequence to generate high quality training data for speech-driven 3D face animation models. }
\label{fig:mead_example}
\end{figure*}

\subsection{Datasets}
We utilize the popular VOCASET \cite{vocaset} to train and test different methods in our experiments, as well as the 3D-MEAD dataset. Both contain 3D facial animations paired with English utterances. VOCASET contains 255 unique sentences, which are partially shared among different speakers, yielding $480$ animation sequences from 12 unique speakers. Those 12 speakers are split into $8$ unique training, $2$ unique validation, and $2$ unique testing speakers. Each sequence is captured at 60 fps, resampled to 30 fps, and ranges between $3$ and $4$ seconds. We use the same training, validation, and testing splits as VOCA and FaceFormer, which we similarly refer to as VOCA-Train, VOCA-Val, and VOCA-Test. For 3D-MEAD, there are $43$ unique speakers, where each speaker has $40$ unique sequences, yielding a total of $1680$ sequences. We randomly split the dataset into $27$, $8$, and $8$ training, validation, and test speakers. To align with VOCASET, we subsample the training set to only containe $8$ speakers. We refer to each split as 3D-MEAD-Train, 3D-MEAD-Val, 3D-MEAD-Test. In both datasets, face meshes are composed of $5023$ vertices of the FLAME~\cite{FLAME} topology. To train on the downstream task, we combine these two datasets together and treat VOCA-Train and 3D-MEAD-Train as a single dataset. 

\subsection{Training}
We implement the popular state-of-the-art transformer-based model CodeTalker \cite{codetalker}, and train it a combined dataset of 3D-MEAD and VOCASET. This combined dataset has $16$ training speakers, so we increase the one-hot style encoding to be of size $16$. We optimize the network with Adam \cite{ba2015adam} and a learning rate of $\num{1e-4}$ and a batch size of $1$. The network is trained for $100$ epochs across three random seeds, and we report the average results using the weights from the last epoch in training.

\subsection{Results and Discussion}
To evaluate the results of our model, we test on the popular VOCASET benchmark \cite{vocaset} using the lip vertex error (LVE). The lip vertex error calculates the deviation of the lip position in a sequence with respect to the ground truth. More specifically, it is the 
maximal L2 error of all lip vertices for each frame and averaged over all frames.
Using the augmented data generated by our method, we are able to improve from a lip vertex error of $\num{3.13e-5}$ to $\num{2.85e-5}$ on the VOCASET benchmark, an $8.8\%$ improvement.

As previously mentioned, 3D facial animation models require the neutral face mesh for their training and inference. This is because they are trained to predict \textit{vertex offsets} rather than the absolute vertex positions. In practice, vertex offsets are generated by taking a sequence of facial meshes and subtracting the neutral mesh. 
It is therefore vital that our face tracker accurately disentangles expression and neutral meshes. We can confidently establish that our method is able to perform this task effectively given the positive results obtained.

\newpage
\newcommand{\nowimg}[1]{\includegraphics[width=1.8cm]{figures/now/#1}}

\begin{figure}[p]
\tablefont
\setlength{\tabcolsep}{1pt}
\renewcommand{\arraystretch}{1}
\centering
\begin{tabular}{c c c c }

\multicolumn{4}{c}{Input image} \\
\multicolumn{4}{c}{\nowimg{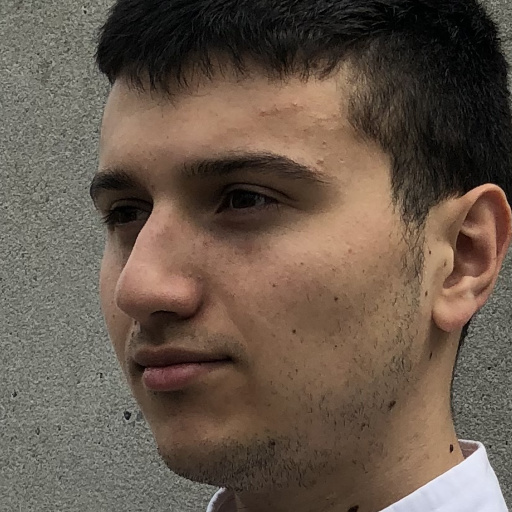}} \\[0.2cm]

\nowimg{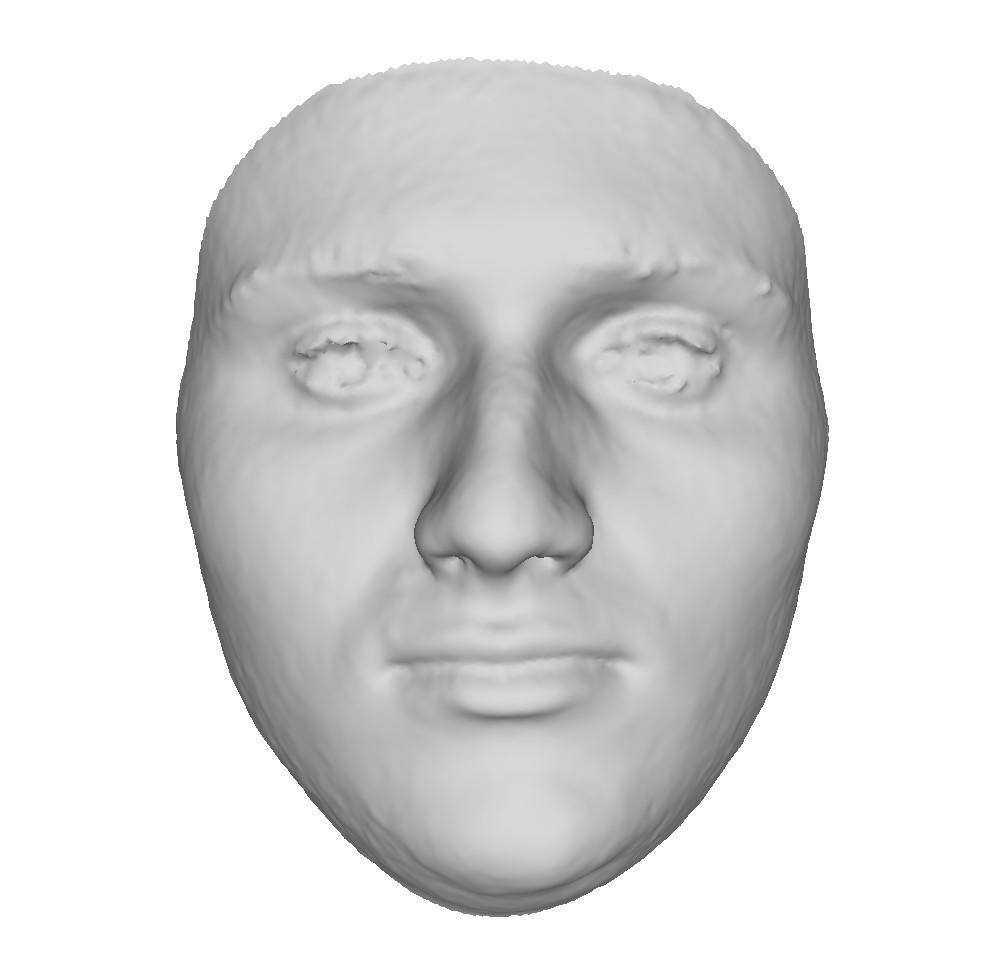} & \nowimg{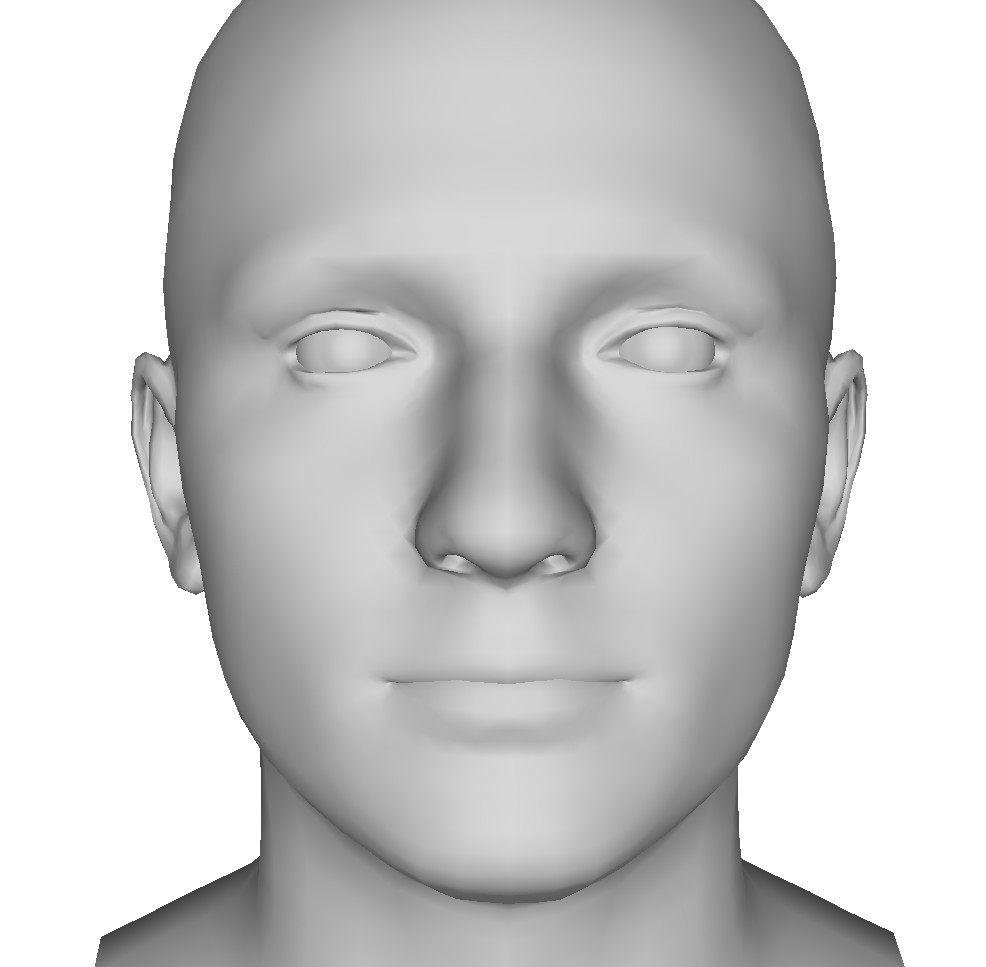} & \nowimg{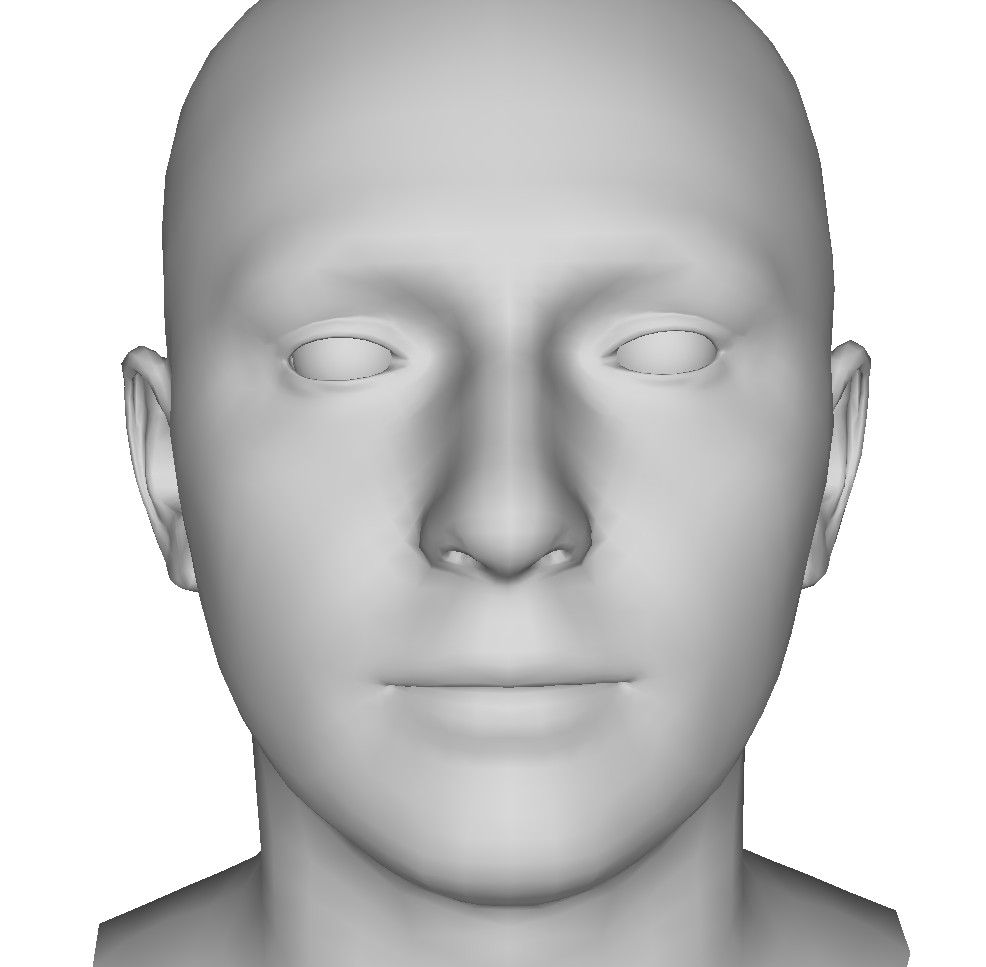} & \nowimg{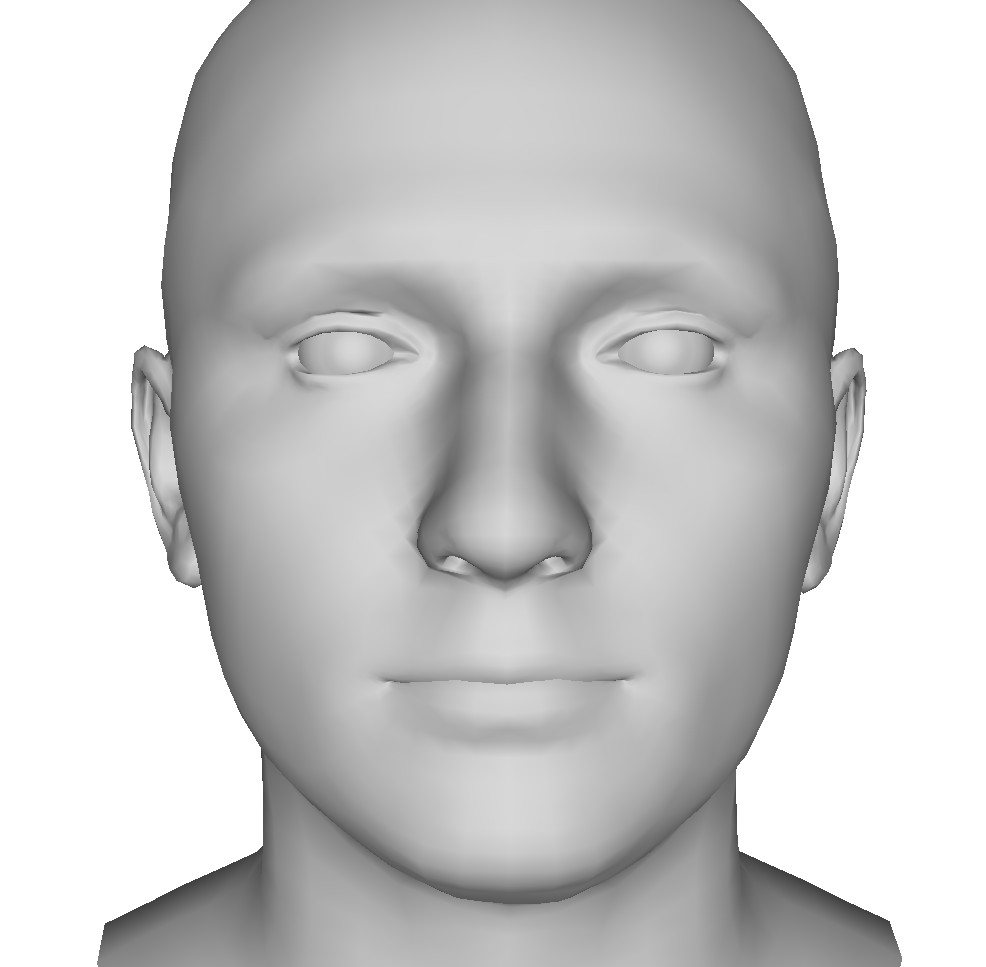}\\

\nowimg{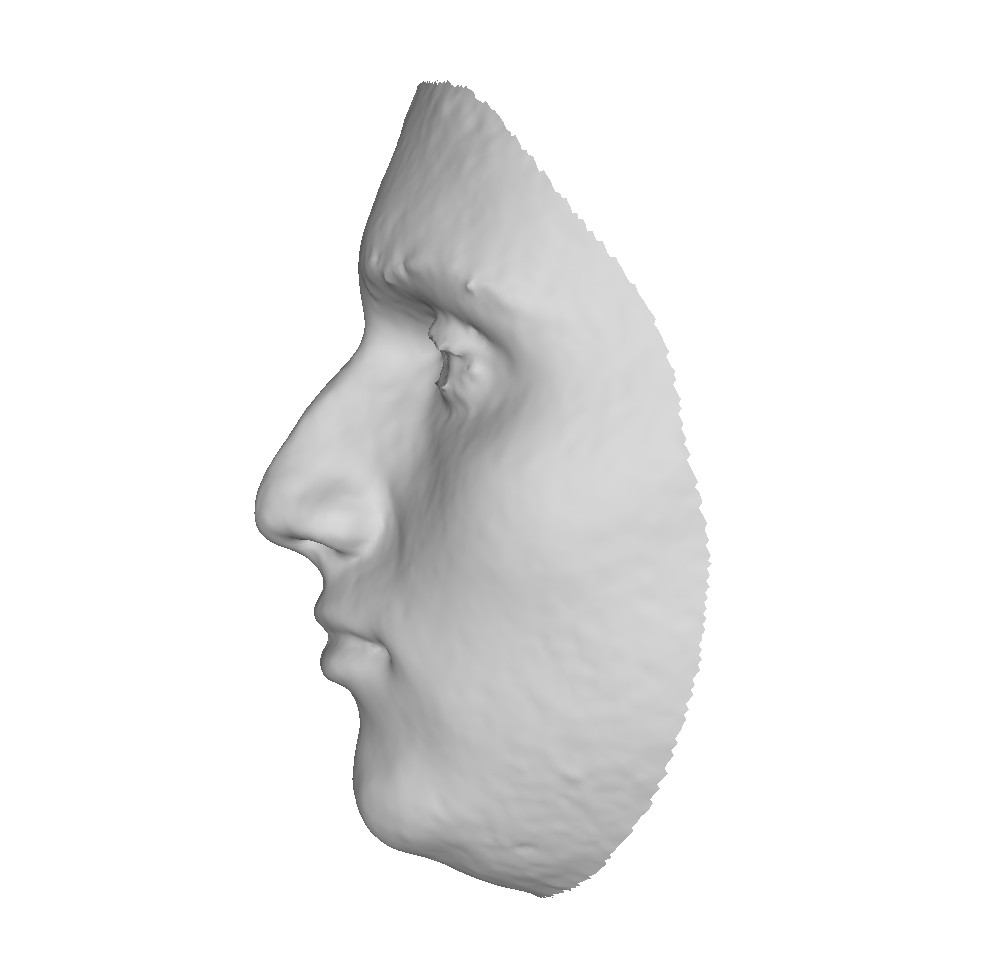} & \nowimg{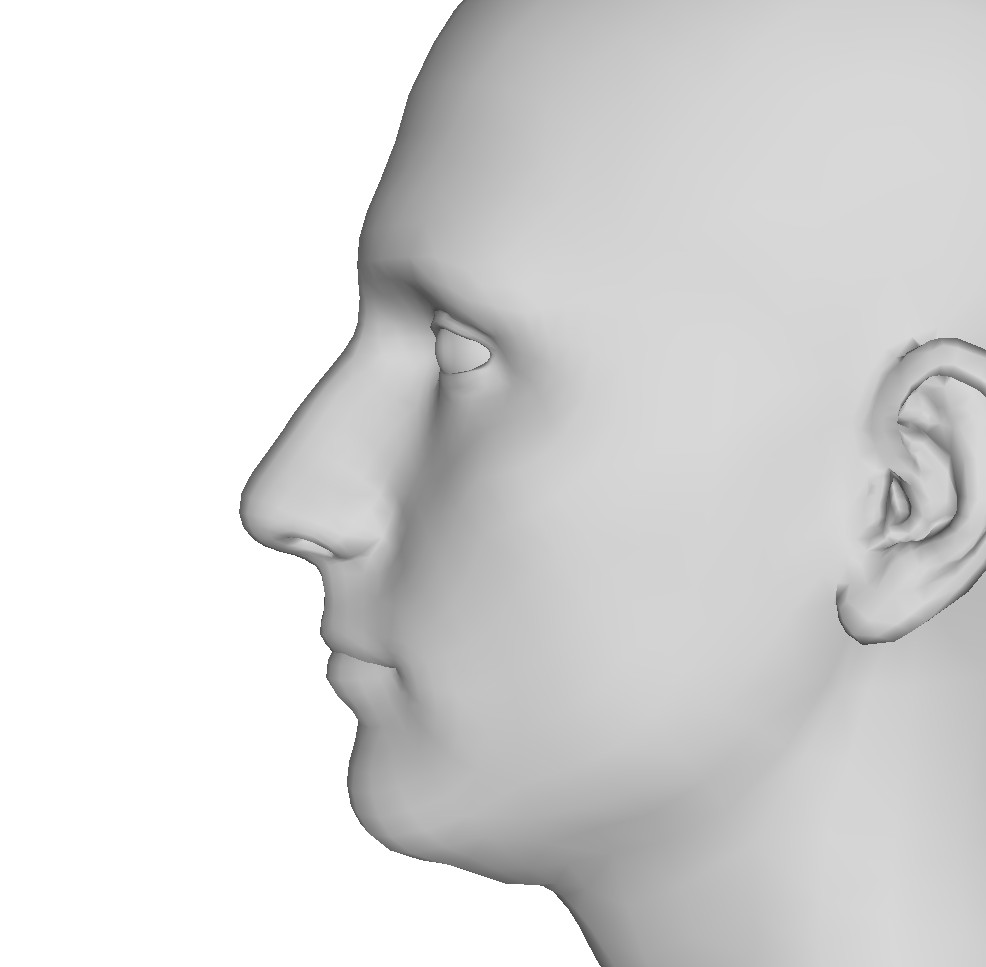} & \nowimg{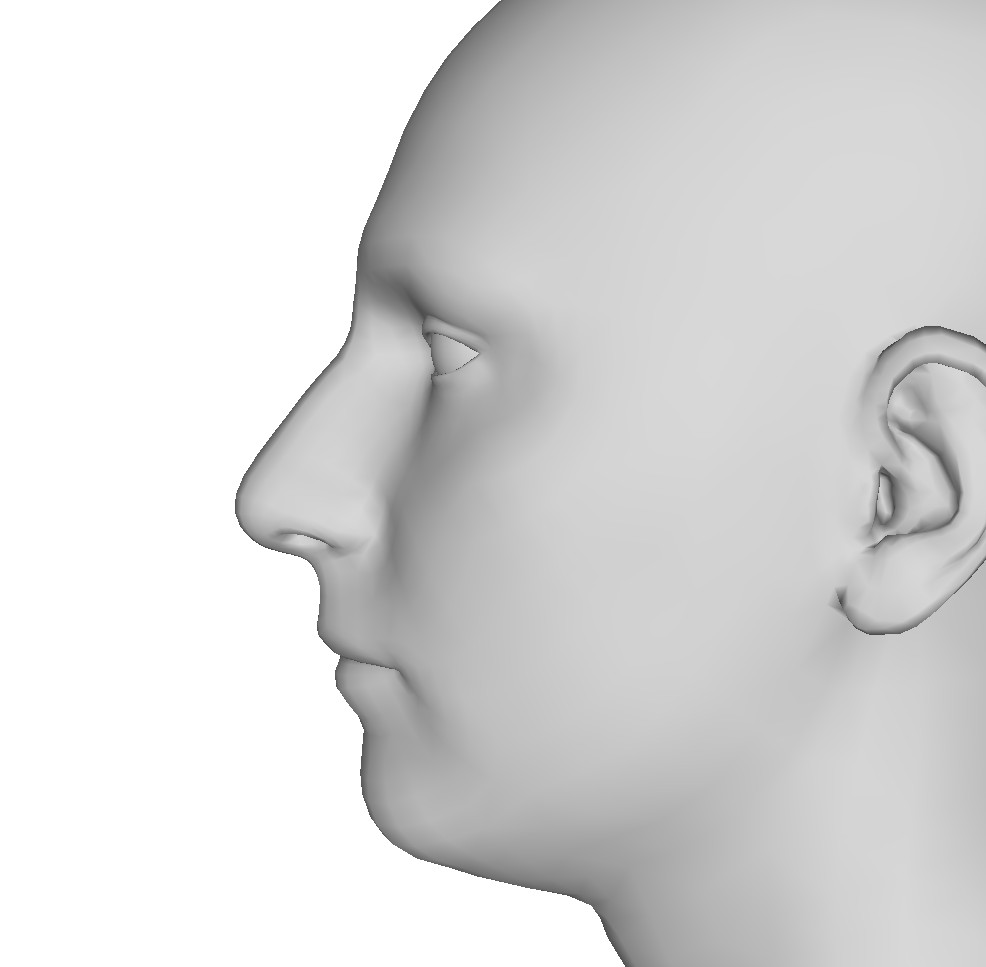} & \nowimg{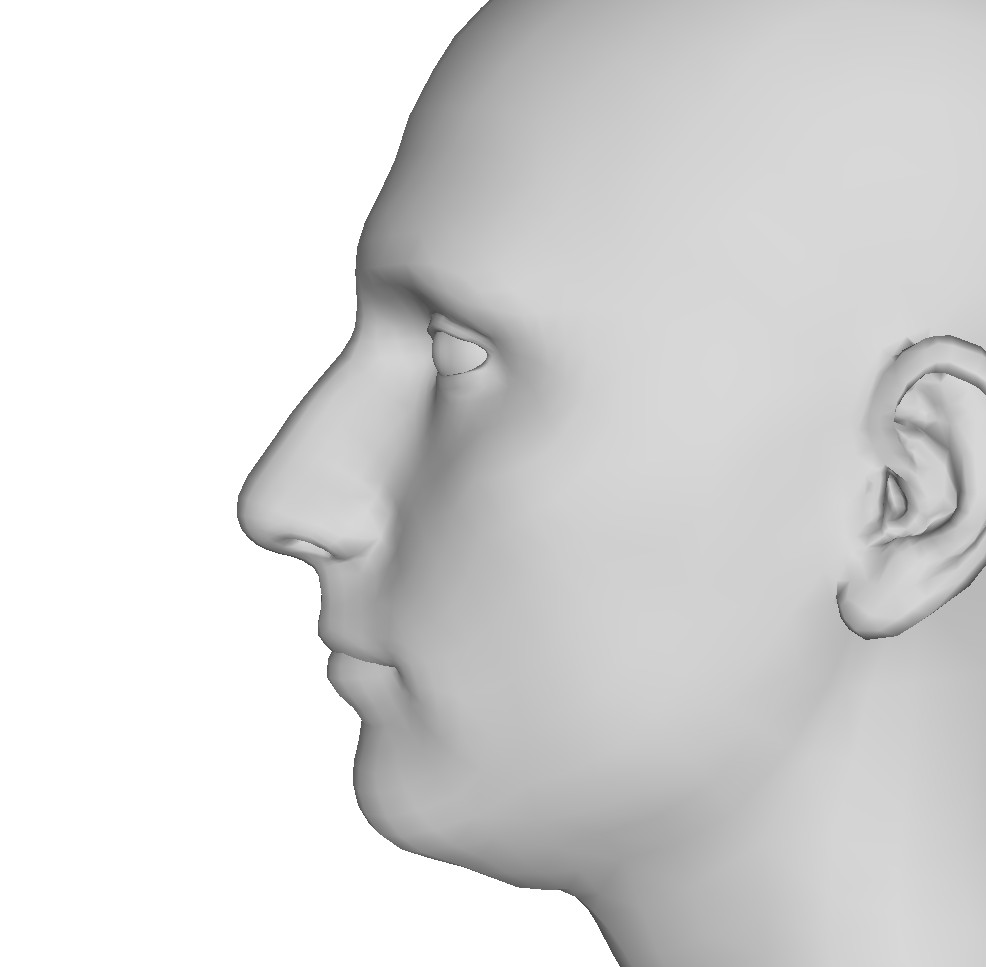}\\

\nowimg{subject_0/0_gt.jpg}  & \nowimg{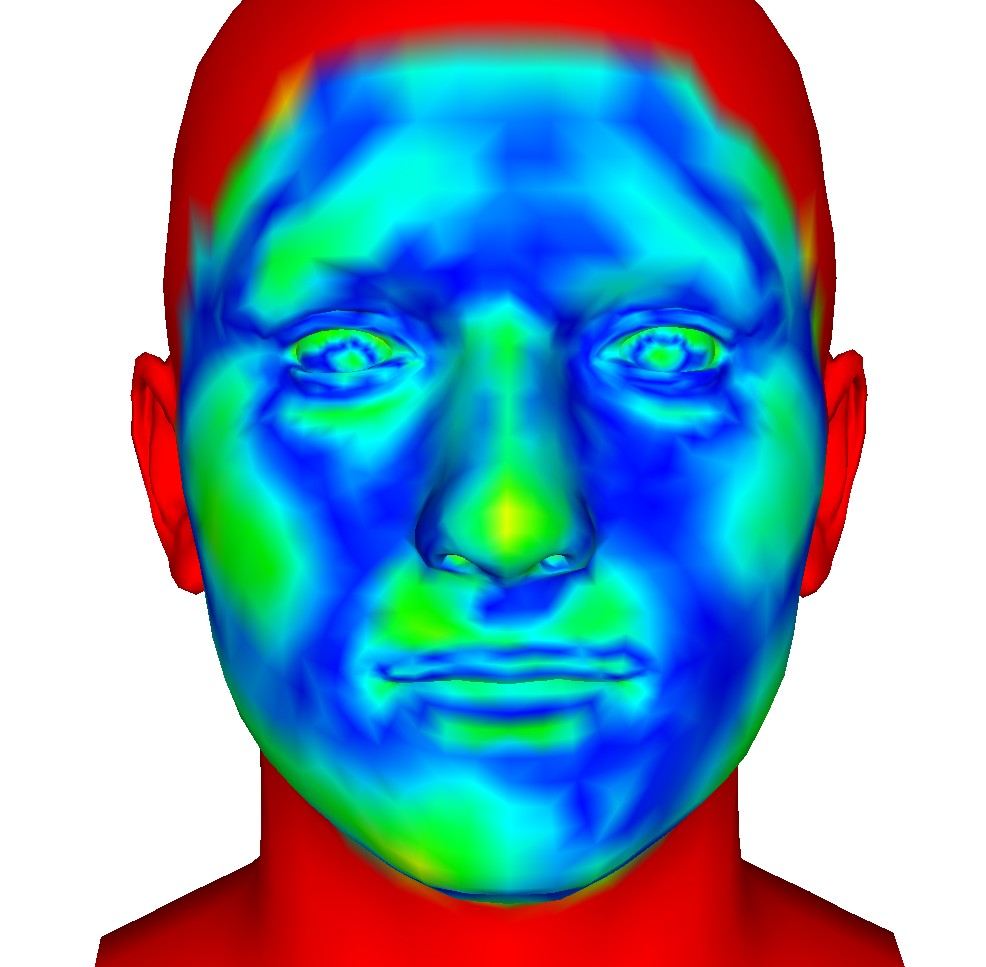} & \nowimg{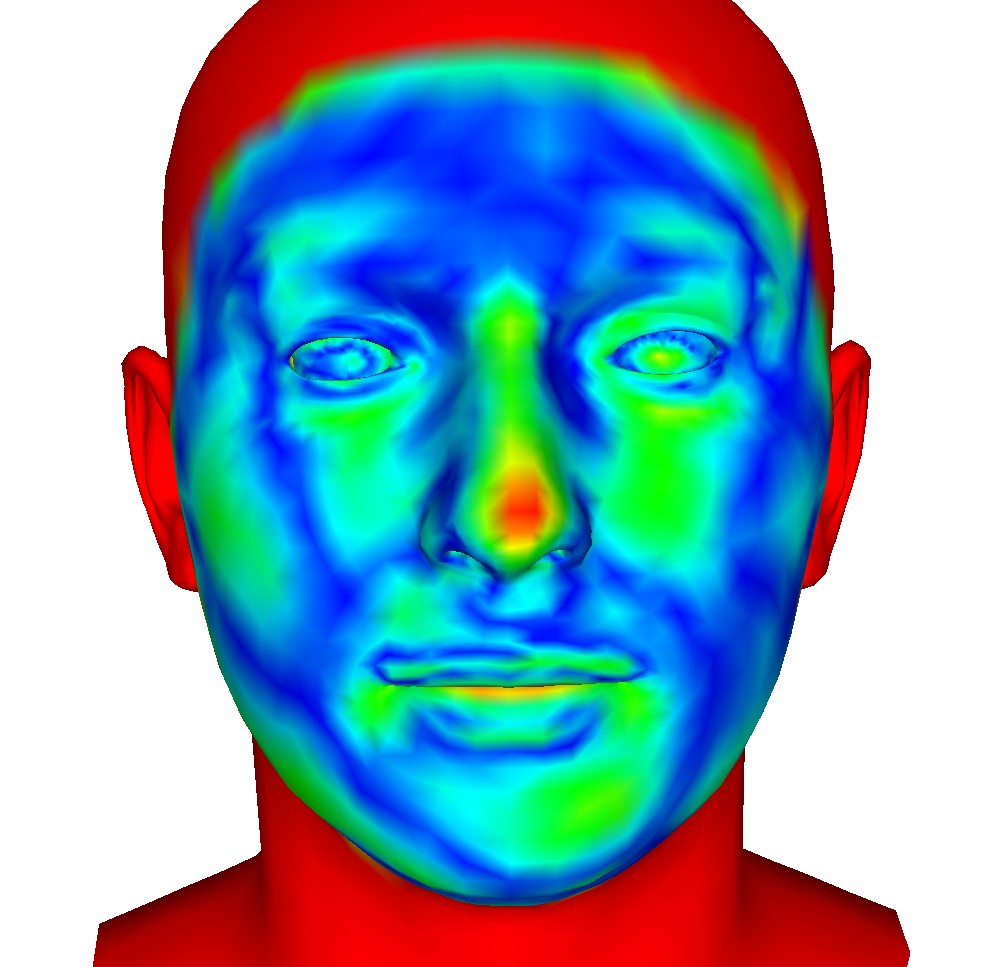} & \nowimg{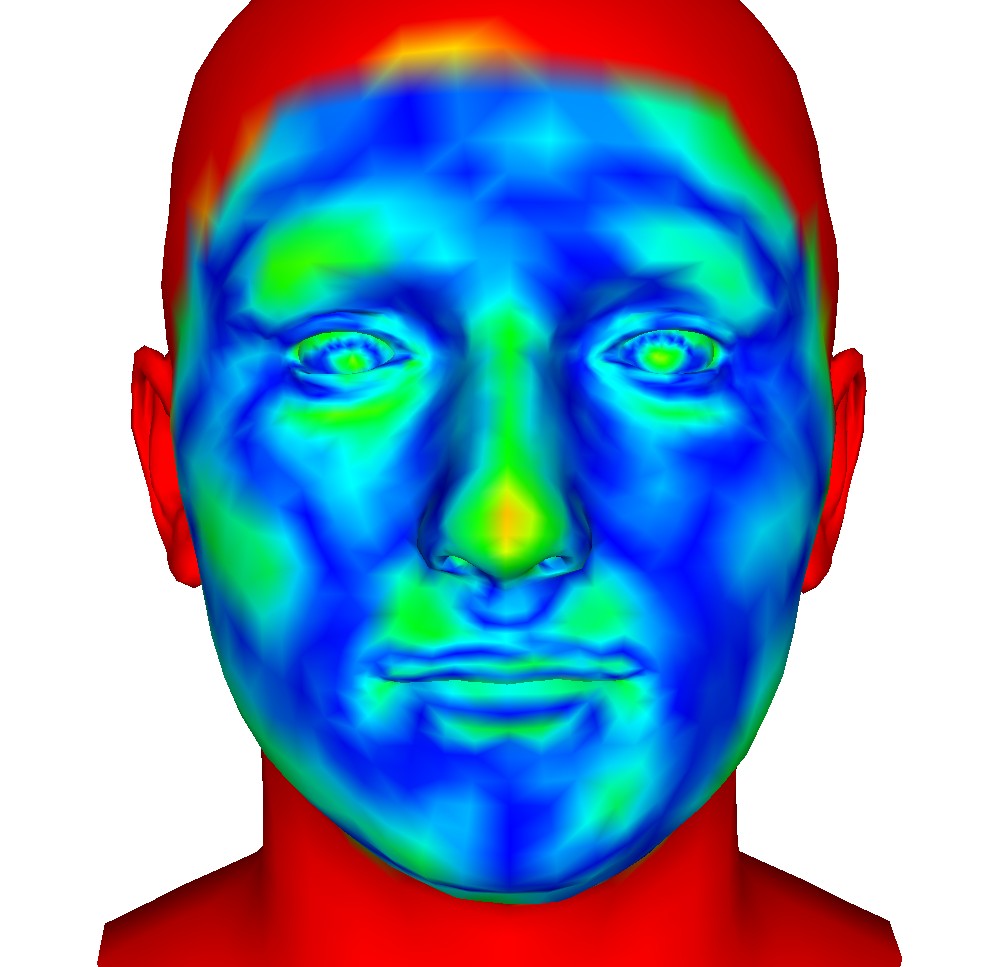} \\[0.15cm]

GT & MICA only & w/o MICA & Ours \\
&&&\\
\hline
&&&\\
\multicolumn{4}{c}{Input image} \\
\multicolumn{4}{c}{\nowimg{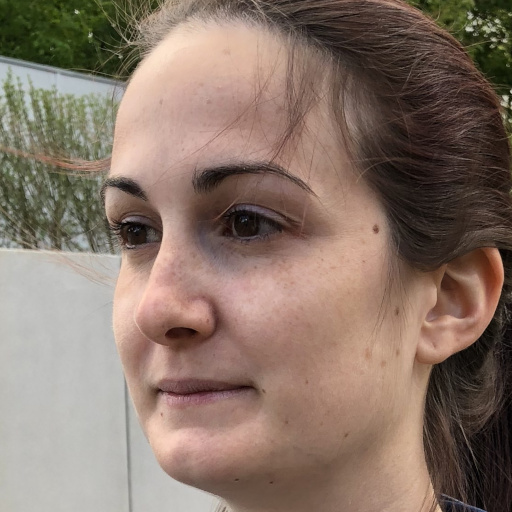}} \\[0.2cm]

\nowimg{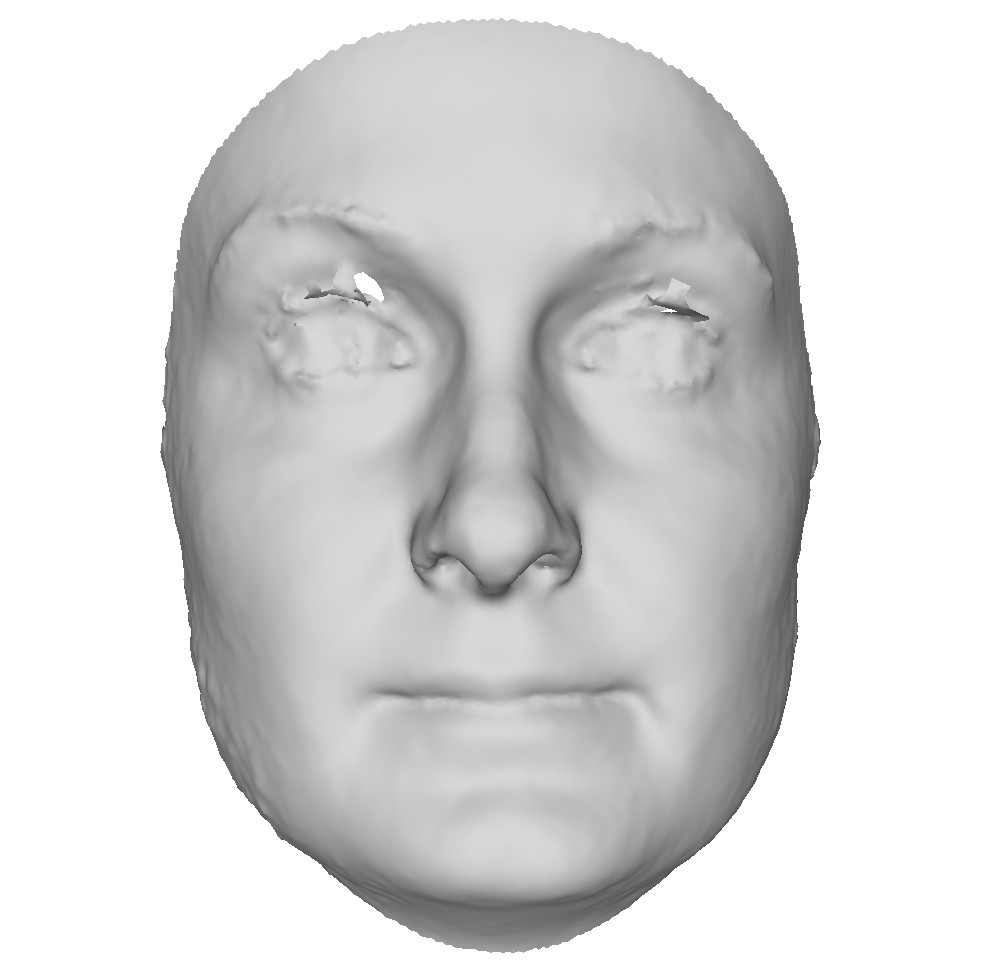} & \nowimg{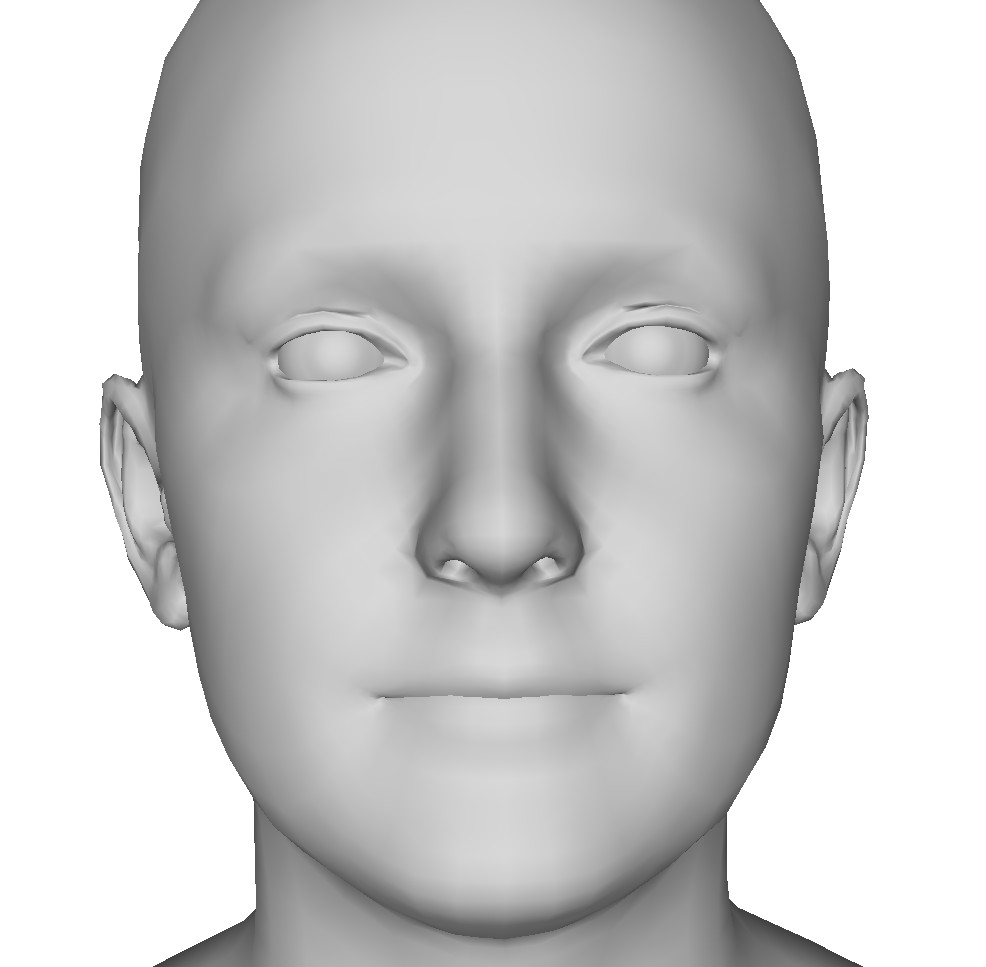} & \nowimg{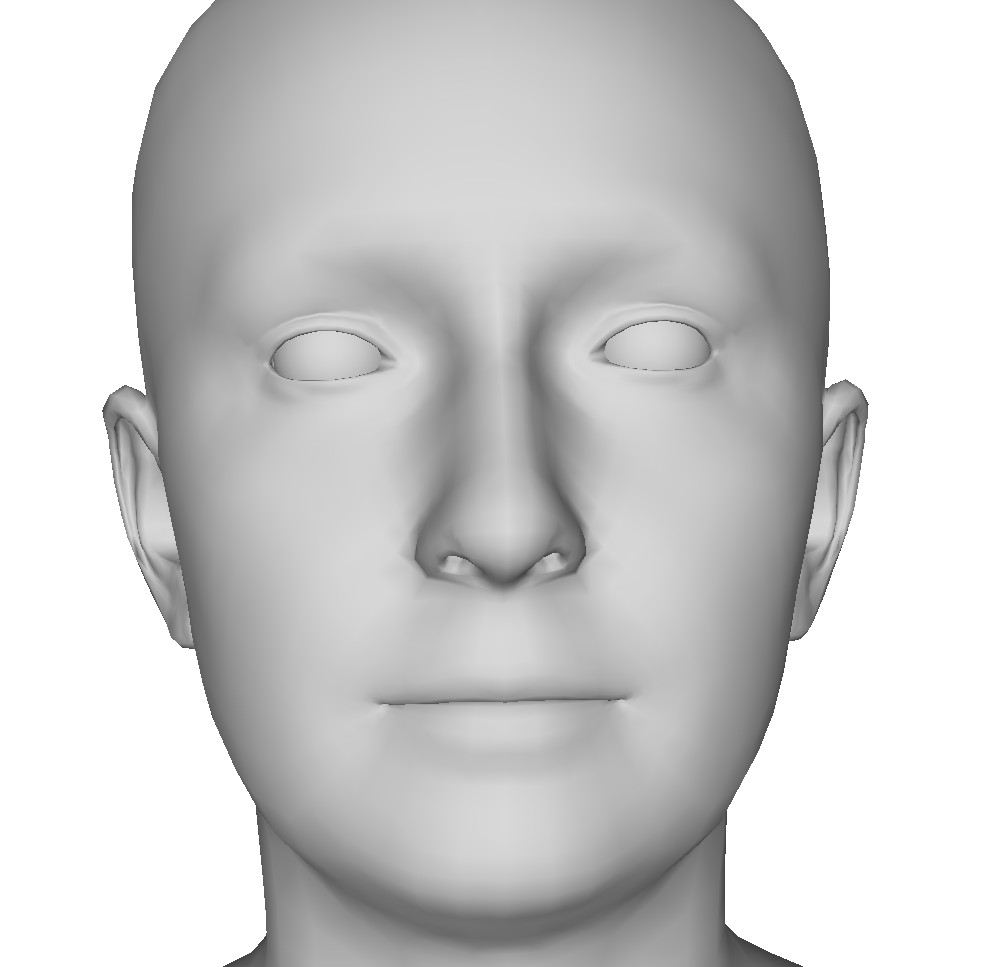} & \nowimg{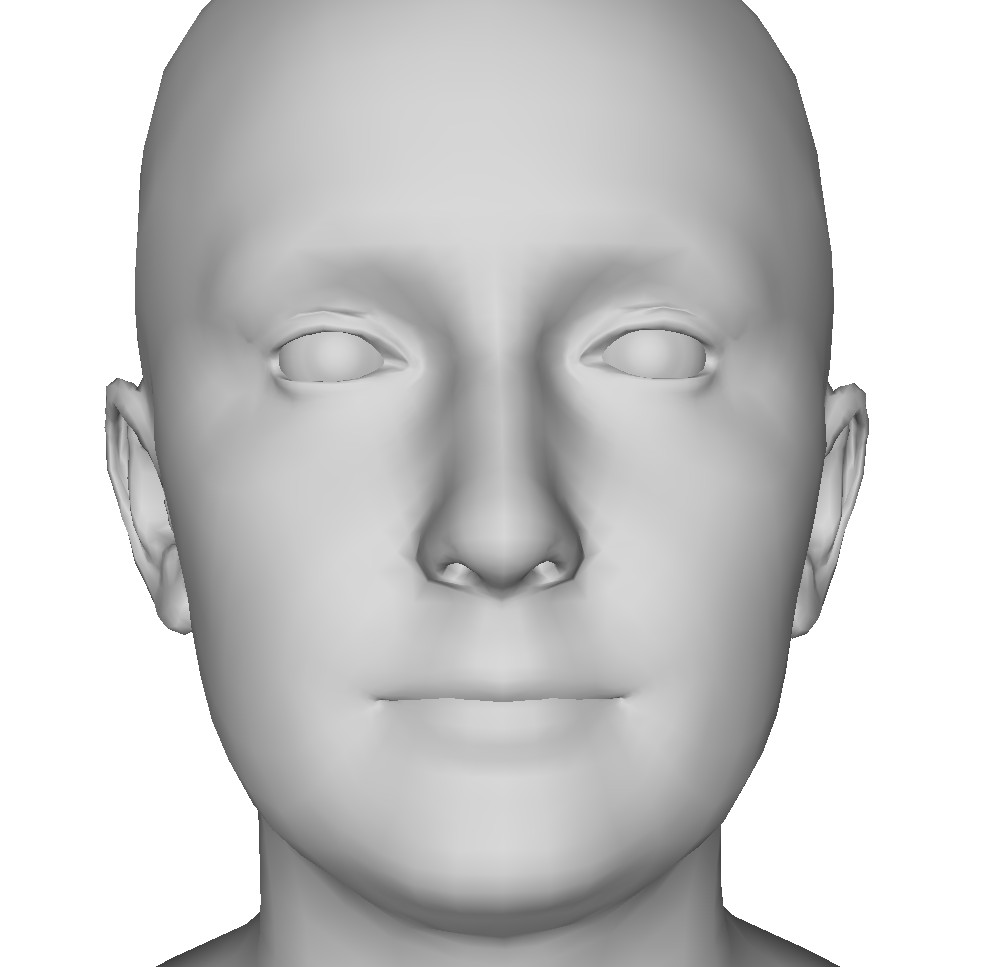}\\

\nowimg{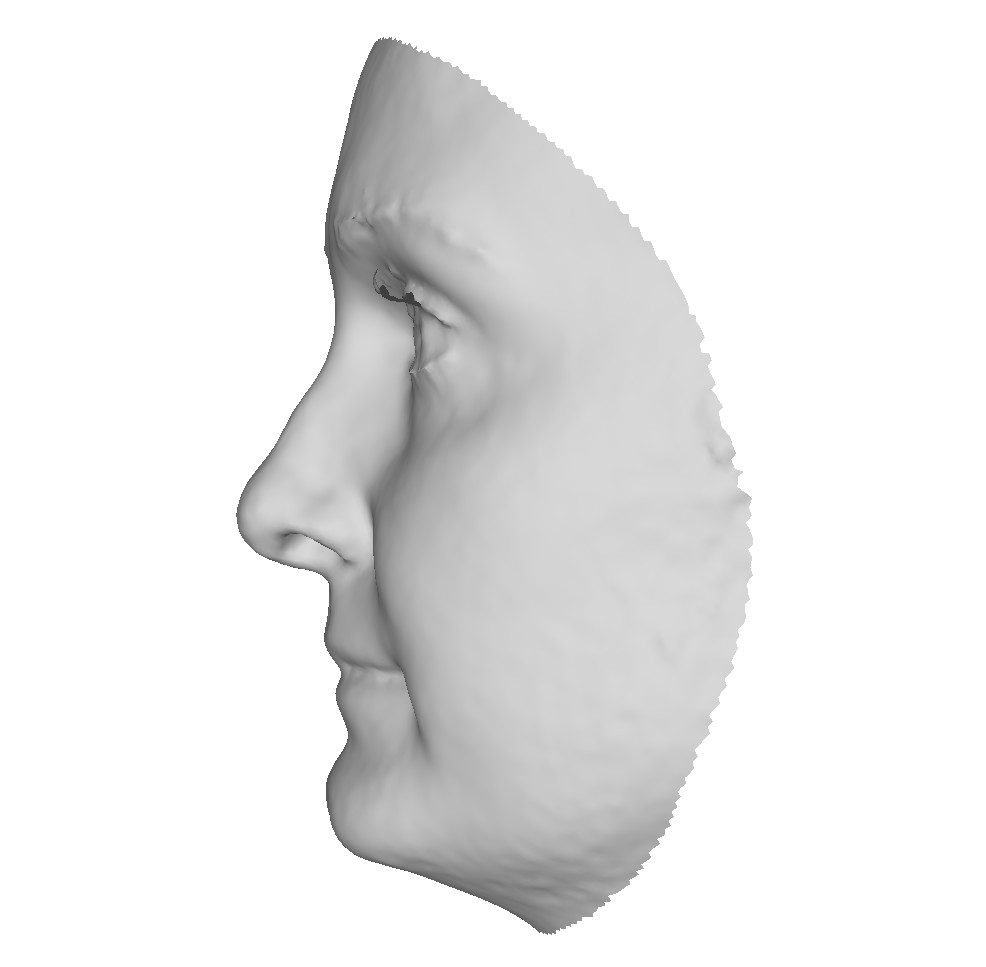} & \nowimg{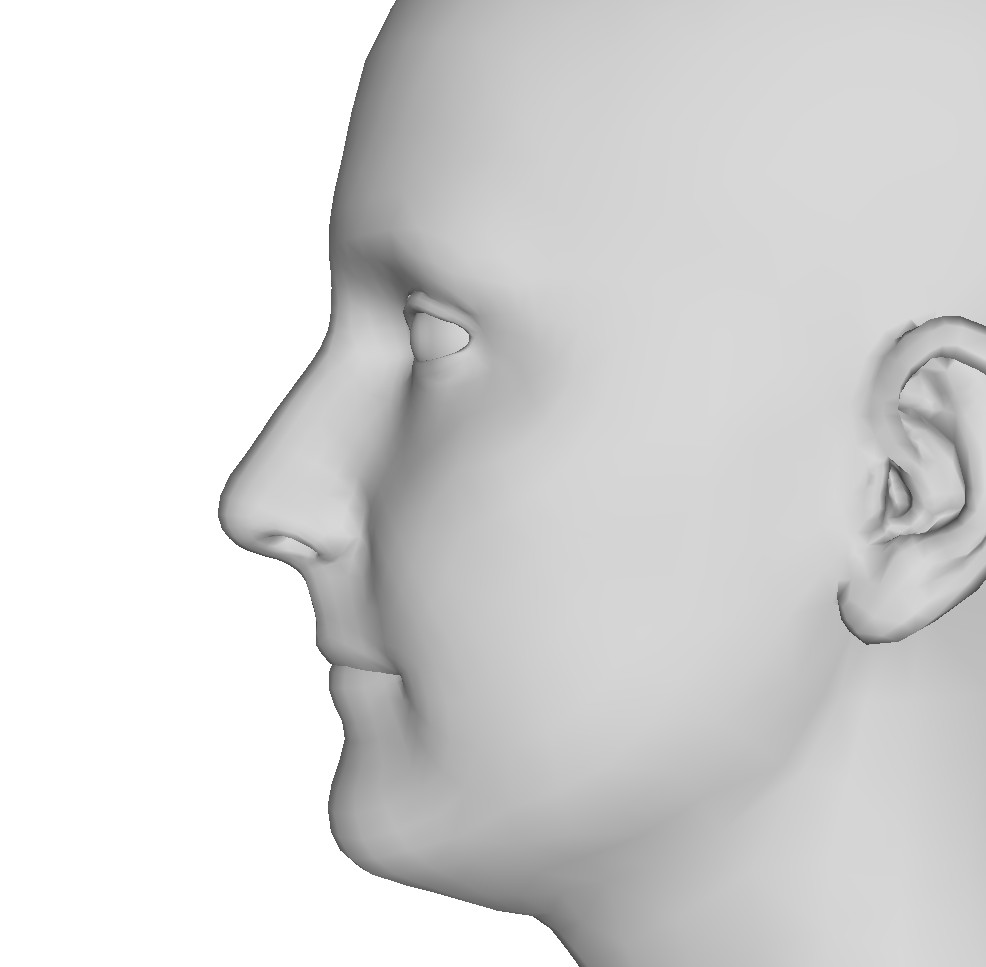} & \nowimg{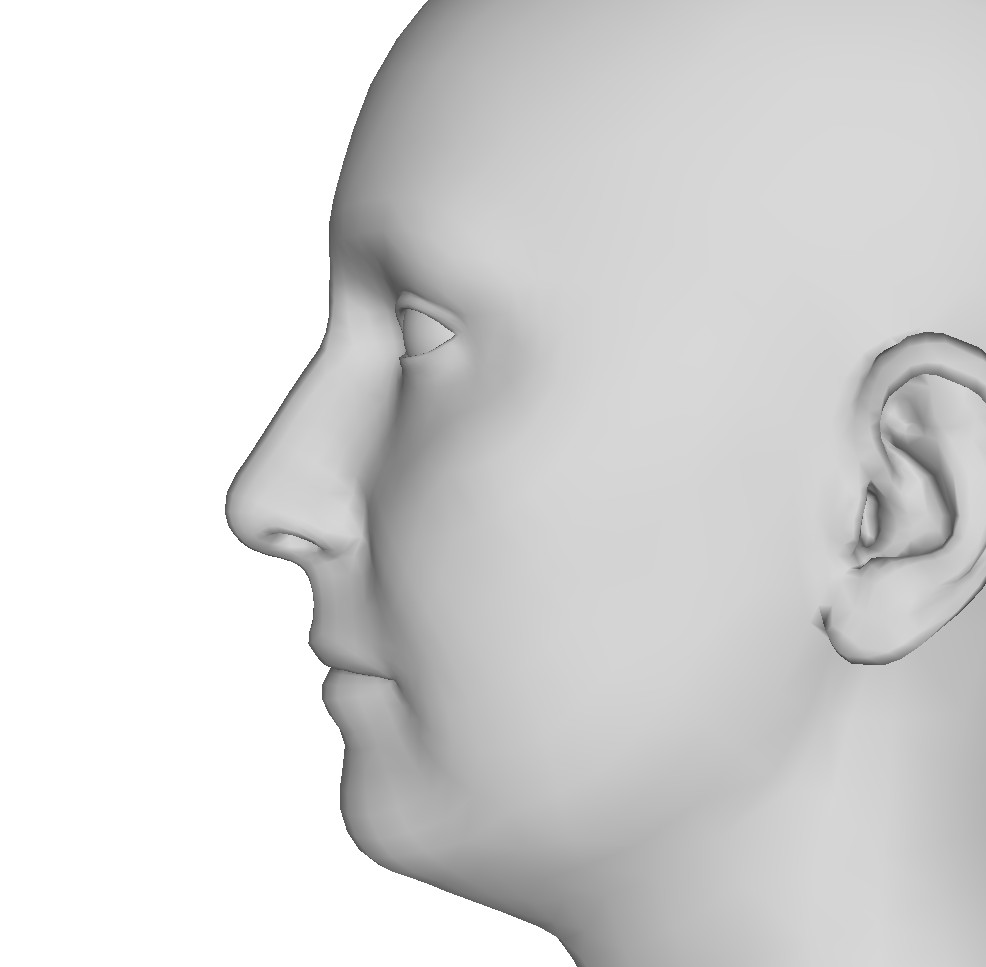} & \nowimg{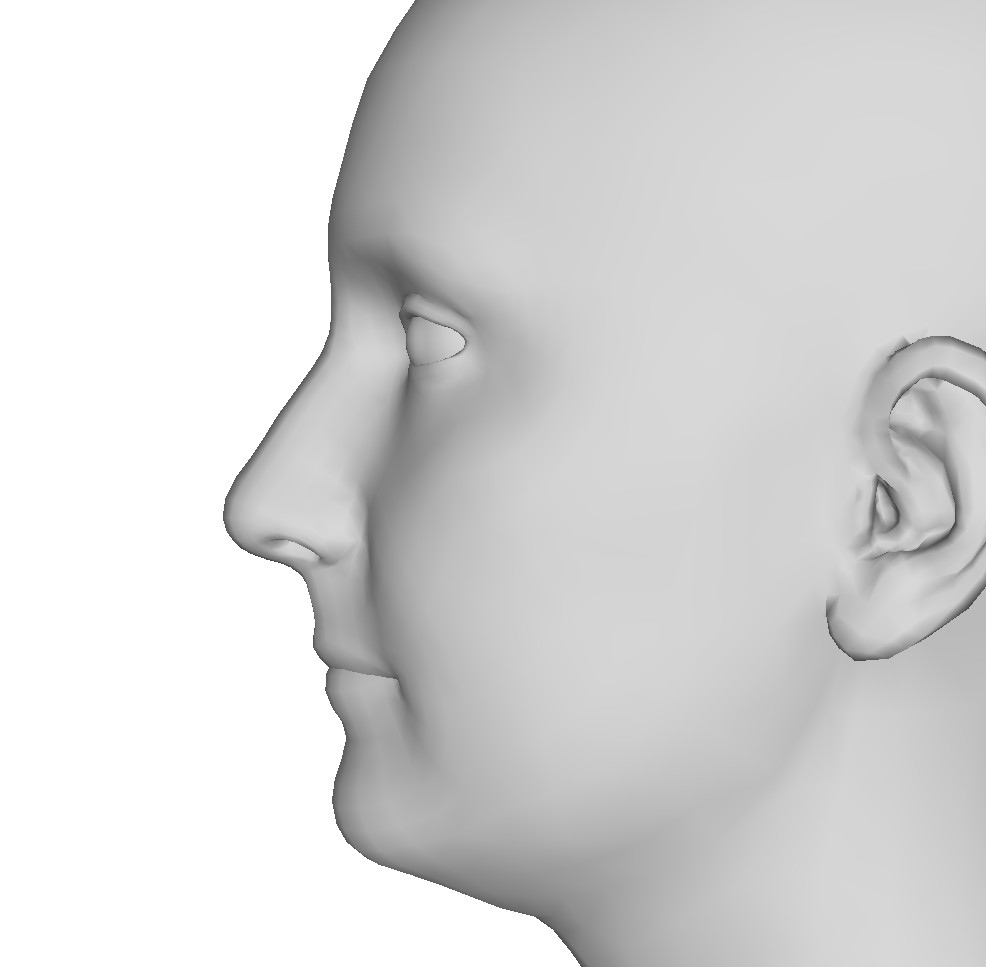}\\

\nowimg{subject_1/1_gt.jpg}  & \nowimg{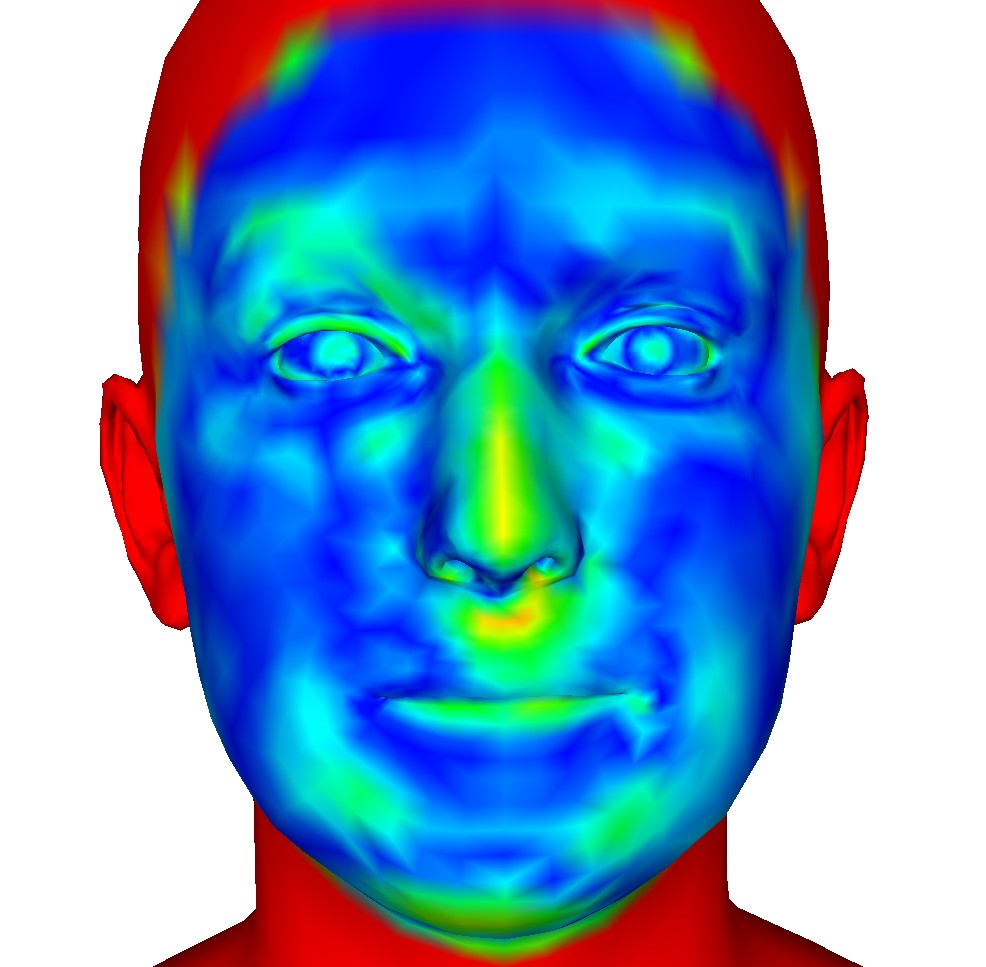} & \nowimg{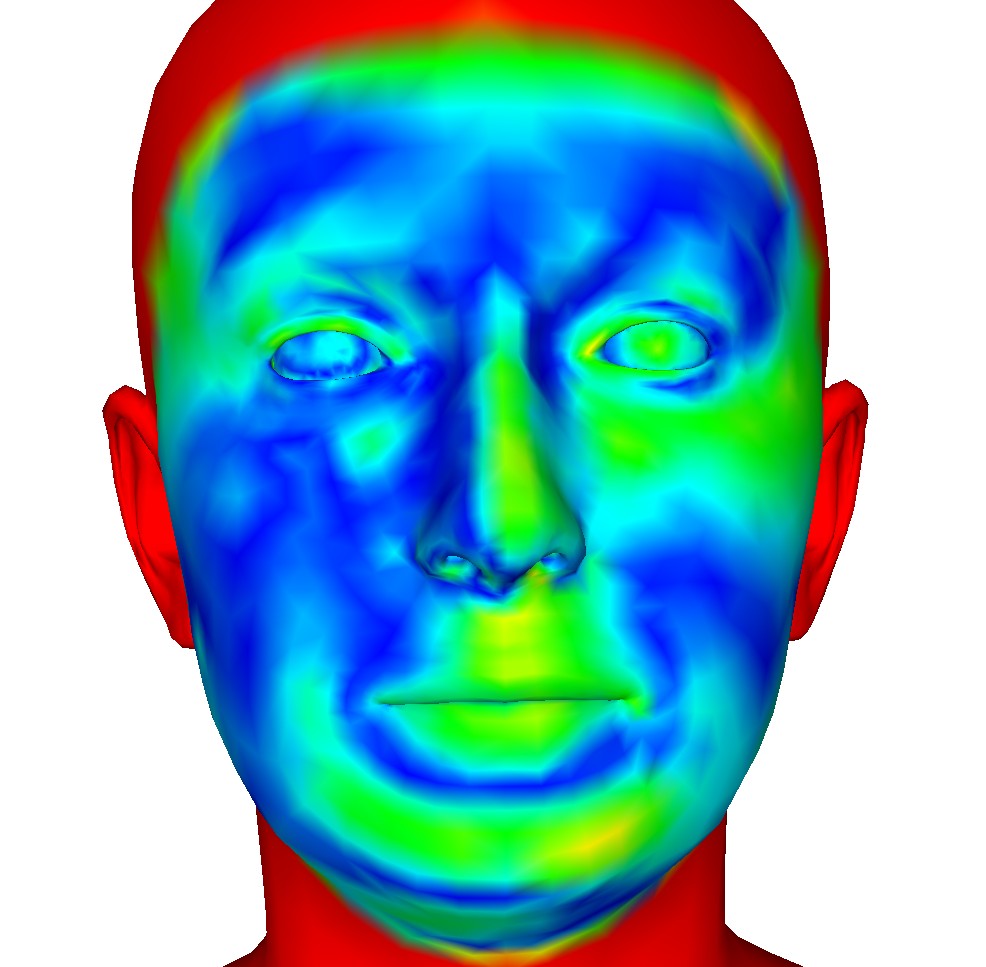} & \nowimg{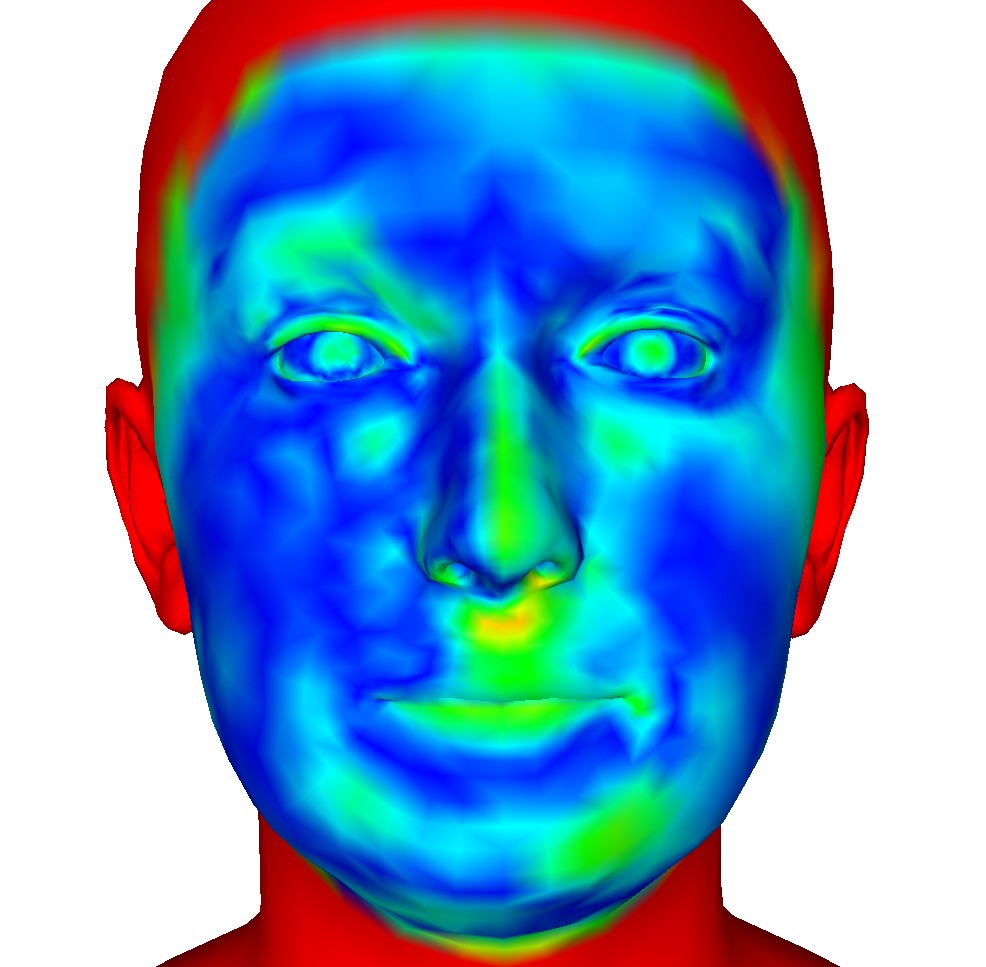} \\[0.3cm]

GT & MICA only & w/o MICA & Ours \\

\end{tabular}
\vspace{0.3cm}
\caption{Ablations of our 3D model fitting module on the NoW validation set (single-view). Figures show qualitative results of MICA predictions (MICA only), without MICA prediction (w/o MICA) and the full model fitting pipeline (Ours). Comparing to the ground truth scan, our full model with MICA template prediction produces more accurate results than without MICA template, which is visible in the 3D visualizations (top two rows) and the error plot (bottom row), where cold colors represent lower error. Our model is also able to improve on the MICA template reconstruction.  }
\label{fig:now_qualitative_single}
\end{figure}

\begin{figure}[p]
\tablefont
\setlength{\tabcolsep}{1pt}
\renewcommand{\arraystretch}{1}
\centering
\begin{tabular}{c c c c }

\multicolumn{4}{c}{Input images} \\
\multicolumn{4}{c}{ \begin{tabular}{c c c}
    \nowimg{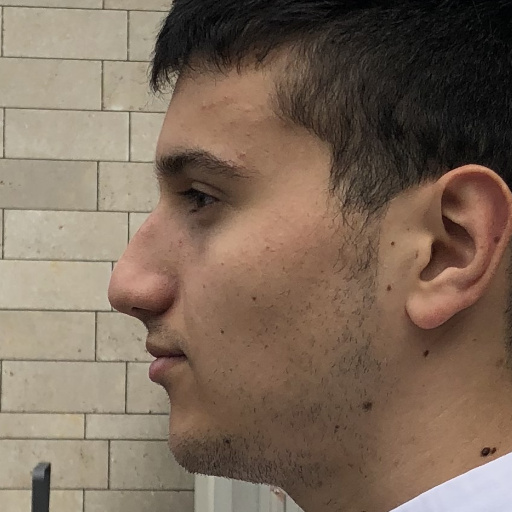} &   \nowimg{subject_0/img_1.jpg} & \nowimg{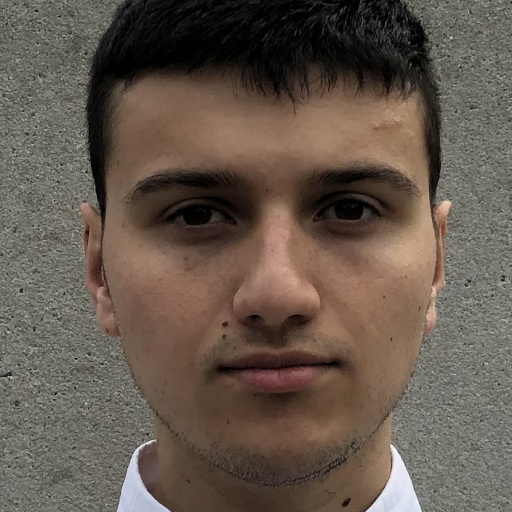}
\end{tabular} } \\[1.0cm]

\nowimg{subject_0/0_gt.jpg} & \nowimg{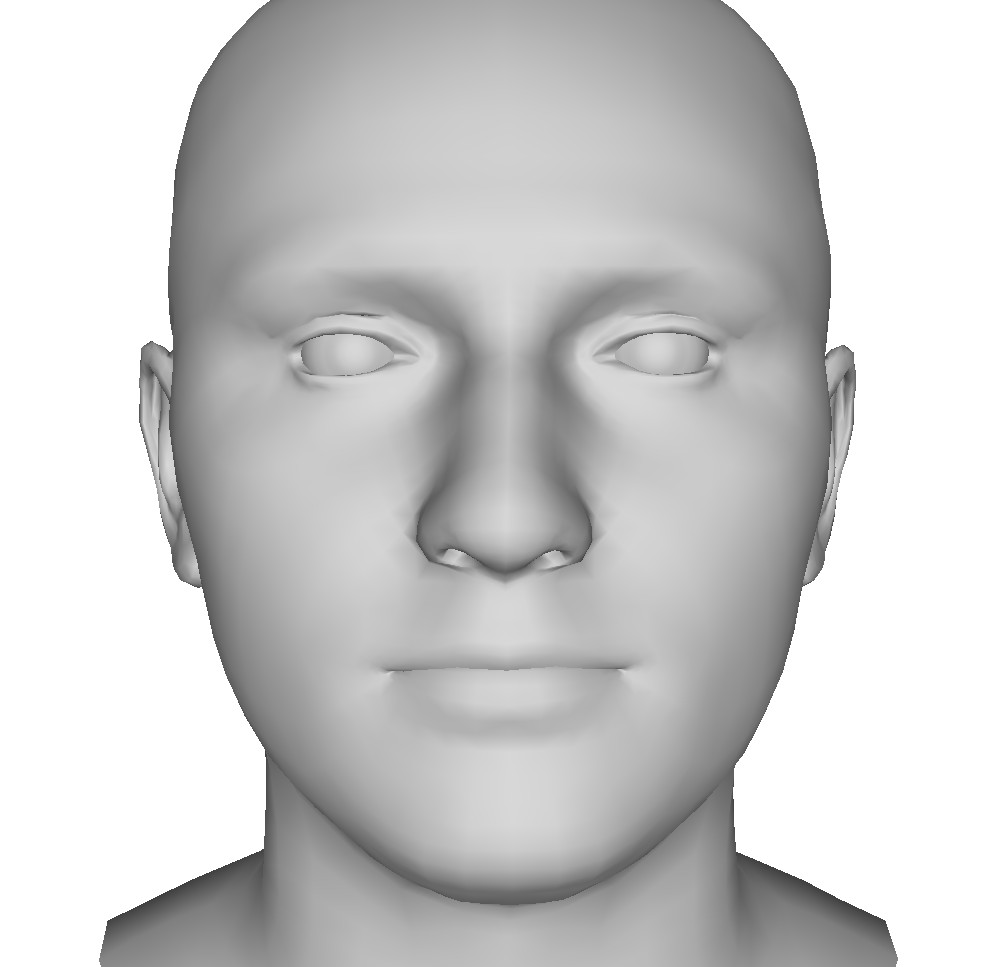} & \nowimg{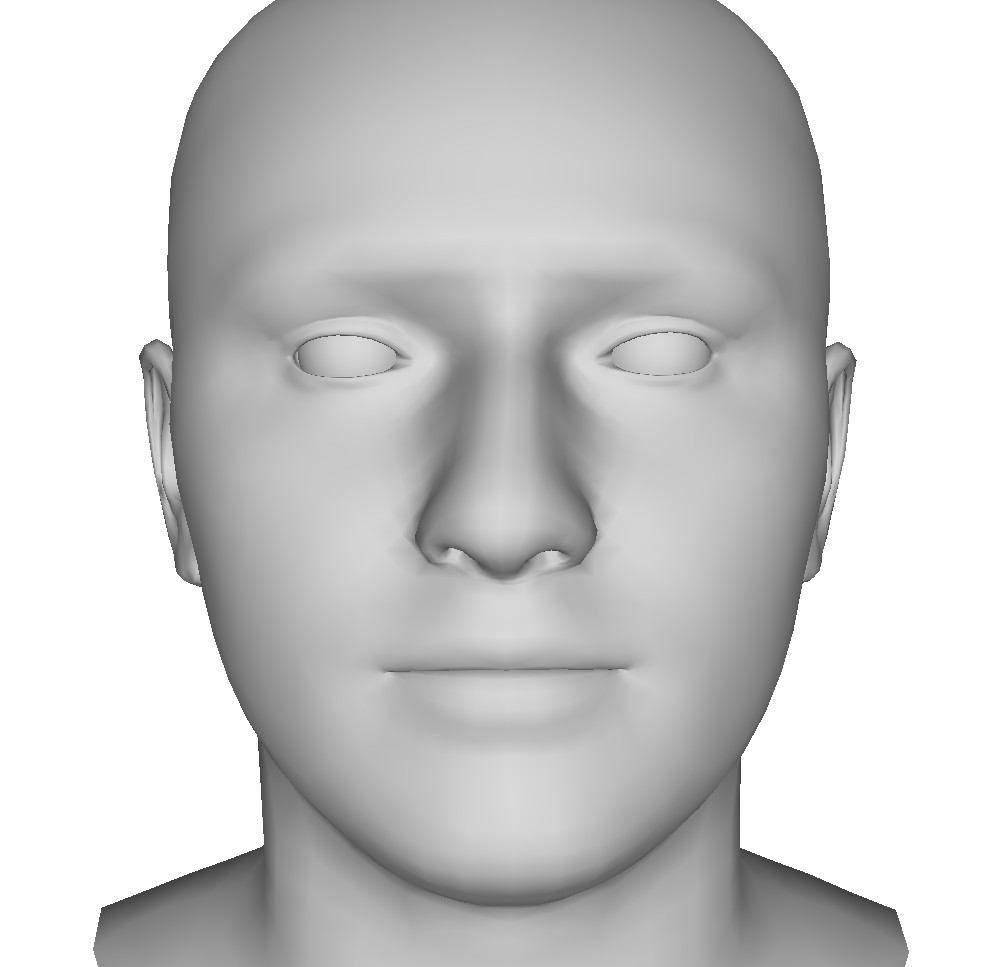} & \nowimg{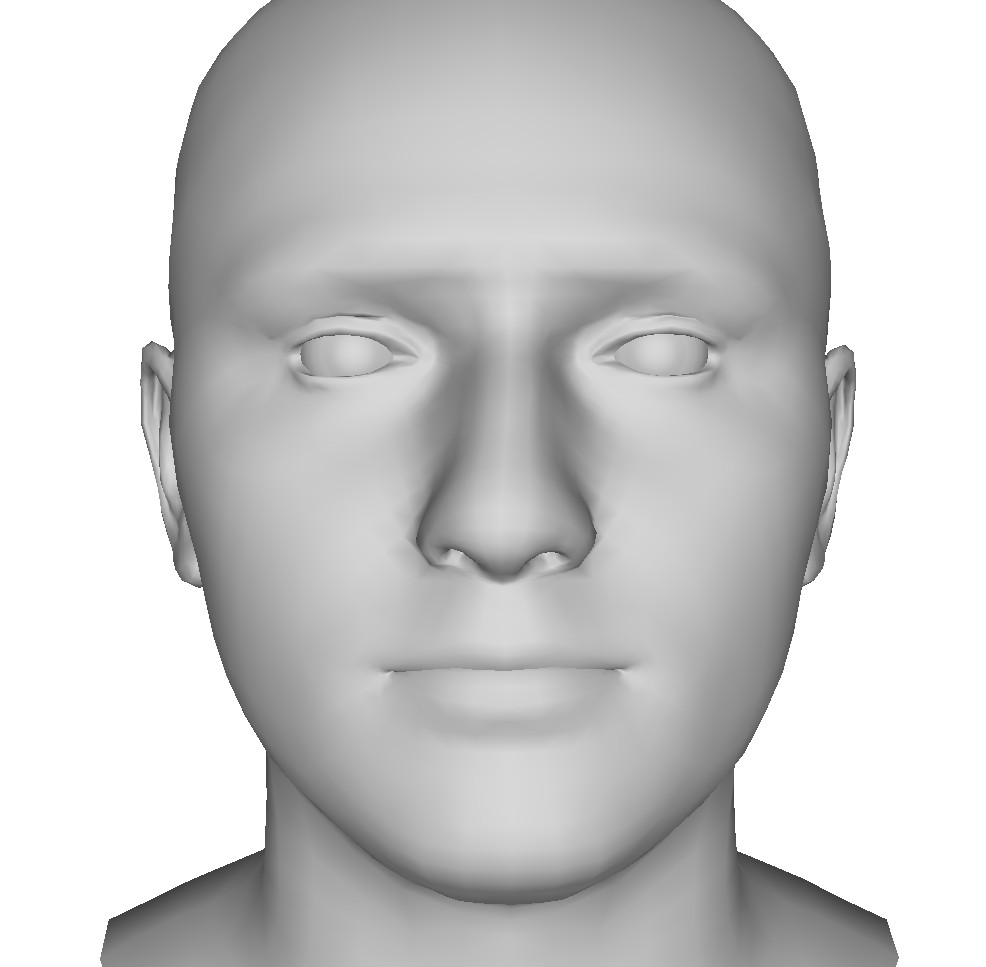}\\

\nowimg{subject_0/0_gt_1.jpg} & \nowimg{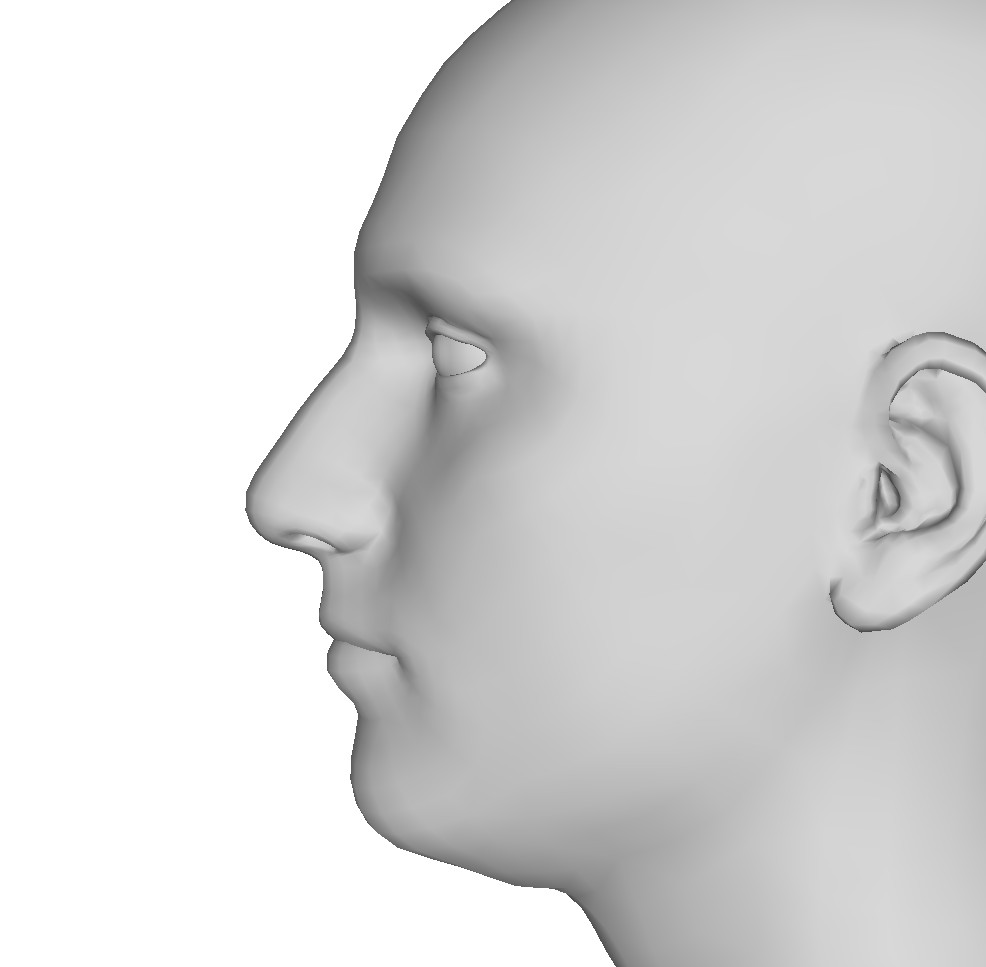} & \nowimg{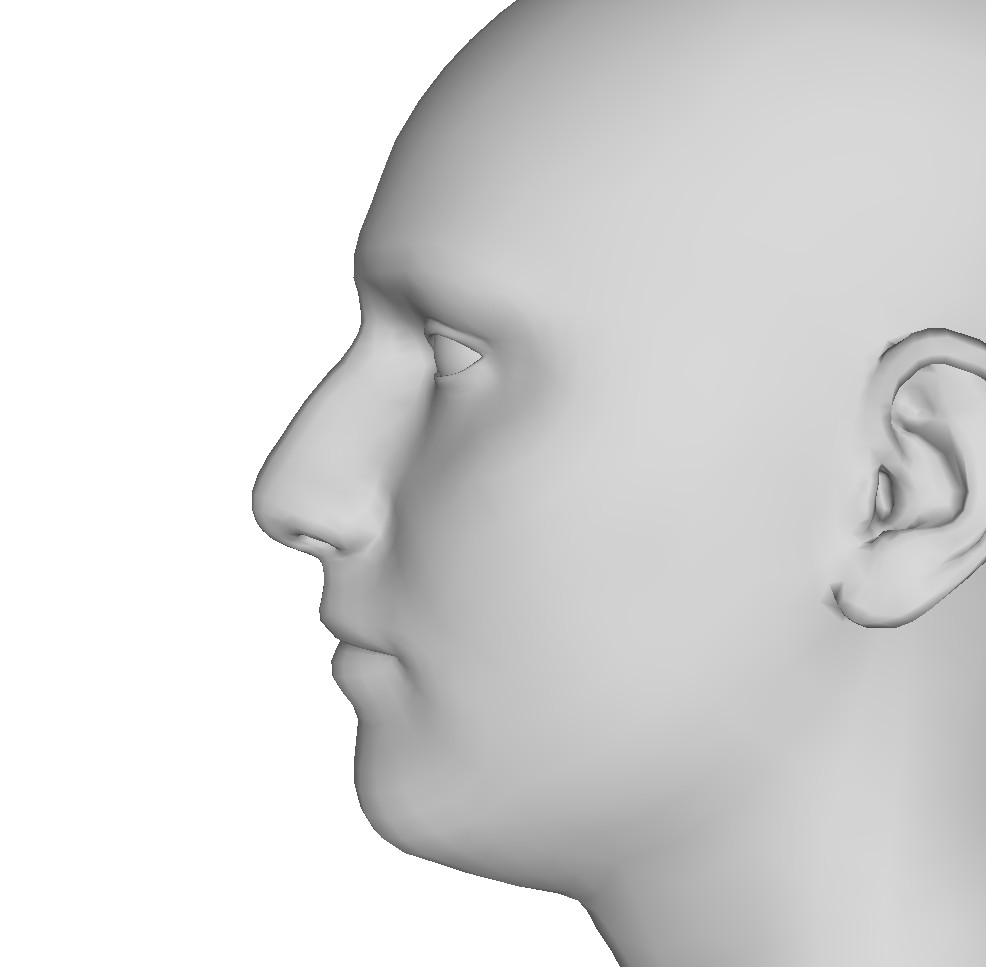} & \nowimg{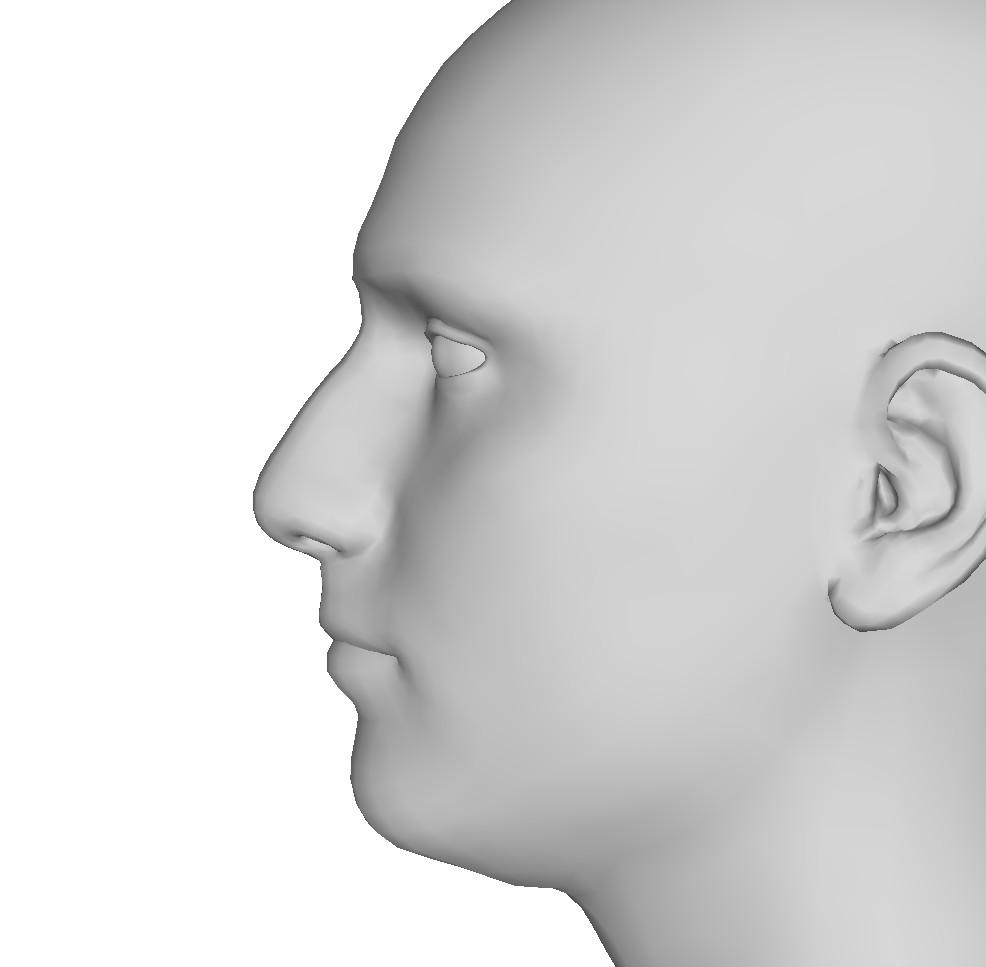}\\

\nowimg{subject_0/0_gt.jpg}  & \nowimg{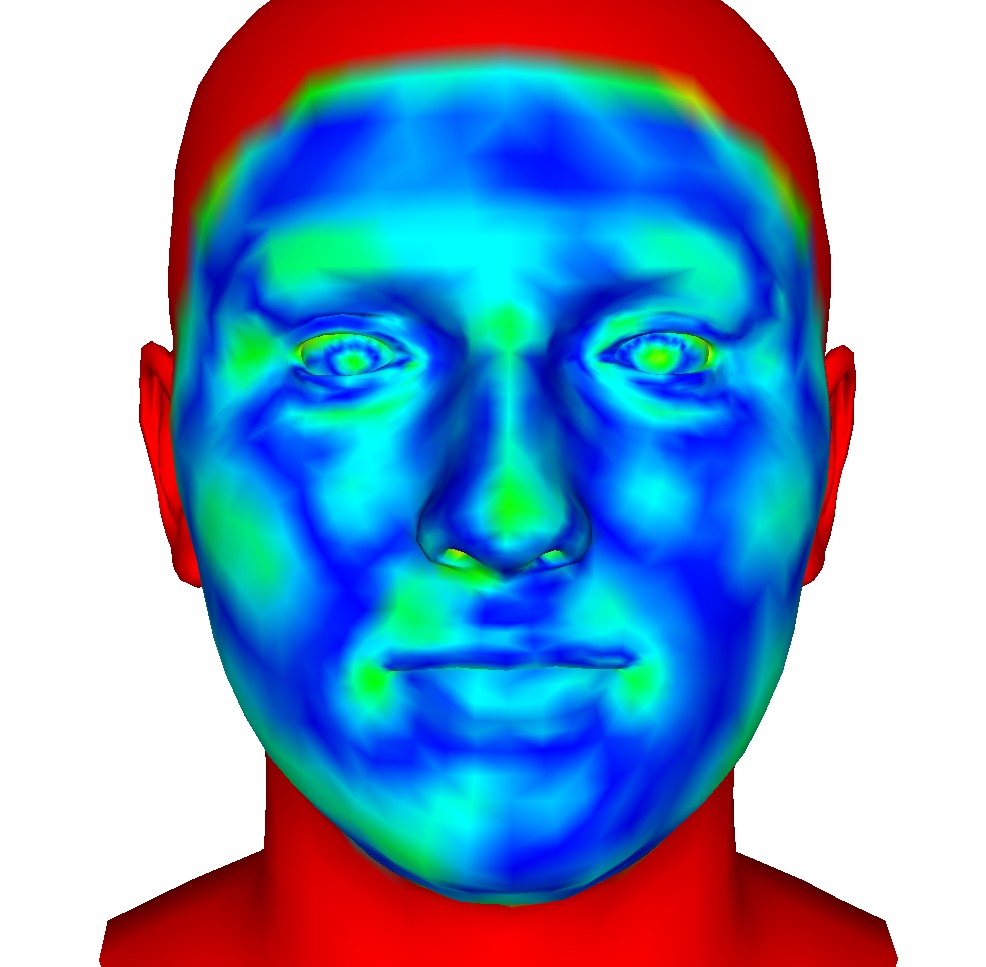} & \nowimg{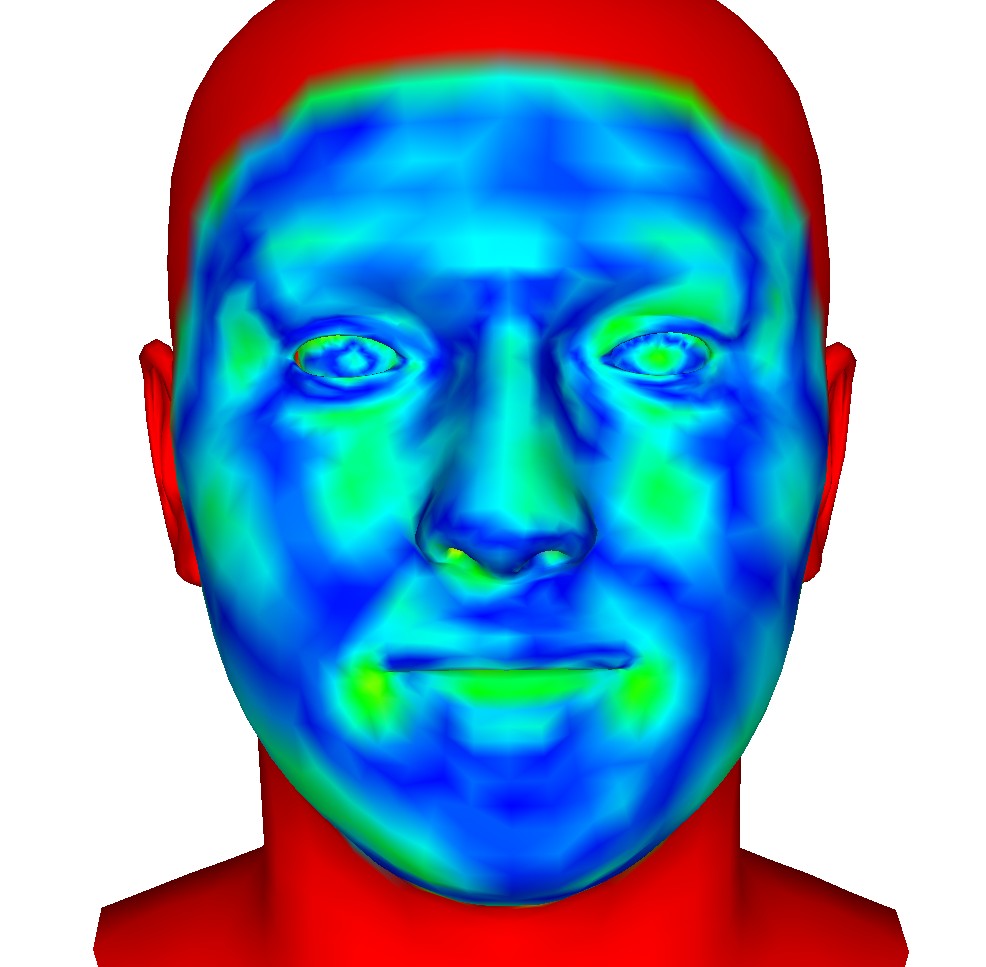} & \nowimg{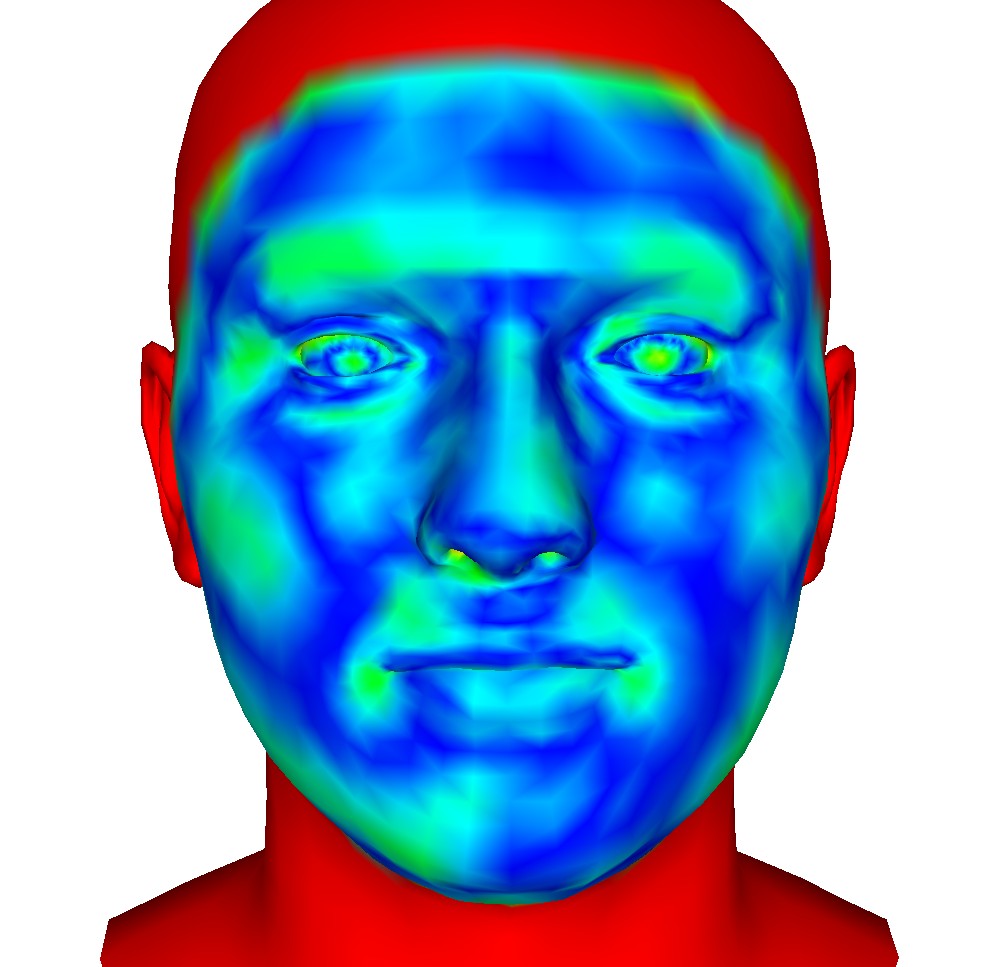} \\[0.15cm]

GT & w/o $\delta_d$ & w/o MICA & Ours \\
&&&\\
\hline
&&&\\
\multicolumn{4}{c}{Input images} \\
\multicolumn{4}{c}{ \begin{tabular}{c c c}
    \nowimg{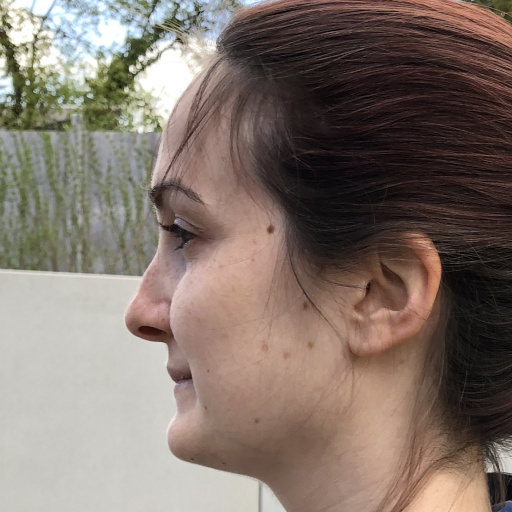} &   \nowimg{subject_1/img_1.jpg} & \nowimg{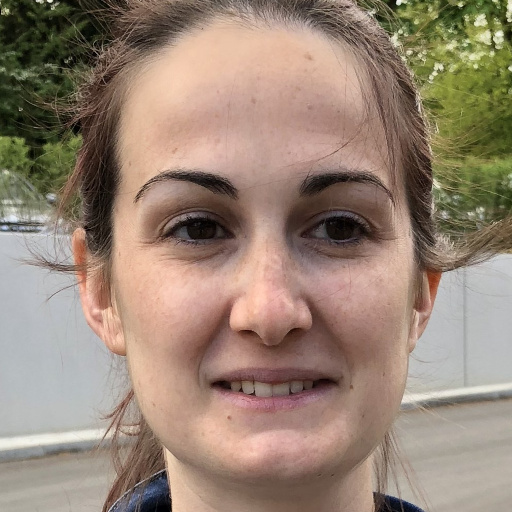}
\end{tabular} } \\[1.0cm]

\nowimg{subject_1/1_gt.jpg} & \nowimg{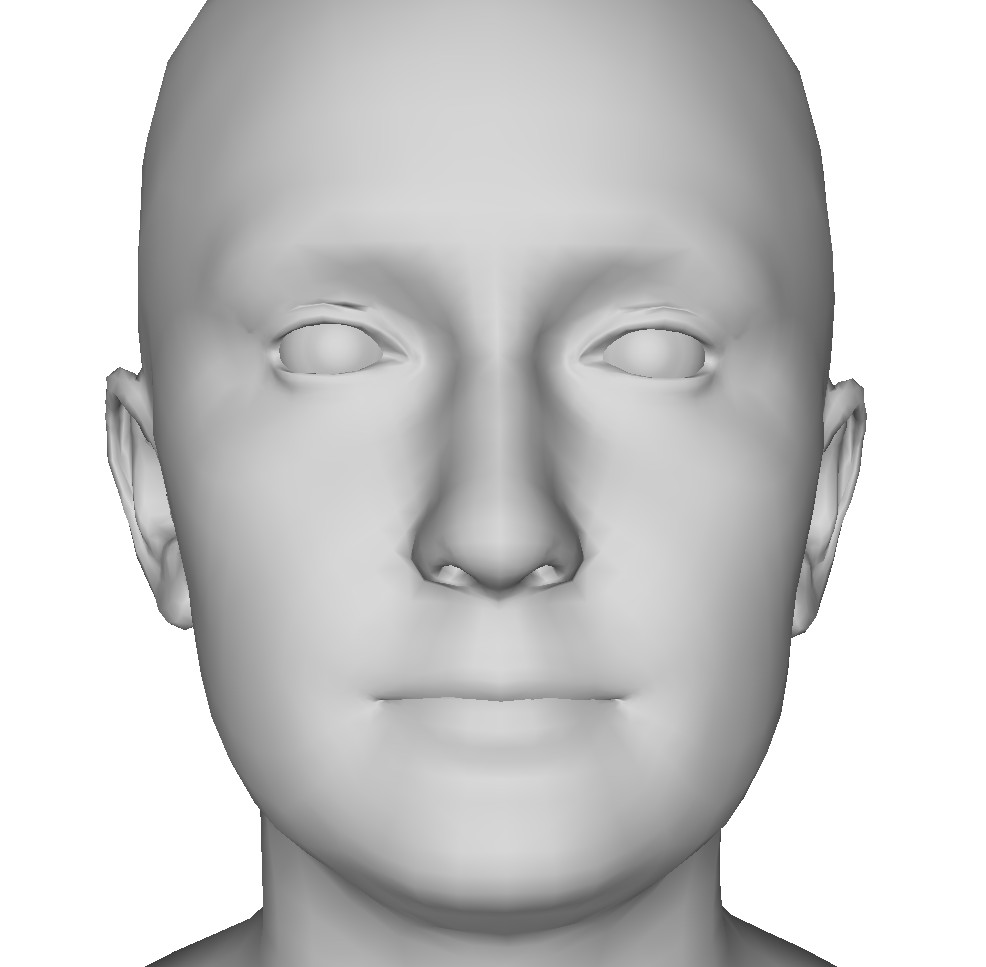} & \nowimg{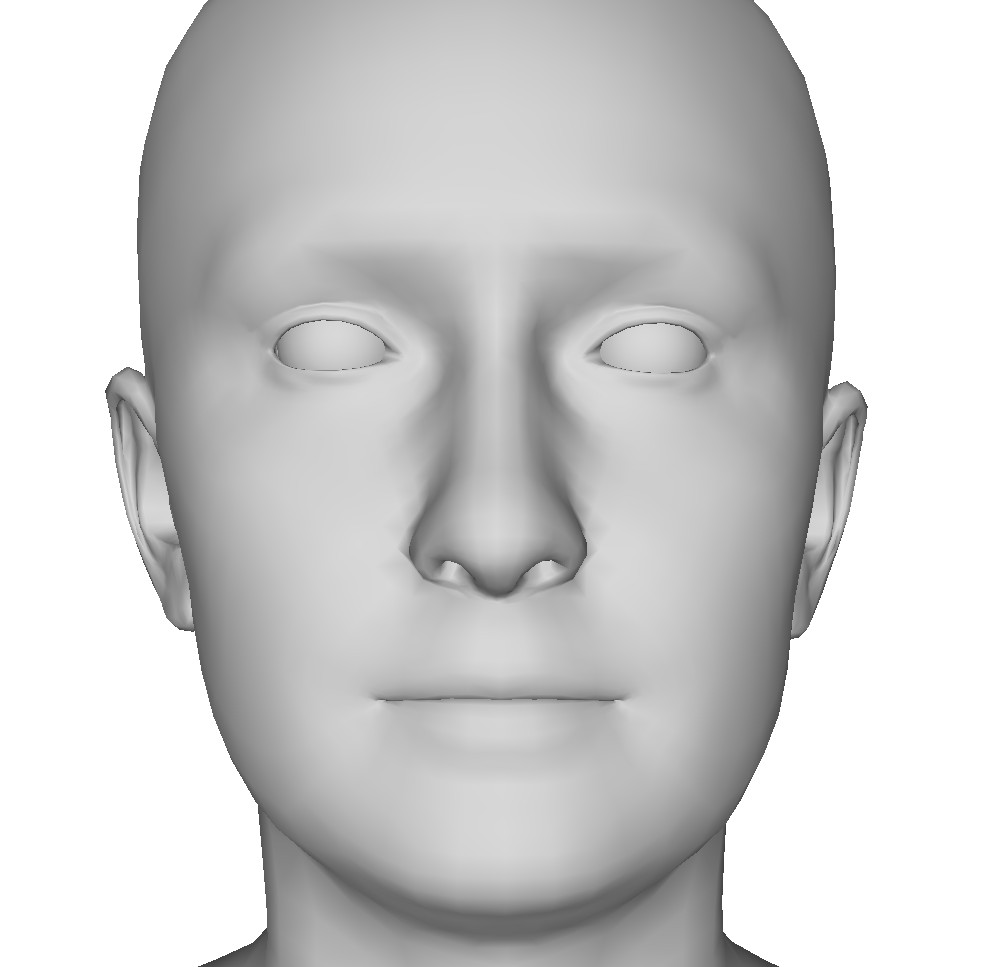} & \nowimg{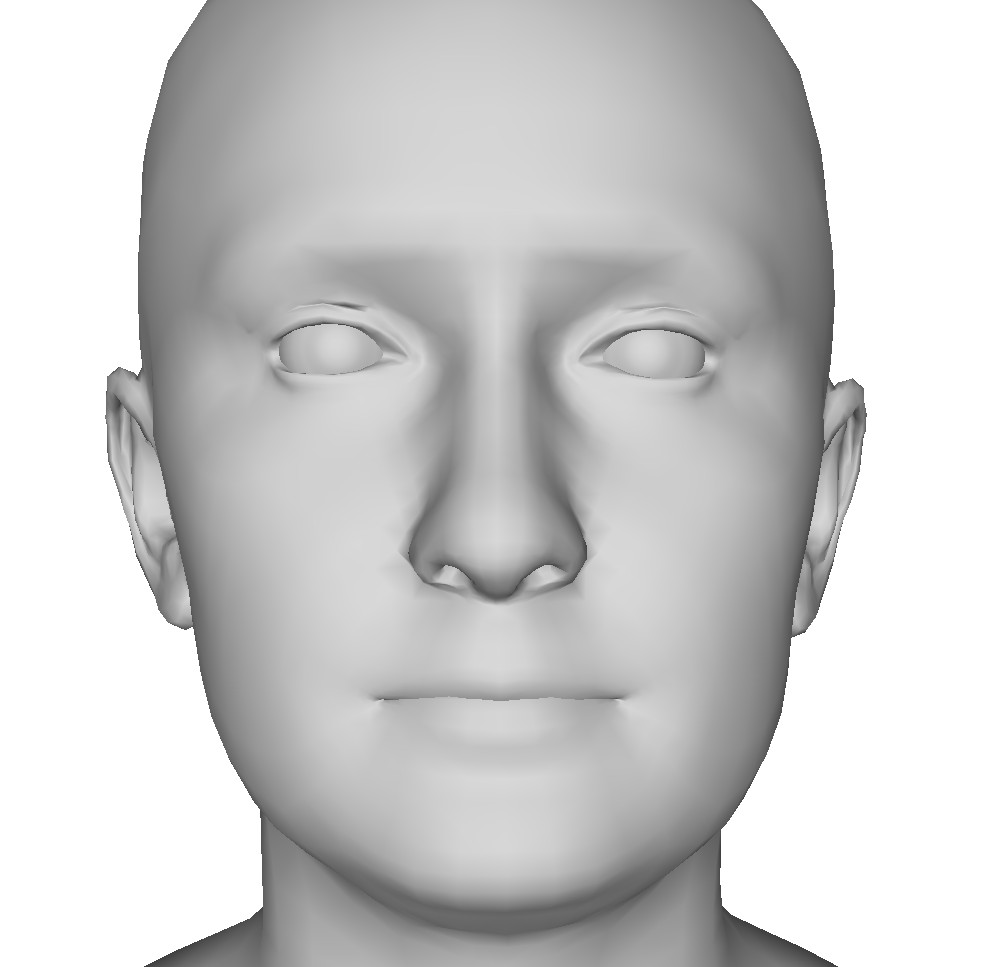}\\

\nowimg{subject_1/1_gt_1.jpg} & \nowimg{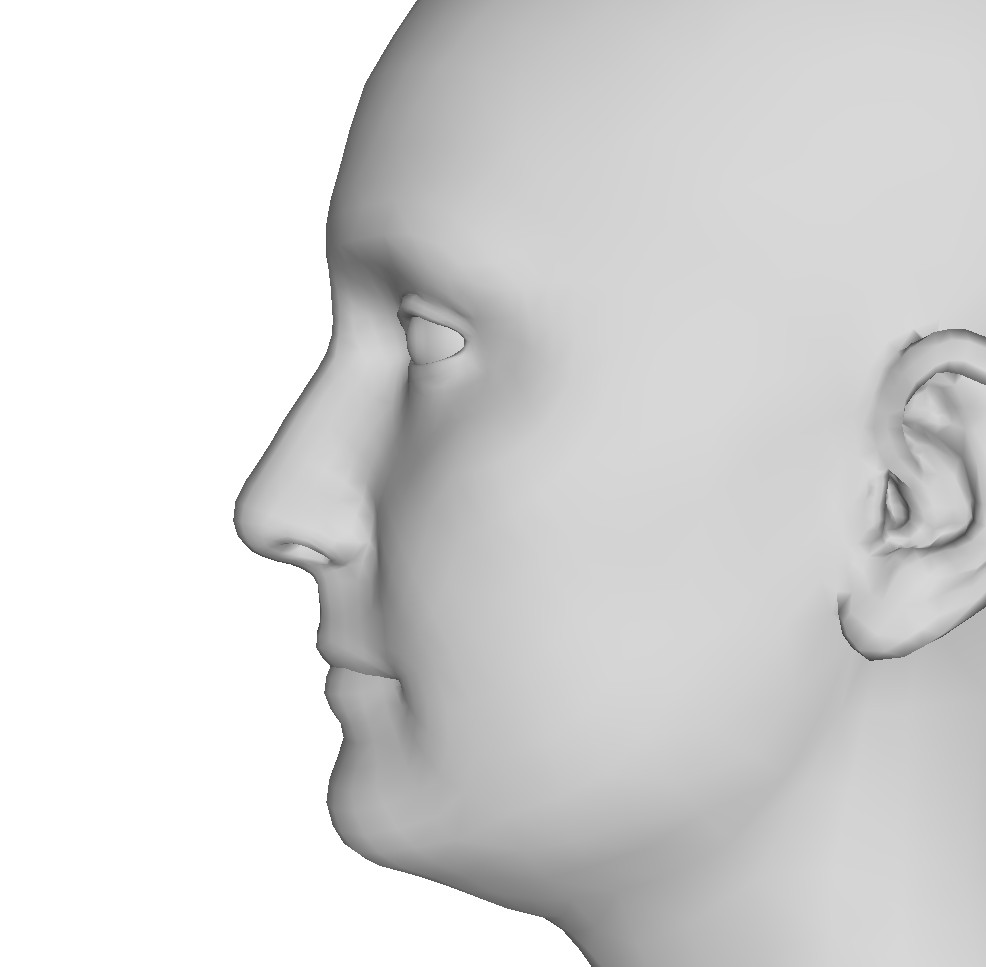} & \nowimg{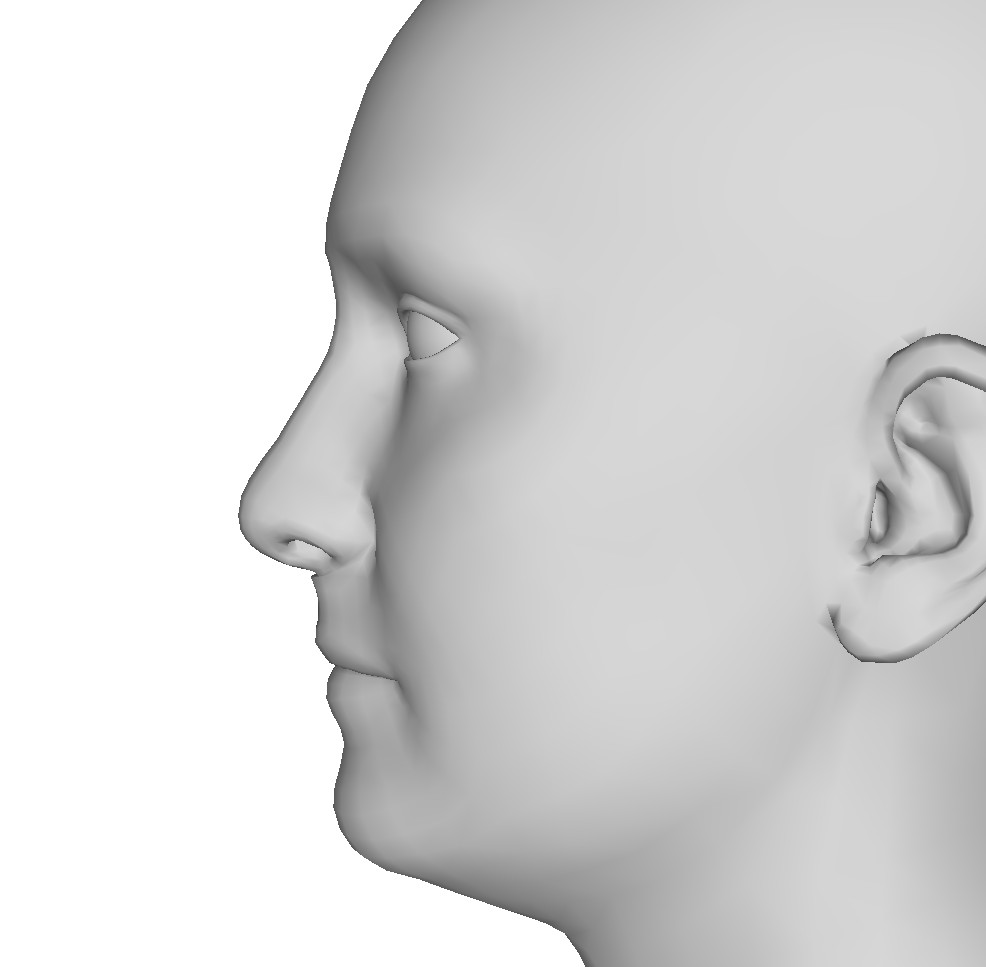} & \nowimg{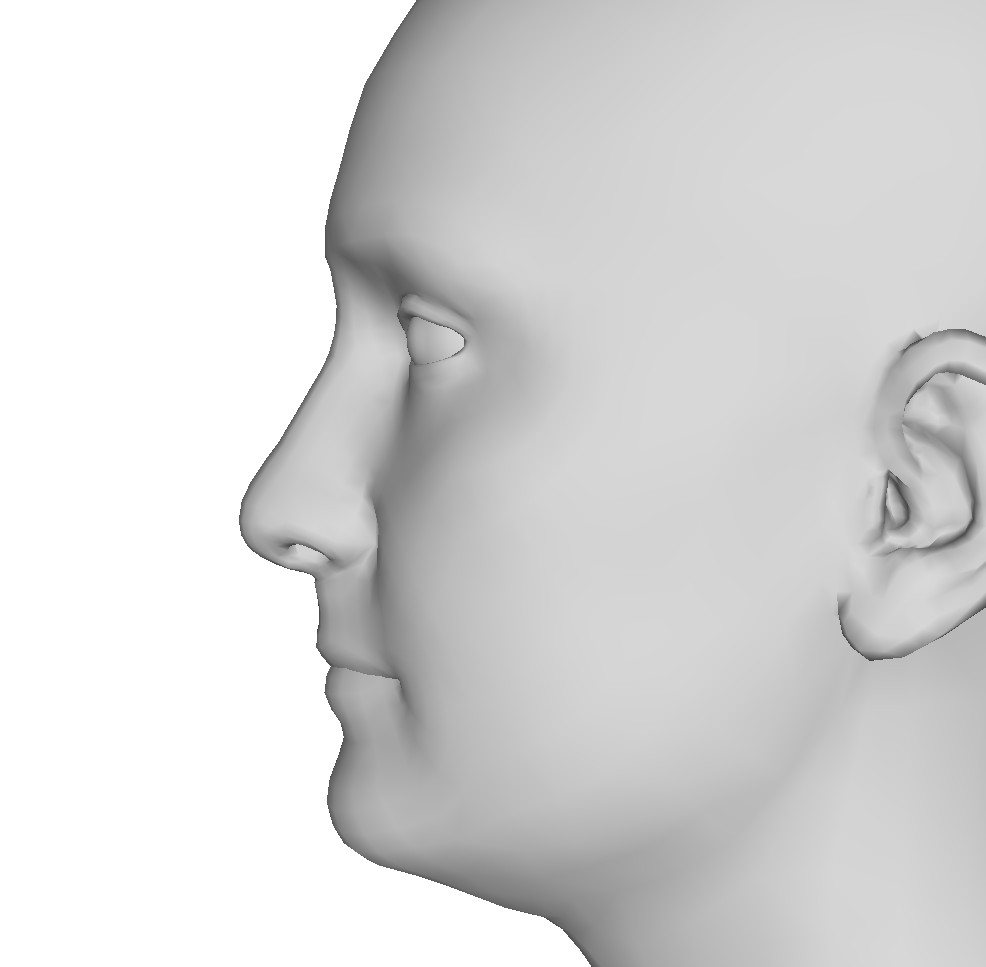}\\

\nowimg{subject_1/1_gt.jpg}  & \nowimg{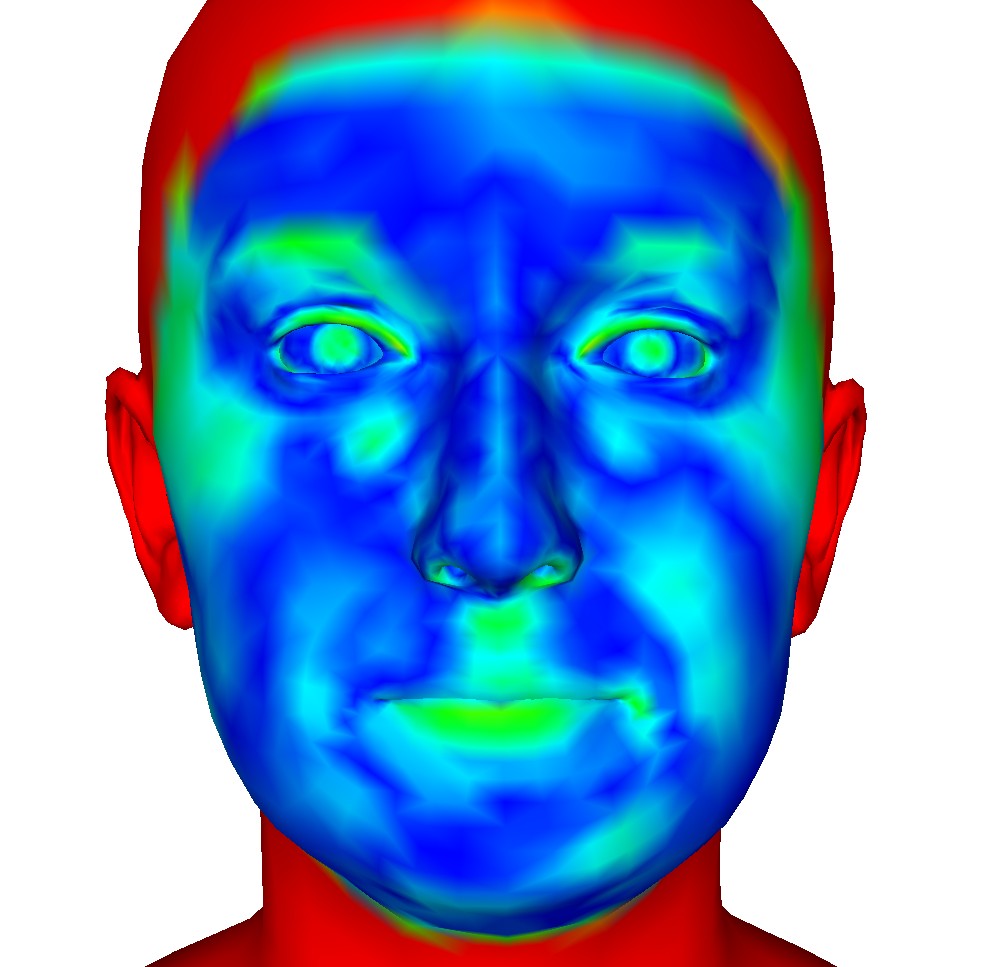} & \nowimg{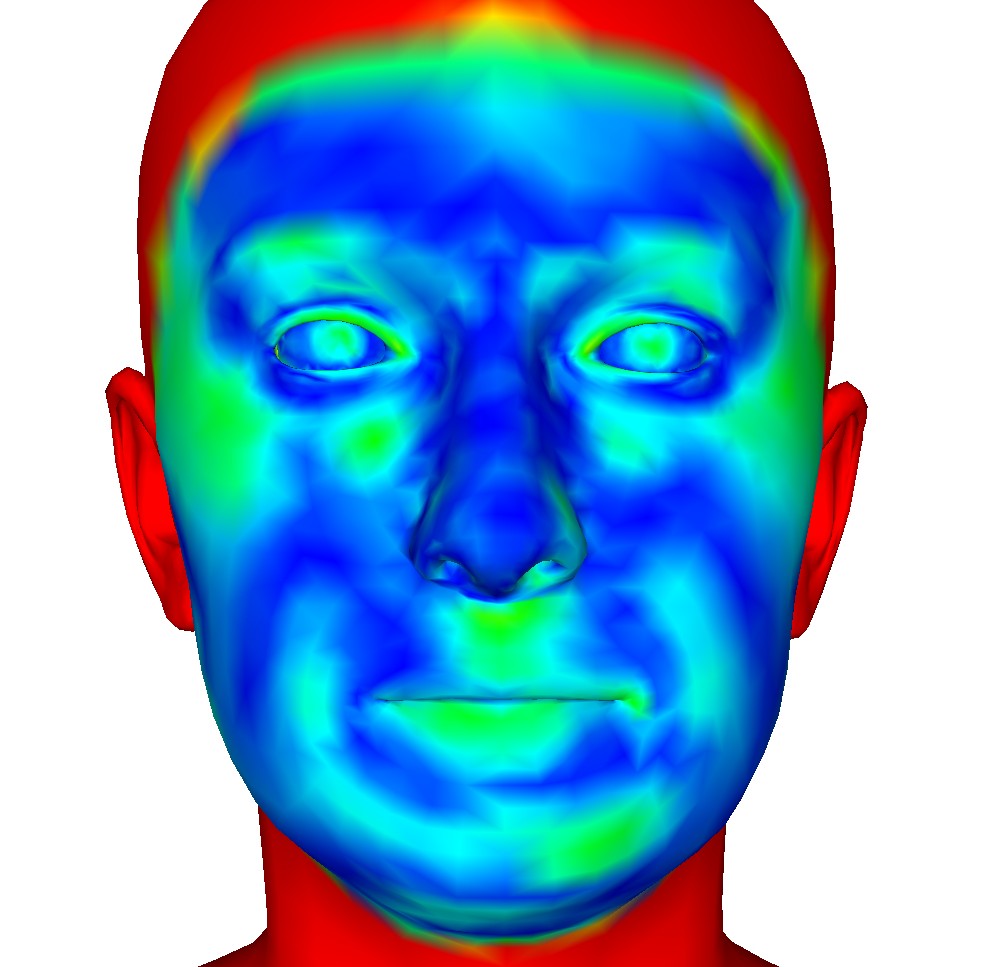} & \nowimg{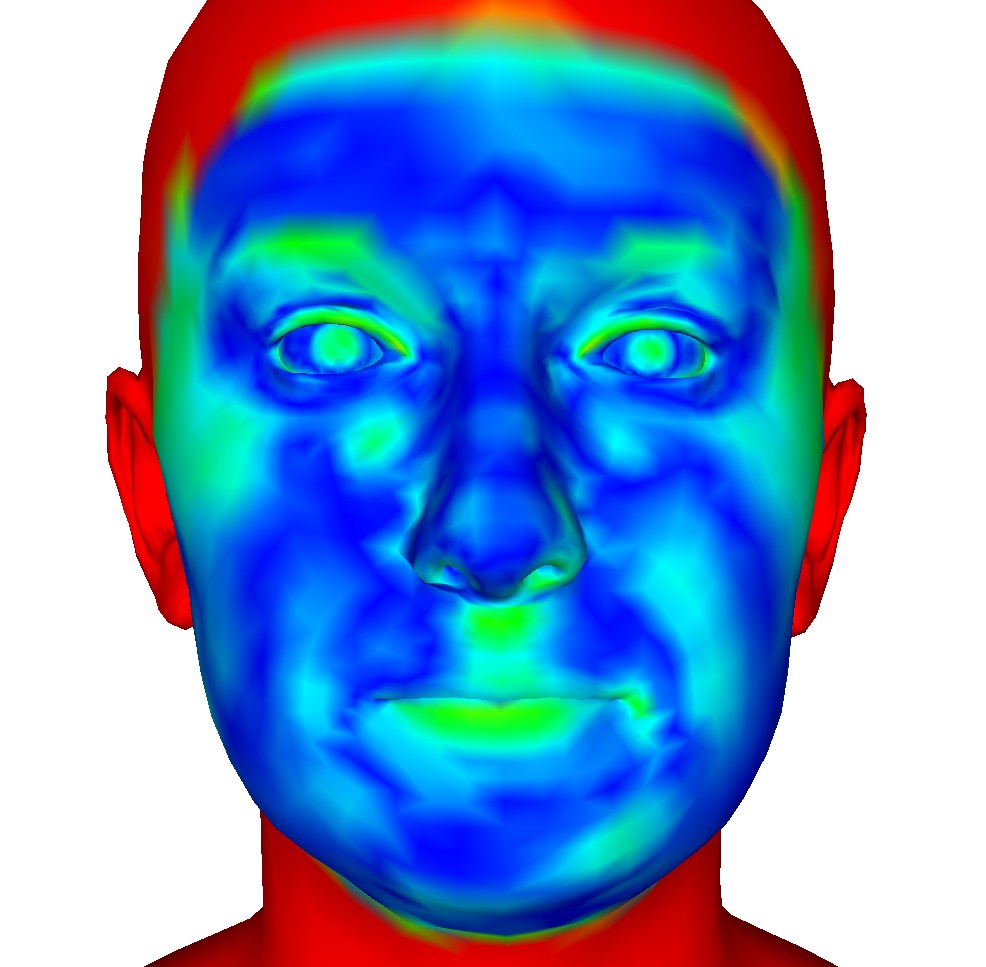} \\[0.3cm]

GT & w/o $\delta_\text{d}$ & w/o MICA & Ours \\

\end{tabular}
\vspace{0.3cm}
\caption{Ablations of our 3D model fitting module on the NoW validation set (multi-view). Figures show qualitative results without per-vertex deformations (w/o $\delta_\text{t}$), without MICA prediction (w/o MICA) and the full model fitting pipeline (Ours). Multiple views allow us to enable per-vertex deformations. Comparing to the ground truth scan, our full model with per-vertex deformations produces more accurate results in the nose region, which is visible in the 3D visualizations (top two rows) and the error plot (bottom row), where cold colors represent lower error. The MICA template prediction aids the accurate disentanglement of expression and neutral head shape. }
\label{fig:now_qualitative_multi}
\end{figure}

\newcommand{\ffhqimg}[1]{\includegraphics[width=2cm]{figures/ffhq/#1}}
\newcommand{\ffhqrow}[1]{\ffhqimg{gt/#1.jpg} & \ffhqimg{flowface/align/#1.jpg} & \ffhqimg{flowface/fit/#1.jpg} & \ffhqimg{flowface/recon/#1.jpg} & \ffhqimg{hrn/#1.jpg} & \ffhqimg{deca/#1.jpg} & \ffhqimg{sadrnet/#1.jpg} & \ffhqimg{3ddfa/#1.jpg} 
 }

\begin{figure*}[th]
\tablefont
\setlength{\tabcolsep}{1pt}
\renewcommand{\arraystretch}{0}
\centering
\begin{tabular}{c | c c c | c | c | c | c}
\ffhqrow{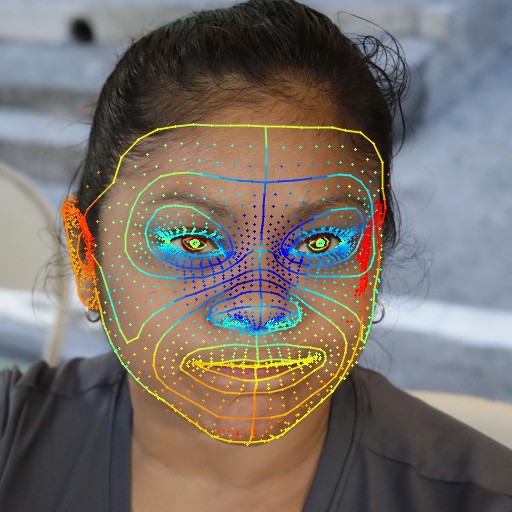} \\
\ffhqrow{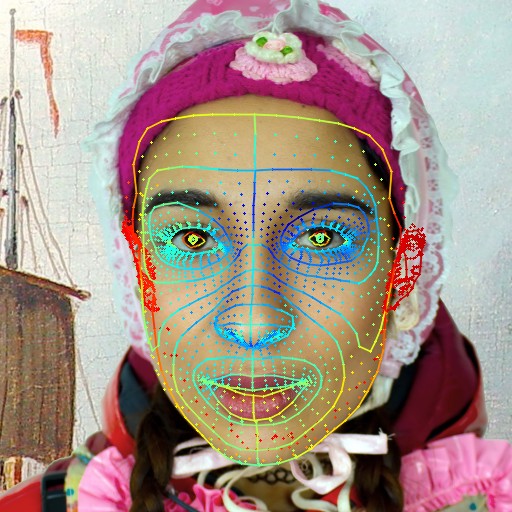} \\
\ffhqrow{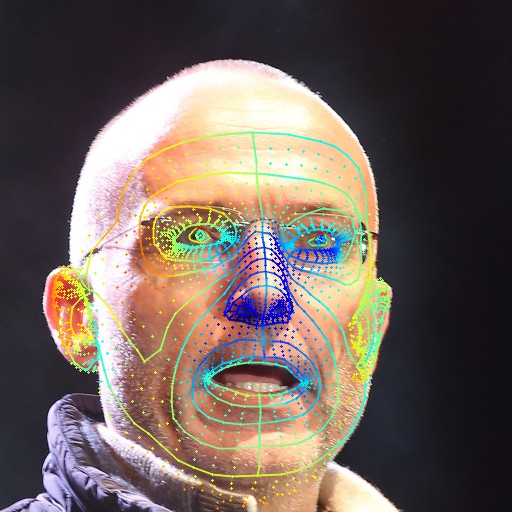} \\
\ffhqrow{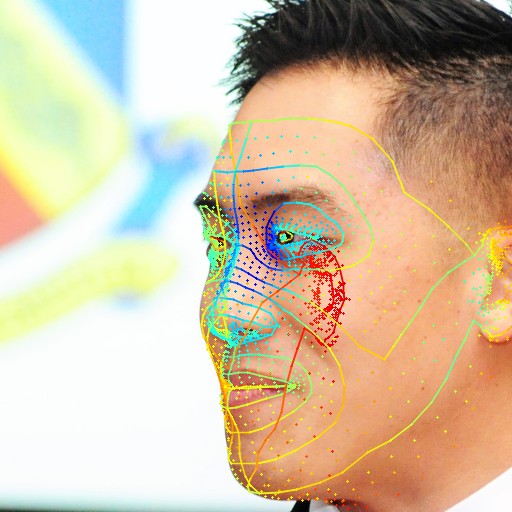} \\
\ffhqrow{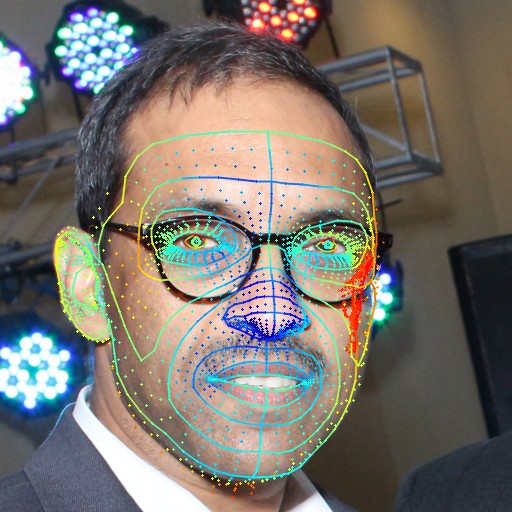} \\
\ffhqrow{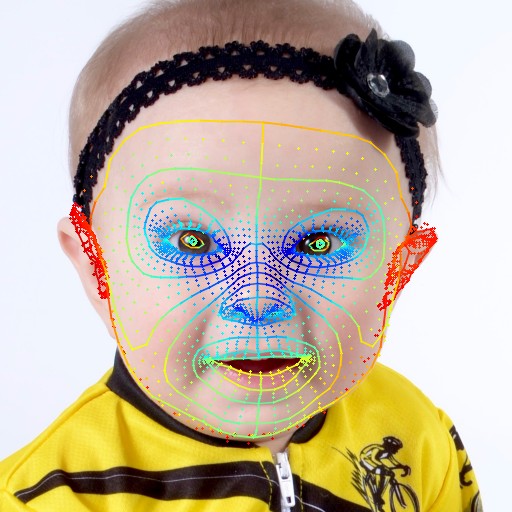} \\
\ffhqrow{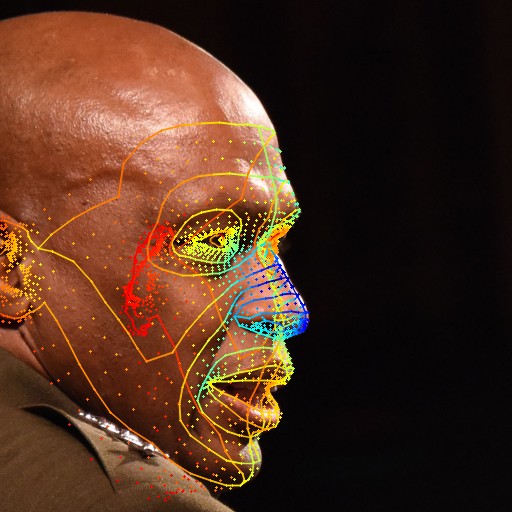} \\
\ffhqrow{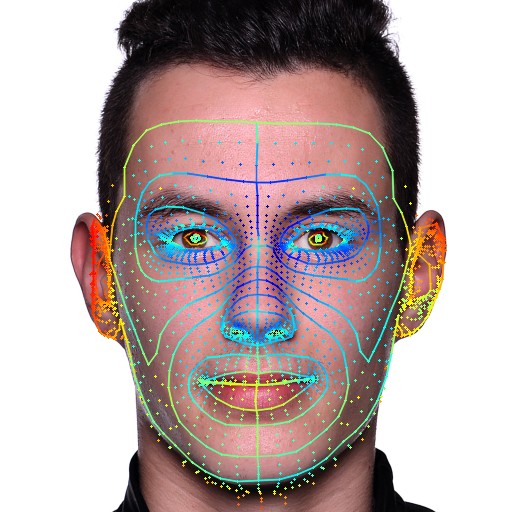} \\
\ffhqrow{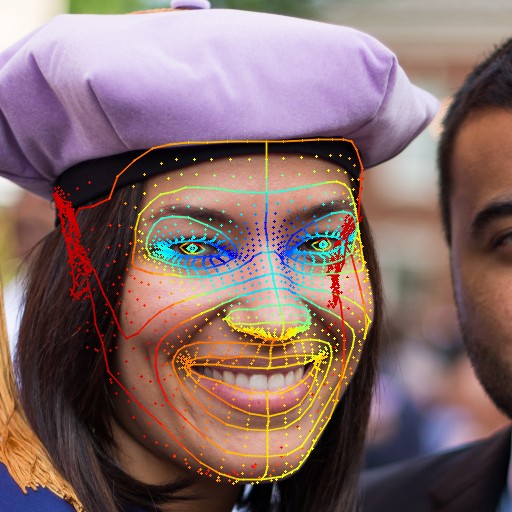} \\
\ffhqrow{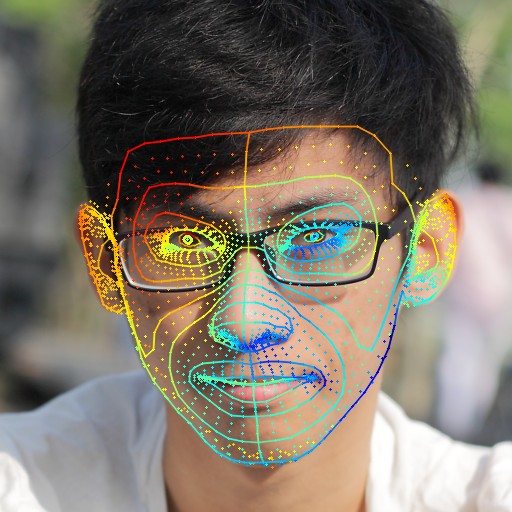} \\[0.2cm]
(a) & (b) & (c) & (d) & (e) & (f) & (g) & (h) \\[0.1cm]
\end{tabular}
\caption{Qualitative results on in-the-wild images. (a) shows the ground truth image, (b) our 2D alignment, (c) and (d) our reconstruction, (e) shows reconstructions from HRN~\cite{hrn}, (f) DECA~\cite{deca}, (g) SADRNet~\cite{sadrnet} and (h) 3DDFAv2~\cite{3ddfa_v2}. Despite being trained only on in-the-lab images, our 2D alignment module produces pixel-accurate alignment. The model fitter uses this alignment to produce accurate 3D reconstruction, even from single images. This shows that our tracker generalizes well to images with challenging occlusions, lighting.  }
\label{fig:ffhq_qualitative_1}
\end{figure*}

\begin{figure*}[th]
\tablefont
\setlength{\tabcolsep}{1pt}
\renewcommand{\arraystretch}{0}
\centering
\begin{tabular}{c | c c c | c | c | c | c }
\ffhqrow{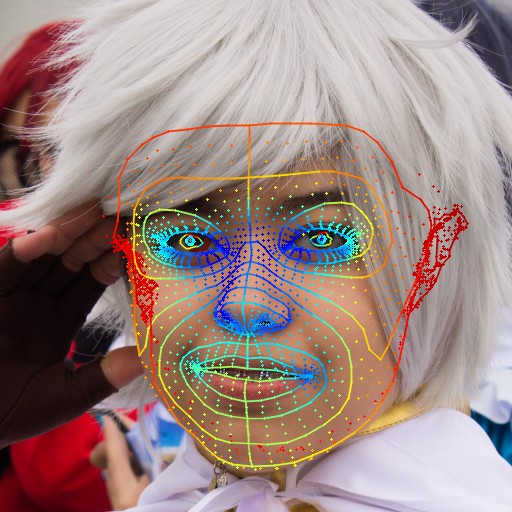} \\
\ffhqrow{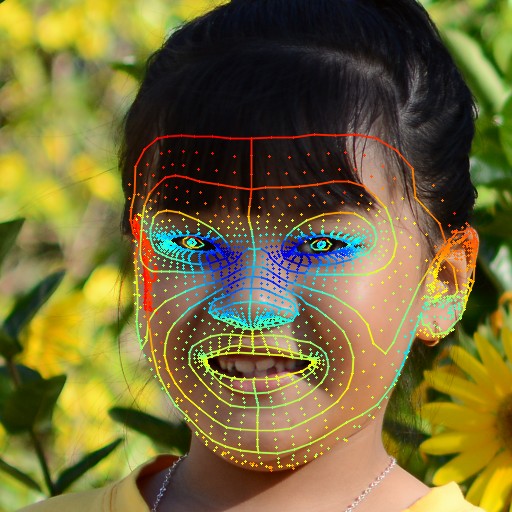} \\
\ffhqrow{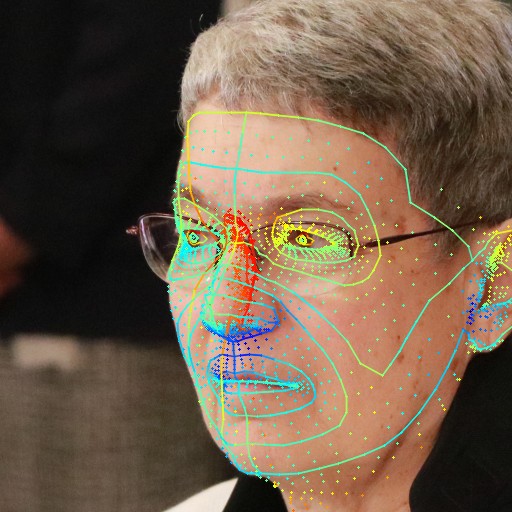} \\
\ffhqrow{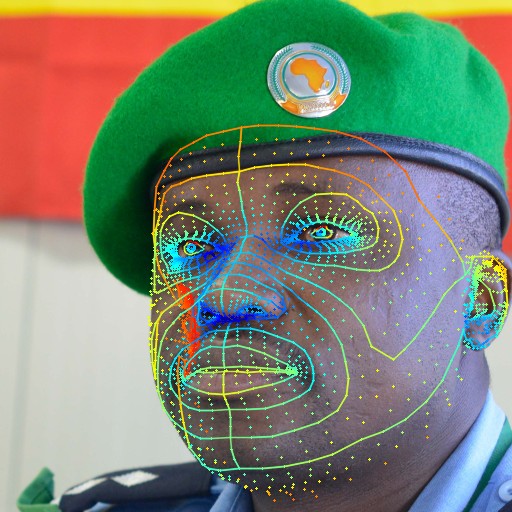} \\
\ffhqrow{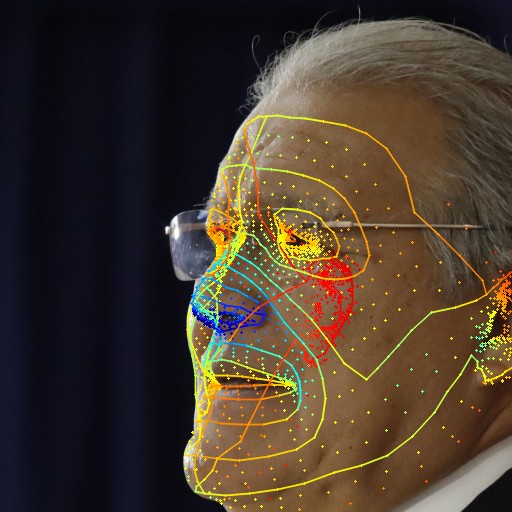} \\
\ffhqrow{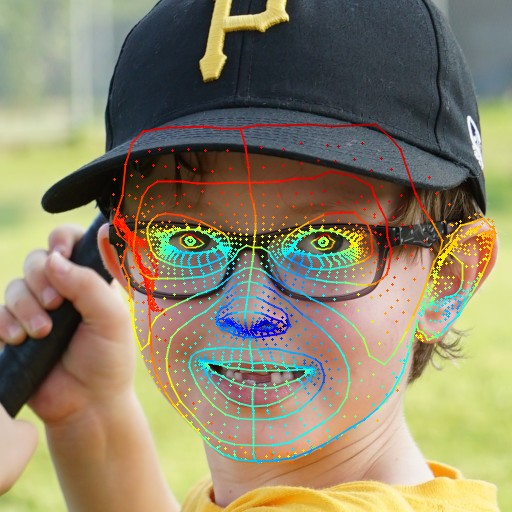} \\
\ffhqrow{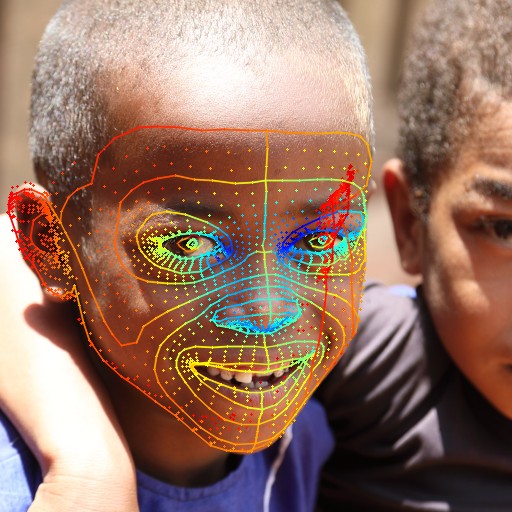} \\
\ffhqrow{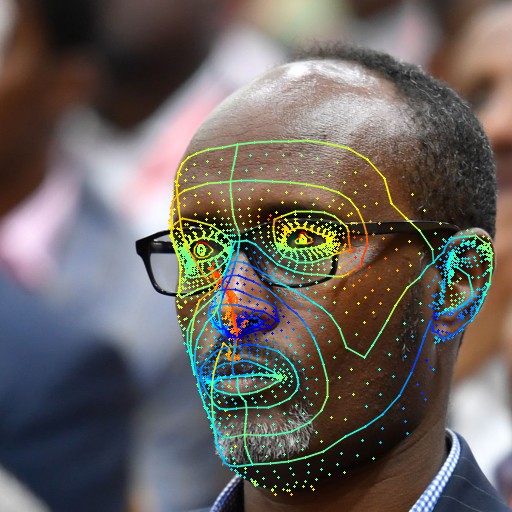} \\
\ffhqrow{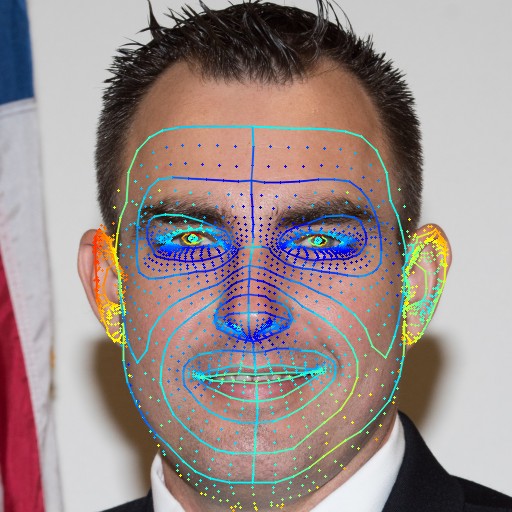} \\
\ffhqrow{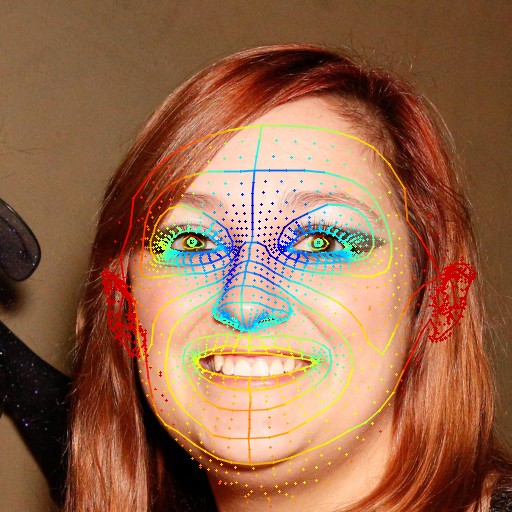} \\[0.2cm]
(a) & (b) & (c) & (d) & (e) & (f) & (g) & (h) \\[0.1cm]
\end{tabular}
\caption{Qualitative results on in-the-wild images. (a) shows the ground truth image, (b) our 2D alignment, (c) and (d) our reconstruction, (e) shows reconstructions from HRN~\cite{hrn}, (f) DECA~\cite{deca}, (g) SADRNet~\cite{sadrnet} and (h) 3DDFAv2~\cite{3ddfa_v2}. Despite being trained only on in-the-lab images, our 2D alignment module produces pixel-accurate alignment. The model fitter uses this alignment to produce accurate 3D reconstruction, even from single images. This shows that our tracker generalizes well to images with challenging occlusions, lighting.  }
\label{fig:ffhq_qualitative_2}
\end{figure*}


\end{document}